\PassOptionsToPackage{unicode,hypertexnames=false}{hyperref}
\PassOptionsToPackage{hyphens}{url}
\PassOptionsToPackage{dvipsnames,svgnames,x11names}{xcolor}
\documentclass[
  ngerman,
  english,
  onecolumn]{article}
\usepackage{xcolor}
\usepackage{amsmath,amssymb}
\usepackage{iftex}
\ifPDFTeX
  \usepackage[T1]{fontenc}
  \usepackage[utf8]{inputenc}
  \usepackage{textcomp} 
\else 
  \usepackage{unicode-math} 
  \defaultfontfeatures{Scale=MatchLowercase}
  \defaultfontfeatures[\rmfamily]{Ligatures=TeX,Scale=1}
\fi
\usepackage{lmodern}
\ifPDFTeX\else
\fi
\IfFileExists{upquote.sty}{\usepackage{upquote}}{}
\IfFileExists{microtype.sty}{
  \usepackage[]{microtype}
  \UseMicrotypeSet[protrusion]{basicmath} 
}{}
\makeatletter
\@ifundefined{KOMAClassName}{
  \IfFileExists{parskip.sty}{%
    \usepackage{parskip}
  }{
    \setlength{\parindent}{0pt}
    \setlength{\parskip}{6pt plus 2pt minus 1pt}}
}{
  \KOMAoptions{parskip=half}}
\makeatother
\makeatletter
\ifx\paragraph\undefined\else
  \let\oldparagraph\paragraph
  \renewcommand{\paragraph}{
    \@ifstar
      \xxxParagraphStar
      \xxxParagraphNoStar
  }
  \newcommand{\xxxParagraphStar}[1]{\oldparagraph*{#1}\mbox{}}
  \newcommand{\xxxParagraphNoStar}[1]{\oldparagraph{#1}\mbox{}}
\fi
\ifx\subparagraph\undefined\else
  \let\oldsubparagraph\subparagraph
  \renewcommand{\subparagraph}{
    \@ifstar
      \xxxSubParagraphStar
      \xxxSubParagraphNoStar
  }
  \newcommand{\xxxSubParagraphStar}[1]{\oldsubparagraph*{#1}\mbox{}}
  \newcommand{\xxxSubParagraphNoStar}[1]{\oldsubparagraph{#1}\mbox{}}
\fi
\makeatother

\usepackage{color}
\usepackage{fancyvrb}

\DefineVerbatimEnvironment{Highlighting}{Verbatim}{commandchars=\\\{\}}
\usepackage{framed}
\definecolor{shadecolor}{RGB}{241,243,245}
\newenvironment{Shaded}{\begin{snugshade}}{\end{snugshade}}

\newcommand{\BuiltInTok}[1]{\textcolor[rgb]{0.00,0.23,0.31}{#1}}
\newcommand{\CharTok}[1]{\textcolor[rgb]{0.13,0.47,0.30}{#1}}
\newcommand{\CommentTok}[1]{\textcolor[rgb]{0.37,0.37,0.37}{#1}}

\newcommand{\ControlFlowTok}[1]{\textcolor[rgb]{0.00,0.23,0.31}{\textbf{#1}}}
\newcommand{\DataTypeTok}[1]{\textcolor[rgb]{0.68,0.00,0.00}{#1}}
\newcommand{\DecValTok}[1]{\textcolor[rgb]{0.68,0.00,0.00}{#1}}

\newcommand{\ExtensionTok}[1]{\textcolor[rgb]{0.00,0.23,0.31}{#1}}

\newcommand{\ImportTok}[1]{\textcolor[rgb]{0.00,0.46,0.62}{#1}}

\newcommand{\KeywordTok}[1]{\textcolor[rgb]{0.00,0.23,0.31}{\textbf{#1}}}
\newcommand{\NormalTok}[1]{\textcolor[rgb]{0.00,0.23,0.31}{#1}}
\newcommand{\OperatorTok}[1]{\textcolor[rgb]{0.37,0.37,0.37}{#1}}
\newcommand{\OtherTok}[1]{\textcolor[rgb]{0.00,0.23,0.31}{#1}}
\newcommand{\PreprocessorTok}[1]{\textcolor[rgb]{0.68,0.00,0.00}{#1}}

\newcommand{\SpecialCharTok}[1]{\textcolor[rgb]{0.37,0.37,0.37}{#1}}
\newcommand{\SpecialStringTok}[1]{\textcolor[rgb]{0.13,0.47,0.30}{#1}}
\newcommand{\StringTok}[1]{\textcolor[rgb]{0.13,0.47,0.30}{#1}}
\newcommand{\VariableTok}[1]{\textcolor[rgb]{0.07,0.07,0.07}{#1}}

\usepackage{longtable,booktabs,array}
\usepackage{calc}    
\usepackage{etoolbox}
\makeatletter
\patchcmd\longtable{\par}{\if@noskipsec\mbox{}\fi\par}{}{}
\makeatother
\IfFileExists{footnotehyper.sty}{\usepackage{footnotehyper}}{\usepackage{footnote}}
\makesavenoteenv{longtable}
\usepackage{graphicx}
\makeatletter
\newsavebox\pandoc@box
\newcommand*\pandocbounded[1]{
  \sbox\pandoc@box{#1}%
  \Gscale@div\@tempa{\textheight}{\dimexpr\ht\pandoc@box+\dp\pandoc@box\relax}%
  \Gscale@div\@tempb{\linewidth}{\wd\pandoc@box}%
  \ifdim\@tempb\p@<\@tempa\p@\let\@tempa\@tempb\fi
  \ifdim\@tempa\p@<\p@\scalebox{\@tempa}{\usebox\pandoc@box}%
  \else\usebox{\pandoc@box}%
  \fi%
}
\def\fps@figure{htbp}
\makeatother

\usepackage[backend=biber, style=authoryear, doi=true, url=true, eprint=true,
            isbn=false, language=ngerman, sorting=nyt, maxbibnames=10,
            uniquename=init, giveninits=true, natbib=true]{biblatex}
\DeclareSourcemap{
  \maps[datatype=bibtex]{
    \map{
      \step[fieldsource=doi, final]
      \step[fieldset=url, null]
      \step[fieldset=eprint, null]
      \step[fieldset=archiveprefix, null]
      \step[fieldset=eprinttype, null]
    }
  }
}
\ifLuaTeX
\usepackage[bidi=basic,shorthands=off]{babel}
\else
\usepackage[bidi=default,shorthands=off]{babel}
\fi
\ifPDFTeX
\else
\babelfont{rm}[
  Extension     = .otf,
  UprightFont   = *-regular,
  ItalicFont    = *-italic,
  BoldFont      = *-bold,
  BoldItalicFont= *-bolditalic,
]{texgyretermes}
\fi
\ifLuaTeX
  \usepackage{selnolig} 
\fi

\providecommand{\tightlist}{%
  \setlength{\itemsep}{0pt}\setlength{\parskip}{0pt}}

\usepackage{arxiv}
\usepackage{orcidlink}
\usepackage{dirtree}
\usepackage{csquotes}
\usepackage{multicol}
\AtBeginDocument{\selectlanguage{english}}
\DefineVerbatimEnvironment{Highlighting}{Verbatim}{commandchars=\\\{\},fontsize=\small}
\makeatletter
\renewcommand{\@maketitle}{%
  \vbox{%
    \hsize\textwidth
    \linewidth\hsize
    \vskip 0.1in
    \@toptitlebar
    \centering
    {\LARGE\sc \@title\par}
    \@bottomtitlebar
    \textsc{\runninghead}\\
    \vskip 0.1in
    \def\And{\end{tabular}\hfil\linebreak[0]\hfil%
      \begin{tabular}[t]{c}\fontencoding{TU}\selectfont\bfseries\rule{\z@}{24\p@}\ignorespaces}%
    \def\AND{\end{tabular}\hfil\linebreak[4]\hfil%
      \begin{tabular}[t]{c}\fontencoding{TU}\selectfont\bfseries\rule{\z@}{24\p@}\ignorespaces}%
    \begin{tabular}[t]{c}\fontencoding{TU}\selectfont\bfseries\rule{\z@}{24\p@}\@author\end{tabular}%
    \vskip 0.4in \@minus 0.1in \center{\today} \vskip 0.2in
  }%
}
\newenvironment{zusammenfassung}
  {\centerline{\large \bfseries \scshape Zusammenfassung}\begin{quote}}
  {\end{quote}}
\makeatother
\makeatletter
\@ifpackageloaded{tcolorbox}{}{\usepackage[skins,breakable]{tcolorbox}}
\@ifpackageloaded{fontawesome5}{}{\usepackage{fontawesome5}}
\definecolor{quarto-callout-note-color}{HTML}{0758E5}
\definecolor{quarto-callout-note-color-frame}{HTML}{4582ec}
\makeatother
\makeatletter
\@ifpackageloaded{caption}{}{\usepackage{caption}}
\AtBeginDocument{%
\ifdefined\contentsname
  \renewcommand*\contentsname{Inhaltsverzeichnis}
\else
  \newcommand\contentsname{Inhaltsverzeichnis}
\fi
\ifdefined\listfigurename
  \renewcommand*\listfigurename{Abbildungsverzeichnis}
\else
  \newcommand\listfigurename{Abbildungsverzeichnis}
\fi
\ifdefined\listtablename
  \renewcommand*\listtablename{Tabellenverzeichnis}
\else
  \newcommand\listtablename{Tabellenverzeichnis}
\fi
\ifdefined\figurename
  \renewcommand*\figurename{Abbildung}
\else
  \newcommand\figurename{Abbildung}
\fi
\ifdefined\tablename
  \renewcommand*\tablename{Tabelle}
\else
  \newcommand\tablename{Tabelle}
\fi
\ifdefined\abstractname
  \renewcommand*\abstractname{Abstract}
\else
  \newcommand\abstractname{Abstract}
\fi
}
\@ifpackageloaded{float}{}{\usepackage{float}}
\makeatother
\makeatletter
\@ifpackageloaded{caption}{}{\usepackage{caption}}
\makeatother
\usepackage{amsthm}
\theoremstyle{definition}
\newcommand*\examplename{Example}
\newtheorem{example}{\examplename}[section]
\usepackage{bookmark}
\IfFileExists{xurl.sty}{\usepackage{xurl}}{} 
\hypersetup{
  pdftitle={Time-Series Forecasting in Safety-Critical Environments: An Open-Source Package for EU-AI-Act-Compliant Development / Zeitreihenprognose in sicherheitskritischen Umgebungen: Ein Open-Source-Paket für die KI-VO-konforme Entwicklung},
  pdflang={en},
  colorlinks=true,
  linkcolor={blue},
  filecolor={Maroon},
  citecolor={Blue},
  urlcolor={Blue},
  pdfcreator={LaTeX via pandoc}}

\newcommand{\runninghead}{Working Paper }
\title{Time-Series Forecasting in Safety-Critical Environments:\\
An Open-Source Package for EU-AI-Act-Compliant Development\\[1.2ex]
{\large\itshape Zeitreihenprognose in sicherheitskritischen Umgebungen:\\
Ein Open-Source-Paket für die KI-VO-konforme Entwicklung}}
\def\asep{\\\\\\ } 
\author{\textbf{Thomas
Bartz-Beielstein}~\orcidlink{0000-0002-5938-5158}\\\\\href{mailto:bartzbeielstein@gmail.com}{bartzbeielstein@gmail.com}\asep\textbf{Eva
Bartz}~\orcidlink{0009-0005-2989-2385}\\\\Bartz \& Bartz GmbH, 51643
Gummersbach,
Deutschland\\\\\href{mailto:eva.bartz@bartzundbartz.de}{eva.bartz@bartzundbartz.de}}
\date{}
\begin{document}
\maketitle

\renewcommand*\figurename{Figure}
\renewcommand*\tablename{Table}
\renewcommand*\examplename{Example}


\begin{abstract}
With \texttt{spotforecast2-safe} we present an
integrated \emph{Compliance-by-Design} approach to Python-based point
forecasting of time series in safety-critical environments. A review of
the relevant open-source tooling shows that existing compliance solutions
operate consistently outside of the library to be used, e.g., as
scanners, templates, or runtime layers. \texttt{spotforecast2-safe}
takes the inverse approach and anchors the requirements of Regulation
(EU) 2024/1689 (the EU AI Act, in German: \emph{KI-VO}), of IEC 61508,
of the ISA/IEC 62443 standards series, and of the Cyber Resilience Act
\emph{within} the library: in application-programming-interface
contracts, persistence formats, and the verification pipeline of the
release process. The
approach is operationalised by four non-negotiable code-development rules
(zero dead code, deterministic processing, fail-safe handling, minimal
dependencies) together with the corresponding process rules (model card,
executable docstrings, documented release procedure,
Common-Platform-Enumeration (CPE)
identifier, REUSE-conformant licensing, structured audit log). Interactive
visualisation, hyperparameter tuning and automated machine learning, as well as
deep-learning and large-language-model backends are deliberately
excluded, because each of these components either enlarges the attack
surface, introduces non-determinism, or impairs reproducibility. Every
article of the EU AI Act that is relevant to the library is discussed
in detail. A
bidirectional traceability matrix maps every regulatory provision onto
the corresponding mechanism in the code. An end-to-end example of
European-market electricity generation, transmission, and consumption
forecasting demonstrates the application. The package is open-source and
available under Affero General Public License (AGPL) 3.0-or-later.

\end{abstract}

\bigskip
\noindent\textit{Note:} This is the English version of the paper.
A German version (\emph{Zeitreihenprognose in sicherheitskritischen
Umgebungen}) follows further down in the same PDF. Both versions
share a single bibliography, printed once at the end of the document.

\section{Introduction}\label{sec-introduction-en}

The practice of software development in the field of artificial
intelligence (AI\footnote{In what follows we use AI as an umbrella term
  covering machine learning, learning with deep neural networks, and
  language models.}) and in the field of critical
infrastructure\footnote{The definition of KRITIS (critical infrastructures) follows §2 of the
  German law on the Federal Office for Information Security and on the
  information security of facilities (BSI-Gesetz - BSIG) \citep{deut25a} and §2 of the
  framework law for the resilience of critical installations
  (KRITIS-Dachgesetz - KRITISDachG), together with the IT Security Act
  2.0 and the act transposing the NIS-2 directive. KRITIS therefore
  refers to a service for the supply of the general public in sectors
  whose failure or impairment would lead to significant supply shortages
  or to a threat to public safety \citep{kritisDachG26}.} has, until
now, evolved largely in separate software ecosystems. AI-based software
development for time-series forecasting favours feature-rich libraries
that bundle visualisation, hyperparameter search, probabilistic
deep-learning backends, and interactive dashboards into a single
installable package. Such broad libraries are excellently suited to
exploratory work and innovative research. However, this wealth of
features turns into a liability the moment the resulting forecasting
model is deployed in an environment that must withstand audits for
safety-critical systems. In the following, this is referred to as the
``inherent conflict between research and safety'', because the
requirements on software development differ between research and
critical infrastructure.

Visualisation tools, for instance, require the inclusion of a
larger number of additional graphics libraries. The same holds for the
hyperparameter-tuning layer and for deep-learning models, which
additionally exhibit very short release cycles, so that the long-term
reproducibility that regulators demand is not easily ensured.

A concrete review of the 2026 Python forecasting landscape confirms this
tension. The three dominant Python libraries, \texttt{sktime}
\citep{loni19a}, \texttt{Darts} \citep{herz22a}, and \texttt{skforecast}
\citep{scip24a}, ship extensive functionality that goes far beyond
forecasting per se. The classical libraries (\texttt{statsmodels},
\texttt{pmdarima}) are leaner but offer neither dedicated fail-safe
semantics for invalid input nor a fixed identifier under the \emph{Common Platform Enumeration}
(CPE) for vulnerability tracking.

\begin{figure}

\centering{

\pandocbounded{\includegraphics[keepaspectratio]{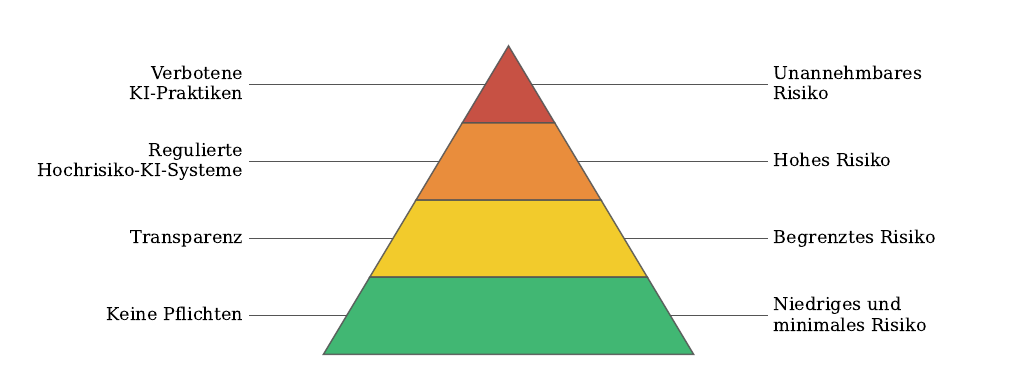}}

}

\caption{\label{fig-risk-pyramid-en}The four risk tiers of the EU AI Act
in their canonical pyramid depiction \citep{euro24a}. The depth of the
regulatory intervention rises with the risk level: outright prohibition
at the top, no obligations at the base. \texttt{spotforecast2-safe} is
designed for use within systems on the second-highest tier
(\emph{high risk}).}

\end{figure}

When such software libraries are deployed in safety-critical
environments, the regulatory requirements that arise, among others,
from Regulation (EU) 2024/1689 on artificial intelligence (referred
to below as the ``EU AI Act'' or, in German, \emph{KI-VO}) must be
taken into account. The EU AI Act follows a \emph{risk-based} approach:
the regulatory intervention is tailored to the concrete risk level, not
to the technology as such \citep{stet24a}.
Figure~\ref{fig-risk-pyramid-en} arranges the four resulting tiers
(unacceptable, high, limited, and low or minimal risk) in their
canonical depiction as a \emph{pyramid of risk}.

In heavily simplified form, the EU AI Act defines the four risk
tiers as follows:

\begin{itemize}
\tightlist
\item
  \emph{Unacceptable risk.} The EU AI Act prohibits manipulative AI
  systems using subliminal techniques, the targeted exploitation of
  vulnerable groups, social scoring by public authorities, as well as
  real-time remote biometric identification in publicly accessible
  spaces for law-enforcement purposes (subject to narrow exceptions).
\item
  \emph{High risk.} This category covers AI systems acting as a safety
  component of, or being themselves, a product regulated by existing EU
  product-safety legislation (toys, aviation, motor vehicles, medical
  devices, lifts), as well as systems deployed in the use cases listed
  in Annex III of the EU AI Act. Providers must conduct an ex-ante
  conformity assessment, register the system in an EU database, and
  implement seven substantive requirements: EU-1.\ human oversight,
  EU-2.\ technical robustness and safety, EU-3.\ privacy and data
  governance, EU-4.\ transparency, EU-5.\ diversity, non-discrimination
  and fairness, EU-6.\ social and environmental well-being, and EU-7.\
  accountability. These seven requirements derive from the
  \emph{Ethics Guidelines for Trustworthy AI} of the High-Level Expert
  Group on Artificial Intelligence (AI-HLEG) \citep{hleg19a} and are
  operationalised by the \emph{Assessment List for Trustworthy AI}
  (ALTAI) \citep{hleg20a, stet24a}.
\item
  \emph{Limited risk.} For chatbots, emotion-recognition systems,
  biometric categorisation, and for systems that generate or manipulate
  image, audio, or video content (so-called \emph{deepfakes}),
  transparency obligations apply: users must be able to recognise that
  they are interacting with an AI system.
\item
  \emph{Low or minimal risk.} No additional obligations. The EU AI Act
  encourages voluntary codes of conduct under which providers of
  non-high-risk systems may adopt the requirements imposed on high-risk
  systems.
\end{itemize}

\begin{longtable}[]{@{}
  >{\raggedright\arraybackslash}p{(\linewidth - 4\tabcolsep) * \real{0.3529}}
  >{\raggedright\arraybackslash}p{(\linewidth - 4\tabcolsep) * \real{0.2353}}
  >{\raggedright\arraybackslash}p{(\linewidth - 4\tabcolsep) * \real{0.4118}}@{}}
\caption{Semantic mapping of EU AI Act articles to the seven high-risk
requirements. EU-5 (diversity and non-discrimination) and EU-6 (social
and environmental well-being) are not addressed by an own EU-AI-Act row
in Table~\ref{tbl-compliance-en}, because they do not trigger an
immediate implementation obligation for the software package presented
here.}\label{tbl-kivo-eu-mapping-en}\tabularnewline
\toprule\noalign{}
\begin{minipage}[b]{\linewidth}\raggedright
\textbf{EU AI Act article}
\end{minipage} & \begin{minipage}[b]{\linewidth}\raggedright
\textbf{High-risk requirement}
\end{minipage} & \begin{minipage}[b]{\linewidth}\raggedright
\textbf{Rationale}
\end{minipage} \\
\midrule\noalign{}
\endfirsthead
\toprule\noalign{}
\begin{minipage}[b]{\linewidth}\raggedright
\textbf{EU AI Act article}
\end{minipage} & \begin{minipage}[b]{\linewidth}\raggedright
\textbf{High-risk requirement}
\end{minipage} & \begin{minipage}[b]{\linewidth}\raggedright
\textbf{Rationale}
\end{minipage} \\
\midrule\noalign{}
\endhead
\bottomrule\noalign{}
\endlastfoot
Art.~9 risk management & EU-7 & Accountability: risk-management
system as a procedural evidence framework \\
Art.~10 data governance & EU-3 & Privacy and data governance \\
Art.~11 technical documentation & EU-7 & Accountability through
documentation \\
Art.~12 record-keeping obligations & EU-7 & Accountability through
logs \\
Art.~13 transparency & EU-4 & Transparency \\
Art.~14 human oversight & EU-1 & Human oversight \\
Art.~15 accuracy, robustness, cybersecurity & EU-2 & Technical
robustness and safety \\
Art.~17 quality-management system & EU-7 & Accountability
(umbrella) \\
Art.~18--21 record-keeping obligations of providers & EU-7 &
Accountability \\
Art.~72 post-market monitoring & EU-7 & Accountability \\
Art.~73 reporting & EU-7 & Accountability \\
\end{longtable}

The open-source software library \texttt{spotforecast2-safe} presented
in this report is designed as a component for forecasting applications
in safety-critical environments, such as load forecasting in grid
operations. Whether such an application is a high-risk system within the
meaning of the EU AI Act depends on the concrete deployment.
Table~\ref{tbl-kivo-eu-mapping-en} shows the semantic mapping of the EU
AI Act articles to the seven high-risk requirements (EU-1 to EU-7). The
EU AI Act \citep{euro24a} obliges the provider of a high-risk AI system
to demonstrate compliance with the requirements of Chapter III
Section 2 (Art.~16(a) of the EU AI Act). The deployer bears the
separate duties of Art.~26 of the EU AI Act. When AI-research-grade software packages are used, a
deployer\footnote{Depending on the context, a deployer may additionally
  take on the role of a provider.} inherits none of these demonstrations
from the software library itself. Such demonstrations may have to be
re-established at considerable expense.

By contrast, \texttt{spotforecast2-safe} shifts this evidence-bearing
inside the software library, so that the deployer inherits it. To this
end, the design principles described in
Chapter~\ref{sec-principles-en} are introduced. The
discipline of translating regulatory provisions into implementable,
verifiable software requirements while maintaining a bidirectional
traceability between every provision and the mechanism that fulfills it
is known as \emph{Legal-Requirements-Engineering} (LRE) \citep{otto07a,
brea08a}. \citet{kose24a} provides a systematic survey of the field.
LRE ensures that all relevant statutes, directives, recitals,
explanatory background, and standards are integrated into the software
development process to deliver sustained legal compliance
\citep{hens24a}. LRE means that legal requirements should be considered
from the very design stage of an AI system. This avoids tedious and
potentially expensive adjustments at later stages of development
(\emph{compliance-by-design}). Chapter~\ref{sec-lre-en} shows how the
\texttt{spotforecast2-safe} software applies the LRE discipline.

Compliance-monitoring libraries such as \texttt{Deepchecks}
\citep{chor22a} and
\texttt{Evidently~AI}\footnote{\url{https://github.com/evidentlyai/evidently}}
partially cover the evaluation side of the robustness requirement under
Art.~15 EU~AI~Act. They sit on top of an already finished
forecasting software and alter neither the provenance trail, nor
determinism, nor the dependency footprint of the software itself.
Chapter~\ref{sec-deepchecks-en} and Chapter~\ref{sec-evidently-en} (in
the appendix) show how the \texttt{spotforecast2-safe} software
presented here can be combined with the compliance-monitoring features
of \texttt{Deepchecks} and \texttt{Evidently~AI}.

A software library that rests on a software stack with dozens of
unpinned dependencies is not unconditionally uncertifiable. The cost of
constructing the required evidence trail grows with every
additional dependency.

\begin{figure}

\centering{

\pandocbounded{\includegraphics[keepaspectratio]{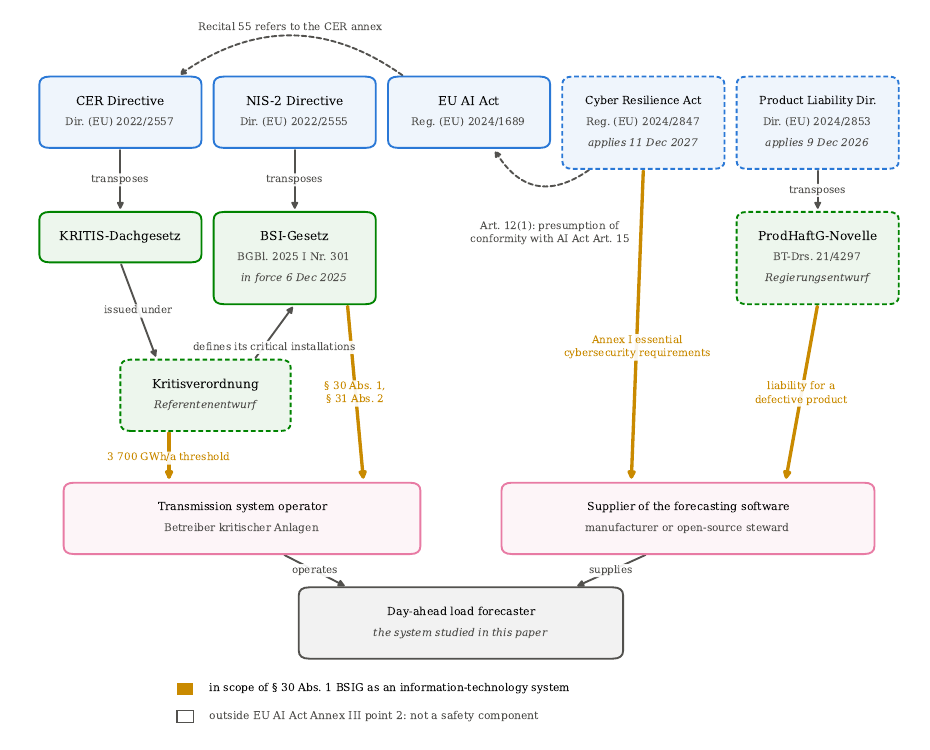}}

}

\caption{\label{fig-regulatory-map-en}Simplified representation of
selected legal acts bearing on an AI load-forecasting system, and the
actors they bind. \emph{Blue} marks Union legal acts, \emph{green} their
German counterparts, \emph{pink} the two actors. Solid arrows denote a
derivation, i.e., a directive transposed into German law, or a statutory
ordinance issued on the basis of an act. The EU AI Act and the Cyber
Resilience Act have no German counterpart because an EU regulation
applies directly in EU member states; hence no arrow leads downwards
from them. Dashed arrows denote one legal act referring to another.
\emph{Ochre} arrows denote the point at which an obligation attaches and
are each labelled with the provision that creates it. A \emph{dashed}
outline marks a legal act that is still a draft or not yet applicable.
Source: \citet{bart26o}.}

\end{figure}%

The requirements EU-1 to EU-7 defined by the EU AI Act are
not the only regulatory metrics against which an AI forecasting library
must be measured if it is to be deployed in safety-critical
environments. Figure~\ref{fig-regulatory-map-en}, taken from
\citet{bart26o}, shows, in strongly simplified form, the relevant
regulatory instruments. In what
follows, particular emphasis falls on the \emph{Cyber Resilience Act}
(CRA) and the \emph{Product Liability Directive} (PLD), both of which
act directly on software and AI systems in the EU.

\emph{Cyber Resilience Act.} In industrial deployment contexts, a
forecasting library runs as a software component on level 2 or 3 of the
Purdue reference model\footnote{A reference model for automation and
  industrial networks.} (ISA-95 / IEC~62264), where the
dominant cybersecurity framework is the ISA/IEC~62443 standards series
\citep{iec62443_1_1}.\footnote{The standards series distributes
  responsibilities among asset owners (62443-2-1
  \citep{iec62443_2_1}), service providers (62443-2-4
  \citep{iec62443_2_4}), system integrators (62443-3-2
  \citep{iec62443_3_2}, 62443-3-3 \citep{iec62443_3_3}), and component
  manufacturers (62443-4-1 \citep{iec62443_4_1}, 62443-4-2
  \citep{iec62443_4_2}). The latter two parts are directly applicable
  to a software vendor.} Within the European Union, the 62443
series, and EN IEC~62443-4-1 in particular, is regarded as a promising
basis for the harmonised standards to be developed under the
standardisation request M/606 of Commission Implementing Decision
C(2025)~618 of 3~February~2025 \citep{euM606} under the now-in-force
Cyber Resilience Act \citep{euCRA24}. The Decision itself
designates no existing standard as a reference standard. The 62443
series furthermore shapes the risk-management duties under Directive (EU)
2022/2555 (\emph{NIS-2 Directive}) \citep{euNIS22} as well as the
Critical Entities Resilience Directive (CER Directive)
\citep{euRCE22}. \citet{fluc24a} characterises 62443-2-1 as the
``equivalent of ISO/IEC~27001 for cyber-physical systems (CPS)''. The
part of the standards series directly binding for
\texttt{spotforecast2-safe} is 62443-4-1.

\emph{Civil-law product liability.} Beyond the public-law conformity
and cybersecurity frameworks, the new Product Liability Directive (Directive (EU) 2024/2853)
\citep{euPLD24} extends strict product
liability to software and AI systems: these are, for the first time,
explicitly captured as a ``product'' within the meaning of liability
law, including integrated digital services and security-relevant
post-market updates. The directive replaces the 1985 Product Liability
Directive 85/374/EEC and applies to products placed on the market or
put into service after 9~December~2026. By that date the Member States
must also have completed national transposition. In Germany, the
directive is being implemented by the draft act on the modernisation of
product-liability law \citep{breg26a}, which at the time of this report
is undergoing parliamentary procedure (Bundestag printed paper
21/4297, first reading on 4~March~2026). Open-source software developed
or made available outside of a commercial activity remains exempt from
product liability. For a library released under the Affero General Public License
(AGPL-3.0-or-later) such
as \texttt{spotforecast2-safe}, liability arises only where it is made
available on the Union market in the course of a commercial activity,
but there it carries the same duties of care and documentation that
the CRA and the EU AI Act impose. At the time of writing this report,
neither instrument bound the package: the CRA applies only from
11~December~2027, and the Product Liability Directive covers products
placed on the market after 9~December~2026.

\emph{Open-source carve-outs.} There is an almost unsurveyable thicket
of regulations that providers and deployers of AI models or systems may
have to take into account. The most prominent are Art.~50 and 51 of the
EU AI Act, the EU General Data Protection Regulation (GDPR), the Data
Act, and the violation of copyright, personality, or labour rights.
This contribution deliberately confines itself, as a first foray of
free and open-source software (FOSS) into KRITIS, to the regulations
mentioned at the outset.

\emph{EU AI Act.} On top of the FOSS
exemption of the PLD, the EU AI Act and the CRA each contain their
own, mutually independent special rules for FOSS that are relevant
for \texttt{spotforecast2-safe} as a library released under
AGPL-3.0-or-later. Art.~2(12) of the EU AI Act \citep{euro24a}
excludes AI systems released under a free or open-source licence in
principle from the regulation's scope, but lifts that exception again
for high-risk AI systems within the meaning of Annex~III, for
prohibited practices under Art.~5 of the EU AI Act, and for
transparency-bound systems under Art.~50 of the EU AI Act. Whether
that exception falls away in the typical KRITIS use case of a load
forecast turns solely on whether the system is put into service as a
\emph{safety component} in the management and operation of the
electricity supply within the meaning of Annex~III No.~2 of the EU AI
Act. Recital~55 construes that term narrowly: safety components
directly protect the physical integrity of critical infrastructure or
the health and safety of persons, but are precisely not required for
the system itself to function. Water-pressure monitoring and the
operation of fire-alarm systems in cloud computing centres can be
named here as examples. A day-ahead load forecast as a rule satisfies
neither element: it protects nothing directly, and it is precisely
required for operation. Such a forecast therefore sits alongside
critical infrastructure rather than inside its safety perimeter. It
is ordinarily not a high-risk system under Annex~III No.~2, the FOSS
exception of
Art.~2(12) of the EU AI Act remains applicable, and the substantive
requirements discussed in Chapter~\ref{sec-ki-vo-9-en} through
Chapter~\ref{sec-ki-vo-15-en} do not bind \texttt{spotforecast2-safe}
by operation of law. Assessing the concrete use case nevertheless
remains the provider's task: where the software library is embedded in a system
that does satisfy the elements of Recital~55, such as an automatic
grid-protection or load-shedding scheme\footnote{Grid protection
refers to the protective technology of the transmission and
distribution grid: protection relays (distance, differential, and
overcurrent protection) that detect a fault such as a short circuit
within milliseconds and automatically disconnect the affected line or
asset from the grid via circuit breakers, before people are
endangered or equipment is damaged. Automatic load shedding is the
last line of defence against a grid collapse: when the grid frequency
falls below critical thresholds (staged from about 49.0~Hz in the
continental European synchronous area, under-frequency load
shedding), relays automatically disconnect consumer load from the
grid in stages to restore the balance between generation and
consumption and to prevent a blackout. This happens \emph{autonomously},
without a human decision at the moment of intervention. Both
automatic schemes satisfy the two elements of Recital~55: they
directly protect the physical integrity of the grid and the safety of
persons, but are not required for the functioning of the system
itself, since in normal operation the grid runs without any
deployment of the protection scheme. By definition, the scheme acts
only in the event of a fault. They would therefore be safety components within the
meaning of Annex~III No.~2 of the EU AI Act, whereas the day-ahead
load forecast is the mirror image: it protects nothing directly and
is precisely required for operation.}, the exception falls away and
the obligations attach to the provider of that system. Even without
such an embedding, a load forecast informs decisions taken by
operators of critical installations, and determinism,
reproducibility, and auditability are the properties on which an
audit of such a system would rest.
\texttt{spotforecast2-safe} therefore meets those requirements
deliberately and \emph{in advance}, irrespective of whether they are already
owed in the individual case. That is precisely the package's
compliance-by-design approach. Independently thereof,
Art.~25(4), first subparagraph, second sentence, of the EU AI Act explicitly does not
extend the obligation of a written value-chain agreement between
component supplier and provider of a high-risk system to tools,
services, processes, or components made available under a free or
open-source licence, provided they do not concern general-purpose AI
(GPAI) models. The library therefore owes its downstream provider no
contractually regulated cooperation, but supplies the
cooperation-enabling preconditions through open provenance,
documented interfaces, and reproducible releases (see
Chapter~\ref{sec-principles-en}). The GPAI-specific FOSS exemptions in
Art.~53(2) and Art.~54(6) of the EU AI Act are not applicable to a
classical regression library such as \texttt{spotforecast2-safe},
because \texttt{spotforecast2-safe} does not place any GPAI model in
the sense of Art.~3 No.~63 of the EU AI Act on the market.

\emph{CRA.} The CRA \citep{euCRA24}
complements this scheme via Art.~24 CRA, which provides for reduced
obligations for ``open-source software stewards'' (typically
not-for-profit or community-economic organisations that maintain
FOSS projects on a long-term basis) relative to the manufacturer
duties of Art.~13~ff.\ CRA (a documented cybersecurity policy,
coordinated vulnerability handling, cooperation with market
surveillance) and excludes the imposition of administrative fines
except for reporting duties.

\emph{NIS-2.} NIS-2 \citep{euNIS22} provides
no explicit FOSS exemption. Its supply-chain duty under
Art.~21(2)(d) NIS-2 binds the regulated KRITIS operator and not the
open-source maintainer. It does propagate de~facto into the
FOSS ecosystem through contractual requirements. The relevant legal
literature systematically analyses this entanglement
\citep{colo25a, vsch25a, kike24a}.

\emph{Trustworthiness.} A separate research line addresses \emph{how}
the trustworthiness of a high-risk AI system is to be assessed.
\citet{stet24a} propose a trustworthiness-assurance framework that
decomposes the EU AI Act requirements through a six-step iterative
process into verifiable properties and connects AI-system engineering
with the assurance argumentation of functional safety. The present
report is compatible with this framework: every process rule in
Chapter~\ref{sec-principles-en} is an argumentation step, and the
\emph{Mechanism} column of Table~\ref{tbl-compliance-en} provides the
substantiation.

\emph{Artificial intelligence and fundamental rights.} The
legal-scholarship background is set out by \citet{raue25a}, who examines
the interplay between the EU AI Act and the Charter of Fundamental
Rights of the European Union, anti-discrimination law, and data
protection law. This work is the authoritative reference when a
deployer has to translate the technical evidence documented here into
a Fundamental Rights Impact Assessment under Art.~27 of the
EU AI Act.

Table~\ref{tbl-normen-einleitung-en} summarises the legal acts,
standards, and reference models cited in the preceding paragraphs by
regulatory level.

{

\begin{longtable}{p{0.15\textwidth} p{0.21\textwidth} p{0.56\textwidth}}

\caption{\label{tbl-normen-einleitung-en}Systematic overview of the
cited legal acts, standards, and reference models. EU AI Act =
Regulation (EU) 2024/1689; CRA = Regulation (EU) 2024/2847 (Cyber
Resilience Act); NIS-2 = Directive (EU) 2022/2555; CER = Directive (EU)
2022/2557; CPS = cyber-physical system; ISMS = information-security
management system; IACS = industrial automation and control systems.}

\tabularnewline

\\
\hline
\textbf{Domain} & \textbf{Legal act / standard} & \textbf{Subject} \\
\hline
\endfirsthead
\multicolumn{3}{l}{\itshape (continued from \tablename~\ref{tbl-normen-einleitung-en})}\\
\hline
\textbf{Domain} & \textbf{Legal act / standard} & \textbf{Subject} \\
\hline
\endhead
\hline
\multicolumn{3}{r}{\itshape (continued on next page)}\\
\endfoot
\hline
\endlastfoot
EU regulation (AI) & EU AI Act & High-risk AI systems: data governance,
transparency, accuracy, robustness, and cybersecurity. \\
EU regulation (product safety) & Cyber Resilience Act (CRA) & Products
with digital elements: essential cybersecurity requirements (Annex~I
Part~I) and vulnerability handling throughout the support period of, as
a rule, at least five years (Art.~13(8), Annex~I Part~II). \\
EU regulation (network and information security) & NIS-2 Directive &
Risk-management duties of essential and important entities. \\
EU regulation (resilience) & CER Directive & Resilience of critical
entities (energy, transport, water, etc.). \\
EU regulation (civil liability) & Product Liability Directive (PLD,
Directive (EU) 2024/2853) & Strict liability for defective products;
software and AI systems explicitly captured for the first time;
implementation and applicability date 9~December~2026. German
transposition by the draft act on the modernisation of product-liability
law (BT-Drs.\ 21/4297). \\
AI ethics guidelines & Ethics Guidelines for Trustworthy AI \citep{hleg19a} &
Three-component framework (lawful, ethical, robust); seven requirements
that underlie the high-risk obligations EU-1 to EU-7 of the EU AI Act. \\
AI ethics guidelines & Assessment List for Trustworthy AI
\citep[ALTAI,][]{hleg20a} & Operationalisation of the seven requirements
as a self-assessment checklist. \\
AI standards (ISO/IEC) & Six ISO/IEC core standards on AI (TS 4213,
5338, 23894, 24027, 38507, 42001) & Operationalisation of the seven
high-risk requirements through technical and management-system
standards; detailed listing in Table~\ref{tbl-ki-vo-iso-en} and
Table~\ref{tbl-core-standards-en}. \\
Functional safety & IEC 61508-3 & Software for safety-related electrical,
electronic, and programmable systems; basis of the SIL classification. \\
Functional safety & ISO~26262-6 & Road-vehicle-specific adaptation of
IEC~61508 at the software level; basis of the ASIL classification. \\
Industrial cybersecurity & ISA/IEC~62443-1-1 & Terminology, concepts, and
models of the standards series. \\
Industrial cybersecurity & ISA/IEC~62443-2-1 & Cybersecurity-management
system of the asset owner (CPS counterpart of ISO/IEC~27001). \\
Industrial cybersecurity & ISA/IEC~62443-2-4 & Security requirements for
IACS service providers. \\
Industrial cybersecurity & ISA/IEC~62443-3-2 & Risk assessment and system
design for IACS. \\
Industrial cybersecurity & ISA/IEC~62443-3-3 & System-wide security
requirements (seven foundational requirements, four security levels). \\
Industrial cybersecurity & ISA/IEC~62443-4-1 & Secure product-development
lifecycle of the component manufacturer (directly applicable to
\texttt{spotforecast2-safe}). \\
Industrial cybersecurity & ISA/IEC~62443-4-2 & Technical security
requirements for components (software-application profile). \\
Reference model & ISA-95 / IEC~62264 & Functional hierarchy of industrial
automation (Purdue model, levels 0--4). \\
Management system & ISO/IEC~27001 & Information-security management
system (ISMS); used in the text as the counterpart of 62443-2-1. \\
\hline

\end{longtable}

}

The \texttt{spotforecast2-safe} package attempts to resolve the
inherent conflict between research and safety noted at the outset.
The starting point for the development of
\texttt{spotforecast2-safe} is \texttt{skforecast} \citep{scip24a}.
About one third of the safety-critical lines of code are a direct port
from the \texttt{skforecast} source code. Additional code implements
the four code-development rules described in
Chapter~\ref{sec-code-rules-en} (``no dead code'', ``deterministic
transformations'', ``fail-safe handling'', and ``minimal scope''), as
well as the four process rules described in
Chapter~\ref{sec-process-rules-en}. Because of these rules, the feature
scope of the software is reduced. To compensate for this deliberate
reduction, \texttt{spotforecast2-safe} is published as a subset
(``core'') of the broader open-source sister package \texttt{spotforecast2}
\citep{spotforecast2}. The \texttt{spotforecast2} package adds further
features: interactive visualisation based on \texttt{plotly} and
\texttt{matplotlib}, hyperparameter tuning via \texttt{optuna} and
\texttt{spotoptim}, and model-explanation tooling.

The remainder of this report\footnote{This report meets requirements
  similar to those of the software it presents: every code block is
  executed at the time of document generation, and the entire document
  including the bibliography is under version control.} describes the
subset implemented by
\texttt{spotforecast2-safe}. The report pursues three objectives.
First, it situates the package within the landscape of Python
time-series tools and explains which features have been deliberately
omitted from the safety-critical subset and why
(Chapter~\ref{sec-related-en}). Second, it documents the architecture
(Chapter~\ref{sec-architecture-en}), the formal preprocessing and
forecasting algorithms (Chapter~\ref{sec-preprocessing-en} through
Chapter~\ref{sec-testen-en}), and the safety-relevant design decisions
shipped in the most recent versions
(Chapter~\ref{sec-principles-en} through Chapter~\ref{sec-safety-en}).
Third, it maps the design decisions onto individual provisions of the
EU AI Act and of the functional-safety standards
(Chapter~\ref{sec-ki-vo-andere-normen-en}) and closes with a brief
example of electric load forecasting
(Chapter~\ref{sec-example-en}).

\section{Software packages for time-series forecasting}\label{sec-related-en}

The Python time-series-forecasting ecosystem has grown strongly over
the past decade and has also adopted implementations from the
statistical programming language R. Foundational is the
\texttt{statsmodels} package, which provides ARIMA, SARIMAX, and
exponential-smoothing models in line with the classical exposition by
\citet{box15a}. \texttt{scikit-learn} \citep{pedr11a} deliberately
abstains from time-series-specific functionality and delegates such
features to downstream libraries. The most relevant of these libraries
are introduced briefly below.

\begin{itemize}
\tightlist
\item
  \texttt{skforecast} \citep{scip24a, rodr25a} provides a \texttt{Forecaster}
  abstraction that wraps any scikit-learn-compatible regressor into a
  recursive, direct, or multi-series forecaster. It is the closest
  conceptual relative of the package documented here, because
  \texttt{spotforecast2-safe} builds on \texttt{skforecast}. From it we adopt
  the architecture of recursive forecasting, the rolling-origin
  splitter, and substantial parts of the modules \texttt{\_binner},
  \texttt{\_differentiator}, and the residuals infrastructure. We do not
  adopt the visualisation subsystem (which depends on
  \texttt{matplotlib} and \texttt{plotly}) or the hyperparameter tuning
  via \texttt{bayesian\_search\_forecaster} (which relies on
  \texttt{optuna} and produces non-deterministic optimisation traces).
  These tools are contained in the \texttt{spotforecast2} package.
\item
  The unified \texttt{Forecaster}/\texttt{BaseEstimator} API of
  \texttt{sktime} \citep{loni19a}, which provides a wide range of models
  and validation schemes, is the most extensive of the libraries
  considered here. However, the API is not fully compatible with that of
  \texttt{scikit-learn}, so that integrating regressors from that
  ecosystem is not straightforward. Moreover, many models in
  \texttt{sktime} rely on \texttt{statsmodels} implementations that are
  not optimised for the requirements of functional safety (e.g., no
  fail-safe handling of invalid inputs). The very breadth of the feature
  set, moreover, enlarges the dependency surface and with it the audit
  surface.
\item
  The Nixtla ecosystem, which targets improvements in runtime
  throughput, implements three complementary libraries:
  \texttt{statsforecast} \citep{garz22a} provides vectorised
  implementations of classical models, trading model flexibility for
  aggressive runtime optimisation via Numba \citep{lam15a}.
  \texttt{mlforecast} extends the same API to gradient-boosting methods.
  \texttt{neuralforecast} offers deep-learning architectures through a
  unified interface. The deep-learning backend immediately violates the
  rule of minimal scope.
\item
  \texttt{Darts} \citep{herz22a} introduces a unified API for various
  time-series models. Its installation footprint is dominated by
  optional dependencies whose behaviour influences the availability of
  model classes at import time. The same applies to \texttt{tsai},
  \texttt{Merlion}, and \texttt{Kats}.
\end{itemize}

None of the libraries listed is deficient from a software-engineering
standpoint. On the contrary, they constitute outstanding research-grade
software, but they are good examples of the inherent conflict between
research and safety noted above. \texttt{spotforecast2-safe} attempts
to resolve this conflict by considering safety \emph{by design} from
the outset. The deployer inherits the required safety demonstrations
from the software and does not have to implement them after the fact.

\section{\texorpdfstring{The \texttt{spotforecast2-safe}
package}{The spotforecast2-safe package}}\label{sec-package-en}

\subsection{Architecture}\label{sec-architecture-en}

The \texttt{spotforecast2-safe} package is organised into three layers.
The lowest layer (\texttt{forecaster/}, \texttt{preprocessing/},
\texttt{splitter/}, \texttt{backtesting/}, \texttt{data/}) contains
the estimator wrappers, transformers, splitters, and the backtesting
driver that derive most directly from \texttt{skforecast}. A middle
layer (\texttt{manager/}) bundles an estimator together with an
\texttt{ExogBuilder}, a configurator, a logger, and a persistence
layer into a deployable unit. The top layer (\texttt{multitask/},
\texttt{processing/}, and \texttt{tasks/}) provides end-to-end
pipelines: \texttt{multitask/} orchestrates multi-target forecasts
through a task hierarchy whose dispatcher \texttt{MultiTask} selects
between lazy, default, predict-only, and clean-up runs,
\texttt{processing/} supplies the accompanying prediction, blending,
and scoring functions, and \texttt{tasks/} composes the lower layers
into console scripts. The module \texttt{submission\_cli.py} adds
check-and-abort helpers for automated submission workflows.

\dirtree{%
.1 src/spotforecast2\_safe/.
.2 forecaster/.
.3 recursive/, wrappers/.
.2 preprocessing/.
.3 repeating\_basis\_function.py.
.3 \_binner.py, \_differentiator.py.
.3 linearly\_interpolate\_ts.py.
.3 exog\_builder.py, imputation.py, outlier.py.
.2 splitter/.
.3 split\_ts\_cv.py, split\_one\_step.py.
.2 backtesting/.
.3 validation.py.
.2 manager/.
.3 logger.py, demo\_metrics.py, features.py.
.3 persistence.py, trainer.py, predictor.py.
.2 multitask/.
.3 base.py, multi.py, runner.py.
.3 strategies.py, guards.py.
.2 configurator/, data/, calendar/, stats/.
.2 processing/, tasks/.
.2 datasets/, downloader/, weather/.
.2 security/, utils/.
.2 exceptions.py, submission\_cli.py.
}

The public interface is fixed by \texttt{\_\_all\_\_} at the top
package level and comprises ten symbols: \texttt{Period},
\texttt{RepeatingBasisFunction}, \texttt{ExogBuilder},
\texttt{LinearlyInterpolateTS}, the four
\texttt{ForecasterRecursive*} model classes
(\texttt{ForecasterRecursiveModel}, \texttt{ForecasterRecursiveLGBM},
\texttt{ForecasterRecursiveXGB}, \texttt{ForecasterRecursiveCatBoost}),
\texttt{ConfigEntsoe}, and the alias \texttt{Config}. The package
version is additionally available as the module attribute
\texttt{\_\_version\_\_}. Every API-breaking change to these symbols is
bound by convention to a Conventional-Commits subject \texttt{feat!:}
and to a major version bump. Private modules\footnote{Anything with a
  leading underscore or in a submodule that is not re-exported at the
  top level.} are explicitly unstable and excluded from
this guarantee.

Adding a new estimator wrapper therefore happens in the directory
\texttt{forecaster/wrappers/}: the wrapper inherits from
\texttt{ForecasterRecursiveModel}, which itself orchestrates
\texttt{fit}, \texttt{predict}, \texttt{backtest}, and saving and
loading. The wrapper is exclusively responsible for
declaring the backend regressor and any backend-specific serialisation
hooks. The exogenous-feature layer, the rolling-origin validation, and
the audit logging are inherited.

The sister package \texttt{spotforecast2} \citep{spotforecast2}
(version 10.9.0) declares \texttt{spotforecast2-safe} as a dependency
and contributes precisely the extensions that are deliberately
excluded from the safe subset: the auto-tuning tasks
(\texttt{OptunaTask}, \texttt{SpotOptimTask}) together with their
search spaces, full variants of the four forecaster classes with
Optuna tuning and SHAP feature importance, plotting functions, and
additional statistics and model-selection tools. The complete
forecasting core, including the retrieval of ENTSO-E data (European
Network of Transmission System Operators for Electricity) and weather
data (\texttt{downloader/}, \texttt{weather/}) on which the audit
narrative of Chapter~\ref{sec-safety-en} builds, resides in
\texttt{spotforecast2-safe} alone. The dependency direction is one-way:
the \texttt{spotforecast2-safe} core does not import from
\texttt{spotforecast2} and is independently installable and testable.
This split makes it possible to release the safe subset on its own
cadence and to certify it independently, while \texttt{spotforecast2}
remains a functional superset for the broader, non-regulated use
case.

\subsection{Preprocessing}\label{sec-preprocessing-en}

The \texttt{preprocessing} sub-package contains four kinds of
transformer:

\begin{enumerate}
\def\labelenumi{\arabic{enumi}.}
\tightlist
\item
  Cyclical encoders map periodic calendar features (hour of day, day
  of week, day of year) into a smooth representation.
\item
  Interpolators reconstruct short gaps under an explicit contract.
\item
  A quantile binner discretises a continuous feature into
  equal-frequency bins.
\item
  A differentiator applies a finite-difference operator.
\end{enumerate}

All four transformers inherit from
\texttt{sklearn.base.TransformerMixin} and are therefore composable
within an \texttt{sklearn} \texttt{Pipeline}.

\subsection{Forecasting}\label{sec-forecasting-en}

The forecasting layer implements the recursive multi-step strategy: a
single scalar one-step-ahead model is trained and then iterated to
produce forecast horizons of arbitrary length \citep{bont13a, scip24a}.
Prediction intervals are computed by default by bootstrap-resampling
the in-sample residuals. This is the same
procedure as in \texttt{skforecast} \citep{rodr25a}. Given a seed, it
is deterministic.

Any scikit-learn-compatible estimator can be deployed by direct
instantiation of \texttt{ForecasterRecursiveModel}. The backend
regressors shipped with the package are LightGBM \citep{ke17a},
XGBoost \citep{chen16a}, and CatBoost \citep{prok18a}, implemented as
\texttt{ForecasterRecursiveLGBM}, \texttt{ForecasterRecursiveXGB},
and \texttt{ForecasterRecursiveCatBoost}.
Gradient-boosting methods of this kind achieve good results on
M5-style datasets \citep{makr22a}. The
decision to restrict the package's backends to
LightGBM, XGBoost, CatBoost, and scikit-learn regressors (and to ship
the three named wrappers as the default models) follows directly from
this empirical evidence.

\subsection{Splitters for generating training and test
data}\label{sec-testen-en}

A naive application of \texttt{sklearn.model\_selection.KFold} to a
time-indexed series leaks the future into the training set.
Time-series-aware splitters remedy this by forcing the training
segment to precede the test segment in time. The package provides two
such splitters as well as a backtesting driver.

\begin{figure}[H]

\centering{

\pandocbounded{\includegraphics[keepaspectratio]{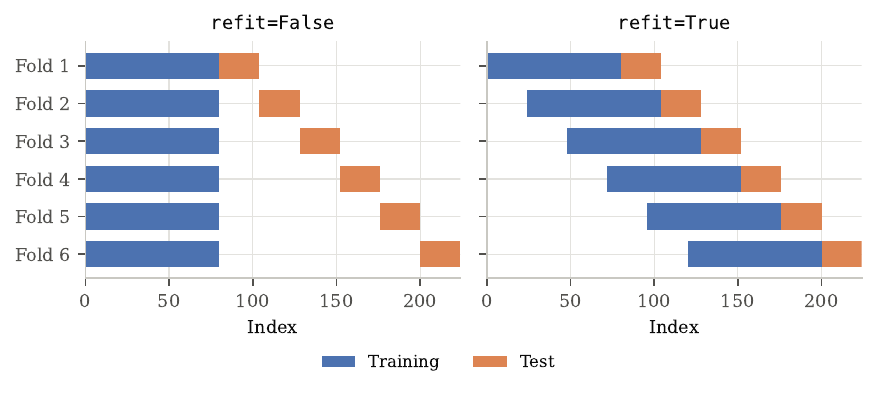}}

}

\caption{\label{fig-folds-en}Rolling-origin backtesting folds of
\texttt{TimeSeriesFold} (initial training 80, step size 24, six
folds). Training segments are shown in blue, held-out test segments
in orange. On the left (\texttt{refit=False}) the forecaster is
trained once on the initial window, after which only the test window
advances by \texttt{steps} observations. On the right
(\texttt{refit=True}) the model is re-trained at every origin on a
sliding window of constant length.}

\end{figure}

\texttt{TimeSeriesFold} realises the rolling-origin protocol. The
splitter decomposes the series into \emph{folds}, temporally ordered
train-test pairs, and yields the test windows.
Figure~\ref{fig-folds-en} contrasts two possible settings.

\texttt{OneStepAheadFold} is a slim specialisation with
\texttt{steps=1}, typically used for evaluating the conditional-mean
regressor itself (rather than the multi-step point forecast).
\texttt{backtesting\_forecaster} composes one of the two splitters
with a \texttt{ForecasterRecursive*} together with a list of metrics
(\texttt{mean\_absolute\_error}, \texttt{mean\_squared\_error},
\texttt{mean\_absolute\_scaled\_error} \citep{hynd06a}) and returns a
pandas DataFrame of fold-wise metric values together with the
concatenated forecast vector. Parallelised refit execution is offered
via \texttt{joblib} and obliges the caller to fix \texttt{n\_jobs}.
This is the only point in the package where parallelism is exposed,
and the fixing is mandatory because OpenMP scheduling is a known
source of cross-platform non-determinism.

\section{Design principles}\label{sec-principles-en}

The design principles fall into two parallel categories.

\begin{itemize}
\tightlist
\item
  The \emph{code-development rules} restrict which constituents the
  software source code may contain. These are enforced at commit time
  by a test or by a linter\footnote{A linter is a static-analysis tool
    that checks the source code, without executing it, for style
    violations, suspicious patterns, and breaches of programming
    conventions.}.
\item
  The \emph{process rules} restrict how the package is developed,
  shipped, and operated. These are met by \texttt{MODEL\_CARD.md}, the
  documented release procedure, and a versioned logging schema.
\end{itemize}

Both categories are required: the code-development rules alone do not
suffice to discharge the regulatory duties of traceability, threat
modelling, data governance, supply-chain integrity, and operational
observability.

\subsection{Code-development rules}\label{sec-code-rules-en}

The four non-negotiable code rules, which at the same time satisfy
the secure-software development practices required by IEC 62443-4-1
\citep{iec62443_4_1}, are:

\begin{tcolorbox}[enhanced jigsaw, arc=.35mm, bottomrule=.15mm, bottomtitle=1mm, breakable, colback=white, colbacktitle=quarto-callout-note-color!10!white, colframe=quarto-callout-note-color-frame, coltitle=black, left=2mm, leftrule=.75mm, opacityback=0, opacitybacktitle=0.6, rightrule=.15mm, title=\textcolor{quarto-callout-note-color}{\faInfo}\hspace{0.5em}{Code rules}, titlerule=0mm, toprule=.15mm, toptitle=1mm]

\begin{itemize}
\tightlist
\item
  CR-1. \emph{No dead code.} ``Dead code'' is code whose removal does
  not affect the program's behaviour. No function, class, or branch
  ships without a test or an executable docstring example reaching
  it.
\item
  CR-2. \emph{Deterministic transformations.} The same input generates
  the same bit-level output. This is implemented by the following
  measures: random-number generators are seeded, dictionary iteration
  order is not relied upon, and \texttt{joblib} parallelism is pinned
  to a fixed \texttt{n\_jobs} argument.
\item
  CR-3. \emph{Fail-safe handling.} Invalid inputs raise an explicit
  exception rather than being silently imputed or coerced. For example,
  NaN values or missing indices raise explicit
  \texttt{ValueError} or \texttt{TypeError} exceptions. Silent
  imputation is available only behind a typed
  \texttt{Literal{[}"raise",\ "ffill\_bfill",\ "passthrough"{]}}
  switch that must be set explicitly.
\item
  CR-4. \emph{Minimal CVE attack surface (CVE: Common Vulnerabilities
  and Exposures).} A short, versioned blocklist (``negative list'') of
  forbidden dependencies (\texttt{plotly}, \texttt{matplotlib},
  \texttt{spotoptim}, \texttt{optuna}, \texttt{torch},
  \texttt{tensorflow}) is implemented by a test that inspects the
  lock files with \texttt{grep}.
\end{itemize}

\end{tcolorbox}

\subsection{Process rules}\label{sec-process-rules-en}

Four further rules govern how the software package is developed,
shipped, and operated. Each is inspected at audit or release time and fulfilled
by an artefact (\texttt{MODEL\_CARD.md}, the \texttt{Makefile}
targets documented in \texttt{RELEASING.md}, or a versioned logging
schema). This split corresponds to the
\emph{trustworthiness-assurance} process proposed by
\citet{stet24a}, in which AI-specific regulatory requirements are
refined into verifiable properties whose evidence is collected along
the entire development lifecycle.

Carrying out the process rules can be automated by
continuous-integration (CI) jobs. Established tooling is available
on GitHub or GitLab, for example, to implement most of this work.
Each rule below therefore names the artefact it demands, the
mechanism by which this package produces it, and an established way
of checking the rule by machine.

The package \texttt{spotforecast2-safe} itself deliberately does not
run these checks on a CI platform but uses the local development
environment by means of Make \citep{free23a}. The entire
verification chain behind the process rules runs
through documented Make targets (\texttt{make verify},
\texttt{make release}, \texttt{make ship}) entirely locally, and is
therefore reproducible by any auditor
without platform access. GitLab serves solely to host source code and
releases, not for security or conformity checking. The sister package
\texttt{spotforecast2} follows the same model. The price of this
independence is stated openly: without a CI identity there is no
keyless signing and no independent execution of the test suite on a
second platform (see Chapter~\ref{sec-ki-vo-13-en}).

\begin{tcolorbox}[enhanced jigsaw, arc=.35mm, bottomrule=.15mm, bottomtitle=1mm, breakable, colback=white, colbacktitle=quarto-callout-note-color!10!white, colframe=quarto-callout-note-color-frame, coltitle=black, left=2mm, leftrule=.75mm, opacityback=0, opacitybacktitle=0.6, rightrule=.15mm, title=\textcolor{quarto-callout-note-color}{\faInfo}\hspace{0.5em}{Process rules}, titlerule=0mm, toprule=.15mm, toptitle=1mm]

\begin{itemize}
\tightlist
\item
  PR-1. \emph{Traceability.} Every public symbol carries a docstring
  that appears in the generated API reference, and the execution of
  the make pipeline (\texttt{make guard}) fails if a public symbol is
  undocumented or its generated stub is stale. Linking a symbol
  to a documented requirement in \texttt{MODEL\_CARD.md} (Sections 2 to
  5) or to a specific clause of Table~\ref{tbl-compliance-en} happens at
  audit time. This supports IEC 61508-3
  Section 7.2 (specification of software-safety requirements)
  \citep{iec61508}, ISO 26262-6 Section 6.4 (specification of software
  safety requirements, ASIL-dependent per Section 4.4)
  \citep{iso26262}, and Art.~9 of the
  EU AI Act (risk management across the entire lifecycle)
  \citep{euro24a}. Orphan symbols, that is, symbols without a
  requirement reference and without a \texttt{MODEL\_CARD} entry, are
  flagged in code review.
\item
  PR-2. \emph{Documented threat model.} The module docstring of every
  network-facing module maintains a STRIDE table\footnote{STRIDE
    is a threat-classification scheme introduced by Microsoft. The
    acronym stands for \emph{spoofing} (impersonation),
    \emph{tampering} (manipulation), \emph{repudiation}
    (deniability), \emph{information disclosure},
    \emph{denial of service}, and \emph{elevation of privilege}.
    A STRIDE table enumerates, per module and data flow, which of
    these six categories applies, which countermeasure is in place,
    and in which source file that countermeasure is realised.},
  currently in \texttt{downloader/entsoe.py},
  \texttt{downloader/resilience.py}, and
  \texttt{weather/\_\_init\_\_.py}. Every change to the attack surface
  requires an updated threat entry. The rule
  is stated in \texttt{CONTRIBUTING.md} and checked at review time.
  The established way to enforce it by machine is a path-filtered
  CI job that refuses a merge request touching a
  network-facing module without an updated threat entry. This satisfies
  IEC 62443-4-1 SR-1 and SR-2 (threat-model-driven security
  requirements) \citep{iec62443_4_1}.
\item
  PR-3. \emph{Supply-chain integrity.} The rule demands two artefacts:
  a software bill of materials (SBOM) and an attestation binding it to
  the artefact actually shipped. The SBOM is produced by
  \texttt{make sbom} as a CycloneDX document. The attestation is
  currently not provided, because releases carry no cryptographic
  signature (see Chapter~\ref{sec-ki-vo-13-en}).
\item
  PR-4. \emph{Structured audit log.} Every operational action emits a
  log record according to a fixed JSON schema under
  \texttt{\textasciitilde{}/spotforecast2\_safe\_models/logs/}. The
  schema is versioned: \texttt{audit\_log\_schema.json} carries its
  version in the \texttt{\$id}, \texttt{SCHEMA\_VERSION} is a pinned
  constant, and \texttt{tests/manager/test\_logger\_schema.py} asserts
  that the two agree. Since release 26.0.0 the records are
  additionally hash-chained: each record carries in the required field
  \texttt{prev\_hash} the SHA-256 value of the previously written
  line, and the function \texttt{verify\_audit\_log()} checks the
  chain offline and reports the first break with its line number.
  Since release 27.0.0 its verification report additionally states
  whether the chain head has been externally anchored (see
  Chapter~\ref{sec-ki-vo-12-en}).
\end{itemize}

\end{tcolorbox}

A further, separately maintained rule concerns data governance: the
package \texttt{spotforecast2-safe} itself ships no training data.
Every forecaster object records its provenance (creation and training
timestamps, library and Python versions) in its persisted state. The
package computes no content hash of the training data. This supports
the provenance requirement of Art.~10 of the EU AI Act (data and data
governance) \citep{euro24a} as far as a library can reach without
insight into the deployer's actual training data. Documenting the
data sources and retrieval times, and assessing representativeness,
bias, and fitness for the intended purpose, remain the responsibility
of the deployer.

A further principle that does not strictly belong to the four rules
but follows from them is that every source file carries an SPDX header
(Software Package Data Exchange) in the format recommended by
REUSE v3.0 \citep{reuse30}. Ported
\texttt{skforecast} files carry both the original BSD-3-Clause notice
and the added AGPL-3.0-or-later statement. This enables the licence-law
resolution as a dual-licence derivation, and no impression of
relicensing third-party software arises.

\section{Safety-critical implementation}\label{sec-safety-en}

In addition to the four code rules and four process rules listed in
Chapter~\ref{sec-principles-en}, whose evidence is mapped row by row
against the regulatory provisions in Table~\ref{tbl-compliance-en},
\texttt{spotforecast2-safe} demonstrates several
community-established quality attributes. These are published via
machine-readable status badges in the package's \texttt{README.md} and
point partly to independent external assessment bodies (PyPI,
pepy.tech, bestpractices.dev) and partly to the authoritative artefacts
in the repository. Table~\ref{tbl-badges-en} gives a complete overview.

\begin{longtable}[]{@{}
  >{\raggedright\arraybackslash}p{(\linewidth - 6\tabcolsep) * \real{0.1304}}
  >{\raggedright\arraybackslash}p{(\linewidth - 6\tabcolsep) * \real{0.2609}}
  >{\raggedright\arraybackslash}p{(\linewidth - 6\tabcolsep) * \real{0.3913}}
  >{\raggedright\arraybackslash}p{(\linewidth - 6\tabcolsep) * \real{0.2174}}@{}}
\caption{Complete list of status badges in the
\protect\texttt{README.md} of \protect\texttt{spotforecast2-safe},
grouped by the categories used in the repository. The groups
\emph{Version \& License} and \emph{Downloads} serve only package
discoverability and tracking; the groups \emph{Quality},
\emph{Scores}, and \emph{Status} are conformity or
quality signals. OpenSSF = Open Source
Security Foundation; SPDX =
Software Package Data Exchange. CR-X refers to code rule X in
Chapter~\ref{sec-code-rules-en}.}\label{tbl-badges-en}\tabularnewline
\toprule\noalign{}
\begin{minipage}[b]{\linewidth}\raggedright
\textbf{Group}
\end{minipage} & \begin{minipage}[b]{\linewidth}\raggedright
\textbf{Badge}
\end{minipage} & \begin{minipage}[b]{\linewidth}\raggedright
\textbf{Meaning}
\end{minipage} & \begin{minipage}[b]{\linewidth}\raggedright
\textbf{Source / reference}
\end{minipage} \\
\midrule\noalign{}
\endfirsthead
\toprule\noalign{}
\begin{minipage}[b]{\linewidth}\raggedright
\textbf{Group}
\end{minipage} & \begin{minipage}[b]{\linewidth}\raggedright
\textbf{Badge}
\end{minipage} & \begin{minipage}[b]{\linewidth}\raggedright
\textbf{Meaning}
\end{minipage} & \begin{minipage}[b]{\linewidth}\raggedright
\textbf{Source / reference}
\end{minipage} \\
\midrule\noalign{}
\endhead
\bottomrule\noalign{}
\endlastfoot
Version \& License & Python Version & Minimum Python version for the
package installation & \texttt{pyproject.toml}; python.org \\
Version \& License & PyPI Version & Current package version on the Python
Package Index & PyPI \\
Version \& License & License & Licence statement (AGPL-3.0-or-later) &
\texttt{LICENSE}; SPDX \\
Downloads & PyPI Downloads & Monthly download counts & PyPI \\
Downloads & Total Downloads & Cumulative download statistics &
pepy.tech \\
Quality & EU AI Act & Regulatory deployability under the EU AI Act &
\texttt{MODEL\_CARD.md} \\
Quality & Dependencies & Minimised dependency surface (CR-4) &
\texttt{pyproject.toml} \\
Quality & Audit & Whitebox audit status & \texttt{MODEL\_CARD.md} \\
Quality & Reliability & Fail-safe behaviour (CR-3) &
\texttt{MODEL\_CARD.md} \\
Quality & Security & Documented security policy &
security policy of the package documentation \\
Quality & Documentation & Published API reference and documentation &
documentation site of the package \\
Scores & OpenSSF Best Practices & Best-practices maturity level
(\emph{passing} / \emph{silver} / \emph{gold}); project ID 11932 &
bestpractices.dev \\
Status & Maintenance & Active project maintenance & project repository \\
Status & Code style: black & Deterministic source-code style formatter
(\texttt{black}) & github.com/psf/black \\
\end{longtable}

\subsection{Quality}\label{quality-en}

\subsubsection{Fail-safe NaN handling}\label{sec-fail-safe-nan-en}

A tabular measurement may be missing, e.g., because a weather station
reports no data for an hour, because a power meter has been restarted,
or because an import error left a cell empty. In Python such a gap is
marked by the special value \texttt{NaN} (``Not a Number''),
analogous to a blank field on a questionnaire that has not yet been
answered. What matters for downstream processing is how the next-stage
software reacts to such a field: should it estimate the empty value
and continue, or should it stop processing and request a decision from
the deployer?

The Python time-series libraries established in this space mostly
follow the pattern of ``silently keep going''. For exploratory work this
stance is convenient, because it permits rapid prototyping with messy
data. For deployment in safety-critical environments, e.g.,
electric load forecasting for a grid operator, it is exactly the case
that Art.~15 of the EU AI Act, IEC 61508, and ISO 26262 seek to avoid:
an operator may dispatch on the basis of an apparently complete
forecast without noticing that part of the input data was missing and
was substituted by the package without enquiry.

A concrete example makes the difference clear:

\begin{example}[Fail-safe
vs.~silent]\protect\hypertarget{exm-fail-safe-nan-en}{}\label{exm-fail-safe-nan-en}

Suppose a grid
dispatcher is to forecast hourly load for the coming day, and the
weather API returns no temperature value for a single hour, say
03:00. Under the silent approach, the forecasting package would
interpolate the value (linearly) between 02:00 and 04:00, return the
load forecast with an apparently normal confidence interval, and the
dispatcher would schedule the 03:00 hour as usual. The gap in the
measurements would be visible only in the preprocessing log. Moreover,
someone must take the trouble to read these log files. Under the
fail-safe approach, by contrast, the gap becomes a visible event: the weather
adapter rejects the incomplete weather row with an explicit error
message, the CI pipeline raises an alarm, and the
deployer must take a documented decision: switch to a fallback model,
flag the 03:00 hour as invalid, or close the gap deliberately and
record that decision in the audit log.

\end{example}

\texttt{spotforecast2-safe} inverts the default of the libraries
listed above. The abort is the desired outcome, not seen as a fault:
the error becomes visible, reaches the CI pipeline, and forces the
deployer to take an explicit decision. The central point is that the
deployer cannot, in an audit, claim to have overlooked the data gap. The mechanism therefore
addresses Art.~15 of the EU AI Act (accuracy and robustness)
\citep{euro24a}, the fail-safe principle of IEC 61508-3
\citep{iec61508}, and the design principles for software unit
implementation of ISO 26262-6 Section 8.4 \citep{iso26262} directly.
How a gap is treated after that explicit decision is shown by
Step 3 in Chapter~\ref{sec-example-en}.

\subsection{Testing}\label{testing-en}

\subsubsection{Verification pipeline}\label{continuous-integration-en}

The package's verification pipeline is documented in
\texttt{RELEASING.md} and bundled in the \texttt{Makefile} target
\texttt{make\ verify}, which runs four gates in sequence: \texttt{guard}
(API documentation coverage, freshness of the generated stubs, REUSE
lint, and the blocklist check), \texttt{lint} (deterministic formatting
and static analysis), \texttt{test} (the full test suite with an
enforced coverage floor), and \texttt{doc} (Quarto rendering of the
documentation including the executable examples). Every release
requires a complete \texttt{verify} run, and \texttt{make\ release}
refuses to run on a dirty working tree. Continuous integration is the
established way to additionally bind the same pipeline to every git
push.

\subsubsection{Code coverage}\label{code-coverage-en}

Line coverage is measured on every run of \texttt{make\ test} and
enforced against the floor \texttt{fail\_under\ =\ 68} fixed in
\texttt{pyproject.toml}, below which the test run fails. The test suite
collects more than 3000 tests and currently measures 75.99\,\% line
coverage.
\texttt{CONTRIBUTING.md} names 80\,\% line coverage as the aim for new
code.

\subsubsection{SPDX and REUSE conformance}\label{sec-spdx-reuse-en}

\emph{SPDX} is a standardised,
machine-readable format for licence and copyright metadata in source
files. The specification was developed by the Linux Foundation and
ratified as the standard ISO/IEC 5962:2021 \citep{iso5962}.
SPDX header lines such as \texttt{SPDX-FileCopyrightText} and
\texttt{SPDX-License-Identifier} allow an SBOM toolchain to
reconstruct a package's licence situation from
the sources without manual inspection. The \emph{REUSE Specification}
of the Free Software Foundation Europe, used here in version 3.0
\citep{reuse30}, operationalises SPDX into three verifiable
requirements:

\begin{itemize}
\tightlist
\item
  \begin{enumerate}
  \def\labelenumi{(\roman{enumi})}
  \tightlist
  \item
    every file carries an \texttt{SPDX-FileCopyrightText} and an
    \texttt{SPDX-License-Identifier} entry in the header,
  \end{enumerate}
\item
  \begin{enumerate}
  \def\labelenumi{(\roman{enumi})}
  \setcounter{enumi}{1}
  \tightlist
  \item
    the full texts of all licences used in the project are placed in
    the directory \texttt{LICENSES/} and
  \end{enumerate}
\item
  \begin{enumerate}
  \def\labelenumi{(\roman{enumi})}
  \setcounter{enumi}{2}
  \tightlist
  \item
    compliance is verified automatically by the tool
    \texttt{reuse\ lint}.
  \end{enumerate}
\end{itemize}

The licence traceability thus guaranteed addresses the documentation
duties for software components and third-party tools under Art.~11 in
conjunction with Annex~IV No.~1 of the EU AI Act \citep{euro24a}
and supports the SM (security management) practice of IEC 62443-4-1
(inventory of third-party components) \citep{iec62443_4_1}. The REUSE
lint is also part of the \texttt{guard} gate of \texttt{make\ verify},
so licence conformance is verified again on every verification run and
before every release.

\subsection{Scores}\label{scores-en}

\subsubsection{OpenSSF Best Practices Badge and Scorecard}\label{openssf-best-practices-badge-en}

The Open Source Security Foundation (OpenSSF)\footnote{\url{https://openssf.org}}
maintains two independent programmes for assessing the security
maturity of open-source projects.

\begin{itemize}
\tightlist
\item
  The \emph{OpenSSF Best Practices Badge}
  programme\footnote{\url{https://www.bestpractices.dev}}
  inspects project-side submitted evidence on documentation, source
  accessibility, security reporting channels, dependency management,
  and test and build automation against the maturity levels
  \emph{passing}, \emph{silver}, and \emph{gold}.
  \texttt{spotforecast2-safe} is registered under project ID 11932 and
  demonstrates the \emph{passing} maturity level.
\item
  The \emph{OpenSSF Scorecard}\footnote{\url{https://scorecard.dev}}
  complements this self-attestation programme with an automated
  assessment of the project supply chain against a fixed
  list of check points, including branch protection, token scoping,
  automated dependency updates, static security analysis, and signed
  releases. No Scorecard assessment currently exists for
  \texttt{spotforecast2-safe}. The evidence for supply-chain practice
  rests instead on the Best Practices badge and the SDL mapping.
\end{itemize}

\subsection{Status}\label{status-en}

The badges of the \emph{Status} group (maintenance activity and a
fixed source-code formatter,
\texttt{black}\footnote{\url{https://github.com/psf/black}}) do
not constitute a conformance demonstration in the sense of the
EU AI Act or IEC 62443. They do make visible that the
package is actively maintained and that its source code is subject
to a deterministic style formatter, which together with
\texttt{ruff}, \texttt{isort}, and \texttt{mypy} is already anchored
in Chapter~\ref{sec-principles-en} as part of the code-development
rules.

\subsection{Quarantining of corrupted caches}\label{quarantine-en}

The weather-data client stores API responses as Parquet files in a
user-configurable directory. The cache reader catches
\texttt{(OSError,\ ValueError)} errors, writes a message at
\texttt{WARNING} level, and renames the corrupted file to
\texttt{\textless{}cache\textgreater{}.corrupt-\textless{}epoch\textgreater{}}
so that an operator can recover it forensically. A missing cache file
remains silent and returns \texttt{None}.

\subsection{CPE identifier for SBOM integration}\label{sec-cpe-sbom-en}

An SBOM is a machine-readable list
of all components of a software package\footnote{Comparable to the
  ingredients list on a food package.}. It enumerates which third-party
libraries, in which versions, and under which licences are contained
in the shipped software. On this basis a deployer can determine, without source-code
analysis, which components are present in a given installation.
This is the prerequisite for any subsequent vulnerability assessment.

The CPE is the standardised
identifier under which a single software product appears in such a
bill of materials\footnote{Comparable to a barcode.}. The CPE 2.3
specification of the U.S. National Institute of Standards and
Technology \citep{nistir7695} prescribes a fixed format:
\texttt{cpe:2.3:a:\textless{}vendor\textgreater{}:\textless{}product\textgreater{}:\textless{}version\textgreater{}:...}.
A typical identifier for the present package is therefore:\newline
\texttt{cpe:2.3:a:sequential\_parameter\_optimization:spotforecast2\_safe:*:*:*:*:*:*:*:*}.

With a stable CPE identifier, a vulnerability scanner can determine
automatically that a particular version of a
product is affected as soon as a CVE is reported under the same CPE,
in particular through the catalogue of Known Exploited
Vulnerabilities (KEV) maintained by the U.S.\ Cybersecurity and
Infrastructure Security Agency (CISA) \citep{cisaKEV}.

The distinguishing feature of \texttt{spotforecast2-safe} is that the
CPE identifier is not guessed by downstream tools but is fixed
within the package itself. Most Python packages rely on the heuristics of scanners such as
\texttt{pip-audit} or \texttt{safety} and declare no CPE. By contrast, the file
\texttt{src/spotforecast2\_safe/utils/cpe.py} contains the canonical
CPE string, and \texttt{tests/test\_cpe.py} verifies it in a
round-trip (serialisation followed by deserialisation yields an
identical object). This elevates the identifier to a first-class API contract:
a substantial software change requires a Conventional-Commits subject
\texttt{feat!:}, which forces a major version bump and
appears as an entry in the \texttt{CHANGELOG.md}. A deployer can
therefore rely on the CPE listed in their SBOM remaining stable
across the entire minor-version line.

From a regulatory perspective, this approach matches the three duties
that the fixed CPE directly serves:

\begin{itemize}
\tightlist
\item
  Annex~I Part~II No.~1 CRA \citep{euCRA24} requires the manufacturer
  of a product with digital elements to produce a software bill of
  materials.
\item
  Art.~72 of the EU AI Act \citep{euro24a} obliges providers of
  high-risk AI systems to perform post-market monitoring, which
  presupposes a reliable attribution of CVE reports to the own
  product.
\item
  The process area \emph{SM} (Security Management) of IEC 62443-4-1
  \citep{iec62443_4_1} requires an inventory of third-party
  components, which in this mechanism is the SBOM.
\end{itemize}

\section{EU AI Act requirements and applicable standards}\label{sec-ki-vo-andere-normen-en}

The analysis that follows describes the requirements the EU AI Act
imposes on high-risk systems and their providers. As set out in
Chapter~\ref{sec-introduction-en}, those obligations do not bind
\texttt{spotforecast2-safe} by operation of law in the simple
forecasting case. The package meets the requirements in advance and
by design decision. The addressee of the
obligations is always the provider of the system into which the library
is embedded, if that system satisfies the elements of a high-risk
system. The sections that follow are to be read in that sense.

\begin{figure}

\centering{

\includegraphics[width=0.9\linewidth,height=\textheight,keepaspectratio,alt={Venn diagram of two sets of operationalisation-essential and suitability-essential AI standards. The intersection contains six core standards: ISO/IEC 4213, ISO/IEC 5338, ISO/IEC 23894, ISO/IEC 24027, ISO/IEC 38507, and ISO/IEC 42001.}]{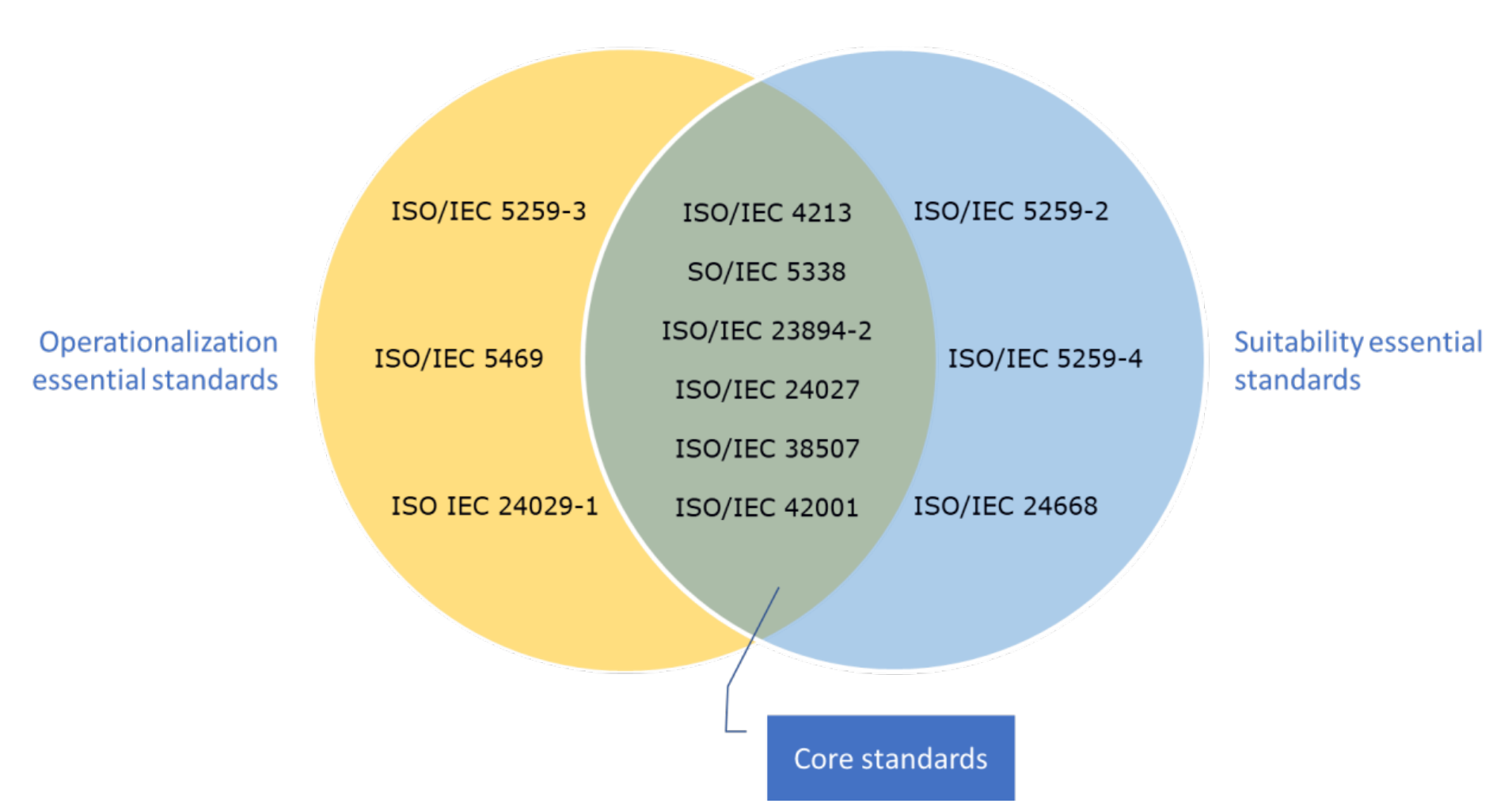}

}

\caption{\label{fig-core-standards-en}Relationship between the groups
of operationalisation-essential and suitability-essential AI standards
together with the central core set. The six core standards form the
intersection of the two essential groups (nine standards each).
Source: \citet{publ21a}, Figure~18, p.~52; © European Union 2021,
licensed under Creative Commons Attribution 4.0 International (CC BY
4.0); original figure reproduced with attribution under the licence
terms.}

\end{figure}

Building on \citet{publ21a}, \citet{stet24a} compile which standards of
ISO/IEC and the European Telecommunications Standards Institute
(ETSI) operationalise the individual requirements of the
EU AI Act for high-risk systems. Table~\ref{tbl-ki-vo-iso-en} adopts
this inventory. From this inventory a tighter core can be distilled:
\citet{publ21a} compress the twelve operationalisation- and
suitability-essential standards into six so-called \emph{core
standards} (Figure~\ref{fig-core-standards-en}) that are present in
both essential groups and therefore form the smallest common
denominator of the maturity and operationalisation analysis carried
out. These are the following six standards: ISO/IEC
4213, ISO/IEC 5338, ISO/IEC 23894, ISO/IEC 24027, ISO/IEC 38507, and
ISO/IEC 42001. Table~\ref{tbl-core-standards-en} explains the six
core standards and refers in each case to the ISO/IEC
primary source.

\begin{longtable}{p{0.33\columnwidth} p{0.6\columnwidth}}
\caption{\label{tbl-ki-vo-iso-en}Mapping of the EU AI Act's
high-risk-system requirements to the relevant ISO/IEC and ETSI
operationalisation standards. The listing follows \citet{stet24a},
Table~1 (there following the JRC report 125952 \citep{publ21a}).
Many of the standards listed are still under
development. TS = Technical Specification; SAI = Securing
Artificial Intelligence (ETSI Industry Specification Group).}\\
\hline
\textbf{EU AI Act requirement} & \textbf{Relevant operationalisation standards} \\
\hline
\endfirsthead
\multicolumn{2}{l}{\itshape (continued from \tablename~\ref{tbl-ki-vo-iso-en})}\\
\hline
\textbf{EU AI Act requirement} & \textbf{Relevant operationalisation standards} \\
\hline
\endhead
\hline
\multicolumn{2}{r}{\itshape (continued on next page)}\\
\endfoot
\hline
\endlastfoot
Art.~9: Risk-management system & ISO/IEC 5338; ISO/IEC 5469; ISO/IEC 23894.2; ISO/IEC 38507; ISO/IEC 42001 \\
Art.~10: Data and data governance & ISO/IEC TS 4213; ISO/IEC 5259-2, -3, -4; ISO/IEC 5338; ISO/IEC 5469; ISO/IEC 23894.2; ISO/IEC 24027; ISO/IEC 24029-1; ISO/IEC 24668; ISO/IEC 38507; ISO/IEC 42001; ETSI SAI 002, 005 \\
Art.~11: Technical documentation & ISO/IEC 23894.2; ISO/IEC 24027; ISO/IEC 42001 \\
Art.~12: Record-keeping obligations & ISO/IEC 23894.2 \\
Art.~13: Transparency to users & ISO/IEC 23894.2; ISO/IEC 24028; ISO/IEC 38507; ISO/IEC 42001 \\
Art.~14: Human oversight & ISO/IEC 23894.2; ISO/IEC 38507; ISO/IEC 42001 \\
Art.~15: Accuracy, robustness, cybersecurity & ISO/IEC TS 4213; ISO/IEC 5338; ISO/IEC 5469; ISO/IEC 23894.2; ISO/IEC 24029-1; ISO/IEC 24668; ISO/IEC 42001; ETSI SAI 002, 003, 005, 006 \\
Art.~17: Quality-management system & ISO/IEC 5259-3, -4; ISO/IEC 5338; ISO/IEC 23894.2; ISO/IEC 24029-1; ISO/IEC 38507; ISO/IEC 42001 \\
\end{longtable}

\begin{longtable}[]{@{}
  >{\raggedright\arraybackslash}p{(\linewidth - 4\tabcolsep) * \real{0.2644}}
  >{\raggedright\arraybackslash}p{(\linewidth - 4\tabcolsep) * \real{0.5517}}
  >{\raggedright\arraybackslash}p{(\linewidth - 4\tabcolsep) * \real{0.1839}}@{}}
\caption{Explanation of the six core standards from \citet{publ21a}.
The standards present in the intersection of the operationalisation-
and suitability-essential groups represent the smallest common
denominator of the ISO/IEC standards identified as particularly
relevant to the EU AI Act. SC 42 = ISO/IEC JTC 1/SC 42 \emph{Artificial
intelligence}; SC 40 = JTC 1/SC 40 \emph{IT Service Management and IT
Governance}; PDCA =
Plan--Do--Check--Act.}\label{tbl-core-standards-en}\tabularnewline
\toprule\noalign{}
\begin{minipage}[b]{\linewidth}\raggedright
\textbf{Standard}
\end{minipage} & \begin{minipage}[b]{\linewidth}\raggedright
\textbf{Explanation}
\end{minipage} & \begin{minipage}[b]{\linewidth}\raggedright
\textbf{Reference}
\end{minipage} \\
\midrule\noalign{}
\endfirsthead
\toprule\noalign{}
\begin{minipage}[b]{\linewidth}\raggedright
\textbf{Standard}
\end{minipage} & \begin{minipage}[b]{\linewidth}\raggedright
\textbf{Explanation}
\end{minipage} & \begin{minipage}[b]{\linewidth}\raggedright
\textbf{Reference}
\end{minipage} \\
\midrule\noalign{}
\endhead
\bottomrule\noalign{}
\endlastfoot
ISO/IEC TS 4213:2022 & Standardised methods for measuring the
classification performance of machine-learning models, systems, and
algorithms (Technical Specification, 33 pages). & \citet{isoi22a} \\
ISO/IEC 5338:2023 & Lifecycle processes for AI systems; adopts
ISO/IEC/IEEE 15288 and 12207 and adds AI-specific processes from
ISO/IEC 22989 and 23053. & \citet{isoi23a} \\
ISO/IEC 23894:2023 & Guideline for AI risk management,
methodologically aligned with ISO 31000; with concrete implementation
examples across the entire AI lifecycle. & \citet{isoi23b} \\
ISO/IEC TR 24027:2021 & Technical report on bias in AI systems and
AI-supported decision-making; measurement procedures and a typology
of bias sources (cognitive, data, engineering bias). & \citet{isoi21a} \\
ISO/IEC 38507:2022 & IT-governance implications of AI use by
organisations; addresses the governing body and is, in deviation from
the other core standards, maintained by SC 40 (not SC 42). &
\citet{isoi22b} \\
ISO/IEC 42001:2023 & Certifiable management-system standard for AI
(PDCA-based); requirements for policies, risk assessment,
lifecycle steering, and supplier oversight, the ``ISO 27001 for
AI''. & \citet{isoi23c} \\
\end{longtable}

Table~\ref{tbl-core-standards-mapping-en} maps each of the six core
standards to the design principles of Chapter~\ref{sec-principles-en}
(CR-1--CR-4 code rules, PR-1--PR-4 process rules) and to the supporting
\texttt{spotforecast2-safe} artefacts. The mapping refers
exclusively to mechanisms already documented in this report. It does
not invent new controls but links the existing body of evidence to
the ISO/IEC vocabulary.

\begin{longtable}[]{@{}
  >{\raggedright\arraybackslash}p{(\linewidth - 4\tabcolsep) * \real{0.2727}}
  >{\raggedright\arraybackslash}p{(\linewidth - 4\tabcolsep) * \real{0.2727}}
  >{\raggedright\arraybackslash}p{(\linewidth - 4\tabcolsep) * \real{0.4545}}@{}}
\caption{Mapping of the six core standards from
Figure~\ref{fig-core-standards-en} to the design principles in
Chapter~\ref{sec-principles-en} and the supporting
\protect\texttt{spotforecast2-safe} artefacts. ISO/IEC TS 4213 and
ISO/IEC TR 24027 are classification-related; for regression-based
forecasting only their methodology, respectively their bias-source
taxonomy, transfers. ISO/IEC 38507 addresses the deployer's
governing body; the package artefacts listed provide the substrate
on which governance decisions can be made, not the governance
structure itself. CR = code rule; PR = process rule;
PDCA = Plan--Do--Check--Act; SBOM = Software Bill of Materials;
CPE = Common Platform Enumeration; STRIDE = Spoofing, Tampering,
Repudiation, Information disclosure, Denial of service, Elevation of
privilege.}\label{tbl-core-standards-mapping-en}\tabularnewline
\toprule\noalign{}
\begin{minipage}[b]{\linewidth}\raggedright
\textbf{Standard}
\end{minipage} & \begin{minipage}[b]{\linewidth}\raggedright
\textbf{Supporting rules}
\end{minipage} & \begin{minipage}[b]{\linewidth}\raggedright
\textbf{Package mechanisms and evidence}
\end{minipage} \\
\midrule\noalign{}
\endfirsthead
\toprule\noalign{}
\begin{minipage}[b]{\linewidth}\raggedright
\textbf{Standard}
\end{minipage} & \begin{minipage}[b]{\linewidth}\raggedright
\textbf{Supporting rules}
\end{minipage} & \begin{minipage}[b]{\linewidth}\raggedright
\textbf{Package mechanisms and evidence}
\end{minipage} \\
\midrule\noalign{}
\endhead
\bottomrule\noalign{}
\endlastfoot
ISO/IEC 42001:2023 \citep{isoi23c} & PR1--PR4, CR1, CR4 &
Policy and objective framework in \texttt{MODEL\_CARD.md} (Sections 1
and 4) and \texttt{CONTRIBUTING.md}; risk assessment via
STRIDE (PR2); technical controls via CR1--CR4; PDCA cycle through
the \texttt{Makefile} targets \texttt{verify}, \texttt{preflight},
\texttt{release}, and \texttt{ship} documented in
\texttt{RELEASING.md}; supplier oversight via \texttt{uv.lock} and
\texttt{tests/test\_prohibited\_dependencies.py}; lifecycle evidence
via release tags and audit log
(\texttt{manager/logger.py},
\texttt{audit\_log\_schema.json}). \\
ISO/IEC 23894:2023 \citep{isoi23b} & PR2, CR3, PR1, PR4 &
Risk identification through module-wise STRIDE in
\texttt{CONTRIBUTING.md} and the security policy of the package
documentation; risk analysis
through CR3 fail-safe contracts (typed \texttt{on\_missing} switch)
and CR2 determinism; risk treatment through CR4 dependency
blocklist; monitoring through the structured audit log (PR4);
communication via \texttt{MODEL\_CARD.md} Section 6 and
\texttt{CHANGELOG.md}. \\
ISO/IEC 5338:2023 \citep{isoi23a} & CR1, CR2, PR1, PR3, PR4 &
Technical-management processes via the documented release procedure
and Conventional-Commits; development through the pre-commit chain
(\texttt{black}, \texttt{isort}, \texttt{ruff}) and
\texttt{test\_docstring\_examples\_*.py} (CR1); verification via
the enforced coverage floor (\texttt{fail\_under\ =\ 68}) and
\texttt{test\_linearly\_interpolate\_ts.py}; deployment
via versioned PyPI wheel plus CycloneDX SBOM (PR3, unsigned);
operation via audit log (PR4); deprecation via
\texttt{feat!:} major bumps in the \texttt{CHANGELOG.md}. \\
ISO/IEC TS 4213:2022 \citep{isoi22a} & CR2, PR1 & Methodological
discipline in performance measurement: rolling-origin evaluation
(Figure~\ref{fig-folds-en}), error-measure reporting
(Figure~\ref{fig-residuals-en}), and a documented evaluation metric
in \texttt{MODEL\_CARD.md} Section 7; reproducibility through CR2
determinism; PR1 links every metric to a \texttt{MODEL\_CARD}
requirement. \\
ISO/IEC TR 24027:2021 \citep{isoi21a} & CR2, CR3, PR1, PR4 &
Engineering bias eliminated by CR2 determinism; data bias addressed
by CR3 prohibition of silent imputation
(\texttt{on\_missing="raise"} default); cognitive bias sources in
\texttt{MODEL\_CARD.md} Section 6 (known limits, contraindicated
applications); bias measurement via residual analysis
(Figure~\ref{fig-residuals-en}) as the regression counterpart to the
24027 measurement specification for classifiers. \\
ISO/IEC 38507:2022 \citep{isoi22b} & PR1, PR3, PR4 (indirectly) &
The package supplies the substrate of the governance decision, not
the governance itself: \texttt{MODEL\_CARD.md} Sections 6 to 9 as a
decision basis; audit log (PR4) for operational traceability for
the governing body; CPE identifier in \texttt{utils/cpe.py} plus
CycloneDX SBOM from \texttt{make sbom} for supply-chain
risk; the responsibilities table in \texttt{MODEL\_CARD.md} Section 1
makes roles visible. \\
\end{longtable}

\subsection{The obligations of the EU AI Act in detail}\label{ki-vo-und-isoiec-25059-en}

The code and process rules of Chapter~\ref{sec-principles-en} map
onto individual provisions of the EU AI Act, IEC 61508-3,
ISO 26262-6, the ISA/IEC 62443 standards series, and the CRA. The
articles of the EU AI Act allow the obligations of the provider of a
high-risk AI system to be mapped onto the six core standards of
Table~\ref{tbl-core-standards-en}. These mappings are explained in
the following chapters. In detail, they concern

\begin{itemize}
\tightlist
\item
  Article 4 of the EU AI Act (AI literacy), which is discussed in
  Chapter~\ref{sec-ki-vo-4-en},
\item
  Article 9 of the EU AI Act (risk management), which is discussed in
  Chapter~\ref{sec-ki-vo-9-en},
\item
  Article 10 of the EU AI Act (data and data governance), which is
  discussed in Chapter~\ref{sec-ki-vo-10-en},
\item
  Article 11 of the EU AI Act (technical documentation), which is
  discussed in Chapter~\ref{sec-ki-vo-11-en},
\item
  Article 12 of the EU AI Act (record-keeping), which is discussed in
  Chapter~\ref{sec-ki-vo-12-en},
\item
  Article 13 of the EU AI Act (transparency towards users), which is
  discussed in Chapter~\ref{sec-ki-vo-13-en},
\item
  Article 14 of the EU AI Act (human oversight), which is discussed
  in Chapter~\ref{sec-ki-vo-14-en},
\item
  Article 15 of the EU AI Act (accuracy, robustness, cybersecurity),
  which is discussed in Chapter~\ref{sec-ki-vo-15-en},
\item
  Article 16 of the EU AI Act (obligations of providers), which is
  discussed in Chapter~\ref{sec-ki-vo-16-en},
\item
  Article 17 of the EU AI Act (quality-management system), which is
  discussed in Chapter~\ref{sec-ki-vo-17-en},
\item
  Article 18 of the EU AI Act (retention of documentation), which is
  discussed in Chapter~\ref{sec-ki-vo-18-en},
\item
  Article 72 of the EU AI Act (post-market monitoring), which is
  discussed in Chapter~\ref{sec-ki-vo-72-en}, and
\item
  Article 73 of the EU AI Act (reporting of serious incidents), which
  is discussed in Chapter~\ref{sec-ki-vo-73-en}.
\end{itemize}

\subsubsection{Article 4 of the EU AI Act (AI literacy)}\label{sec-ki-vo-4-en}

Article 4 of the EU AI Act \citep{euro24a} obliges providers and
deployers to ensure that their staff and other persons operating or
using an AI system on their behalf possess a sufficient level of AI
literacy, calibrated to the use context and to the groups of persons
on which the system acts. \texttt{spotforecast2-safe} is designed so
that a process engineer without formal training in machine learning
can attain those requirements by reading the package documentation
alone. Three mechanisms contribute:

\begin{itemize}
\tightlist
\item
  Every public transformer and every forecaster carries executable
  docstring examples, so that the intended behaviour can be learnt by
  running small, self-contained snippets and is not dependent on the
  study of external manuals.
\item
  Typed contracts make the pre- and post-conditions of every call
  visible at the call site. This lowers the expert knowledge required
  to read a training script correctly.
\item
  The structured audit log (PR-4, Chapter~\ref{sec-principles-en})
  discloses, line by line, the sequence of decisions taken by a
  running forecaster, so that the deployer's staff can review the
  operational behaviour after the fact.
\end{itemize}

The model-related \texttt{MODEL\_CARD.md} complements
these mechanisms by recording the intended deployment frame and the
known limits in a language with which non-specialist operating
personnel can also work. Taken together, these design choices lower
the AI-literacy threshold the deployer must reach in order to
supervise the system responsibly, and thereby support the obligation
of Art.~4 of the EU AI Act directly.

\subsubsection{Article 9 of the EU AI Act (risk management)}\label{sec-ki-vo-9-en}

Article 9 of the EU AI Act \citep{euro24a} obliges the provider of a
high-risk AI system to establish, implement, document, and maintain a
risk-management system (Risk Management System, RMS) (paragraph 1).
This is to be understood as a continuous, iterative process across
the entire lifecycle and comprises four steps (paragraph 2):
identification and analysis of known and reasonably foreseeable risks
(Art.~9(2)(a) of the EU AI Act); estimation and evaluation of risks
in the intended use and under reasonably foreseeable misuse
(Art.~9(2)(b) of the EU AI Act); evaluation of further risks from the
post-market monitoring system under Art.~72 of the EU AI Act
(Art.~9(2)(c) of the EU AI Act); and the adoption of suitable,
targeted risk-management measures (Art.~9(2)(d) of the EU AI Act).
The measures must take into account the combined interplay of the
requirements of Chapter III Section 2 (paragraph 4) and must be such
that the residual risks remaining are acceptable (paragraph 5). The
system must be tested against pre-defined metrics and probabilistic
thresholds (paragraphs 6 and 8), in particular with regard to minors
and other vulnerable groups (paragraph 9). Under parallel regulation
it may be combined with existing Union risk-management processes
(paragraph 10).

For a forecasting library embedded as a component, a \emph{shared
responsibility} applies here as in all the following articles: the
provider or deployer of the embedded high-risk system remains the
primary addressee of the respective duty, while the library
contributes the technical infrastructure with which that duty can
be met. For the RMS this means that \texttt{spotforecast2-safe}
covers the two provider-side duties that a library can address on
its own, namely the risk-mitigating design of its own code
(Art.~9(3) and Art.~9(5)(a) of the EU AI Act) and the supply of the
technical information on which the provider builds its own RMS
(Art.~9(5)(c) of the EU AI Act). Three mechanisms carry this
work.

\emph{Threat-model-based risk identification}
(Art.~9(2)(a) and (b) of the EU AI Act). For every network-facing
module (\texttt{downloader/entsoe.py},
\texttt{downloader/resilience.py}, \texttt{weather/\_\_init\_\_.py})
a STRIDE table is maintained in the module docstring, and the rule
is stated in \texttt{CONTRIBUTING.md}. Every change of the attack
surface requires an updated threat entry in the same merge request.
The machine enforcement of this rule via a path-filtered CI job is
described in Chapter~\ref{sec-process-rules-en}. This represents the identification and
evaluation steps for the intended use and reasonably foreseeable
misuse of the network-mediated part of the library. For the
data-side part, the typed \texttt{on\_missing} contract (see
Chapter~\ref{sec-code-rules-en}, CR-3) acts as misuse early-warning: a
missing or non-numerical input is not silently imputed but raised
to a documentation-required exception.

\emph{Design-side risk mitigation} (Art.~9(2)(d) and Art.~9(5)(a)
and (b) of the EU AI Act). The four code rules of
Chapter~\ref{sec-principles-en} (CR-1 \emph{No dead code}, CR-2
\emph{Determinism}, CR-3 \emph{Fail-safe}, CR-4 \emph{Minimal CVE
attack surface}) and the four process rules (PR-1
\emph{Traceability}, PR-2 \emph{Threat model}, PR-3
\emph{Supply-chain integrity}, PR-4 \emph{Structured audit log}) are
the concrete risk-management measures anchored in source code and the
development pipeline. They eliminate certain classes of risk
through the design of the library (Art.~9(5)(a) of the EU AI Act)
rather than through downstream usage notices, and are, in the
sense of Art.~9(4) of the EU AI Act, explicitly aligned with the
interplay of the remaining requirements of Chapter III Section 2:
the mapping is laid out row by row in
Table~\ref{tbl-compliance-en} and summarised visually in
Figure~\ref{fig-compliance-map-en}.

\emph{Continuous testing and feedback from monitoring}
(Art.~9(2)(c) and Art.~9(6) and (8) of the EU AI Act). The log
instrumentation described in Chapter~\ref{sec-ki-vo-72-en} feeds
events observed after placing on the market back into risk
analysis. The executable docstrings run via
\texttt{tests/test\_docstring\_examples\_*.py}, the coverage floor
fixed in \texttt{pyproject.toml} and enforced by \texttt{make test}
(\texttt{fail\_under\ =\ 68}),
and the rolling-origin cross-validation \texttt{split\_ts\_cv}
together form the test layer required by Art.~9(6) and (8) against
pre-defined metrics. Static security analysis and automated
dependency monitoring are the established tools for extending that
test layer with security-oriented checks across the whole
lifecycle.

\subsubsection{Article 10 of the EU AI Act (data governance)}\label{sec-ki-vo-10-en}

Article 10 of the EU AI Act \citep{euro24a} obliges the provider of
a high-risk AI system to data-governance measures across training,
validation, and test datasets (paragraphs 1--4) and contains an
exception rule for special categories of personal data (paragraph 5)
together with a scope limitation for systems developed without model
training (paragraph 6). For a forecasting library that does not itself ship
training data, the shared responsibility of
Chapter~\ref{sec-ki-vo-9-en} applies: above all, the package
prevents the substantive requirements from being silently bypassed.
Three mechanisms carry this support.

\emph{Documented design choices and assumptions} (Art.~10(2)(a)
and (d) of the EU AI Act). The structured \texttt{MODEL\_CARD.md}
framework (see Chapter~\ref{sec-ki-vo-11-en}), with sections
\emph{Intended Use and Scope}, \emph{Data and Operational Design
Domain}, and \emph{Evaluation}, forces the
deployer to record the relevant design choices and the
assumptions about the input data in writing.

\emph{Data provenance and preprocessing} (Art.~10(2)(b) and (c)
of the EU AI Act). Every \texttt{Forecaster} object persists in
its state the metadata of the fitting run: creation and fit
timestamps, training range, and library and Python versions (see
Chapter~\ref{sec-example-en}, Step 9). For
live-data paths, the adapters \texttt{downloader/entsoe.py} for
ENTSO-E \citep{entsoe24} and \texttt{weather/client.py}
for Open-Meteo \citep{openmeteo24} are available. Preprocessing with
data-changing semantics (linear interpolation, fill imputation,
outlier marking) is never inserted into the pipeline
automatically. It takes effect only through an explicit call by
the deployer (see Chapter~\ref{sec-code-rules-en}, CR-3), and the
fill imputation \texttt{get\_missing\_weights} returns alongside
the filled frame a weight series that excludes filled rows and
their lag tail from model training with sample weight zero (see
Chapter~\ref{sec-example-en}, Step 3). Annotation and labelling do not arise in the
regression-based forecasting use case.

\emph{Detection of data gaps} (Art.~10(2)(h) of the EU AI Act).
The fail-safe check \texttt{check\_y} rejects every target
series with missing values before fitting and thus turns a
silently-bypassed data gap into a discoverable, documentation-
required event. The deployer decision thereby enforced
(Chapter~\ref{sec-fail-safe-nan-en}) operationalises the ``identification of relevant data gaps or
shortcomings'' required by Art.~10(2)(h) of the EU AI Act.

For the bias requirements (Art.~10(2)(f) and (g) of the EU AI Act)
and the representativeness duty (Art.~10(3) of the EU AI Act),
assessment lies with the deployer, because biases arise only from
the interplay of training data and deployment scenario. The library
supports verifiability through deterministic transformations
(CR-2, see Chapter~\ref{sec-ki-vo-15-en}), so that repeated
experiments on the same dataset deliver bit-identical results and
deviations can be attributed exclusively to data differences. The
representativeness can be checked empirically by the deployer with
the ported rolling-origin cross-validation \texttt{split\_ts\_cv}.
Special categories of personal data (Art.~10(5) of the EU AI Act)
do not arise in the typical load-forecasting use case. If the
package is used exclusively for inference with an already-trained
model, the scope of Art.~10(2) to (5) of the EU AI Act narrows
under Art.~10(6) of the EU AI Act to the test dataset. This is a
configuration that the \texttt{Forecaster} persistence mechanism
expressly supports through its serialised training range and
recorded version information.

\subsubsection{Article 11 of the EU AI Act (technical
documentation)}\label{sec-ki-vo-11-en}

The open-source tool \texttt{quartodoc} \citep{chow25a} generates a
searchable HTML API reference from a Python library's docstrings
inside the Quarto publication system \citep{alla26a} and provides,
for \texttt{spotforecast2-safe}, the link between the executable
docstrings in the source code and the technical documentation
required by Annex IV of the EU AI Act.

Article 11 of the EU AI Act \citep{euro24a} read together with
Annex IV of the EU AI Act requires that, for every high-risk AI
system, technical documentation be drawn up before placing on the
market and kept up to date, enabling a competent authority or a
notified body to perform conformity assessment. For
\texttt{spotforecast2-safe}, three artefacts carry this duty: the
\texttt{MODEL\_CARD.md}, versioned in the repository, following the
taxonomy of the Hugging Face Model Card Guidebook \citep{ozon22a}
with the sections \emph{Uses}, \emph{Bias, Risks, and Limitations},
\emph{How to Get Started}, \emph{Evaluation}, \emph{Environmental
Impact}, \emph{Glossary}, and \emph{Citation}, as well as the mapping
to the EU AI Act and IEC 62443; the API reference generated
automatically from docstrings via \texttt{quartodoc} under
\texttt{docs/}; and
the present report itself, which records the architectural design
decisions and their mapping to the regulatory provisions. The
narrative documentation complements the machine-verifiable evidence
of Chapter~\ref{sec-ki-vo-13-en} and Chapter~\ref{sec-ki-vo-15-en}
without replacing it.

\subsubsection{Article 12 of the EU AI Act (record-keeping
obligations)}\label{sec-ki-vo-12-en}

Article 12 of the EU AI Act \citep{euro24a} obliges the provider to
design a high-risk AI system so that it records events (``logs'')
technically and automatically across the entire lifecycle
(paragraph 1). The logging must capture those events (paragraph 2)
that serve to identify risks within the meaning of Art.~79(1) of the
EU AI Act and substantial modifications (Art.~12(2)(a) of the
EU AI Act), enable post-market monitoring under Art.~72 of the
EU AI Act (Art.~12(2)(b) of the EU AI Act), and support the
oversight of systems covered by Art.~26(5) of the EU AI Act
(Art.~12(2)(c) of the EU AI Act). For the special class of remote
biometric identification systems (Annex III No.~1(a) of the
EU AI Act), paragraph 3 additionally requires minimum content:
duration of use, reference database, input data of a hit, and
persons involved.

The implementation in the package is anchored in process rule PR-4
(see Chapter~\ref{sec-process-rules-en}) as a structured audit log
and concretely realised in the module
\texttt{src/spotforecast2\_safe/manager/logger.py}. The entry point
is the function \texttt{setup\_logging()}, which always sets up a
\emph{console handler} with a plain-text formatter for human
readability (configurable level, default \texttt{INFO}) and, as soon
as the caller passes a log directory \texttt{log\_dir}, additionally
a \emph{file handler} with the corresponding
class's \texttt{JsonAuditFormatter}, which persists at
\texttt{INFO} level independently of the console verbosity. The
log file is placed with a timestamped name
(\texttt{sf2-safe-logger\_YYYYMMDD\_HHMMSS.log}) in that directory,
so that every run produces its own log, sortable by start time. The
bundled tasks pass
\texttt{\textasciitilde{}/spotforecast2\_safe\_models/logs/} by
default. If the log file cannot be created, \texttt{setup\_logging()}
aborts with an \texttt{OSError} instead of silently falling back to
console-only logging. In a safety-critical workflow a missing
audit trail is not an acceptable state.
This satisfies Art.~12(1) of the EU AI Act (automatic logging
across the lifecycle) in a deployer-reproducible form of a
JSON-lines file.

The schema of individual log records is fixed in
\texttt{src/spotforecast2\_safe/manager/audit\_log\_schema.json}
(version 2.0.0) and comprises seven mandatory fields:
\texttt{schema\_version} (constant ``2.0.0''),
\texttt{prev\_hash} (the SHA-256 value of the preceding log line,
with a documented genesis value of 64 zeros in the first record of
a file), \texttt{timestamp\_utc} (ISO 8601 UTC with microsecond
precision and trailing ``Z''), \texttt{logger} (Python logger name),
\texttt{level}
(\texttt{DEBUG}/\texttt{INFO}/\texttt{WARNING}/\texttt{ERROR}/\texttt{CRITICAL}),
\texttt{event} (short slug such as \texttt{task\_start},
\texttt{fit}, \texttt{predict}, \texttt{fetch}), and
\texttt{message} (formatted plain text); in addition there are the
optional fields \texttt{task} (name of the originating top-level
task), \texttt{context} (structured supplement from
\texttt{extra=\{...\}}), and \texttt{exception} (traceback, if
\texttt{exc\_info} was set). Risk events under Art.~12(2)(a) of the
EU AI Act are identified in this schema by the combination of
\texttt{level\ \textgreater{}=\ ERROR}, a descriptive
\texttt{event} slug, and the \texttt{exception} field. Any change
to the schema itself is a \emph{substantial modification} within
the meaning of the provision and is bound by convention to a
Conventional-Commits subject \texttt{feat!:} and a major version
bump (see Chapter~\ref{sec-ki-vo-13-en}). This binding is upheld by
hand at review time. A comparison job that checks the schema
version against the package version can enforce it by machine. The
traceability of substantial modifications required by
Art.~12(2)(a) of the EU AI Act is thereby mapped from the release
path onto the log schema. The UTC timestamps with microsecond
precision and the constant \texttt{schema\_version} permit the
deployment-spanning correlation of events for post-market
monitoring (Art.~72 of the EU AI Act) required by Art.~12(2)(b).
The \texttt{task} field enables the reconstruction of operational
running required by Art.~12(2)(c).

The schema secures the form of the log records, not their content. A
deployer holding the files could subsequently edit, delete, or
reorder individual records, and it is precisely the deployer who,
under Art.~26(5) of the EU AI Act, oversees the operation from which
an authority is to reconstruct the system's behaviour. Since release
26.0.0 the package meets this gap with a hash chain: the required
field \texttt{prev\_hash} chains every record to the exact bytes of
its predecessor, and the public function
\texttt{verify\_audit\_log()} walks a file offline and reports the
first break with its line number. Changes, deletions, and
reorderings within a retained file thereby become detectable. The
limit of the mechanism is documented just as plainly: an unkeyed
chain can be recomputed by an adversary able to rewrite the whole
file. Evidential force towards third parties arises only once the
operator deposits the file name, the record count, and the chain
head in an independently controlled store, for instance via an
RFC 3161 timestamp or a transparency log. Like retention under
Art.~18 of the EU AI Act (Chapter~\ref{sec-ki-vo-18-en}), this
anchoring deliberately remains an operator duty, because the library
is offline-deterministic and performs no network calls. Since
release 27.0.0 the verification report treats this limit as a
finding in its own right: \texttt{verify\_audit\_log()} accepts a
previously deposited chain head as an anchor and reports its status,
and without a presented anchor the report names the missing
anchoring explicitly instead of certifying the intact chain alone.
The repository's make targets \texttt{anchor} and
\texttt{anchor-verify} automate the deposit via an RFC 3161
timestamp, with the network call living in the operator tooling
rather than in the library. Example~\ref{exm-fahrtenbuch-en}
illustrates mechanism and limit.

\begin{example}[Logbook with
fingerprints]\protect\hypertarget{exm-fahrtenbuch-en}{}\label{exm-fahrtenbuch-en}

The hash-chained audit log resembles a logbook in which every new
line contains a fingerprint of the previous line. If a line is later
changed, deleted, or reordered, the fingerprints no longer match and
the check raises an alarm. This protects only the inside of the
book. Whoever holds the whole book can rewrite it completely, with
fresh fingerprints that all match. A remedy is to deposit the
latest fingerprint outside the logbook, for instance by publishing
it as a newspaper notice or by lodging it with a notary. This
deposit is the external anchoring of the chain head.

\end{example}

Article 12(3) of the EU AI Act is not substantively applicable
because \texttt{spotforecast2-safe} is a regression-based
forecasting library and not a remote biometric identification
system under Annex III No.~1(a) of the EU AI Act. Should a deployer
wish to record an interval-oriented capture for their own audit
profile, the schema provides a technical basis for it.

\subsubsection{Article 13 of the EU AI Act (transparency)}\label{sec-ki-vo-13-en}

Article 13 of the EU AI Act \citep{euro24a} obliges the provider of
a high-risk AI system to design the system so that the deployer can
interpret and use its output appropriately, and to supply
instructions for use that include, among other things, the intended
purpose, performance characteristics, pre-determined changes, and
necessary maintenance and updating measures. The package treats this
duty as an executable rather than a purely narrative specification.
Four mechanisms together form the machine-verifiable part of the
audit trail.

\emph{Executable docstrings.} Every public symbol ships with a
docstring whose examples are
executed by the test suite via
\texttt{test\_docstring\_examples\_*.py} on every test run. This
rules out a common failure mode in which documentation and
behaviour drift apart unnoticed. Executable docstrings document
the intended purpose and the expected behaviour of every public
symbol and thereby directly address Art.~13(3)(b)(i) of the
EU AI Act (intended purpose) and (vii) (information for the
appropriate interpretation of the output).

\emph{Product identity via the CPE identifier.} The CPE 2.3
identifier fixed in the package, described in detail in
Chapter~\ref{sec-cpe-sbom-en} \citep{nistir7695}, identifies
vendor and product in every derived SBOM and thereby satisfies
the identity part of Art.~13(3)(a)
of the EU AI Act. The cybersecurity benefit it provides
(automated resolution against CVE reports and the CISA KEV
catalogue) is discussed in Chapter~\ref{sec-ki-vo-15-en}
alongside the other cybersecurity mechanisms.

\emph{Conventional Commits and the release procedure.} The commit
message format enforced by pre-commit corresponds to Conventional
Commits. A subject \texttt{fix:} indicates, by convention, a patch
version bump, \texttt{feat:} a minor bump, \texttt{feat!:} or
\texttt{fix!:} a major bump. The release itself is a documented
sequence of three \texttt{Makefile} targets,
\texttt{make\ preflight}, \texttt{make\ release\ VERSION=X.Y.Z},
and \texttt{make\ ship\ VERSION=X.Y.Z}, which
\texttt{RELEASING.md} describes in full. It checks the working
tree, refuses the release as soon as a commit since the last tag
is not readable as a Conventional Commit, regenerates the
\texttt{CHANGELOG.md}, sets the tag, and publishes a wheel on
PyPI. This pipeline addresses Art.~13(3)(c)
of the EU AI Act (pre-determined changes by the provider at the
time of conformity assessment) and Art.~13(3)(e) of the
EU AI Act (software updates as part of maintenance and care
measures). At the same time it operationalises IEC 62443-4-1
SUM-4 (secure delivery of updates) \citep{iec62443_4_1}, though
only partially, because the wheel is unsigned.

\emph{Narrative documentation as a complement.} The
machine-verifiable mechanisms (docstrings, CPE,
Conventional Commits) are complemented, not replaced, by narrative
documentation: the present report, the \texttt{MODEL\_CARD.md},
and the \texttt{CHANGELOG.md} (see Chapter~\ref{sec-ki-vo-11-en}).
The shipped \texttt{MODEL\_CARD.md} contains, among others, the
sections \emph{Intended Use and Scope}, \emph{Data and Operational
Design Domain}, \emph{Evaluation}, \emph{Model Transparency}, and
\emph{How to Audit} and is therefore itself a
transparency artefact within the meaning of Art.~13 of the
EU AI Act.

\subsubsection{Article 14 of the EU AI Act (human
oversight)}\label{sec-ki-vo-14-en}

The same mechanisms as in Chapter~\ref{sec-ki-vo-4-en} also address
the provider-side counterpart in Art.~14(4) of the EU AI Act
\citep{euro24a}, according to which a high-risk AI system must be
provided so that the persons entrusted with human oversight are
enabled to understand the capabilities and limits of the system
(Art.~14(4)(a) of the EU AI Act), to monitor its operation and to
detect anomalies, malfunctions, and unexpected performance (also
Art.~14(4)(a) of the EU AI Act), and to interpret its output
correctly (Art.~14(4)(c) of the EU AI Act).

Executable docstrings,
typed contracts, and \texttt{MODEL\_CARD.md} satisfy the
understanding and interpretation duties. The structured audit log
satisfies the monitoring and anomaly-detection duty. Article 4 of
the EU AI Act establishes the deployer-side competence duty,
Article 14(4) of the EU AI Act the provider-side design duty that
makes that competence operationally attainable in the first place.
The artefacts of \texttt{spotforecast2-safe} satisfy both duties
from the same source basis.

\subsubsection{Article 15 of the EU AI Act (accuracy, robustness, and
cybersecurity)}\label{sec-ki-vo-15-en}

Article 15 of the EU AI Act \citep{euro24a} describes accuracy,
robustness, and cybersecurity for high-risk AI systems. Paragraph 1
requires that these properties be delivered consistently
``throughout their lifecycle'', paragraph 4 demands resilience
against errors, faults, and inconsistencies. On the implementation
side of \texttt{spotforecast2-safe}, this consistency requirement
is met through a consistently deterministic execution path secured
on three layers. At the estimator level, LightGBM and XGBoost are
always instantiated with a \texttt{random\_state} passed through by
the wrapper. In addition, the options \texttt{deterministic=True}
and \texttt{force\_col\_wise=True} are set for LightGBM to remove
the difference between row-wise and column-wise histogram
construction on multi-core systems. At the feature level, the order
of concatenation of exogenous features is Python-version-stable,
because the \texttt{ExogBuilder} stores its components in a list
rather than a dictionary. At the I/O level, a Parquet round-trip
preserves the dtypes but not the frequency of the DatetimeIndex. The
cache reader therefore compares value-based, not frequency-based,
and the dedicated test suite
\texttt{tests/test\_weather\_client.py::TestWeatherServiceCache}
fixes this behaviour. The consistent performance across the
lifecycle required by Art.~15(1) of the EU AI Act is thus evidenced
in a machine-verifiable, deployer-reproducible form.

In addition to robustness, the cybersecurity requirement under
Art.~15(5) of the EU AI Act addresses the defence and tracking of
vulnerabilities. The CPE 2.3 identifier fixed in the package,
described in detail in Chapter~\ref{sec-cpe-sbom-en}
\citep{nistir7695}, is the technical link that downstream
vulnerability scanners use to assign a CVE reported under the same
CPE deterministically to the installed package version. Without
this fixing, scanners would have to rely on heuristic
package-name matches, with the well-known false-positive and
miss risks.

Beyond the binding obligations of the regulation, the AI-specific
extension of the \emph{Systems and software Quality Requirements
and Evaluation} (SQuaRE) standards series, ISO/IEC 25059:2023
\citep{isoiec25059}, provides the quality vocabulary that the harmonised standards of
the European Committee for Standardisation (CEN) and the European
Committee for Electrotechnical Standardisation (CENELEC) expected
under Art.~40 of the EU AI Act are likely to adopt. Its quality model refines the
sub-characteristics functional correctness, robustness, and user
controllability of ISO/IEC 25010 for AI systems and is therefore
the natural technical counterpart to Art.~15 of the EU AI Act. The
2023 edition is currently being revised: the enquiry on the Draft
International Standard ISO/IEC 25059:2025 has closed, and
an EU-AI-Act-aligned European edition (Preliminary European Norm,
prEN ISO/IEC 25059) was submitted to public enquiry via the Joint
Technical Committee 21 (JTC 21) of CEN-CENELEC in February~2026.

\subsubsection{Article 16 of the EU AI Act (provider
duties)}\label{sec-ki-vo-16-en}

Article 16 of the EU AI Act bundles the provider-side duties for
high-risk AI systems into a catalogue. The catalogue combines
substantive product requirements, organisational requirements on
the provider, and procedural requirements on the placing-on-the-
market and post-market path. To these is added Art.~16(b) of the
EU AI Act, which requires an identifying provider mark on the
product itself, on its packaging, or on the accompanying
documentation.

Because the catalogue refers predominantly to other articles of the
regulation, Art.~16(a) of the EU AI Act is, so far as its obligations
apply at all, for \texttt{spotforecast2-safe} satisfied through the sections
already treated on Art.~9 (Chapter~\ref{sec-ki-vo-9-en}), Art.~10
(Chapter~\ref{sec-ki-vo-10-en}), Art.~11
(Chapter~\ref{sec-ki-vo-11-en}), Art.~12
(Chapter~\ref{sec-ki-vo-12-en}), Art.~13
(Chapter~\ref{sec-ki-vo-13-en}), and Art.~15
(Chapter~\ref{sec-ki-vo-15-en}) of the EU AI Act. The duty under
Art.~16(c) of the EU AI Act is covered by
Chapter~\ref{sec-ki-vo-17-en}, that under Art.~16(d) of the
EU AI Act by Chapter~\ref{sec-ki-vo-18-en}, and that
under Art.~16(j) of the EU AI Act by
Chapter~\ref{sec-ki-vo-73-en}. Within the shared responsibility
(Chapter~\ref{sec-ki-vo-9-en}), the package directly covers three
duty clusters at component level.

\emph{Identity and contact} (Art.~16(b) of the EU AI Act). The
package name, the AGPL-3.0-or-later licence, the maintainer
address, and the vulnerability reporting channel are fixed in
\texttt{pyproject.toml}, in the PyPI listing, and in the security
policy of the package documentation and are republished with every
release. The CPE 2.3 identifier described in
Chapter~\ref{sec-cpe-sbom-en} is the machine-readable form of the
same statement and appears in every derived software bill of
materials. A provider that integrates the library into a
high-risk AI system can adopt the provenance information required
by Art.~16(b) of the EU AI Act deterministically from the
package, instead of maintaining it manually.

\emph{On-demand evidence} (Art.~16(k) in conjunction with
Art.~16(a) of the EU AI Act). The technical documentation
described in Chapter~\ref{sec-ki-vo-11-en}
(\texttt{MODEL\_CARD.md}, the \texttt{quartodoc} API reference,
and the present report), the row-by-row mapping in
Table~\ref{tbl-compliance-en}, and the graphical summary in
Figure~\ref{fig-compliance-map-en} provide the provider with the
evidence to be presented on official request under Art.~16(k) of
the EU AI Act. Because all evidence is under version control and
reproducible via the annotated release tag, the provider can, for
any shipped package version, reconstruct the corresponding
conformity evidence.

\emph{Corrections and accessibility of the technical
communication} (Art.~16(j) and Art.~16(l) of the EU AI Act). The
release procedure described in
Chapter~\ref{sec-ki-vo-13-en} delivers the corrective measures
required by Art.~16(j) of the EU AI Act in the form of
temporally traceable, tagged releases with a corresponding
\texttt{CHANGELOG.md} entry. For the accessibility of the
technical communication under Art.~16(l) of the EU AI Act, the
API documentation generated with \texttt{quartodoc} is shipped
as a pure HTML page with semantic markup (heading levels,
alternative texts on graphics, searchable plain text).
Executable docstrings additionally provide the function
description in searchable plain-text form. The package thereby
supports the two accessibility directives (EU) 2016/2102 (public
bodies) and (EU) 2019/882 (European Accessibility Act) at the
documentation level, which a library can reach on its own.

Some duties of Art.~16 of the EU AI Act lie beyond the reach of a
library and remain reserved to the provider: Art.~16(e) of the
EU AI Act concerns the operational retention of the logs
automatically generated by the deployed system (the package
generates these via the audit-log schema, see
Chapter~\ref{sec-ki-vo-12-en}; their retention falls to the
provider); Art.~16(f) of the EU AI Act requires undergoing a
conformity assessment procedure under Art.~43 of the EU AI Act for
the embedded overall system; Art.~16(g) of the EU AI Act the EU
declaration of conformity under Art.~47 of the EU AI Act; Art.~16(h)
of the EU AI Act the CE marking under Art.~48 of the EU AI Act;
Art.~16(i) of the EU AI Act registration in the EU database under
Art.~49(1) of the EU AI Act. These duties presuppose a complete
product and cannot be satisfied by a component. The
machine-readable conformity evidence described above accelerates the
provider's discharging of these procedures.

\subsubsection{Article 17 of the EU AI Act (quality-management
system)}\label{sec-ki-vo-17-en}

The EU AI Act organises the provider-side duties around Art.~17 of
the EU AI Act. This requires a documented quality-management
system (QMS) that integrates the individual substantive duties
(risk management in Art.~9 of the EU AI Act, data governance in
Art.~10 of the EU AI Act, technical documentation in Art.~11 of
the EU AI Act, record-keeping in Art.~12 of the EU AI Act,
transparency in Art.~13 of the EU AI Act, accuracy and robustness
in Art.~15 of the EU AI Act) with the provider-side process duties
(record-keeping of the provider under Art.~18 of the EU AI Act,
post-market monitoring under Art.~72 of the EU AI Act, and
reporting of serious incidents under Art.~73 of the EU AI Act).
Table~\ref{tbl-compliance-en} therefore contains an ``EU AI Act
Art.~17 row'' whose mechanism column points to the mechanisms used
to satisfy the respective sub-duty and cross-references the
adjacent rows. A deployer preparing a conformity assessment can
read the ``EU AI Act Art.~17 row'' as an index into the rest of
the table.

A mechanism not explicitly required by Art.~17 of the EU AI Act,
but indispensable for a software provider's QMS, is licence
traceability. \texttt{spotforecast2-safe} satisfies it through
REUSE v3.0 conformance (detailed in
Chapter~\ref{sec-spdx-reuse-en}): every source file, every data
fixture, and every test carries an SPDX header;
\texttt{uv\ run\ reuse\ lint} is integrated into pre-commit and is
run on every commit; \texttt{REUSE.toml} resolves the licence of
the few externally licensed files. From this conformance, the
deployer can at any time generate a \texttt{SPDX-3.0-JSON}
document via \texttt{reuse\ spdx} and feed it into their own QMS.
Article 13 of the EU AI Act does not itself raise the licence
question. A downstream deployer must clarify it before
integrating the library, and the provider's QMS supplies the
answer already at release time.

\subsubsection{Article 18 of the EU AI Act (retention of
documentation)}\label{sec-ki-vo-18-en}

The retention obligation of Art.~18 of the EU AI Act primarily
concerns the technical documentation under Art.~11 of the EU AI
Act and the QMS under Art.~17 of the EU AI Act. The market
surveillance authorities competent for it (in Germany the
Bundesnetzagentur) must be able to inspect the high-risk AI system
at any time. The retention period is 10 years after placing on the
market or putting into service (Art.~3 No.~9 and No.~11 of the
EU AI Act, respectively). In view of the forthcoming Product
Liability Act, longer retention periods are advisable.

\subsubsection{Article 72 of the EU AI Act (post-market
monitoring)}\label{sec-ki-vo-72-en}

Article 72 of the EU AI Act \citep{euro24a} obliges the provider of
a high-risk AI system to set up and document a monitoring system
appropriate to the risk and the technology (paragraph 1) that
actively and systematically collects, documents, and analyses
performance data over the entire lifetime of the system
(paragraph 2), is based on a written monitoring plan that is part
of the technical documentation under Annex IV (paragraph 3), and
can be integrated into existing monitoring systems from other EU
harmonisation legal acts (paragraph 4). The monitoring system must
enable the provider to perform a continuous assessment of conformity
with the requirements of Chapter III Section 2 of the EU AI Act.
Operationally sensitive data of a law-enforcement authority acting
as a deployer is exempt.

Within the shared responsibility (Chapter~\ref{sec-ki-vo-9-en}),
\texttt{spotforecast2-safe} provides the technical infrastructure
from which the required monitoring system can be fed, organised
into three coordinated data sources.

\emph{Operational log instrumentation.} The audit-log schema
v2.0.0 described in Chapter~\ref{sec-ki-vo-12-en} produces, per
forecasting run, a timestamp-named JSON file under
\texttt{\textasciitilde{}/spotforecast2\_safe\_models/logs/} with
\texttt{timestamp\_utc} at microsecond resolution, an event slug
(\texttt{event}), log level, the name of the originating task
(\texttt{task}), and a free-form context dictionary. A monitoring
system can collect these files via arbitrary log-aggregation
tools (ELK stack, Grafana Loki, Splunk) and thereby implement the
active and systematic performance capture across the lifetime
required by Art.~72(2) of the EU AI Act, without the package
having to ship its own telemetry infrastructure.

\emph{Lifecycle metadata of the deployment instance.} Every
\texttt{Forecaster} object persists its fitting metadata and the
model hyperparameters (see
Chapter~\ref{sec-ki-vo-10-en}); the CPE identifier (see
Chapter~\ref{sec-cpe-sbom-en}) identifies the package version
used in every derived SBOM; the release procedure (see
Chapter~\ref{sec-ki-vo-13-en}) documents in
\texttt{CHANGELOG.md} machine-readably how successive versions
differ from one another. Together, these artefacts supply the
monitoring system with the information necessary for the
continuous conformity assessment required by Art.~72(2) of the
EU AI Act, because the provider can, for every stored log entry,
reconstruct the exact combination of package version, training
data, and configuration.

\emph{Supply-chain monitoring.} The CPE identifier (see
Chapter~\ref{sec-cpe-sbom-en}) and the CycloneDX SBOM from
\texttt{make\ sbom} allow the package and its resolved dependency
graph to be resolved continuously against the National
Vulnerability Database and the CISA KEV catalogue. Static security
analysis and automated dependency monitoring usually bind that
resolution to every push. This monitoring
covers the dimension of
``interaction with other AI systems'' contemplated within
Art.~72(2) of the EU AI Act, in so far as that dimension is
mediated through third-party components, and feeds the same event
chain that also serves Art.~73 of the
EU AI Act. For the provider, this means that the monitoring duty
under Art.~72 of the EU AI Act and the reporting duty under
Art.~73 of the EU AI Act rest on the same event chain and need
not be maintained in two separate systems.

The monitoring plan required under Art.~72(3) of the EU AI Act is
part of the technical documentation and finds its written
expression in the \texttt{MODEL\_CARD.md} (see
Chapter~\ref{sec-ki-vo-11-en}). The German law implementing the
EU AI Act (KI-Marktüberwachungs- und Innovationsförderungsgesetz) is, as of 1~April~2026, in the Bundesrat's statement
following the federal-government briefing. Adoption is expected by
the end of 2026. Among the planned supplementary information, a
template for Art.~72(3) of the EU AI Act may be issued.

The Commission has not made use of the empowerment under Art.~72(3)
of the EU AI Act to issue a model. There are deliberations
about deleting this empowering provision in a future revision of
the EU AI Act. The background is the idea of preserving maximum
flexibility for the development and deployment of high-risk AI.

For a provider that simultaneously bears duties under other EU
harmonisation legal acts (e.g., the NIS-2 Directive, or for medical
high-risk products of the Medical Device Regulation (MDR) or the
In-vitro Diagnostic Medical Devices Regulation (IVDR)), paragraph 4
permits the integration
of the EU AI Act requirements into the existing monitoring system,
or the integration of the QMS from other sectoral legal provisions
into the EU AI Act QMS, in order to avoid double documentation
(Recital 81 of the EU AI Act). The unified log and release trail of
the package supports this integration because it is
machine-readable and can be fed into any existing system for
security information and event management (SIEM) or
device-surveillance system.

\subsubsection{Article 73 of the EU AI Act (reporting of serious
incidents)}\label{sec-ki-vo-73-en}

Article 73 of the EU AI Act \citep{euro24a} obliges the provider of
a high-risk AI system to report every serious incident to the
market surveillance authority of the affected Member State
(paragraph 1). The deadlines are graded: at the latest 15 days
after becoming aware in the standard case (paragraph 2), at the
latest 2 days for widespread infringements or incidents within the
meaning of Art.~3 No.~49(b) of the EU AI Act (paragraph 3), at the
latest 10 days in the case of death (paragraph 4). An initially
incomplete first report may be supplemented by a complete follow-up
report (paragraph 5). After the report, the provider promptly
conducts its own investigations (risk assessment and corrective
measures) and cooperates with the competent authorities.
Unilateral changes to the system that could impair a subsequent
root-cause analysis are prohibited (Art.~73(6) of the EU AI Act).

Within the shared responsibility (Chapter~\ref{sec-ki-vo-9-en}),
the primary addressee of the reporting duty is the provider of the
embedded system, regularly the deployer-integrator that puts the
library to use. The package provides three mechanisms that enable
the provider to meet the deadlines of Art.~73 of the EU AI Act
technically and to settle the investigation and correction chain
required by Art.~73(6) of the EU AI Act in a verifiable manner.

\emph{Confidential reporting channel and version policy.} The
security policy of the package documentation establishes the
maintainer's email address as the primary channel for
confidential reporting. As a second route, confidential issues in
the project repository are available. Public issues are explicitly
excluded. The published response commitments (acknowledgement
within 24 hours, initial assessment within 3 working days,
coordinated public disclosure only after a correction has been
made available) are temporally compatible with the 2-day and
10-day deadlines of Art.~73(3) and (4) of the EU AI Act. The
security policy additionally documents the version policy
(only the current release receives security updates, and there is
no long-term-support branch), so that the provider can clarify,
before reporting, on which version the investigation will be
based. The responsibilities table in \texttt{MODEL\_CARD.md}
Section 1 names the maintainer responsible for releases and makes
the cooperation axis required by Art.~73(6) of the EU AI Act
traceable.

\emph{Early-warning signals.} The CPE-anchored SBOM (see
Chapter~\ref{sec-cpe-sbom-en}) makes the package and its
dependencies deterministically resolvable for vulnerability
scanners and is thus the data basis from which providers and
maintainers draw early warnings before an incident reaches market
surveillance. That resolution is usually run continuously by static security
analysis and automated dependency alerts. In IEC 62443-4-1 terminology,
this early warning fills the process area \emph{DM} (Defect
Management). In EU AI Act terminology it creates the knowledge
state on the basis of which the provider can date the start of
the reporting deadlines at all.

\emph{Verifiable corrective measure.} After the triage of a
report, the release procedure (see Chapter~\ref{sec-ki-vo-13-en})
produces a new release with the corresponding
correction. The accompanying CycloneDX SBOM (see
Chapter~\ref{sec-cpe-sbom-en}) allows the recipient to
reconcile against the reported vulnerability. Releases currently
carry no cryptographic signature. The
\texttt{CHANGELOG.md} entry references the report's identifier, so
that the verification of an investigation-based correction chain
required by Art.~73(6) of the EU AI Act (commit trail through
release to publication log) is produced without additional
manual effort.

For cases in which the provider or deployer simultaneously bears
duties under other EU legal acts (NIS-2 Directive, MDR/IVDR for
medical high-risk products, CRA Annex II for products with digital
elements) and corresponding reporting duties under the EU AI Act,
Art.~73(9) and (10) of the EU AI Act reduces the reporting matter
to Art.~3 No.~49(c) of the EU AI Act. The mechanisms described
above cover this reduced scope equally, because the advisory,
release, and SBOM trails remain identical. What the library cannot
do is take the substantive decision whether a concrete incident
constitutes a serious incident within the meaning of Art.~3 No.~49
of the EU AI Act.

\subsection{Relation to the Cyber Resilience Act and the NIS-2
Directive}\label{sec-cra-nis2-en}

The CRA \citep{euCRA24}, in force since 10~December~2024 and with
its substantive duties applicable from 11~December~2027, imposes on
manufacturers of ``products with digital elements'' essential
cybersecurity requirements (Annex I Part I) and a
vulnerability-handling duty running for the support period of, as a
rule, at least five years (Art.~13(8) in conjunction with Annex I
Part II). A forecasting library shipped as a wheel
on PyPI is a CRA product with digital elements as soon as it is
made available on the Union market in the course of a commercial
activity. The Commission Implementing Decision C(2025)~618 of
3~February~2025 (standardisation request M/606) commissions CEN, CENELEC, and ETSI to develop
41 harmonised standards; type-C product-category standards are
foreseen by 30~October~2026, type-B technical-measures standards
by 30~October~2027; two horizontal standards (Annex~I entries 1 and
15) are due as early as 30~August~2026. EN IEC 62443-4-1 is among the expected type-B
reference standards. An adaptation to 62443-4-1 already in place
today is therefore at the same time the most cost-effective path to
CRA conformity assessment in 2027. NIS-2 \citep{euNIS22} already
applies. Art.~21(2)(e) NIS-2 obliges essential and important
entities to take measures for ``security in the procurement,
development, and maintenance of network and information systems''.

For the German KRITIS context, this set of duties can be made
concrete. The NIS-2 Directive regulates entities rather than
systems and makes no demand of a forecasting model as such. A
transmission system operator above the threshold of the draft
Kritisverordnung \citep{bund26a} is, however, an
operator of critical installations within the meaning of the
BSI-Gesetz \citep{deut25a}. § 30(1) BSIG, the German transposition
of the risk-management catalogue in Art.~21 NIS-2, obliges it to
prevent disruptions to the availability, integrity, and
confidentiality of every information-technology system, component,
and process it uses to deliver its services. A productively
deployed day-ahead forecasting service is such a system. If it is
material to the functioning of a critical installation, the duty to
deploy attack-detection systems under § 31(2) BSIG applies in
addition. The duties attach to the operator and reach the supplier
of the forecasting library through the supply-chain security that
the same catalogue of measures folds into the operator's risk
management (Art.~21(2)(d) NIS-2). In that reading, the artefacts
required by the process rules of
Chapter~\ref{sec-process-rules-en} are exactly the supplier-side
evidence such an operator would ask for: the software bill of
materials (PR-3), the documented threat model (PR-2), and the
structured audit log (PR-4). During the period documented here,
none of these duties bound the package itself, because
\texttt{spotforecast2-safe} is supplied open source outside a
commercial activity, and its maintainer is not an entity within the
meaning of the directive. This reading is therefore anticipatory in
the same sense as the classification under the CRA and the Product
Liability Directive in Chapter~\ref{sec-introduction-en}.

\subsection{Legal-Requirements-Engineering}\label{sec-lre-en}

\begin{figure}

\centering{

\pandocbounded{\includegraphics[keepaspectratio]{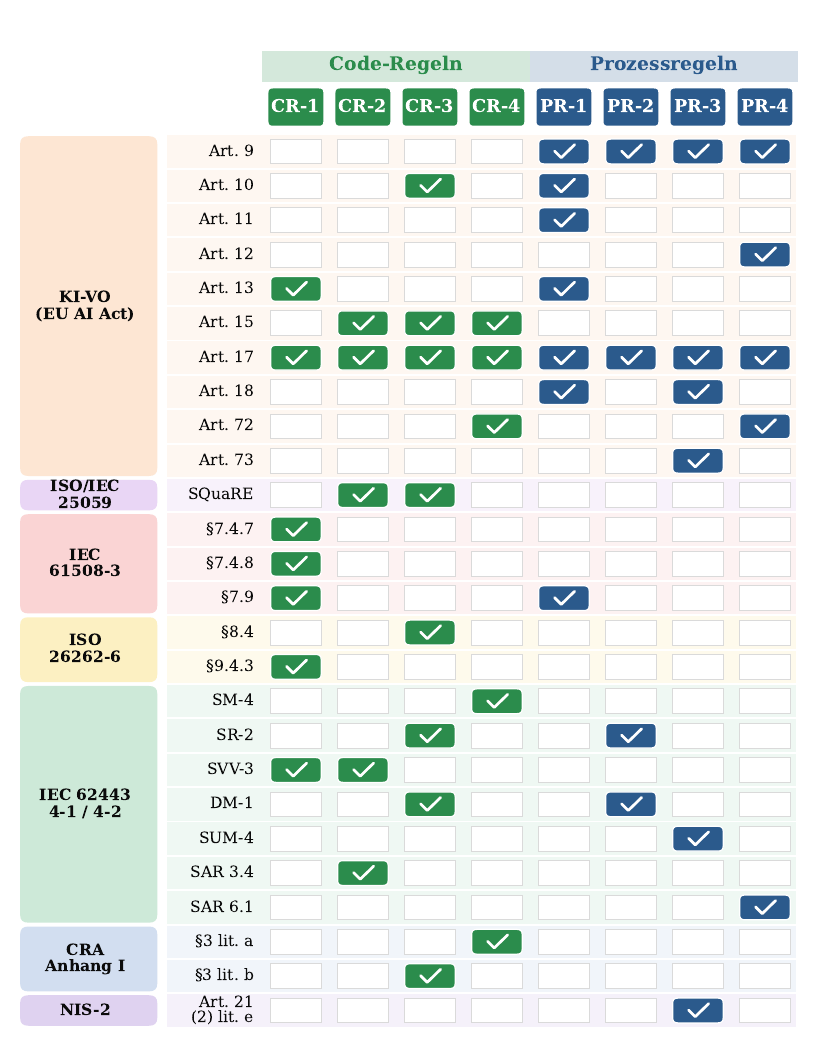}}

}

\caption{\label{fig-compliance-map-en}Compliance matrix: mapping of
the code rules (CR) and process rules (PR) of
Chapter~\ref{sec-principles-en} onto the regulatory provisions of
the preceding table. A tick in a cell indicates that the
implementation named in the mechanism column of
Table~\ref{tbl-compliance-en} instantiates that rule. Green-marked
rows denote code rules, blue rows process rules; the coloured
bands at the top group the provisions by rule family.}

\end{figure}

Table~\ref{tbl-compliance-en} sets out the LRE traceability matrix
for the present package. Each table row names a clause or
requirement together with the mechanism in the package that
addresses it. Figure~\ref{fig-compliance-map-en} summarises the
mapping visually.

{\small

\begin{longtable}{p{0.28\textwidth} p{0.64\textwidth}}

\caption{\label{tbl-compliance-en}Mapping of regulatory provisions
to package mechanisms. EU AI Act = Regulation (EU) 2024/1689;
IEC61508 = IEC 61508-3:2010; ISO26262 = ISO 26262-6:2018;
IEC62443 = ISA/IEC 62443 standards series; CRA = Regulation (EU)
2024/2847 (Cyber Resilience Act); NIS-2-RL = Directive (EU)
2022/2555.}

\tabularnewline

\\
\hline
\textbf{Provision} & \textbf{Mechanism} \\
\hline
\endfirsthead
\multicolumn{2}{l}{\itshape (continued from \tablename~\ref{tbl-compliance-en})}\\
\hline
\textbf{Provision} & \textbf{Mechanism} \\
\hline
\endhead
\hline
\multicolumn{2}{r}{\itshape (continued on next page)}\\
\endfoot
\hline
\endlastfoot
EU AI Act Art.~9 (risk management across the entire lifecycle) & The four process rules of Chapter~\ref{sec-principles-en} form the risk-management framework; STRIDE threat model per network-facing module; threat entries are updated in the same merge request as the corresponding code change. \\
EU AI Act Art.~10 (data governance) & Explicit \texttt{on\_missing} contract; no silent imputation; provenance of the training data recorded in \texttt{MODEL\_CARD.md}. \\
EU AI Act Art.~11 and Annex~IV (technical documentation) & \texttt{MODEL\_CARD.md} maintained in the repository for intended purpose, data provenance, metrics, and limits; automatically generated API documentation via quartodoc. \\
EU AI Act Art.~12 (record-keeping / automatic logs) & Double-handler logger: console plain text, file sink JSON to a fixed schema (\texttt{audit\_log\_schema.json}); \texttt{SCHEMA\_VERSION} is derived from the schema and pinned by a test. Schema changes are bound by convention to a Conventional-Commits subject \texttt{feat!:} and a major bump; the binding is upheld at review time. \\
EU AI Act Art.~13 (transparency) & Executable docstrings for every public symbol; fixed public API in \texttt{\_\_all\_\_}; CHANGELOG produced by \texttt{git-cliff} in the course of \texttt{make release}. \\
EU AI Act Art.~15 (accuracy, robustness, cybersecurity) & Deterministic transformations; blocklist of forbidden dependencies; CPE identifier for SBOM. \\
ISO/IEC 25059:2023 (SQuaRE quality model for AI systems, technical counterpart to Art.~15 EU AI Act) & Functional correctness through deterministic transformations; robustness through typed configuration and the fail-safe \texttt{on\_missing} contract; controllability through the prohibition of silent auto-adaptation documented in Chapter~\ref{sec-principles-en}. \\
EU AI Act Art.~17 (quality-management system as umbrella provision) & QMS documented as: (i) compliance strategy via the process rules of Chapter~\ref{sec-principles-en}; (ii) design control in \texttt{CONTRIBUTING.md}; (iii) verification and validation in the \texttt{make verify} pipeline; (iv) data management in the Art.~10 row; (v) risk management in the Art.~9 row; (vi) post-market monitoring and incident reporting in the rows on Art.~72 and Art.~73; (vii) record-keeping in the rows on Art.~12 and Art.~18; (viii) resource and responsibility framework in \texttt{MODEL\_CARD.md} Section 1. \\
EU AI Act Art.~18 (provider record-keeping: documentation, logs, conformity evidence) & The repository itself is the records archive: Conventional-Commits history, release tags, \texttt{CHANGELOG.md}, release-specific \texttt{MODEL\_CARD.md}, \texttt{uv.lock} snapshot per tag. Records are retained over the ten-year period of Art.~18 EU AI Act via the git history and the published PyPI releases. \\
EU AI Act Art.~72 (post-market monitoring) & Structured audit log under \texttt{\~{}/spotforecast2\_safe\_models/logs/}; CPE identifier enables vulnerability tracking after release. \\
EU AI Act Art.~73 (reporting of serious incidents to authorities) & Confidential reporting channel per the security policy of the package documentation (email, confidential issues); CPE-anchored SBOM as the early-warning data basis; CHANGELOG entries reference the report's identifier; coordinated disclosure with delivery of the fix via PyPI. \\
IEC 61508-3 Section 7.4.7 (requirements for testing software modules) & One test file per module; coverage floor \texttt{fail\_under\ =\ 68}, enforced by \texttt{make test}; 80\,\% aim for new code per \texttt{CONTRIBUTING.md}. \\
IEC 61508-3 Section 7.4.8 (requirements for software integration tests) & Top-level console scripts (\texttt{tasks/}) are exercised end-to-end in the test suite. \\
IEC 61508-3 Section 7.9 (software verification) & Executable docstrings; \texttt{make verify} fails if docstring examples fail. \\
ISO 26262-6 Section 8.4 (design principles, including initialisation of variables) & Fail-safe: uninitialised or NaN inputs raise an exception rather than being defaulted to zero. \\
ISO 26262-6 Section 9.4.3 (component tests) & One test file per module; pytest configuration in \texttt{pyproject.toml} in strict mode. \\
IEC 62443-4-1 SM-4 (third-party components) & Pinned \texttt{uv.lock}; blocklist of forbidden dependencies; CPE identifier; dual SPDX headers on ported \texttt{skforecast} code. \\
IEC 62443-4-1 SR-2 (threat-model-driven requirements) & Fail-safe \texttt{on\_missing} contract; tight exception handling at the network boundary. \\
IEC 62443-4-1 SVV-3 (security tests) & Executable docstrings; per-module unit tests; coverage floor \texttt{fail\_under\ =\ 68} (\texttt{make test}). \\
IEC 62443-4-1 DM-1 (management of security-related issues) & Confidential reporting channel per the security policy of the package documentation; tight, CVE-relevant exception handling; tracking via \texttt{CHANGELOG.md}. \\
IEC 62443-4-1 SUM-4 (delivery of security-related updates) & \emph{Partial:} versioned PyPI wheel via \texttt{make release}/\texttt{ship}; Conventional-Commits gate; CHANGELOG. Releases currently carry no cryptographic signature. \\
IEC 62443-4-2 SAR 3.4 (integrity of the software) & Deterministic transformations; pinned CPE; Parquet round-trip contract. \\
IEC 62443-4-2 SAR 6.1 (accessibility of the audit log) & Double-handler logger with file-based audit sink under \texttt{\~{}/spotforecast2\_safe\_models/logs/}. \\
CRA Annex~I No.~1(3)(a) (no known exploitable vulnerabilities) & CPE identifier; CycloneDX SBOM for reconciliation against the National Vulnerability Database; \texttt{grep} check of the lock files against forbidden dependencies. \\
CRA Annex~I No.~1(3)(b) (secure by default) & Fail-safe default \texttt{on\_missing="raise"}; no implicit network calls on import. \\
NIS-2 Directive Art.~21(2)(e) (security in procurement, development, maintenance) & SDL evidence, summarised in \texttt{MODEL\_CARD.md} Section 10. \\

\end{longtable}

}

\section{Example: forecasting electric load}\label{sec-example-en}

This section walks through the complete \texttt{spotforecast2-safe}
pipeline on a synthetic load series that reproduces the daily and
weekly patterns of an ENTSO-E market zone. Synthetic rather
than live data are used so that document generation remains
deterministic and offline-capable. The module
\texttt{downloader/entsoe.py} and an Open-Meteo adapter
\citep{openmeteo24} provide the corresponding live-data paths for
production deployment.

Each of the nine steps below consists of a
short explanation of what happens and why, together with a code
block that is executed live when the report is rendered. Where the
production path requires network access, an additional, non-executed
listing shows the real call. For all computations we use exclusively
the interface of the \texttt{spotforecast2-safe} library. Training
runs through the same configuration and pipeline entry point
(\texttt{ConfigMulti}, \texttt{MultiTask}) that production operation
uses, so that the actual estimator (LightGBM) never has to be
called directly. Each step closes with a sentence that assigns
it to a code or process rule and its legal basis.
Table~\ref{tbl-step-compliance-en} summarises this mapping at the
end of the section.

The figures of this report, by contrast, are produced with the
plot functions of the sister package \texttt{spotforecast2}, which
build on \texttt{matplotlib} and \texttt{plotly}, which are on the
dependency blocklist of code rule CR-4
(see Chapter~\ref{sec-code-rules-en}), so plotting functionality is
excluded from \texttt{spotforecast2-safe}. The documentation and
evaluation layer to which this report belongs is not part of the safe
core and may use the sister package. \texttt{spotforecast2-safe}
itself remains free of plotting code, and the blocklist check of CR-4
ensures that this remains so.

\subsection{Step 1: turn on the audit log}\label{schritt-1-audit-protokoll-einschalten-en}

The library provides a structured audit sink for record-keeping. A
subsequent supervisory authority can use those records to
reconstruct which data were processed when. If the caller passes a
log directory \texttt{log\_dir}, the library writes, in addition to
the console output, a file in which every line is a JSON object
conforming to a versioned schema (mandatory fields include
\texttt{schema\_version}, \texttt{prev\_hash},
\texttt{timestamp\_utc}, and \texttt{event}). The records are linked
into a hash chain via the \texttt{prev\_hash} field, whose
verification Step 9 demonstrates. Without that argument it logs to
the console only.
The file sink is pinned to the INFO level so that the audit trail
remains complete even when the operator quietens the console. In
production the directory defaults to
\texttt{\textasciitilde{}/spotforecast2\_safe\_models/logs/}. Here we
use a local working directory that is emptied at the start so that
every render begins from scratch. The library deliberately has no
built-in rotation or retention period: retaining the logs remains an
operator duty (Art.~18 of the EU AI Act,
Chapter~\ref{sec-ki-vo-18-en}). Domain events are contributed by the
caller through the \texttt{extra} argument. We mark the start of the
tutorial run with an event of our own and, as a check, read the
first entry of the file back. This practice instantiates the
record-keeping obligation of Art.~12 of the EU AI Act
(Chapter~\ref{sec-ki-vo-12-en}) and process rule PR-4. The logger's
module docstring additionally names IEC 62443-4-2 SAR 6.1 and
IEC 61508-3 §7.4.7 as the applicable standard requirements.

\begin{Shaded}
\begin{Highlighting}[]
\ImportTok{import}\NormalTok{ json}
\ImportTok{import}\NormalTok{ logging}
\ImportTok{import}\NormalTok{ shutil}
\ImportTok{from}\NormalTok{ pathlib }\ImportTok{import}\NormalTok{ Path}
\ImportTok{from}\NormalTok{ spotforecast2\_safe.manager.logger }\ImportTok{import}\NormalTok{ setup\_logging}
\NormalTok{artifacts }\OperatorTok{=}\NormalTok{ Path(}\StringTok{"\_paper\_artifacts"}\NormalTok{)}
\NormalTok{shutil.rmtree(artifacts, ignore\_errors}\OperatorTok{=}\VariableTok{True}\NormalTok{)}
\NormalTok{logger, log\_path }\OperatorTok{=}\NormalTok{ setup\_logging(level}\OperatorTok{=}\NormalTok{logging.INFO, log\_dir}\OperatorTok{=}\NormalTok{artifacts }\OperatorTok{/} \StringTok{"logs"}\NormalTok{)}
\NormalTok{logger.info(}
    \StringTok{"Tutorial run starting."}\NormalTok{,}
\NormalTok{    extra}\OperatorTok{=}\NormalTok{\{}\StringTok{"event"}\NormalTok{: }\StringTok{"tutorial\_start"}\NormalTok{, }\StringTok{"task"}\NormalTok{: }\StringTok{"lazy"}\NormalTok{,}
           \StringTok{"context"}\NormalTok{: \{}\StringTok{"seed"}\NormalTok{: }\DecValTok{2026}\NormalTok{, }\StringTok{"horizon\_h"}\NormalTok{: }\DecValTok{24}\NormalTok{\}\},}
\NormalTok{)}
\NormalTok{record }\OperatorTok{=}\NormalTok{ json.loads(log\_path.read\_text(encoding}\OperatorTok{=}\StringTok{"utf{-}8"}\NormalTok{).splitlines()[}\DecValTok{0}\NormalTok{])}
\BuiltInTok{print}\NormalTok{(}\SpecialStringTok{f"Audit file:     }\SpecialCharTok{\{}\NormalTok{log\_path}\SpecialCharTok{\}}\SpecialStringTok{"}\NormalTok{)}
\BuiltInTok{print}\NormalTok{(}\SpecialStringTok{f"Schema version: }\SpecialCharTok{\{}\NormalTok{record[}\StringTok{\textquotesingle{}schema\_version\textquotesingle{}}\NormalTok{]}\SpecialCharTok{\}}\SpecialStringTok{, Event: }\SpecialCharTok{\{}\NormalTok{record[}\StringTok{\textquotesingle{}event\textquotesingle{}}\NormalTok{]}\SpecialCharTok{\}}\SpecialStringTok{"}\NormalTok{)}
\end{Highlighting}
\end{Shaded}

\begin{verbatim}
2026-08-13 18:58:13,818 - sf2-safe-logger - INFO - Persistent logging initialized at: _paper_artifacts/logs/sf2-safe-logger_20260813_185813.log
2026-08-13 18:58:13,818 - sf2-safe-logger - INFO - Tutorial run starting.
Audit file:     _paper_artifacts/logs/sf2-safe-logger_20260813_185813.log
Schema version: 2.0.0, Event: audit_log_init
\end{verbatim}

\subsection{Step 2: the data basis}\label{schritt-2-datengrundlage-en}

The \texttt{spotforecast2-safe} forecasting process does not generate
data, it retrieves them. The
target load comes as aggregated actual load from the ENTSO-E
transparency platform. The weather covariates come from the
Open-Meteo archive and from the
Open-Meteo weather forecast: over the horizon, covariates are
themselves forecasts, not measurements. The following listing shows
the production retrieval path\footnote{It is deliberately not executed
  at render time because both calls require network access and, for
  ENTSO-E, an API key.}:

\begin{Shaded}
\begin{Highlighting}[]
\ImportTok{import}\NormalTok{ os}
\ImportTok{from}\NormalTok{ spotforecast2\_safe.downloader.entsoe }\ImportTok{import}\NormalTok{ download\_new\_data}
\ImportTok{from}\NormalTok{ spotforecast2\_safe.data.fetch\_data }\ImportTok{import}\NormalTok{ fetch\_weather\_data}
\NormalTok{download\_new\_data(api\_key}\OperatorTok{=}\NormalTok{os.environ[}\StringTok{"ENTSOE\_API\_KEY"}\NormalTok{], country\_code}\OperatorTok{=}\StringTok{"FR"}\NormalTok{)}
\NormalTok{weather }\OperatorTok{=}\NormalTok{ fetch\_weather\_data(cov\_start}\OperatorTok{=}\StringTok{"2025{-}01{-}01"}\NormalTok{, cov\_end}\OperatorTok{=}\StringTok{"2025{-}03{-}31"}\NormalTok{)}
\end{Highlighting}
\end{Shaded}

For the report we replace this retrieval path with a synthetic
image: a load series as the sum of four simple components, namely a
slight upward trend (e.g., rising power consumption across the
quarter), a daily cycle peaking at midday, a weekday surcharge (load
is higher on weekdays than at weekends), and normally distributed
measurement noise. The series comes from the package function
\texttt{make\_synthetic\_load} and is generated with a fixed seed
(starting value) so that it is bit-identical at every render and the
report is produced offline. The index is a contiguous UTC hourly
sequence and thus exactly the data format that
\texttt{spotforecast2-safe} expects. The documented choice of the
data basis operationalises the data governance of Art.~10 of the
EU AI Act (Chapter~\ref{sec-ki-vo-10-en}). The fixed seed implements
code rule CR-2 \emph{Determinism}.

\begin{Shaded}
\begin{Highlighting}[]
\ImportTok{import}\NormalTok{ numpy }\ImportTok{as}\NormalTok{ np}
\ImportTok{import}\NormalTok{ pandas }\ImportTok{as}\NormalTok{ pd}
\ImportTok{from}\NormalTok{ spotforecast2\_safe.data }\ImportTok{import}\NormalTok{ make\_synthetic\_load}
\NormalTok{y }\OperatorTok{=}\NormalTok{ make\_synthetic\_load(seed}\OperatorTok{=}\DecValTok{2026}\NormalTok{)}
\NormalTok{y.head(}\DecValTok{3}\NormalTok{)}
\end{Highlighting}
\end{Shaded}

\begin{verbatim}
2025-01-01 00:00:00+00:00    47.103439
2025-01-01 01:00:00+00:00    47.757509
2025-01-01 02:00:00+00:00    47.089588
Freq: h, Name: load, dtype: float64
\end{verbatim}

\subsection{Step 3: handle data gaps explicitly}\label{schritt-3-datenluxfccken-explizit-behandeln-en}

Real sensor data occasionally contain gaps, for instance when a
meter restarts or a transmission fails. A safety-critical library
must not paper over such gaps silently. \texttt{spotforecast2-safe}
therefore refuses to fit on an incomplete target series: the input
check \texttt{check\_y}, which precedes every training run, rejects
missing values with an explicit exception
(Chapter~\ref{sec-fail-safe-nan-en}, CR-3 \emph{Fail-safe}). We
simulate a measurement outage on the morning of
14~February~2025 and continue working with this series:

\begin{Shaded}
\begin{Highlighting}[]
\ImportTok{from}\NormalTok{ spotforecast2\_safe.preprocessing.checking }\ImportTok{import}\NormalTok{ check\_y}
\NormalTok{y\_gapped }\OperatorTok{=}\NormalTok{ y.copy()}
\NormalTok{y\_gapped.loc[}\StringTok{"2025{-}02{-}14 09:00"}\NormalTok{:}\StringTok{"2025{-}02{-}14 14:00"}\NormalTok{] }\OperatorTok{=}\NormalTok{ np.nan}
\ControlFlowTok{try}\NormalTok{:}
\NormalTok{    check\_y(y\_gapped)}
\ControlFlowTok{except} \PreprocessorTok{ValueError} \ImportTok{as}\NormalTok{ err:}
    \BuiltInTok{print}\NormalTok{(}\SpecialStringTok{f"ValueError: }\SpecialCharTok{\{}\NormalTok{err}\SpecialCharTok{\}}\SpecialStringTok{"}\NormalTok{)}
\end{Highlighting}
\end{Shaded}

\begin{verbatim}
ValueError: `y` has missing values.
\end{verbatim}

The abort forces the deployer to take an explicit decision on how
the gap is to be treated. Here \texttt{spotforecast2-safe} differs
fundamentally from the standard imputation offered by tools such as
imputeTS \citep{mori17a}, mice \citep{buur11a}, or the
\texttt{SimpleImputer} of scikit-learn \citep{pedr11a}.
\citet{mori15a} compare such methods for univariate time series.
These tools return only the imputed series: the inserted values are
invented quantities that were never measured, indistinguishable from
real measurements afterwards, and they enter model training as
apparent observations. For this reason the statistical literature
treats missing values as an inference problem in its own right,
whose treatment depends on the missingness mechanism and must be
documented \citep{rubi76a, littl19a}.

\texttt{spotforecast2-safe}, by contrast, separates imputing from
using. The function \texttt{get\_missing\_weights} does fill the
gap, because feature construction requires a gap-free time index. In
addition, however, it returns a weight series that is zero inside
the gap and for \texttt{window\_size} subsequent rows, because the
lag features of those rows would otherwise ingest filled values.
This series enters training as a sample weight: an imputed hour
carries no training signal but remains countable in the audit log.
Exactly this mechanism also runs automatically in the pipeline of
Step 5: the default imputation method \texttt{"weighted"} calls
\texttt{get\_missing\_weights} internally and passes the weight
series to the forecaster as \texttt{weight\_func}. We demonstrate
the mechanism up front on the series with the outage. With the
168 lags of the model of Step 5, the gap additionally
devalues the 168 subsequent hours:

\begin{Shaded}
\begin{Highlighting}[]
\ImportTok{from}\NormalTok{ spotforecast2\_safe.preprocessing }\ImportTok{import}\NormalTok{ get\_missing\_weights}
\NormalTok{y\_gapped\_filled, gap\_weights }\OperatorTok{=}\NormalTok{ get\_missing\_weights(}
\NormalTok{    y\_gapped.to\_frame(), window\_size}\OperatorTok{=}\DecValTok{168}
\NormalTok{)}
\BuiltInTok{print}\NormalTok{(}\SpecialStringTok{f"Missing hours:         }\SpecialCharTok{\{}\BuiltInTok{int}\NormalTok{(y\_gapped.isna().}\BuiltInTok{sum}\NormalTok{())}\SpecialCharTok{\}}\SpecialStringTok{"}\NormalTok{)}
\BuiltInTok{print}\NormalTok{(}\SpecialStringTok{f"Hours with weight 0:   }\SpecialCharTok{\{}\BuiltInTok{int}\NormalTok{((gap\_weights }\OperatorTok{==} \DecValTok{0}\NormalTok{).}\BuiltInTok{sum}\NormalTok{())}\SpecialCharTok{\}}\SpecialStringTok{ "}
      \SpecialStringTok{f"(gap plus 168 subsequent hours)"}\NormalTok{)}
\end{Highlighting}
\end{Shaded}

\begin{verbatim}
Missing hours:         6
Hours with weight 0:   174 (gap plus 168 subsequent hours)
\end{verbatim}

\begin{figure}

\centering{

\pandocbounded{\includegraphics[keepaspectratio]{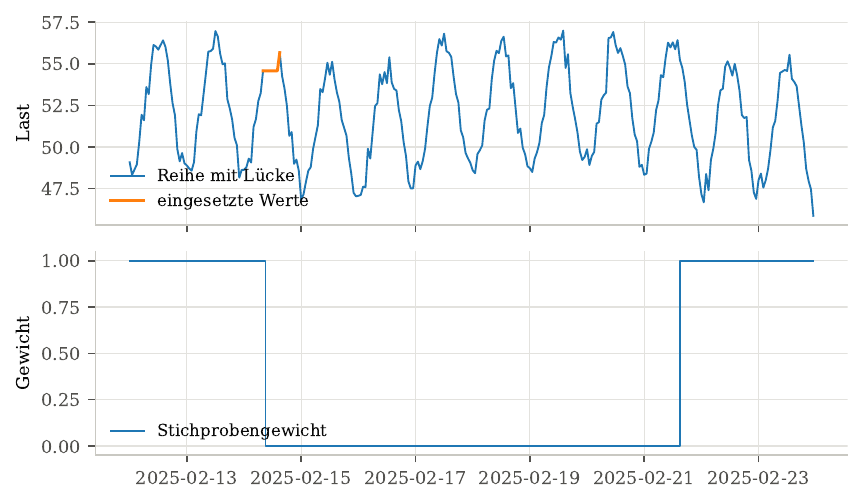}}

}

\caption{\label{fig-imputation-weights-en}Explicit gap treatment with
\texttt{get\_missing\_weights} on the simulated six-hour measurement
outage (14 February 2025, 09:00 to 14:00 UTC). Top panel: series with
the gap (\emph{blue}) and
the values inserted by the fill (\emph{orange}). Bottom panel: companion
weight series; the sample weight is zero inside the gap and for the
168 subsequent hours (the lag window of the model of Step 5), so
that inserted values carry no training signal.}

\end{figure}

Figure~\ref{fig-imputation-weights-en} shows both return values: at
the top the series with the gap and the inserted values, below the
companion weight series, whose zero zone covers the gap and the
subsequent lag window. The six-hour outage thus devalues a good week
of training data. Standard imputation conceals exactly this
consequence.

No separate up-front imputation is needed after this, and the step
therefore deliberately ends without a further processing call: the
gap remains in the series, and the pipeline of Step 5 handles it
automatically through its default imputation method, following
exactly the mechanism just shown. The deployer's decision consists
in the documented choice of this method, not in inventing values.
This discipline operationalises Art.~10 of the EU AI Act (data
governance, Chapter~\ref{sec-ki-vo-10-en}) and CR-3
\emph{Fail-safe}.

\subsection{Step 4: fix the forecast
origin}\label{schritt-4-prognoseursprung-festlegen-en}

For time series, the split into training and evaluation data must
not be random, because future information would otherwise leak into
training. In production, moreover, there is no static train-test
split at all: the model is retrained daily on a rolling window of
recent history (in \texttt{ConfigMulti}, \texttt{train\_size}
governs this span, with a default of three years), and the ``test
set'' is always tomorrow's target day. The tutorial therefore
freezes a single forecast origin reproducibly: training runs
chronologically on the first two months, and March is held back as
the target days for the evaluation in Steps 7 and 8. The
measurement outage of Step 3 thus lies within the training range.
This chronological split is the simplest variant of the
representativeness check required by Art.~10(2)(f) and (g) of the
EU AI Act (Chapter~\ref{sec-ki-vo-10-en}).

\begin{Shaded}
\begin{Highlighting}[]
\NormalTok{y\_train }\OperatorTok{=}\NormalTok{ y\_gapped.loc[}\StringTok{"2025{-}01{-}01"}\NormalTok{:}\StringTok{"2025{-}03{-}01"}\NormalTok{]}
\NormalTok{y\_eval  }\OperatorTok{=}\NormalTok{ y\_gapped.loc[}\StringTok{"2025{-}03{-}02"}\NormalTok{:]}
\BuiltInTok{print}\NormalTok{(}\SpecialStringTok{f"Training range:   }\SpecialCharTok{\{}\NormalTok{y\_train}\SpecialCharTok{.}\NormalTok{index[}\DecValTok{0}\NormalTok{]}\SpecialCharTok{.}\NormalTok{date()}\SpecialCharTok{\}}\SpecialStringTok{ to }\SpecialCharTok{\{}\NormalTok{y\_train}\SpecialCharTok{.}\NormalTok{index[}\OperatorTok{{-}}\DecValTok{1}\NormalTok{]}\SpecialCharTok{.}\NormalTok{date()}\SpecialCharTok{\}}\SpecialStringTok{ "}
      \SpecialStringTok{f"(}\SpecialCharTok{\{}\BuiltInTok{len}\NormalTok{(y\_train)}\SpecialCharTok{\}}\SpecialStringTok{ hours)"}\NormalTok{)}
\BuiltInTok{print}\NormalTok{(}\SpecialStringTok{f"Evaluation range: }\SpecialCharTok{\{}\NormalTok{y\_eval}\SpecialCharTok{.}\NormalTok{index[}\DecValTok{0}\NormalTok{]}\SpecialCharTok{.}\NormalTok{date()}\SpecialCharTok{\}}\SpecialStringTok{ to }\SpecialCharTok{\{}\NormalTok{y\_eval}\SpecialCharTok{.}\NormalTok{index[}\OperatorTok{{-}}\DecValTok{1}\NormalTok{]}\SpecialCharTok{.}\NormalTok{date()}\SpecialCharTok{\}}\SpecialStringTok{ "}
      \SpecialStringTok{f"(}\SpecialCharTok{\{}\BuiltInTok{len}\NormalTok{(y\_eval)}\SpecialCharTok{\}}\SpecialStringTok{ hours)"}\NormalTok{)}
\end{Highlighting}
\end{Shaded}

\begin{verbatim}
Training range:   2025-01-01 to 2025-03-01 (1440 hours)
Evaluation range: 2025-03-02 to 2025-03-31 (720 hours)
\end{verbatim}

\subsection{Step 5: configure and train the
pipeline}\label{schritt-5-pipeline-konfigurieren-und-trainieren-en}

In production, training is not wired up by hand but started through
the same configuration entry point that daily operation uses: a
\texttt{ConfigMulti} object bundles all decisions (horizon, feature
window, seed, cache directory), and the class \texttt{MultiTask}
runs the preparation, checking, and training stages on it. For
one-call use, \texttt{multitask.runner.run(...)} additionally wraps
the same chain in a single function. We configure only two domain
quantities: the forecast horizon of 24 hours and the feature window
of 168 hours. \emph{Lags} are the past values that the model uses as
input: with \texttt{lags=168} the model sees, for every hour, the
last full week as context, which is decisive for the weekly
seasonality (weekday versus weekend). ``Recursive'' means that the
forecast for one hour serves in turn as input for the next hour.
Everything else is a library default: the LightGBM estimator with a
fixed seed and pinned parallelism (\texttt{n\_jobs=1}), the outlier
detection, whose findings enter the pipeline log purely as records
and alter no values, and the weighted imputation of Step 3, which
now automatically fills the outage left in place there and keeps it
out of training via the weight series. A
hyperparameter search deliberately does not take place. The daily
optimisation of production operation remains outside this tutorial.
This step implements CR-2 (seeded estimator, \texttt{n\_jobs=1}) and
addresses the accuracy and robustness requirement of Art.~15 of the
EU AI Act (Chapter~\ref{sec-ki-vo-15-en}).

\begin{Shaded}
\begin{Highlighting}[]
\ImportTok{from}\NormalTok{ spotforecast2\_safe.configurator }\ImportTok{import}\NormalTok{ ConfigMulti}
\ImportTok{from}\NormalTok{ spotforecast2\_safe.multitask }\ImportTok{import}\NormalTok{ MultiTask}
\NormalTok{cfg }\OperatorTok{=}\NormalTok{ ConfigMulti(}
\NormalTok{    predict\_size}\OperatorTok{=}\DecValTok{24}\NormalTok{,}
\NormalTok{    lags\_consider}\OperatorTok{=}\NormalTok{[}\DecValTok{168}\NormalTok{],}
\NormalTok{    window\_size}\OperatorTok{=}\DecValTok{168}\NormalTok{,}
\NormalTok{    use\_exogenous\_features}\OperatorTok{=}\VariableTok{False}\NormalTok{,}
\NormalTok{    use\_outlier\_detection}\OperatorTok{=}\VariableTok{True}\NormalTok{,}
\NormalTok{    auto\_save\_models}\OperatorTok{=}\VariableTok{False}\NormalTok{,}
\NormalTok{    data\_frame\_name}\OperatorTok{=}\StringTok{"synthetic\_load"}\NormalTok{,}
\NormalTok{    cache\_home}\OperatorTok{=}\NormalTok{artifacts }\OperatorTok{/} \StringTok{"cache"}\NormalTok{,}
\NormalTok{)}
\NormalTok{mt }\OperatorTok{=}\NormalTok{ MultiTask(cfg, task}\OperatorTok{=}\StringTok{"lazy"}\NormalTok{,}
\NormalTok{               dataframe}\OperatorTok{=}\NormalTok{y\_train.to\_frame(), data\_test}\OperatorTok{=}\NormalTok{y\_eval.to\_frame())}
\NormalTok{mt.prepare\_data().detect\_outliers().impute().build\_exogenous\_features()}
\NormalTok{result }\OperatorTok{=}\NormalTok{ mt.run(task}\OperatorTok{=}\StringTok{"lazy"}\NormalTok{)}
\NormalTok{fc }\OperatorTok{=}\NormalTok{ mt.results[}\StringTok{"lazy"}\NormalTok{][}\StringTok{"load"}\NormalTok{][}\StringTok{"forecaster"}\NormalTok{]}
\BuiltInTok{print}\NormalTok{(}\SpecialStringTok{f"Training window:  }\SpecialCharTok{\{}\NormalTok{fc}\SpecialCharTok{.}\NormalTok{training\_range\_[}\DecValTok{0}\NormalTok{]}\SpecialCharTok{.}\NormalTok{date()}\SpecialCharTok{\}}\SpecialStringTok{ to "}
      \SpecialStringTok{f"}\SpecialCharTok{\{}\NormalTok{fc}\SpecialCharTok{.}\NormalTok{training\_range\_[}\DecValTok{1}\NormalTok{]}\SpecialCharTok{.}\NormalTok{date()}\SpecialCharTok{\}}\SpecialStringTok{"}\NormalTok{)}
\BuiltInTok{print}\NormalTok{(}\SpecialStringTok{f"Estimator:        }\SpecialCharTok{\{}\BuiltInTok{type}\NormalTok{(fc.estimator)}\SpecialCharTok{.}\VariableTok{\_\_name\_\_}\SpecialCharTok{\}}\SpecialStringTok{, "}
      \SpecialStringTok{f"Lags: }\SpecialCharTok{\{}\NormalTok{fc}\SpecialCharTok{.}\NormalTok{lags}\SpecialCharTok{.}\NormalTok{size}\SpecialCharTok{\}}\SpecialStringTok{, Seed: }\SpecialCharTok{\{}\NormalTok{cfg}\SpecialCharTok{.}\NormalTok{random\_state}\SpecialCharTok{\}}\SpecialStringTok{"}\NormalTok{)}
\BuiltInTok{print}\NormalTok{(}\SpecialStringTok{f"Gap weights:      }\SpecialCharTok{\{}\StringTok{\textquotesingle{}passed\textquotesingle{}} \ControlFlowTok{if}\NormalTok{ mt}\SpecialCharTok{.}\NormalTok{weight\_func }\KeywordTok{is} \KeywordTok{not} \VariableTok{None} \ControlFlowTok{else} \StringTok{\textquotesingle{}none\textquotesingle{}}\SpecialCharTok{\}}\SpecialStringTok{"}\NormalTok{)}
\end{Highlighting}
\end{Shaded}

\begin{verbatim}
Training window:  2025-01-01 to 2025-03-01
Estimator:        LGBMRegressor, Lags: 168, Seed: 314159
Gap weights:      passed
\end{verbatim}

With \texttt{use\_exogenous\_features=False} the run deliberately
stays target-only and thus offline-capable. In production the same
pipeline stage \texttt{build\_exogenous\_features} constructs the
covariates inside the library, namely weather from Open-Meteo
together with calendar, holiday, and solar features including their
cyclical encoding. Nobody writes feature code by hand. The
following, non-executed listing shows the corresponding
configuration switches:

\begin{Shaded}
\begin{Highlighting}[]
\NormalTok{cfg\_prod }\OperatorTok{=}\NormalTok{ ConfigMulti(}
\NormalTok{    use\_exogenous\_features}\OperatorTok{=}\VariableTok{True}\NormalTok{,      }\CommentTok{\# weather (Open{-}Meteo) and calendar}
\NormalTok{    include\_holiday\_features}\OperatorTok{=}\VariableTok{True}\NormalTok{,    }\CommentTok{\# public{-}holiday calendar}
\NormalTok{    include\_ephemeris\_features}\OperatorTok{=}\VariableTok{True}\NormalTok{,  }\CommentTok{\# solar position, daylight duration}
\NormalTok{    on\_weather\_failure}\OperatorTok{=}\StringTok{"raise"}\NormalTok{,       }\CommentTok{\# CR{-}3: no silent continuation}
\NormalTok{)}
\end{Highlighting}
\end{Shaded}

\subsection{Step 6: produce a
forecast}\label{schritt-6-vorhersage-erzeugen-en}

We ask the trained forecaster for a day-ahead forecast, the first
24 hours of the evaluation period. Twenty-four hours is a common
day-ahead forecast horizon in load forecasting, because grid
operators typically have to schedule one day in advance. The model
starts with the last training hours as the initial lag window and
shifts this window recursively, hour by hour. The pipeline has
already produced the same forecast as part of its result package
(\texttt{future\_pred}, together with the in-sample fit, metrics,
and model identity): a deployer never receives a bare series of
numbers but an interpretable package, which is the information basis
that Art.~13 of the EU AI Act (Chapter~\ref{sec-ki-vo-13-en})
requires for transparency towards the operator. Comparing the two
calls at the same time shows the determinism of CR-2 in action: same
state, same output.

\begin{Shaded}
\begin{Highlighting}[]
\NormalTok{y\_eval\_24h }\OperatorTok{=}\NormalTok{ y\_eval.iloc[:}\DecValTok{24}\NormalTok{]}
\NormalTok{y\_pred }\OperatorTok{=}\NormalTok{ fc.predict(steps}\OperatorTok{=}\DecValTok{24}\NormalTok{)}
\NormalTok{y\_pred.name }\OperatorTok{=} \StringTok{"forecast"}
\BuiltInTok{print}\NormalTok{(y\_pred.head(}\DecValTok{3}\NormalTok{))}
\BuiltInTok{print}\NormalTok{(}\SpecialStringTok{f"Identical to future\_pred from Step 5: "}
      \SpecialStringTok{f"}\SpecialCharTok{\{}\BuiltInTok{bool}\NormalTok{((y\_pred }\OperatorTok{==}\NormalTok{ result[}\StringTok{\textquotesingle{}future\_pred\textquotesingle{}}\NormalTok{]).}\BuiltInTok{all}\NormalTok{())}\SpecialCharTok{\}}\SpecialStringTok{"}\NormalTok{)}
\end{Highlighting}
\end{Shaded}

\begin{verbatim}
2025-03-02 00:00:00+00:00    48.093535
2025-03-02 01:00:00+00:00    48.347808
2025-03-02 02:00:00+00:00    47.748086
Freq: h, Name: forecast, dtype: float64
Identical to future_pred from Step 5: True
\end{verbatim}

\subsection{Step 7: measure accuracy}\label{schritt-7-genauigkeit-messen-en}

\begin{figure}

\centering{

\pandocbounded{\includegraphics[keepaspectratio]{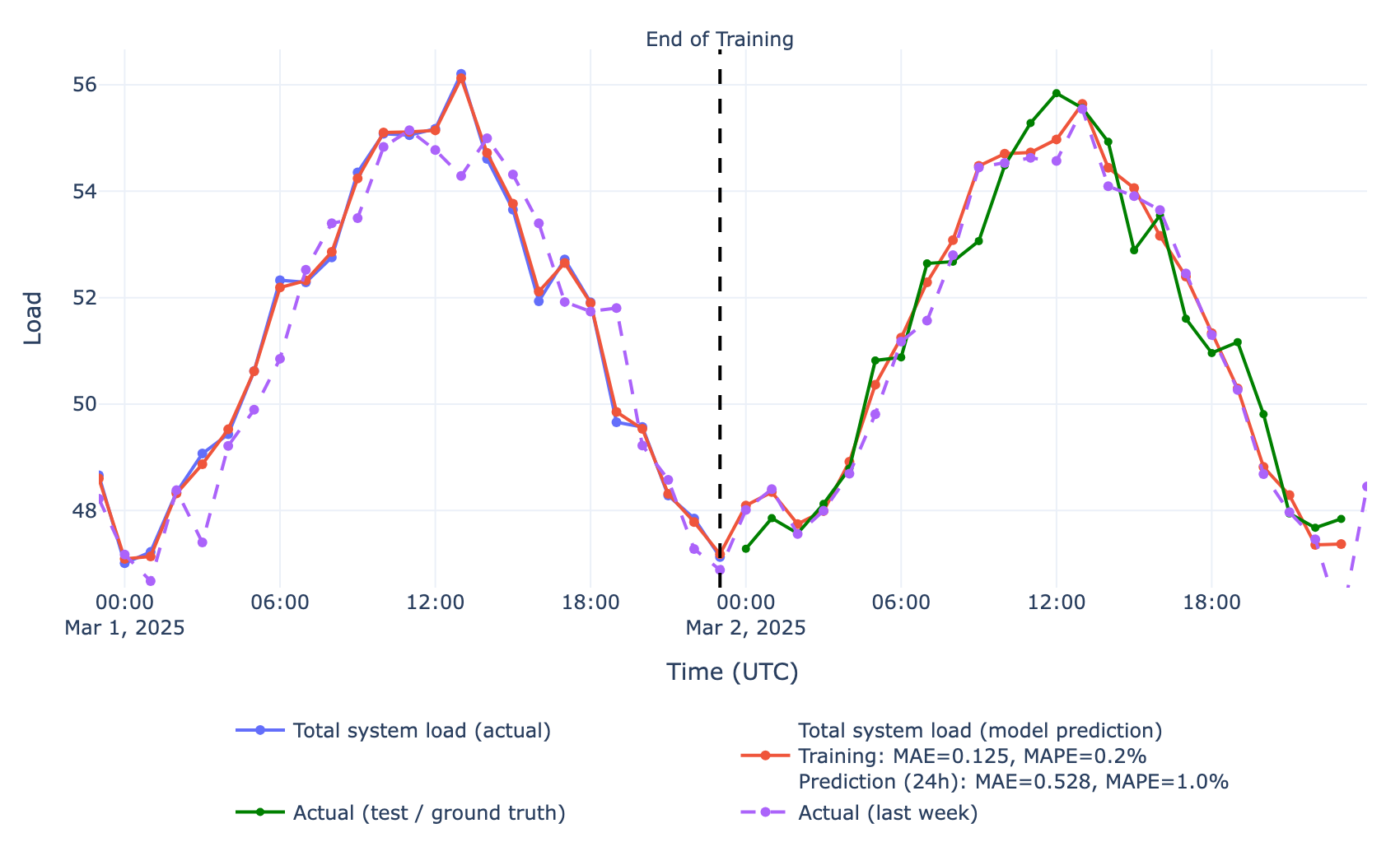}}

}

\caption{\label{fig-forecast24-en}24-hour load forecast as the
unchanged report figure of the production system, generated with
\texttt{make\_plot} from the result package of the pipeline of
Step 5. The window shows the last 24 training hours and the
24-hour horizon, the \emph{dashed} vertical line (End of Training) marks
the end of training. Total system load (actual): observed load up
to the end of training. Total system load (model prediction):
in-sample fit and subsequent 24 h forecast. Actual (test / ground
truth): the actual load that becomes available only after the
target day and against which the day-after scoring computes.
Actual (last week): previous week's load as seasonality context.
The measures in the legend come unchanged from the result package
(\texttt{metrics\_train} and \texttt{metrics\_future}), MAE in
load units, MAPE as a percentage.}

\end{figure}

Three standard measures quantify the deviation between forecast and
the actually observed load. The mean absolute error (MAE) gives the
average deviation in the units of the load series. The root mean squared error (RMSE)
penalises large deviations disproportionately and is likewise scaled
in load units. The mean absolute percentage error (MAPE) gives the
mean percentage deviation. The package function
\texttt{score\_forecasts} returns these measures as a tidy table
with one row per approach and the number of evaluated hours
(\texttt{n}). The legend of Figure~\ref{fig-forecast24-en}, by
contrast, recomputes nothing but takes its measures unchanged from
the result package of the pipeline. In live operation this
step is time-shifted: scoring happens only once the actual load has
been published, that is, on the following day, and subsequent
revisions of the actuals lead to re-scoring over a window of several
weeks. For this per-day evaluation the library provides
\texttt{score\_forecasts\_by\_period} and
\texttt{aggregate\_period\_scores}. Here \texttt{score\_forecasts}
over the frozen target day suffices, complemented by the scoring
that the pipeline itself already supplies (at a 24-hour horizon the
day-based \texttt{metrics\_future\_one\_day} coincides with
\texttt{metrics\_future}). The measures serve the
accuracy requirement of Art.~15 of the EU AI Act
(Chapter~\ref{sec-ki-vo-15-en}). The daily re-evaluation against
published actuals amounts to post-market monitoring under Art.~72 of
the EU AI Act (Chapter~\ref{sec-ki-vo-72-en}) in daily practice.

\begin{Shaded}
\begin{Highlighting}[]
\ImportTok{from}\NormalTok{ spotforecast2\_safe.processing }\ImportTok{import}\NormalTok{ score\_forecasts}
\NormalTok{scores }\OperatorTok{=}\NormalTok{ score\_forecasts(}
\NormalTok{    \{}\StringTok{"Model forecast"}\NormalTok{: y\_pred\}, y\_eval\_24h, metrics}\OperatorTok{=}\NormalTok{(}\StringTok{"mae"}\NormalTok{, }\StringTok{"rmse"}\NormalTok{, }\StringTok{"mape"}\NormalTok{)}
\NormalTok{)}
\BuiltInTok{print}\NormalTok{(scores.}\BuiltInTok{round}\NormalTok{(}\DecValTok{3}\NormalTok{))}
\BuiltInTok{print}\NormalTok{(}\SpecialStringTok{f"Pipeline{-}provided scoring (metrics\_future): "}
      \SpecialStringTok{f"MAE }\SpecialCharTok{\{}\NormalTok{result[}\StringTok{\textquotesingle{}metrics\_future\textquotesingle{}}\NormalTok{][}\StringTok{\textquotesingle{}mae\textquotesingle{}}\NormalTok{]}\SpecialCharTok{:.3f\}}\SpecialStringTok{"}\NormalTok{)}
\end{Highlighting}
\end{Shaded}

\begin{verbatim}
                  mae   rmse   mape   n
Model forecast  0.528  0.626  1.026  24
Pipeline-provided scoring (metrics_future): MAE 0.528
\end{verbatim}

Figure~\ref{fig-forecast24-en} is not an evaluation built
specifically for this report but the report figure of the
production system: \texttt{make\_plot} from
\texttt{spotforecast2.plots} generates it directly from the result
package of Step 5, and the document recomputes not a single measure
for it. The package idea described in Step 6 is thereby delivered
visually, because training fit, 24 h forecast, metrics, and
comparison context appear together in one depiction (Art.~13 of the
EU AI Act, Chapter~\ref{sec-ki-vo-13-en}). Structurally the figure
follows \citet{chag24a}. The dashed trace Actual (last week) shows
the previous week's load as unscored seasonality context, the green
trace Actual (test / ground truth) the actual load published only
after the target day. The labels stay unchanged because the same
figure is delivered as a report in operation.

\subsection{Step 8: rolling-origin
backtest}\label{schritt-8-rolling-origin-backtest-en}

A single train-test split tests the model's quality at only one
point of the time axis. A \emph{rolling-origin backtest} shifts
the split point step by step through the time series and runs a
24-hour forecast for every step. From this, an entire distribution
of accuracy measurements arises that is more robust against
individual outliers.

\begin{figure}

\centering{

\pandocbounded{\includegraphics[keepaspectratio]{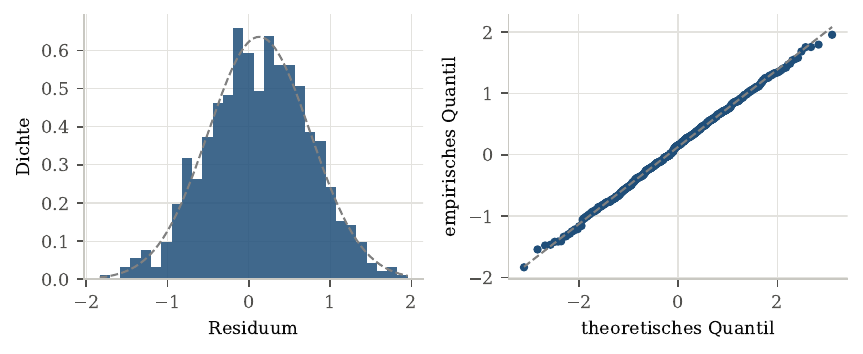}}

}

\caption{\label{fig-residuals-en}Residuals of the 24-hour forecasts
from the rolling-origin backtest in Step 8. Left: histogram with an
overlaid normal density, fitted to the empirical moments. Right:
sorted residuals against standard-normal quantiles (QQ diagram).
The concentration of points around the diagonal speaks for an
approximately normally distributed error structure.}

\end{figure}

In the real system this check does not run
inside operational scoring but inside the daily model selection:
there, a rolling-origin cross-validation on the training window
forms the inner objective of the hyperparameter search. Every
origin uses only data that precede it in time. The tutorial shows it
as a validation instrument over the entire evaluation range. The
input is the training series as filled by the pipeline, extended by
the unchanged evaluation data, because the backtest requires a
gap-free series. The forecaster carries its weight function from
Step 5 with it, so that the filled hours receive no training signal
even when the first fold is refitted. The check
thereby supports the risk management of Art.~9 of the EU AI Act
(Chapter~\ref{sec-ki-vo-9-en}) and, empirically, the
representativeness check required by Art.~10(3) of the EU AI Act
(Chapter~\ref{sec-ki-vo-10-en}).
\texttt{backtesting\_forecaster} returns two objects: the error
aggregated over all folds (\texttt{metrics\_df}, a single row) and
the per-fold forecasts (\texttt{predictions\_df} with a
\texttt{fold} column). The helper function
\texttt{metrics\_per\_fold} reconstructs the per-fold error
distribution from the latter. The forecasts collected in
this way are also the data basis for the residual analysis in
Figure~\ref{fig-residuals-en}.

\begin{Shaded}
\begin{Highlighting}[]
\ImportTok{from}\NormalTok{ spotforecast2\_safe.splitter }\ImportTok{import}\NormalTok{ TimeSeriesFold}
\ImportTok{from}\NormalTok{ spotforecast2\_safe.backtesting }\ImportTok{import}\NormalTok{ backtesting\_forecaster, metrics\_per\_fold}
\NormalTok{y\_bt }\OperatorTok{=}\NormalTok{ pd.concat([mt.df\_pipeline[}\StringTok{"load"}\NormalTok{], y\_eval])}
\NormalTok{cv }\OperatorTok{=}\NormalTok{ TimeSeriesFold(}
\NormalTok{    steps}\OperatorTok{=}\DecValTok{24}\NormalTok{,}
\NormalTok{    initial\_train\_size}\OperatorTok{=}\BuiltInTok{len}\NormalTok{(y\_train),}
\NormalTok{    refit}\OperatorTok{=}\VariableTok{False}\NormalTok{,}
\NormalTok{    fixed\_train\_size}\OperatorTok{=}\VariableTok{False}\NormalTok{,}
\NormalTok{    verbose}\OperatorTok{=}\VariableTok{False}\NormalTok{,}
\NormalTok{)}
\NormalTok{metrics\_df, predictions\_df }\OperatorTok{=}\NormalTok{ backtesting\_forecaster(}
\NormalTok{    forecaster}\OperatorTok{=}\NormalTok{fc,}
\NormalTok{    y}\OperatorTok{=}\NormalTok{y\_bt,}
\NormalTok{    cv}\OperatorTok{=}\NormalTok{cv,}
\NormalTok{    metric}\OperatorTok{=}\StringTok{"mean\_absolute\_error"}\NormalTok{,}
\NormalTok{    show\_progress}\OperatorTok{=}\VariableTok{False}\NormalTok{,}
\NormalTok{    verbose}\OperatorTok{=}\VariableTok{False}\NormalTok{,}
\NormalTok{)}
\NormalTok{mae\_per\_fold }\OperatorTok{=}\NormalTok{ metrics\_per\_fold(predictions\_df, y\_bt, metric}\OperatorTok{=}\StringTok{"mean\_absolute\_error"}\NormalTok{)}
\BuiltInTok{print}\NormalTok{(}\SpecialStringTok{f"Number of folds:  }\SpecialCharTok{\{}\BuiltInTok{len}\NormalTok{(mae\_per\_fold)}\SpecialCharTok{\}}\SpecialStringTok{"}\NormalTok{)}
\BuiltInTok{print}\NormalTok{(}\SpecialStringTok{f"MAE per fold:     mean }\SpecialCharTok{\{}\NormalTok{mae\_per\_fold[}\StringTok{\textquotesingle{}mean\_absolute\_error\textquotesingle{}}\NormalTok{]}\SpecialCharTok{.}\NormalTok{mean()}\SpecialCharTok{:.3f\}}\SpecialStringTok{, "}
      \SpecialStringTok{f"range [}\SpecialCharTok{\{}\NormalTok{mae\_per\_fold[}\StringTok{\textquotesingle{}mean\_absolute\_error\textquotesingle{}}\NormalTok{]}\SpecialCharTok{.}\BuiltInTok{min}\NormalTok{()}\SpecialCharTok{:.3f\}}\SpecialStringTok{, "}
      \SpecialStringTok{f"}\SpecialCharTok{\{}\NormalTok{mae\_per\_fold[}\StringTok{\textquotesingle{}mean\_absolute\_error\textquotesingle{}}\NormalTok{]}\SpecialCharTok{.}\BuiltInTok{max}\NormalTok{()}\SpecialCharTok{:.3f\}}\SpecialStringTok{]"}\NormalTok{)}
\end{Highlighting}
\end{Shaded}

\begin{verbatim}
Number of folds:  30
MAE per fold:     mean 0.518, range [0.378, 0.760]
\end{verbatim}

\subsection{Step 9: provenance, artefacts, and
reproducibility}\label{schritt-9-provenienz-artefakte-und-reproduzierbarkeit-en}

To finish, we inspect the traces the run leaves behind. The trained
forecaster carries its provenance itself: creation and training
timestamps as well as library and Python versions are set when it is
constructed and are serialised with the model.
\texttt{save\_forecaster} stores the model under a deterministic
file name. This is why \texttt{auto\_save\_models=False} is set in
Step 5, since automatic saving would create a new, timestamped file
name per run. The CPE identifier links the artefact to the SBOM of
the release (PR-3). The audit log from Step 1 is hash-chained, and
\texttt{verify\_audit\_log} checks the chain offline and returns the
chain head, which an operator can anchor in an independently
controlled store. When it is presented as an anchor on the next
call, the report confirms that the chain still matches (see
Chapter~\ref{sec-ki-vo-12-en}). Alongside it,
the pipeline keeps a second, package-owned plain-text log under its
cache directory, in which the library stages record their findings:
data shape after preparation, outlier findings, imputation method,
and the training summary. The library offers no manifest writer and
no log rotation. These operator duties, including retention under
Art.~18 of the EU AI Act (Chapter~\ref{sec-ki-vo-18-en}), lie
outside the package. Together, the artefacts shown form the minimum
provenance information for the technical documentation required by
Art.~11 of the EU AI Act (Chapter~\ref{sec-ki-vo-11-en}) and the
record-keeping obligation of Art.~12 of the EU AI Act
(Chapter~\ref{sec-ki-vo-12-en}). PR-3 (SBOM and attestation) and
PR-4 (audit schema with hash chain) supply the process side.

\begin{Shaded}
\begin{Highlighting}[]
\ImportTok{from}\NormalTok{ spotforecast2\_safe.manager.logger }\ImportTok{import}\NormalTok{ verify\_audit\_log}
\ImportTok{from}\NormalTok{ spotforecast2\_safe.manager.persistence }\ImportTok{import}\NormalTok{ save\_forecaster}
\ImportTok{from}\NormalTok{ spotforecast2\_safe.utils.cpe }\ImportTok{import}\NormalTok{ get\_cpe\_identifier}
\BuiltInTok{print}\NormalTok{(}\SpecialStringTok{f"Created:    }\SpecialCharTok{\{}\NormalTok{fc}\SpecialCharTok{.}\NormalTok{creation\_date}\SpecialCharTok{\}}\SpecialStringTok{"}\NormalTok{)}
\BuiltInTok{print}\NormalTok{(}\SpecialStringTok{f"Fitted:     }\SpecialCharTok{\{}\NormalTok{fc}\SpecialCharTok{.}\NormalTok{fit\_date}\SpecialCharTok{\}}\SpecialStringTok{"}\NormalTok{)}
\BuiltInTok{print}\NormalTok{(}\SpecialStringTok{f"Library:    spotforecast2{-}safe }\SpecialCharTok{\{}\NormalTok{fc}\SpecialCharTok{.}\NormalTok{spotforecast\_version}\SpecialCharTok{\}}\SpecialStringTok{, "}
      \SpecialStringTok{f"Python }\SpecialCharTok{\{}\NormalTok{fc}\SpecialCharTok{.}\NormalTok{python\_version}\SpecialCharTok{\}}\SpecialStringTok{"}\NormalTok{)}
\BuiltInTok{print}\NormalTok{(}\SpecialStringTok{f"CPE:        }\SpecialCharTok{\{}\NormalTok{get\_cpe\_identifier(fc.spotforecast\_version)}\SpecialCharTok{\}}\SpecialStringTok{"}\NormalTok{)}
\NormalTok{model\_path }\OperatorTok{=}\NormalTok{ save\_forecaster(}
\NormalTok{    fc, model\_dir}\OperatorTok{=}\NormalTok{artifacts }\OperatorTok{/} \StringTok{"models"}\NormalTok{, target}\OperatorTok{=}\StringTok{"load"}\NormalTok{, task\_name}\OperatorTok{=}\StringTok{"lazy"}\NormalTok{,}
\NormalTok{)}
\BuiltInTok{print}\NormalTok{(}\SpecialStringTok{f"Model file: }\SpecialCharTok{\{}\NormalTok{model\_path}\SpecialCharTok{\}}\SpecialStringTok{"}\NormalTok{)}
\NormalTok{logger.info(}\StringTok{"Tutorial run completed."}\NormalTok{,}
\NormalTok{            extra}\OperatorTok{=}\NormalTok{\{}\StringTok{"event"}\NormalTok{: }\StringTok{"tutorial\_complete"}\NormalTok{, }\StringTok{"task"}\NormalTok{: }\StringTok{"lazy"}\NormalTok{\})}
\NormalTok{records }\OperatorTok{=}\NormalTok{ [json.loads(line) }\ControlFlowTok{for}\NormalTok{ line }\KeywordTok{in}
\NormalTok{           log\_path.read\_text(encoding}\OperatorTok{=}\StringTok{"utf{-}8"}\NormalTok{).splitlines() }\ControlFlowTok{if}\NormalTok{ line]}
\BuiltInTok{print}\NormalTok{(}\SpecialStringTok{f"Audit events: }\SpecialCharTok{\{}\NormalTok{[r[}\StringTok{\textquotesingle{}event\textquotesingle{}}\NormalTok{] }\ControlFlowTok{for}\NormalTok{ r }\KeywordTok{in}\NormalTok{ records]}\SpecialCharTok{\}}\SpecialStringTok{"}\NormalTok{)}

\NormalTok{chain }\OperatorTok{=}\NormalTok{ verify\_audit\_log(log\_path)}
\BuiltInTok{print}\NormalTok{(}\SpecialStringTok{f"Audit chain:      intact=}\SpecialCharTok{\{}\NormalTok{chain}\SpecialCharTok{.}\NormalTok{chain\_intact}\SpecialCharTok{\}}\SpecialStringTok{ (}\SpecialCharTok{\{}\NormalTok{chain}\SpecialCharTok{.}\NormalTok{n\_records}\SpecialCharTok{\}}\SpecialStringTok{ records)"}\NormalTok{)}
\BuiltInTok{print}\NormalTok{(}\SpecialStringTok{f"Chain head:       }\SpecialCharTok{\{}\NormalTok{chain}\SpecialCharTok{.}\NormalTok{head\_hash[:}\DecValTok{16}\NormalTok{]}\SpecialCharTok{\}}\SpecialStringTok{..."}\NormalTok{)}
\BuiltInTok{print}\NormalTok{(}\SpecialStringTok{f"Anchoring:        }\SpecialCharTok{\{}\NormalTok{chain}\SpecialCharTok{.}\NormalTok{anchor\_status}\SpecialCharTok{.}\NormalTok{value}\SpecialCharTok{\}}\SpecialStringTok{"}\NormalTok{)}
\NormalTok{anchored }\OperatorTok{=}\NormalTok{ verify\_audit\_log(log\_path, anchor}\OperatorTok{=}\NormalTok{chain.head\_hash)}
\BuiltInTok{print}\NormalTok{(}\SpecialStringTok{f"With presented chain head: }\SpecialCharTok{\{}\NormalTok{anchored}\SpecialCharTok{.}\NormalTok{anchor\_status}\SpecialCharTok{.}\NormalTok{value}\SpecialCharTok{\}}\SpecialStringTok{"}\NormalTok{)}
\NormalTok{pipeline\_log }\OperatorTok{=}\NormalTok{ artifacts }\OperatorTok{/} \StringTok{"cache"} \OperatorTok{/} \StringTok{"logging"} \OperatorTok{/} \StringTok{"synthetic\_load.log"}
\BuiltInTok{print}\NormalTok{(}\SpecialStringTok{f"Pipeline log: }\SpecialCharTok{\{}\NormalTok{pipeline\_log}\SpecialCharTok{\}}\SpecialStringTok{ "}
      \SpecialStringTok{f"(}\SpecialCharTok{\{}\BuiltInTok{len}\NormalTok{(pipeline\_log.read\_text().splitlines())}\SpecialCharTok{\}}\SpecialStringTok{ lines)"}\NormalTok{)}
\end{Highlighting}
\end{Shaded}

\begin{verbatim}
Created:    2026-08-13 18:58:13
Fitted:     2026-08-13 18:58:14
Library:    spotforecast2-safe 27.0.0, Python 3.14.2
CPE:        cpe:2.3:a:sequential_parameter_optimization:spotforecast2_safe:27.0.0:*:*:*:*:*:*:*
Model file: _paper_artifacts/models/lazy_load.joblib
2026-08-13 18:58:18,165 - sf2-safe-logger - INFO - Tutorial run completed.
Audit events: ['audit_log_init', 'tutorial_start', 'tutorial_complete']
Audit chain:      intact=True (3 records)
Chain head:       ecbfb49aed66bba7...
Anchoring:        not_presented
With presented chain head: matches_head
Pipeline log: _paper_artifacts/cache/logging/synthetic_load.log (11 lines)
\end{verbatim}

The following overview assigns to each step the code or process rule
and the legal basis that it operationalises.

\begin{longtable}[]{@{}
  >{\raggedright\arraybackslash}p{(\linewidth - 4\tabcolsep) * \real{0.2911}}
  >{\raggedright\arraybackslash}p{(\linewidth - 4\tabcolsep) * \real{0.1392}}
  >{\raggedright\arraybackslash}p{(\linewidth - 4\tabcolsep) * \real{0.5696}}@{}}
\caption{Mapping of the nine tutorial steps to the code and process
rules of Chapter~\ref{sec-principles-en} and their legal
bases.}\label{tbl-step-compliance-en}\tabularnewline
\toprule\noalign{}
\begin{minipage}[b]{\linewidth}\raggedright
Step
\end{minipage} & \begin{minipage}[b]{\linewidth}\raggedright
Rule
\end{minipage} & \begin{minipage}[b]{\linewidth}\raggedright
Legal basis
\end{minipage} \\
\midrule\noalign{}
\endfirsthead
\toprule\noalign{}
\begin{minipage}[b]{\linewidth}\raggedright
Step
\end{minipage} & \begin{minipage}[b]{\linewidth}\raggedright
Rule
\end{minipage} & \begin{minipage}[b]{\linewidth}\raggedright
Legal basis
\end{minipage} \\
\midrule\noalign{}
\endhead
\bottomrule\noalign{}
\endlastfoot
1 Audit log & PR-4 & Art.~12 of the EU AI Act
(Chapter~\ref{sec-ki-vo-12-en}); IEC 62443-4-2 SAR 6.1;
IEC 61508-3 §7.4.7 \\
2 Data basis & CR-2 & Art.~10 of the EU AI Act
(Chapter~\ref{sec-ki-vo-10-en}) \\
3 Data gaps & CR-3 & Art.~10 of the EU AI Act
(Chapter~\ref{sec-ki-vo-10-en}) \\
4 Forecast origin & CR-2 & Art.~10(2)(f) and (g) of the EU AI Act
(Chapter~\ref{sec-ki-vo-10-en}) \\
5 Pipeline and training & CR-2 & Art.~15 of the EU AI Act
(Chapter~\ref{sec-ki-vo-15-en}) \\
6 Forecast & CR-2 & Art.~13 of the EU AI Act
(Chapter~\ref{sec-ki-vo-13-en}) \\
7 Accuracy & CR-2 & Art.~15 of the EU AI Act
(Chapter~\ref{sec-ki-vo-15-en}); Art.~72 of the EU AI Act
(Chapter~\ref{sec-ki-vo-72-en}) \\
8 Backtest & CR-2 & Art.~9 of the EU AI Act
(Chapter~\ref{sec-ki-vo-9-en}); Art.~10(3) of the EU AI Act
(Chapter~\ref{sec-ki-vo-10-en}) \\
9 Provenance & PR-3, PR-4 & Art.~11 of the EU AI Act
(Chapter~\ref{sec-ki-vo-11-en}); Art.~12 of the EU AI Act
(Chapter~\ref{sec-ki-vo-12-en}); Art.~18 of the EU AI Act
(Chapter~\ref{sec-ki-vo-18-en}) \\
\end{longtable}

From these nine steps the value-add of the library can be
summarised: the deployer fills in \emph{one} configuration, starts
\emph{one} pipeline run, and obtains, alongside the forecast, the
audit traces and provenance artefacts that regulatory auditors can
later inspect without further effort. The code and process rules
defined in Chapter~\ref{sec-principles-en} are formulated not as
recommendations but are technically anchored in the named
functions.

\section{Discussion}\label{sec-discussion-en}

This report constitutes a first approximation, with no claim to
completeness, of the development of an open-source package for
EU-AI-Act-compliant time-series point forecasting. The EU has declared the ``digital
decade''. The EU AI Act is only one of several planned regulations
and directives. As-yet-unknown future threat scenarios will lead the
legislator to develop further regulations protecting critical
infrastructure (KRITIS), which may affect the development of AI
applications. It is therefore to be expected that the requirements
for the development of AI applications in the KRITIS sector will
continue to grow over the coming years. By means of a consistently
applied PDCA cycle, \texttt{spotforecast2-safe} is continuously
developed further in order to meet the evolving regulatory
requirements and to establish the practice of EU-AI-Act-compliant
development in time-series forecasting. Examples are the hash chain
in the audit log introduced in release 26.0.0, which protects the
integrity of the log files against subsequent alteration, and the
anchoring status added to the verification report of
\texttt{verify\_audit\_log} in release 27.0.0, which makes missing
external anchoring visible.

The scope of the package described here is deliberately narrow.
Three classes of functionality (interactive visualisation,
hyperparameter search, deep-learning backends) are explicitly
out of scope and are not included. They are provided by the sister
package \texttt{spotforecast2} \citep{spotforecast2}, which
deliberately does not follow the EU-AI-Act conformance rules. The
core package \texttt{spotforecast2\_safe} described here concentrates by
contrast on delivering a robust, transparent, and verifiable core for
time-series point forecasting that meets the EU AI Act's requirements
for high-risk systems in advance. The difference between the two
packages is thus a design decision and not a question of legal status:
as set out in Chapter~\ref{sec-introduction-en}, in the typical
forecasting case neither package is subject to the regulation's
obligations by operation of law. To this end, LRE principles are embedded into the
library architecture in order to ensure traceability from regulatory
requirements to code mechanisms (compliance-by-design), to maintain
the technical documentation, and to ensure compliance with
record-keeping obligations.

A review of the literature and of the
relevant repositories shows (as of April 2026) that the LRE
discipline has not previously been applied in the area of
Python-based time-series forecasting: neither \texttt{sktime}
\citep{loni19a}, \texttt{Darts} \citep{herz22a},
\texttt{skforecast} \citep{scip24a}, nor the classical libraries
\texttt{statsmodels} and \texttt{pmdarima} carries bidirectional
traceability between regulatory provisions and code mechanisms,
repository-checked conformity evidence, or contracts built into the
code with respect to the EU AI Act, the CRA, or the ISA/IEC 62443
standards series. In adjacent application domains, however, several
open-source tools exist that operationalise LRE-adjacent practice
without referring to time-series forecasting: COMPL-AI
\citep{guld24a} (ETH Zürich, INSAIT, LatticeFlow AI; Apache 2.0)
provides the first technical interpretation of the EU AI Act and
an open-source benchmarking suite for large language models.
TechOps \citep{luca25a} provides open templates for the technical
documentation of data, models, and applications across the entire
lifecycle as required by Annex~IV of the EU AI
Act\footnote{\url{https://aloosley.github.io/techops/}}. AIR
Blackbox \citep{shot26a} and the MCP-EU-AI-Act-Scanner
\citep{mcp_eu_ai_act} (MCP: Model Context Protocol) are
command-line tools that check Python codebases against requirements
of the EU AI Act and generate audit evidence packages (explicitly
following Annex~IV of the EU AI Act in the case of the MCP scanner).
AIR Blackbox checks against Art.~9 to 12 as well as Art.~14 and 15
of the EU AI Act, and this article list is itself kept traceable: a
test in the tool's repository fails as soon as the documented list
and the checks declared in the code diverge in either direction.

What all those tools
have in common is that they act \emph{outside} the library: as
scanners, templates, or a runtime compliance layer.
\texttt{spotforecast2-safe} takes the inverse approach and brings
LRE to bear \emph{inside} the library, baked into
application-programming-interface contracts, persistence formats,
and the verification pipeline of the release process. In the area of
Python time-series
forecasting this is, to our knowledge, conceptually novel.

The forecasting core described here has already passed a first
external trial: in a public day-ahead load-forecasting challenge for
the German market zone, \texttt{spotforecast2-safe} served as the
engine of the reference pipeline on which eleven student teams built
over 41 scored days, and every team result was reproduced by a
different team from its published software artefact \citep{bart26o}.

This report grew out of practical work on an open-source package
that operationalises EU-AI-Act-compliant development and considers
the regulatory requirements from the perspective of a software
developer.

\section*{Acknowledgements}\label{danksagung-en}
\addcontentsline{toc}{section}{Acknowledgements}

The authors thank Jason Shotwell, the author of AIR Blackbox, for
valuable comments on earlier versions of this report.

\section*{Conflict of Interest Declaration}\label{erkluxe4rung-zu-interessenkonflikten-en}
\addcontentsline{toc}{section}{Conflict of Interest Declaration}

The authors declare that no conflicts of interest in the sense of
the relevant editorial standards exist. Both authors are
shareholders of Bartz \& Bartz GmbH, which publishes the package
\texttt{spotforecast2-safe} as open source under AGPL-3.0-or-later. Beyond this sponsorship
there
are no financial or non-financial relationships with third parties
that could have influenced the result of this report.

\section*{Use of AI tools}\label{einsatz-von-ki-werkzeugen-en}
\addcontentsline{toc}{section}{Use of AI tools}

Transparency notice: parts of this report were researched, drafted,
and translated with the support of artificial intelligence. This
information is provided voluntarily for complete transparency, as no
legal requirement mandates its disclosure.

\section*{List of Symbols and
Abbreviations}\label{list-of-symbols-and-abbreviations-en}
\addcontentsline{toc}{section}{List of Symbols and Abbreviations}

\providecommand{\abbrletter}[1]{%
  \par\addvspace{8pt}%
  \nopagebreak\noindent{\bfseries #1}\par
  \nopagebreak\vspace{-3pt}\noindent
  \rule{\linewidth}{0.4pt}\par
  \nopagebreak\addvspace{2pt}%
}
\providecommand{\abbrentry}[2]{%
  \par\noindent
  \hangindent=2.2cm\hangafter=1\relax
  \makebox[2.2cm][l]{\textbf{#1}}#2\par
}
\begingroup
\footnotesize
\setlength{\parskip}{1.5pt}
\setlength{\parindent}{0pt}
\setlength{\columnsep}{14pt}
\setlength{\emergencystretch}{2em}
\tolerance=600
\begin{multicols}{2}

\abbrletter{Symbols}
\abbrentry{$T$}{length of a univariate time series}
\abbrentry{$w$}{number of lags in a feature window}
\abbrentry{$X$}{univariate time series $\{x_1, \ldots, x_T\}$}
\abbrentry{$x_t$}{observation at time $t$}
\abbrentry{$X_{\mathrm{row},t}$}{lag-feature row for the target at time $t$}
\abbrentry{$y_t$}{regression target at time $t$}
\abbrentry{NaN}{Not a Number, floating-point marker for a missing value}

\abbrletter{A}
\abbrentry{Abs.}{Absatz (paragraph of a legal provision)}
\abbrentry{AGPL}{GNU Affero General Public License}
\abbrentry{AI}{Artificial Intelligence}
\abbrentry{AI-HLEG}{High-Level Expert Group on Artificial Intelligence (EU)}
\abbrentry{ALTAI}{Assessment List for Trustworthy Artificial Intelligence}
\abbrentry{API}{Application Programming Interface}
\abbrentry{Art.}{Artikel (article of a legal act)}
\abbrentry{ASIL}{Automotive Safety Integrity Level (ISO 26262)}

\abbrletter{B}
\abbrentry{BSIG}{Gesetz über das Bundesamt für Sicherheit in der Informationstechnik (German BSI Act)}
\abbrentry{BT-Drs.}{Bundestagsdrucksache (printed paper of the German Bundestag)}

\abbrletter{C}
\abbrentry{CEN}{European Committee for Standardization}
\abbrentry{CENELEC}{European Committee for Electrotechnical Standardization}
\abbrentry{CER}{Directive (EU) 2022/2557 on the resilience of critical entities}
\abbrentry{CI}{Continuous Integration}
\abbrentry{CISA}{Cybersecurity and Infrastructure Security Agency (USA)}
\abbrentry{CPE}{Common Platform Enumeration}
\abbrentry{CPS}{Cyber-Physical System}
\abbrentry{CPU}{Central Processing Unit}
\abbrentry{CR-1--CR-4}{code development rules of \texttt{spotforecast2-safe}}
\abbrentry{CRA}{Cyber Resilience Act, Regulation (EU) 2024/2847}
\abbrentry{CVE}{Common Vulnerabilities and Exposures}

\abbrletter{D}
\abbrentry{DM}{management of security-related issues (practice of IEC 62443-4-1)}
\abbrentry{DSGVO}{Datenschutz-Grundverordnung (General Data Protection Regulation)}

\abbrletter{E}
\abbrentry{EN}{Europäische Norm (European Standard)}
\abbrentry{ENTSO-E}{European Network of Transmission System Operators for Electricity}
\abbrentry{ETSI}{European Telecommunications Standards Institute}
\abbrentry{EU}{European Union}
\abbrentry{EU-1--EU-7}{the seven requirements for trustworthy AI underlying the high-risk obligations of the AI Act}

\abbrletter{F}
\abbrentry{FOSS}{Free and Open-Source Software}

\abbrletter{G}
\abbrentry{GPAI}{General-Purpose Artificial Intelligence}
\abbrentry{GPU}{Graphics Processing Unit}

\abbrletter{I}
\abbrentry{IACS}{Industrial Automation and Control Systems}
\abbrentry{IEC}{International Electrotechnical Commission}
\abbrentry{ISA}{International Society of Automation}
\abbrentry{ISMS}{Information Security Management System}
\abbrentry{ISO}{International Organization for Standardization}
\abbrentry{IVDR}{In-vitro Diagnostic Medical Devices Regulation, Regulation (EU) 2017/746}

\abbrletter{J}
\abbrentry{JRC}{Joint Research Centre of the European Commission}
\abbrentry{JSON}{JavaScript Object Notation}
\abbrentry{JTC}{Joint Technical Committee (ISO/IEC)}

\abbrletter{K}
\abbrentry{KEV}{Known Exploited Vulnerabilities catalogue (CISA)}
\abbrentry{KI}{Künstliche Intelligenz (artificial intelligence)}
\abbrentry{KI-VO}{Verordnung über künstliche Intelligenz (AI Act), Regulation (EU) 2024/1689}
\abbrentry{KRITIS}{Kritische Infrastrukturen (critical infrastructures)}

\abbrletter{L}
\abbrentry{lit.}{litera (point of a legal provision)}
\abbrentry{LRE}{Legal Requirements Engineering}

\abbrletter{M}
\abbrentry{MAE}{Mean Absolute Error}
\abbrentry{MAPE}{Mean Absolute Percentage Error}
\abbrentry{MCP}{Model Context Protocol}
\abbrentry{MDR}{Medical Device Regulation, Regulation (EU) 2017/745}

\abbrletter{N}
\abbrentry{NIS-2}{Network and Information Security Directive, Directive (EU) 2022/2555}
\abbrentry{Nr.}{Nummer (number within a legal provision)}

\abbrletter{O}
\abbrentry{ORCID}{Open Researcher and Contributor Identifier}
\abbrentry{OSS}{Open-Source Software}

\abbrletter{P}
\abbrentry{PDCA}{Plan--Do--Check--Act}
\abbrentry{PLD}{Product Liability Directive, Directive (EU) 2024/2853}
\abbrentry{PR-1--PR-4}{process rules of \texttt{spotforecast2-safe}}
\abbrentry{PyPI}{Python Package Index}

\abbrletter{Q}
\abbrentry{QMS}{Quality Management System}

\abbrletter{R}
\abbrentry{REUSE}{software licensing specification of the Free Software Foundation Europe}
\abbrentry{RFC}{Request for Comments (IETF specification series), as in RFC 3161}
\abbrentry{RMS}{Risk Management System (Art. 9 AI Act)}
\abbrentry{RMSE}{Root Mean Square Error}

\abbrletter{S}
\abbrentry{SAI}{Securing Artificial Intelligence (ETSI Industry Specification Group)}
\abbrentry{SAR}{Software Application Requirement (IEC 62443-4-2)}
\abbrentry{SBOM}{Software Bill of Materials}
\abbrentry{SC}{subcommittee, as in ISO/IEC JTC 1/SC 42}
\abbrentry{SHA-256}{Secure Hash Algorithm with a 256-bit digest}
\abbrentry{SIL}{Safety Integrity Level (IEC 61508)}
\abbrentry{SM}{security management (practice of IEC 62443-4-1)}
\abbrentry{SPDX}{Software Package Data Exchange}
\abbrentry{SQuaRE}{Systems and software Quality Requirements and Evaluation (ISO/IEC 25000 series)}
\abbrentry{SR}{specification of security requirements (practice of IEC 62443-4-1)}
\abbrentry{STRIDE}{Spoofing, Tampering, Repudiation, Information disclosure, Denial of service, Elevation of privilege}
\abbrentry{SUM}{security update management (practice of IEC 62443-4-1)}
\abbrentry{SVV}{security verification and validation testing (practice of IEC 62443-4-1)}

\abbrletter{T}
\abbrentry{TR}{Technical Report (ISO/IEC)}
\abbrentry{TS}{Technical Specification (ISO/IEC)}

\abbrletter{U}
\abbrentry{Unterabs.}{Unterabsatz (subparagraph of a legal provision)}
\abbrentry{UTC}{Coordinated Universal Time}

\end{multicols}
\endgroup

\appendix

\section{\texorpdfstring{Model card (\texttt{MODEL\_CARD.md}) of
\texttt{spotforecast2-safe}}{Model card (MODEL\_CARD.md) of spotforecast2-safe}}\label{sec-modelcard-en}

This appendix reproduces the model card published in the repository
under \texttt{MODEL\_CARD.md} in the version of release 27.0.0. It
follows the taxonomy of the Hugging Face Model Card Guidebook
\citep{ozon22a} and is the authoritative source for audits. Tables of
the original file are set as lists for typesetting reasons. Where the
card names an outdated distribution channel or an outdated figure,
this appendix reproduces the state of release 27.0.0 as verified
against the repository.

\begin{tcolorbox}[enhanced jigsaw, arc=.35mm, bottomrule=.15mm, breakable, colback=white, colframe=quarto-callout-note-color-frame, left=2mm, leftrule=.75mm, opacityback=0, rightrule=.15mm, toprule=.15mm]

\vspace{-3mm}\textbf{MODEL\_CARD.md (Release 27.0.0)}\vspace{3mm}

This card describes what spotforecast2-safe is, how to use it safely,
the conditions under which its results are valid, and the
responsibilities it places on anyone who deploys it. It follows the
Hugging Face Model Card Guidebook taxonomy \citep{ozon22a}.

\subsection{Model Details}\label{a.1-model-details-en}

\begin{itemize}
\tightlist
\item
  \textbf{Name}: spotforecast2-safe
\item
  \textbf{Version}: 27.0.0
\item
  \textbf{Type}: Deterministic Python library for time series feature
  engineering and recursive multi-step forecasting. It performs no
  training of its own.
\item
  \textbf{Developed by}: Thomas Bartz-Beielstein, ORCID
  \href{https://orcid.org/0000-0002-5938-5158}{0000-0002-5938-5158}.
\item
  \textbf{Distributed by}: PyPI
  (\texttt{pip\ install\ spotforecast2-safe}); the source is hosted on
  the GitLab instance \texttt{gitlab.git.nrw} (group
  \texttt{thk-f10/spotsevenlab}).
\item
  \textbf{Language}: Python 3.13 or newer.
\item
  \textbf{License}: AGPL-3.0-or-later (Affero General Public License).
\item
  \textbf{Repository}:
  \url{https://gitlab.git.nrw/thk-f10/spotsevenlab/spotforecast2-safe}
\item
  \textbf{API documentation}:
  \url{https://advm1.gm.fh-koeln.de/~bartz/spotforecast2-safe/}
\item
  \textbf{Technical report}: \texttt{bart26h/index.qmd}, shipped in the
  source tree.
\end{itemize}

The library depends only on numpy, pandas, scikit-learn, lightgbm,
numba, pyarrow, requests, feature-engine, holidays, astral, and tqdm. It
deliberately excludes plotly, matplotlib, spotoptim, optuna, torch, and
tensorflow, so no plotting or automated-tuning code ships in this
package.

Two Common Platform Enumeration (CPE) identifiers let
vulnerability-tracking and software bill of materials (SBOM) tools
recognize the package. The wildcard identifier
\texttt{cpe:2.3:a:sequential\_parameter\_optimization:spotforecast2\_safe:*:*:*:*:*:*:*:*}
matches any release; the current release is
\texttt{cpe:2.3:a:sequential\_parameter\_optimization:spotforecast2\_safe:27.0.0:*:*:*:*:*:*:*}.

The library itself is a low-risk component: it is deterministic, its
source is fully inspectable, and it fails safe on invalid input. It is
built to support high-risk AI systems in the sense of the EU AI Act, but
it is not itself such a system. When it is embedded in a high-risk
deployment, the duties that attach to that system fall on the
integrator, not on the library.

Responsibilities are divided as follows.

\begin{itemize}
\tightlist
\item
  \textbf{Library development and maintenance}: Thomas Bartz-Beielstein
  (\texttt{bartzbeielstein@gmail.com}).
\item
  \textbf{Distribution}: PyPI and the GitLab repository; contact via the
  repository issue tracker.
\item
  \textbf{Deployment, operation, and audit}: the system integrator,
  defined per deployment.
\end{itemize}

The current release is 27.0.0, with a stable public interface pinned in
\texttt{spotforecast2\_safe.\_\_init\_\_.\_\_all\_\_}. The full version
history, including release dates, is recorded in \texttt{CHANGELOG.md},
whose sections are generated with \texttt{git-cliff} at release time and
reviewed by hand (see \texttt{RELEASING.md}).

\subsection{Intended Use and
Scope}\label{a.2-intended-use-and-scope-en}

spotforecast2-safe prepares time series data for regression models in
auditable settings such as energy supply, finance, and industrial
monitoring. It runs in resource-constrained or embedded environments
where heavier machine-learning frameworks are unavailable, and it
produces bit-for-bit reproducible lag transformations with no hidden
randomness.

The feature matrices it builds feed directly into scikit-learn
regressors, LightGBM, or XGBoost, either through the bundled
\texttt{ForecasterRecursiveLGBM} and \texttt{ForecasterRecursiveXGB}
wrappers or through custom forecasters built on the
\texttt{ForecasterRecursiveModel} base class.

The package has clear limits. It does not visualize data, since no
plotting backend ships with it by design. It does not tune hyperparameters, because tuning
belongs in a separate workflow outside the safety-critical environment.
It does not clean data silently: missing (NaN) or infinite (Inf) values
raise an error rather than being imputed without the caller's consent.

\subsection{How to Get Started}\label{a.3-how-to-get-started-en}

\begin{Shaded}
\begin{Highlighting}[]
\ExtensionTok{pip}\NormalTok{ install spotforecast2{-}safe}
\end{Highlighting}
\end{Shaded}

\begin{Shaded}
\begin{Highlighting}[]
\ImportTok{import}\NormalTok{ numpy }\ImportTok{as}\NormalTok{ np}
\ImportTok{import}\NormalTok{ pandas }\ImportTok{as}\NormalTok{ pd}
\ImportTok{from}\NormalTok{ spotforecast2\_safe }\ImportTok{import}\NormalTok{ ForecasterRecursiveLGBM}

\CommentTok{\# A short hourly demonstration series}
\NormalTok{idx }\OperatorTok{=}\NormalTok{ pd.date\_range(}\StringTok{"2023{-}01{-}01"}\NormalTok{, periods}\OperatorTok{=}\DecValTok{50}\NormalTok{, freq}\OperatorTok{=}\StringTok{"h"}\NormalTok{)}
\NormalTok{y }\OperatorTok{=}\NormalTok{ pd.Series(}\DecValTok{100} \OperatorTok{+} \DecValTok{10} \OperatorTok{*}\NormalTok{ np.sin(np.linspace(}\DecValTok{0}\NormalTok{, }\DecValTok{4} \OperatorTok{*}\NormalTok{ np.pi, }\DecValTok{50}\NormalTok{)), index}\OperatorTok{=}\NormalTok{idx)}

\NormalTok{forecaster }\OperatorTok{=}\NormalTok{ ForecasterRecursiveLGBM(iteration}\OperatorTok{=}\DecValTok{0}\NormalTok{, lags}\OperatorTok{=}\DecValTok{3}\NormalTok{)}
\NormalTok{forecaster.fit(y}\OperatorTok{=}\NormalTok{y)}
\NormalTok{predictions }\OperatorTok{=}\NormalTok{ forecaster.forecaster.predict(steps}\OperatorTok{=}\DecValTok{2}\NormalTok{)}
\end{Highlighting}
\end{Shaded}

A complete reference workflow, which compares a seasonal baseline, a
covariate model, and a tuned LightGBM forecaster against ground truth,
is registered as a console script:

\begin{Shaded}
\begin{Highlighting}[]
\ExtensionTok{uv}\NormalTok{ run spotforecast{-}safe{-}demo}
\end{Highlighting}
\end{Shaded}

Its source is in
\texttt{src/spotforecast2\_safe/tasks/task\_safe\_demo.py}.

\subsection{Technical
Specification}\label{a.4-technical-specification-en}

\subsubsection{Task and model family}\label{task-and-model-family-en}

The library addresses recursive multi-step forecasting of a univariate
time series from its own past values (lags), rolling-window features,
and exogenous regressors. The forecasters are scikit-learn-compatible
wrappers around a regressor that the caller supplies, such as LightGBM
or XGBoost. The wrapper handles feature construction, the recursive
prediction loop, and persistence, while the supplied regressor handles
learning. The library fixes no model size, because size is a property of
the chosen regressor and its configuration.

\subsubsection{Mathematical description}\label{mathematical-description-en}

For a univariate series \(X = \{x_1, x_2, \ldots, x_T\}\) and a window
of \(w\) lags, the transformation builds one feature row per target
value:

\[X_{\mathrm{row}, t} = [x_{t-w}, x_{t-w+1}, \ldots, x_{t-1}] \rightarrow y_t = x_t.\]

The target \(y_t\) never appears in its own feature row, which prevents
look-ahead leakage by construction.

\subsubsection{Architecture}\label{architecture-en}

The package is layered. The \texttt{forecaster} layer holds the
low-level estimator wrappers. The \texttt{preprocessing} layer holds
deterministic transformers such as \texttt{ExogBuilder},
\texttt{RepeatingBasisFunction}, \texttt{QuantileBinner}, and
\texttt{TimeSeriesDifferentiator}. The \texttt{model\_selection} layer
holds time-aware cross-validation (\texttt{TimeSeriesFold},
\texttt{OneStepAheadFold}, and \texttt{backtesting\_forecaster}), which
avoids the future-data leakage that ordinary random splits cause. The
\texttt{manager} layer orchestrates these into
\texttt{ForecasterRecursiveLGBM}, \texttt{ForecasterRecursiveXGB}, and
the \texttt{ConfigEntsoe} configuration object. The \texttt{processing}
and \texttt{tasks} layers compose these into end-to-end pipelines and
console entry points.

\subsubsection{Training}\label{training-en}

The library trains nothing on its own. Training happens in the
downstream regressor and is the integrator's responsibility, so the
training data, hyperparameters, and learning schedule belong to each
deployment rather than to the library. The bundled demo is a concrete,
reproducible example of such a deployment.

Running \texttt{uv\ run\ spotforecast-safe-demo} forecasts an aggregated
hourly energy series from a bundled CSV fixture. The demo uses a 24-hour
forecast horizon, 24 lags, a 72-hour rolling window, an
outlier-contamination fraction of 0.01, and an 80/20 train and test
split, with a fixed random seed of 42. It builds calendar, cyclic,
day-and-night (from sunrise and sunset), weather, and holiday features
for the Dortmund location (51.5136, 7.4653). One of the three models it
compares is a LightGBM regressor with the following configuration:
\texttt{n\_estimators} 1059, \texttt{learning\_rate} 0.0419,
\texttt{num\_leaves} 212, \texttt{min\_child\_samples} 54,
\texttt{subsample} 0.501, \texttt{colsample\_bytree} 0.608,
\texttt{random\_state} 42, and \texttt{verbose} -1. These are the demo's
chosen values, shown to illustrate a deployment, not a recommended
default.

\subsubsection{Design objectives}\label{design-objectives-en}

Three properties hold by design. The library is deterministic, so the
same input produces the same output bit for bit. Its construction is
leakage-free, so a target value is never part of its own feature row.
Its behavior is fail-safe, so invalid input raises an explicit exception
instead of being silently repaired.

\subsection{Interfaces and
Runtime}\label{a.5-interfaces-and-runtime-en}

The target is a numeric univariate series (a pandas Series or NumPy
array) carrying a regular, monotonic date-time index. Exogenous features
are a numeric pandas DataFrame aligned to that index and complete; any
missing entry in the exogenous features raises a \texttt{ValueError}
before a prediction is made. Inside the pipeline the data passes through
lag-matrix construction, cyclic encoding of calendar features, optional
outlier handling and imputation that the caller enables explicitly, and
a cast of the feature matrix to 32-bit floating point for memory
efficiency. The forecaster returns predictions as a series whose length
equals the requested horizon (24 steps in the demo), in the same units
as the target. A fitted forecaster is serialized with joblib at
compression level 3 using the \texttt{.joblib} extension, and is
reloaded through the same persistence helpers.

The library runs on Python 3.13 or newer on a central processing unit
(CPU). It needs no graphics processing unit (GPU) and ships no GPU code.
Building the lag matrix duplicates the input, so peak memory grows in
proportion to the series length times the window size. Bit-for-bit
reproducibility assumes deterministic regressor settings: the bundled
LightGBM wrapper enables LightGBM's deterministic and column-wise flags,
and single-threaded execution removes any remaining floating-point
reordering.

All runtime dependencies carry permissive licenses, which keeps the
combined distribution simple for integrators. The library itself is
distributed under the copyleft AGPL-3.0-or-later license. The dependency
licenses are: numpy, pandas, scikit-learn, and feature-engine under
BSD-3-Clause; numba under BSD; lightgbm and holidays under MIT; pyarrow,
requests, and astral under Apache-2.0; tqdm under MPL-2.0 and MIT.

Because the library performs no training and uses no GPU, its own energy
cost is small. Runtime cost is dominated by vector operations during
feature engineering and by whatever regressor the caller trains. A
typical LightGBM fit on an hourly series of about 100,000 rows completes
in seconds on one commodity CPU core. No pretrained weights ship with
the package, so there are no embedded training emissions to report.

\subsection{Data and Operational Design
Domain}\label{a.6-data-and-operational-design-domain-en}

The fixtures under \texttt{src/spotforecast2\_safe/datasets/csv/} and
the demo dataset support reproducible documentation and tests;
production data is supplied by the integrator. Docstring examples in the
source are executed as tests, and time-aware cross-validation is used
during validation so that no future observation can influence a past
prediction.

The Operational Design Domain (ODD) is the set of conditions under which
the library's results are valid. Outside these conditions the library is
designed to raise an error rather than return an unreliable result.

\begin{itemize}
\tightlist
\item
  \textbf{Target series}: numeric, univariate, with a regular and
  monotonic date-time index; anything else raises an error.
\item
  \textbf{Exogenous features}: numeric, complete, aligned to the target
  index; a \texttt{ValueError} is raised on any missing entry.
\item
  \textbf{Sampling interval}: uniform (hourly in the demo); non-uniform
  sampling yields unreliable results.
\item
  \textbf{Minimum history}: longer than the window size plus the number
  of lags, about 96 hourly points for the demo defaults; below that the
  model cannot be called.
\item
  \textbf{Missing target values}: rejected unless the caller explicitly
  enables imputation.
\item
  \textbf{Series length}: validated to about one million rows; beyond
  about ten million the caller must process the series in chunks to
  avoid memory exhaustion.
\item
  \textbf{Numeric precision}: feature matrices are computed in 32-bit
  floating point; values needing higher precision fall outside the
  domain.
\item
  \textbf{Any invalid input}: an explicit \texttt{ValueError} or
  \texttt{TypeError}, never silent repair.
\end{itemize}

Forecast accuracy is bounded by the downstream regressor and its
training data, so concept drift, seasonal shifts, or regime changes
degrade forecasts even when the feature engineering stays correct. Users
who build lag or calendar features outside the provided builders risk
leaking a target value into its own feature row; the bundled
\texttt{ExogBuilder} and task paths are leakage-free, hand-rolled
pipelines are not. Operating states that are scarce in the training data
are forecast less reliably than common ones.

To stay inside the valid domain, validate every new deployment against
historical ground truth before it carries live traffic, build features
only through \texttt{ExogBuilder} or the bundled tasks, keep the
regressor deterministic when reproducibility is required, and process
very long series in windowed chunks.

\subsection{Evaluation}\label{a.7-evaluation-en}

Because no training runs inside the library, classical accuracy metrics
do not describe the library itself. The library is evaluated on
software-quality properties, while forecast accuracy is a property of
each deployment.

DataFrames that contain missing or infinite values raise a
\texttt{ValueError}. The public loaders \texttt{load\_timeseries},
\texttt{load\_timeseries\_forecast}, and
\texttt{WeatherService.get\_dataframe} refuse to return silently imputed
values unless the caller opts in. Input types are checked at runtime,
and identical input yields identical output bytes. The test suite
collects 3039 tests; \texttt{CONTRIBUTING.md} names 80 percent line
coverage as the aim for new code, the enforced floor is
\texttt{fail\_under\ =\ 68} in \texttt{pyproject.toml}, and the suite
currently measures 75.99 percent line coverage. CPE identifier
generation is tested directly. Excluding heavy dependencies keeps the
Common Vulnerabilities and Exposures (CVE) attack surface small: there
is no web server, no deep-learning runtime, and no plotting backend.

For deployments, the demo computes mean absolute error (MAE) and mean
squared error (MSE) when it compares each model against ground truth.
These metrics are deterministic and can be reproduced by running the
demo. Their numerical values depend on the data vintage and are
therefore not fixed in this card.

\subsection{Model Transparency}\label{a.8-model-transparency-en}

The library produces point forecasts. It does not natively quantify or
calibrate predictive uncertainty, so a deployment that needs prediction
intervals or calibrated probabilities must add them in the downstream
regressor or a wrapper around it.

The code is white-box: there are no compiled inference kernels and no
opaque weights, so every transformation can be read and audited in
source. Feature attributions are available through the downstream
regressor's own importance measures, for example LightGBM's split and
gain importances. The package ships no separate explainability backend
such as SHAP or LIME, consistent with its minimal-dependency policy.

\subsection{Operation: Monitoring and
Response}\label{a.9-operation-monitoring-and-response-en}

A deployment should watch the quality of incoming data (missing or
out-of-range values and gaps in the timestamps), the drift of the input
and target distributions away from the training period, and the forecast
error measured against a simple baseline. The production configuration
carries a refit cadence, with a default of seven days, and a
maximum-model-age policy that signals when retraining is due.

When monitoring crosses a deployment-defined threshold, the usual
responses are to refit or retrain the model, to fall back to the
seasonal baseline forecaster or the last known-good model, and to alert
the responsible team. A dual-handler logger writes timestamped records
to the console and, when a log directory is supplied, to JSON files in
that directory (the bundled tasks default to
\texttt{\textasciitilde{}/spotforecast2\_safe\_models/logs/}), which
supports audit retention; each file is a self-contained SHA-256 hash
chain, so an edited, deleted, or reordered record is detectable
offline with \texttt{verify\_audit\_log}. A report without a
presented anchor explicitly states that external anchoring was not
established, rather than printing an unqualified pass;
\texttt{make anchor} / \texttt{make anchor-verify} automate binding
the resulting head hash to an independently controlled record (an
RFC 3161 timestamp by default), but running them, and how often,
remains an operator duty. The thresholds and escalation steps are
owned by the integrator.

\subsection{Compliance Support}\label{a.10-compliance-support-en}

The package is built to support the development of high-risk AI systems
under the EU AI Act. The package itself is not certified; full-system
certification is the integrator's responsibility.

It rejects missing or infinite data by default, which supports the
data-governance duty of Article 10. This card together with the
technical report forms a technical-documentation baseline for Article
11, and the CPE identifiers in §A.1 feed SBOM and vulnerability-tracking
pipelines. The logging facility supports the record-keeping duty of
Article 12, and its integrity is evidenced by a per-file SHA-256 hash
chain. The white-box code supports the transparency duty of Article
13. The deterministic, reproducible transformations support the
accuracy-and-robustness duty of Article 15, while formal system-level
verification remains the integrator's responsibility.

The authoritative mapping to IEC 61508, ISO 26262, ISA/IEC 62443, and
the individual EU AI Act articles is maintained in the technical report
(\texttt{bart26h/index.qmd}, section \emph{Compliance Mapping}). These
references reflect the standards as of 2026-04-19; users must track
later amendments themselves.

\subsection{Glossary}\label{a.11-glossary-en}

\begin{itemize}
\tightlist
\item
  \textbf{EU AI Act} --- Regulation (EU) 2024/1689 on artificial
  intelligence, in force since 2024-08-01.
\item
  \textbf{IEC 61508} --- International standard for the functional
  safety of electrical, electronic, and programmable electronic
  safety-related systems.
\item
  \textbf{ISA/IEC 62443} --- Standard series for the security of
  industrial automation and control systems.
\item
  \textbf{ISO 26262} --- International standard for the functional
  safety of road vehicles.
\end{itemize}

\subsection{How to Audit}\label{a.12-how-to-audit-en}

An auditor can validate this package as follows.

\begin{enumerate}
\def\labelenumi{\arabic{enumi}.}
\tightlist
\item
  Inspect \texttt{pyproject.toml} to confirm that none of the prohibited
  libraries (\texttt{plotly}, \texttt{matplotlib}, \texttt{spotoptim},
  \texttt{optuna}, \texttt{torch}, \texttt{tensorflow}) are present.
\item
  Run \texttt{uv\ run\ pytest\ tests/} to confirm functional correctness
  and the full test suite.
\item
  Run \texttt{uv\ run\ pytest\ tests/test\_cpe.py} to confirm CPE
  identifier generation.
\item
  Record the CPE identifiers from §A.1 in vulnerability-tracking systems
  and supply-chain disclosures.
\item
  Read \texttt{get\_cpe\_identifier} in
  \texttt{src/spotforecast2\_safe/utils/cpe.py} for use in automated
  workflows.
\item
  Run \texttt{uv\ run\ reuse\ lint} to confirm license and
  copyright-header compliance.
\item
  Run \texttt{make\ verify} to execute all of the above as one gate, and
  \texttt{make\ sbom} to regenerate the CycloneDX 1.5 software bill of
  materials with the project CPE injected on the root component.
\item
  Run \texttt{verify\_audit\_log} from
  \texttt{spotforecast2\_safe.manager} against a retained audit log
  file to confirm its SHA-256 hash chain is intact; if no
  \texttt{AuditAnchor} is presented, the returned report explicitly
  states that external anchoring was not established rather than
  printing an unqualified pass.
\item
  Run \texttt{make\ anchor-verify\ LOG=<file-or-directory>} to
  re-verify a previously anchored log fully offline: it checks the
  stored RFC 3161 token against the stored TSA certificate chain,
  then confirms the file's current chain still extends the anchored
  head hash. See the Anchoring Runbook (\texttt{docs/anchoring.qmd})
  for the full workflow, including \texttt{make\ anchor} (which
  requires an explicit \texttt{TSA\_URL}; there is no default).
\end{enumerate}

\subsubsection{Verification environment}\label{verification-environment-en}

Steps 1 to 8 run on a maintainer workstation. This package is no longer
built or tested by a hosted continuous-integration service, so there is
no independent clean-room execution of the test suite on a second
platform, and released artefacts carry no Sigstore signature or
transparency-log entry. The lockfile remains cross-platform:
\texttt{{[}tool.uv{]}\ required-environments} forces dependency
resolution for macOS (arm64, x86-64) and Linux x86-64, so the resolved
versions are guaranteed to have wheels on each target even though the
tests execute on only one of them. The audit-log hash chain itself is
unkeyed and unanchored by this package: it detects edits, deletions,
and reordering within a retained file, but an adversary able to
rewrite the whole file can recompute it, so external anchoring of the
head hash remains the operator's responsibility. Auditors who require
independent verification should run the steps above in their own
environment.

\subsection{Citation, Authors, and
Contact}\label{a.13-citation-authors-and-contact-en}

Maintainer: Thomas Bartz-Beielstein, ORCID
\href{https://orcid.org/0000-0002-5938-5158}{0000-0002-5938-5158},
\texttt{bartzbeielstein@gmail.com}.

\begin{Shaded}
\begin{Highlighting}[]
\VariableTok{@misc}\NormalTok{\{}\OtherTok{spotforecast2safe}\NormalTok{,}
  \DataTypeTok{author}\NormalTok{       = \{Bartz{-}Beielstein, Thomas\},}
  \DataTypeTok{title}\NormalTok{        = \{\{spotforecast2{-}safe\}: Safety{-}critical Subset of \{spotforecast2\}\},}
  \DataTypeTok{year}\NormalTok{         = \{2026\},}
  \DataTypeTok{version}\NormalTok{      = \{27.0.0\},}
  \DataTypeTok{howpublished}\NormalTok{ = \{}\CharTok{\textbackslash{}url}\NormalTok{\{https://gitlab.git.nrw/thk{-}f10/spotsevenlab/spotforecast2{-}safe\}\},}
  \DataTypeTok{note}\NormalTok{         = \{Version 27.0.0, AGPL{-}3.0{-}or{-}later\}}
\NormalTok{\}}
\end{Highlighting}
\end{Shaded}

\textbf{APA}: Bartz-Beielstein, T. (2026). \emph{spotforecast2-safe:
Safety-critical subset of spotforecast2} (Version 27.0.0) {[}Computer
software{]}.
\url{https://gitlab.git.nrw/thk-f10/spotsevenlab/spotforecast2-safe}

The technical report (\texttt{bart26h/index.qmd}) is the long-form
reference for design rationale, compliance mapping, and evaluation
protocol.

\subsection{Disclaimer and
Liability}\label{a.14-disclaimer-and-liability-en}

\textbf{Limitation of liability.} This library is designed with safety
principles and deterministic logic, but it is provided as is, without
warranty of any kind. The authors and contributors accept no liability
for any direct or indirect damage, system failure, or financial loss
arising from its use.

It is the sole responsibility of the system integrator to perform full
system-level safety validation, for example under ISO 26262, IEC 61508,
or the EU AI Act, before deploying this software in a production or
safety-critical environment.

\end{tcolorbox}

\section{External validation}\label{sec-external-validation-en}

\subsection{External validation with
Deepchecks}\label{sec-deepchecks-en}

The feature matrix made up of lags and the basis-function encoder
(RBF, \texttt{RepeatingBasisFunction}) is row-wise
independent. This allows
\texttt{deepchecks.tabular} to be applied as an
external validation stage to \texttt{(X\_train,\ y\_train\_vec)} and
to an evaluation block constructed analogously. The following
listing sketches four representative checks. It is deliberately
\emph{not executed} as part of the document build, because
Deepchecks brings its own transitive dependency space (among others
\texttt{plotly}, \texttt{pkg\textbackslash{}\_resources}, specific
\texttt{scikit-learn} scorer names) that collides with the minimal
dependency closure of \texttt{spotforecast2-safe}. The consequence
following from the placement in Chapter~\ref{sec-related-en} is an
organisational one: Deepchecks is operated by a \emph{downstream}
job that brings its own environment and whose outputs (HTML reports
and reduced JSON summaries) are fed as audit artefacts into the
evidence chain.

\begin{Shaded}
\begin{Highlighting}[]
\CommentTok{\# Non-executing listing -- Deepchecks as an external validation stage.}
\CommentTok{\# Run in a separate environment with its own dependency pins.}

\ImportTok{from}\NormalTok{ deepchecks.tabular }\ImportTok{import}\NormalTok{ Dataset}
\ImportTok{from}\NormalTok{ deepchecks.tabular.checks }\ImportTok{import}\NormalTok{ (}
\NormalTok{    FeatureDrift,}
\NormalTok{    TrainTestPerformance,}
\NormalTok{    RegressionErrorDistribution,}
\NormalTok{    WeakSegmentsPerformance,}
\NormalTok{)}

\CommentTok{\# Build the feature matrix for the evaluation range analogously to X\_train}
\NormalTok{X\_eval\_rows, y\_eval\_rows }\OperatorTok{=}\NormalTok{ [], []}
\ControlFlowTok{for}\NormalTok{ t }\KeywordTok{in} \BuiltInTok{range}\NormalTok{(}\BuiltInTok{len}\NormalTok{(y\_train), }\BuiltInTok{len}\NormalTok{(y\_all)):}
\NormalTok{    X\_eval\_rows.append(}
\NormalTok{        np.concatenate([y\_all.values[t }\OperatorTok{{-}}\NormalTok{ lags : t], exog\_rbf.iloc[t].values])}
\NormalTok{    )}
\NormalTok{    y\_eval\_rows.append(y\_all.values[t])}
\NormalTok{X\_eval }\OperatorTok{=}\NormalTok{ pd.DataFrame(np.vstack(X\_eval\_rows), columns}\OperatorTok{=}\NormalTok{feature\_names)}
\NormalTok{y\_eval\_vec }\OperatorTok{=}\NormalTok{ pd.Series(y\_eval\_rows, name}\OperatorTok{=}\StringTok{"load"}\NormalTok{)}

\NormalTok{train\_tbl }\OperatorTok{=}\NormalTok{ pd.concat(}
\NormalTok{    [X\_train.reset\_index(drop}\OperatorTok{=}\VariableTok{True}\NormalTok{), y\_train\_vec.reset\_index(drop}\OperatorTok{=}\VariableTok{True}\NormalTok{)], axis}\OperatorTok{=}\DecValTok{1}
\NormalTok{)}
\NormalTok{eval\_tbl }\OperatorTok{=}\NormalTok{ pd.concat(}
\NormalTok{    [X\_eval.reset\_index(drop}\OperatorTok{=}\VariableTok{True}\NormalTok{), y\_eval\_vec.reset\_index(drop}\OperatorTok{=}\VariableTok{True}\NormalTok{)], axis}\OperatorTok{=}\DecValTok{1}
\NormalTok{)}
\NormalTok{train\_ds }\OperatorTok{=}\NormalTok{ Dataset(train\_tbl, label}\OperatorTok{=}\StringTok{"load"}\NormalTok{, cat\_features}\OperatorTok{=}\NormalTok{[])}
\NormalTok{eval\_ds }\OperatorTok{=}\NormalTok{ Dataset(eval\_tbl, label}\OperatorTok{=}\StringTok{"load"}\NormalTok{, cat\_features}\OperatorTok{=}\NormalTok{[])}

\CommentTok{\# 1. Feature drift between training and evaluation windows}
\NormalTok{fd }\OperatorTok{=}\NormalTok{ FeatureDrift().run(train\_ds, eval\_ds)}
\CommentTok{\# fd.value: \{feature\_name: \{"Drift score": float, "Method": str\}\}}

\CommentTok{\# 2. Train-vs-test performance -- overfitting indicator}
\NormalTok{ttp }\OperatorTok{=}\NormalTok{ TrainTestPerformance().run(train\_ds, eval\_ds, model)}
\CommentTok{\# ttp.value: DataFrame with columns Dataset/Metric/Value}

\CommentTok{\# 3. Residual distribution with kurtosis (complement to fig-residuals)}
\NormalTok{red }\OperatorTok{=}\NormalTok{ RegressionErrorDistribution().run(eval\_ds, model)}
\CommentTok{\# red.value: float (kurtosis of the residuals)}

\CommentTok{\# 4. Weak segments of the model}
\NormalTok{wsp }\OperatorTok{=}\NormalTok{ WeakSegmentsPerformance().run(eval\_ds, model)}
\CommentTok{\# wsp.value["weak\_segments\_list"]: DataFrame of the most unfavourable feature ranges}
\end{Highlighting}
\end{Shaded}

Time-series phenomena (autocorrelation of neighbouring lags and
the expected trend or seasonal drift between training and
evaluation windows) are flagged by Deepchecks as warnings and
must be classified domain-specifically by auditors.
Rolling-origin correctness (Figure~\ref{fig-folds-en}), determinism
(CR-2, Chapter~\ref{sec-principles-en}), and fail-safe
behaviour (CR-3, Chapter~\ref{sec-principles-en}) lie
outside the Deepchecks scope and are covered by the package's
executable docstring examples (Chapter~\ref{sec-testen-en}) and the
CI workflow.

\subsection{External validation with Evidently AI}\label{sec-evidently-en}

\texttt{Evidently~AI} covers the same tabular validation role as
Deepchecks but shifts the focus from pre-production validation to
production monitoring: reports are typically computed periodically
against a \emph{reference window} and feed dashboards (Prometheus,
Grafana, MLflow). For the regression stage of the example, three
presets are relevant: \texttt{DataDriftPreset} and
\texttt{TargetDriftPreset} for distribution shifts between training
and evaluation block, \texttt{RegressionPreset} for RMSE, MAE,
residual distribution, predicted-vs-actual depiction, and a
per-feature-broken-down error-bias table. One advantage over
Deepchecks: Evidently recognises, via a
\texttt{ColumnMapping(datetime=...)}, the time order of the
evaluation series and plots residuals time-annotated, which makes
seasonal error patterns visible that Figure~\ref{fig-forecast24-en}
and Figure~\ref{fig-residuals-en} taken together do not reveal.
The following listing is, for the same reasons as the Deepchecks
counterpart, \emph{not executed} as part of the build: Evidently
brings its own transitive dependency space (among others
\texttt{plotly}, \texttt{scipy}, and various visualisation and
text-processing packages) that collides with the minimal
dependency closure of \texttt{spotforecast2-safe}.

\begin{Shaded}
\begin{Highlighting}[]
\CommentTok{\# Non-executing listing -- Evidently AI as an external monitoring stage.}
\CommentTok{\# Run in a separate environment; X\_train, X\_eval, y\_train\_vec, y\_eval\_vec}
\CommentTok{\# and model are constructed in the build context of this section.}

\ImportTok{from}\NormalTok{ evidently.report }\ImportTok{import}\NormalTok{ Report}
\ImportTok{from}\NormalTok{ evidently.metric\_preset }\ImportTok{import}\NormalTok{ (}
\NormalTok{    DataDriftPreset,}
\NormalTok{    TargetDriftPreset,}
\NormalTok{    RegressionPreset,}
\NormalTok{)}
\ImportTok{from}\NormalTok{ evidently.pipeline.column\_mapping }\ImportTok{import}\NormalTok{ ColumnMapping}

\CommentTok{\# Reference and comparison windows with target and model prediction}
\NormalTok{reference }\OperatorTok{=}\NormalTok{ X\_train.reset\_index(drop}\OperatorTok{=}\VariableTok{True}\NormalTok{).copy()}
\NormalTok{reference[}\StringTok{"target"}\NormalTok{] }\OperatorTok{=}\NormalTok{ y\_train\_vec.reset\_index(drop}\OperatorTok{=}\VariableTok{True}\NormalTok{).values}
\NormalTok{reference[}\StringTok{"prediction"}\NormalTok{] }\OperatorTok{=}\NormalTok{ model.predict(X\_train)}
\NormalTok{reference[}\StringTok{"timestamp"}\NormalTok{] }\OperatorTok{=}\NormalTok{ y\_train.index[lags:]  }\CommentTok{\# consistent with X\_train rows}

\NormalTok{current }\OperatorTok{=}\NormalTok{ X\_eval.reset\_index(drop}\OperatorTok{=}\VariableTok{True}\NormalTok{).copy()}
\NormalTok{current[}\StringTok{"target"}\NormalTok{] }\OperatorTok{=}\NormalTok{ y\_eval\_vec.reset\_index(drop}\OperatorTok{=}\VariableTok{True}\NormalTok{).values}
\NormalTok{current[}\StringTok{"prediction"}\NormalTok{] }\OperatorTok{=}\NormalTok{ model.predict(X\_eval)}
\NormalTok{current[}\StringTok{"timestamp"}\NormalTok{] }\OperatorTok{=}\NormalTok{ y\_eval.index}

\NormalTok{column\_mapping }\OperatorTok{=}\NormalTok{ ColumnMapping(}
\NormalTok{    target}\OperatorTok{=}\StringTok{"target"}\NormalTok{,}
\NormalTok{    prediction}\OperatorTok{=}\StringTok{"prediction"}\NormalTok{,}
\NormalTok{    datetime}\OperatorTok{=}\StringTok{"timestamp"}\NormalTok{,}
\NormalTok{    numerical\_features}\OperatorTok{=}\NormalTok{feature\_names,}
\NormalTok{)}

\CommentTok{\# Report 1: data and target drift (training vs. evaluation window)}
\NormalTok{drift\_report }\OperatorTok{=}\NormalTok{ Report(metrics}\OperatorTok{=}\NormalTok{[DataDriftPreset(), TargetDriftPreset()])}
\NormalTok{drift\_report.run(}
\NormalTok{    reference\_data}\OperatorTok{=}\NormalTok{reference,}
\NormalTok{    current\_data}\OperatorTok{=}\NormalTok{current,}
\NormalTok{    column\_mapping}\OperatorTok{=}\NormalTok{column\_mapping,}
\NormalTok{)}
\CommentTok{\# drift\_report.as\_dict()["metrics"]           -\textgreater{} JSON summary}
\CommentTok{\# drift\_report.save\_html("evidently\_drift.html")}

\CommentTok{\# Report 2: regression quality with a time-resolved residual plot}
\NormalTok{regression\_report }\OperatorTok{=}\NormalTok{ Report(metrics}\OperatorTok{=}\NormalTok{[RegressionPreset()])}
\NormalTok{regression\_report.run(}
\NormalTok{    reference\_data}\OperatorTok{=}\NormalTok{reference,}
\NormalTok{    current\_data}\OperatorTok{=}\NormalTok{current,}
\NormalTok{    column\_mapping}\OperatorTok{=}\NormalTok{column\_mapping,}
\NormalTok{)}
\CommentTok{\# regression\_report.save\_html("evidently\_regression.html")}
\end{Highlighting}
\end{Shaded}

As with Deepchecks, rolling-origin correctness, determinism, and
fail-safe behaviour remain outside the Evidently scope and are
covered by the package-internal mechanisms
(Chapter~\ref{sec-testen-en}, Chapter~\ref{sec-principles-en}). The
JSON summaries from \texttt{as\_dict()} are suitable as
machine-readable audit artefacts for Art.~15 of the EU AI Act; the
HTML reports as a human-readable supplement to the model card.

The RMSE reported in step 7 is given in the
natural units of the synthetic series. The example presented here
is deliberately kept without weather features so that document
generation remains deterministic and offline-capable;
production-ready results require the \texttt{weather\_client}
adapter together with live access to Open-Meteo
\citep{openmeteo24}.

\clearpage
\selectlanguage{ngerman}
\renewcommand*\figurename{Abbildung}
\renewcommand*\tablename{Tabelle}
\renewcommand*\examplename{Beispiel}
\renewcommand{\thesection}{\arabic{section}}
\setcounter{section}{0}
\setcounter{subsection}{0}
\setcounter{subsubsection}{0}

\begin{zusammenfassung}
Mit \texttt{spotforecast2-safe} wird erstmals ein integrierter
\emph{Compliance-by-Design}-Ansatz für die Python-basierte Punktprognose
von Zeitreihen in sicherheitskritischen Umgebungen vorgestellt. Eine
Sichtung der einschlägigen Open-Source-Werkzeuge zeigt, dass bestehende
Compliance-Lösungen durchgängig außerhalb der zu nutzenden Bibliothek
wirken, z.B. als Scanner, Template oder Runtime-Schicht.
\texttt{spotforecast2-safe} verfolgt den umgekehrten Ansatz und
verankert die Anforderungen der Verordnung (EU) 2024/1689 (KI-VO), der
IEC 61508, der Normreihe ISA/IEC 62443 sowie der
Cyberresilienz-Verordnung innerhalb der Bibliothek: in
Programmierschnittstellen-Verträgen, Persistenzformaten und der
Prüfkette des Release-Verfahrens. Operationalisiert wird dieser Ansatz
durch vier nicht verhandelbare Code-Entwicklungsregeln (zero dead code,
deterministische Verarbeitung, fail-safe-Verarbeitung, minimale
Abhängigkeiten) sowie die zugehörigen Prozessregeln (Model Card,
ausführbare Docstrings, dokumentiertes Release-Verfahren,
Common-Platform-Enumeration- (CPE-) Kennung, REUSE-konforme
Lizenzierung, strukturiertes Audit-Protokoll). Interaktive
Visualisierung, Hyperparameter-Tuning und automatisiertes maschinelles
Lernen sowie Deep-Learning- und Large-Language-Model-Backends sind
bewusst ausgeschlossen, weil jede dieser Komponenten entweder die
Angriffsfläche vergrößert, Nichtdeterminismus einführt oder die
Reproduzierbarkeit beeinträchtigt. Jeder für die Bibliothek einschlägige
Artikel der KI-VO wird ausführlich erörtert. Eine bidirektionale
Rückverfolgungsmatrix bildet jede regulatorische Bestimmung auf den
Mechanismus im Code ab. Ein durchgängiges End-to-End-Beispiel zur
Vorhersage von Stromerzeugung, -übertragung und -verbrauch im
europäischen Markt zeigt die Anwendung. Das Paket ist quelloffen unter
Affero General Public License (AGPL) 3.0-or-later verfügbar.

\textbf{Schlüsselwörter:} Compliance-by-Design, Künstliche Intelligenz,
Kritische Infrastruktur, Open Source, Zeitreihenprognose,
Cyberresilienz-Verordnung, Legal-Requirements-Engineering,
sicherheitskritisches maschinelles Lernen, KI-Verordnung,
deterministisches Rechnen, skforecast, spotforecast2, Legal Aspects of
Computing, Software Engineering, Software Safety, Software Security
\end{zusammenfassung}

\bigskip
\noindent\textit{Hinweis:} Dies ist die deutsche Originalfassung des
Beitrags. Die englische Fassung (\emph{Time-Series Forecasting in
Safety-Critical Environments}) findet sich oberhalb. Beide Fassungen
teilen sich ein gemeinsames Literaturverzeichnis, das einmalig am Ende
des Dokuments gedruckt wird.

\section{Einleitung}\label{sec-introduction}

Die Praxis der Softwareentwicklung im Bereich der künstlichen
Intelligenz (KI\footnote{Im Folgenden verwenden wir KI als Oberbegriff
  für maschinelles Lernen, Lernen mit tiefen neuronalen Netzen sowie
  Sprachmodellen.}) und im Bereich der kritischen
Infrastruktur\footnote{Die Definition von KRITIS (Kritische
  Infrastrukturen) erfolgt nach §2 des Gesetzes über das Bundesamt für
  Sicherheit in der Informationstechnik und über die Sicherheit in der
  Informationstechnik von Einrichtungen (BSI-Gesetz - BSIG)
  \citep{deut25a} und §2 des Dachgesetzes zur Stärkung der physischen
  Resilienz kritischer Anlagen (KRITIS-Dachgesetz - KRITISDachG) und
  IT-Sicherheitsgesetz 2.0 und dem Gesetz zur Umsetzung der
  NIS-2-Richtlinie (NIS: Netz- und Informationssicherheit), also eine
  Dienstleistung zur Versorgung der Allgemeinheit in den Sektoren, deren
  Ausfall oder Beeinträchtigung zu erheblichen Versorgungsengpässen oder
  zur Gefährdung der öffentlichen Sicherheit führen würde
  \citep{kritisDachG26}.} wurden bislang in weitgehend getrennten
Software-Ökosystemen entwickelt. Die KI-basierte Softwareentwicklung zur
Zeitreihenprognose bevorzugt funktionsreiche Bibliotheken, die
Visualisierung, Hyperparametersuche, probabilistische
Deep-Learning-Backends und interaktive Dashboards in einem einzigen
installierbaren Paket zusammenführen. Diese umfangreichen Bibliotheken
sind für exploratives Arbeiten und innovatives Forschen vorzüglich
geeignet. Dieser Funktionsreichtum wird jedoch zum Nachteil, sobald das
daraus hervorgegangene Prognosemodell in einer Umgebung eingesetzt wird,
die einer Auditierung für sicherheitskritische Systeme standhalten muss.
Dies wird im Folgenden als ``inhärenter Zielkonflikt zwischen Forschung
und Sicherheit'' bezeichnet, weil die Anforderungen an die
Softwareentwicklung in der Forschung und in der kritischen Infrastruktur
unterschiedlich sind.

Beispielsweise erfordern Visualisierungstools die Einbindung einer
größeren Anzahl zusätzlicher Grafikbibliotheken. Vergleichbares gilt für
die Hyperparameter-Tuning-Schicht und für Deep-Learning-Modelle, die
zudem häufig sehr kurze Release-Zyklen aufweisen, so dass die
Langzeit-Reproduzierbarkeit, um deren Prüfung die Regulierungsstellen
ersuchen, nicht einfach zu gewährleisten ist.

Eine konkrete Sichtung der Python-Prognoselandschaft des Jahres 2026
bestätigt dieses Spannungsverhältnis. Die drei dominierenden
Python-Bibliotheken \texttt{sktime} \citep{loni19a}, \texttt{Darts}
\citep{herz22a} und \texttt{skforecast} \citep{scip24a} beinhalten
umfangreiche Funktionalitäten, die über die reine Prognose hinausgehen.
Die klassischen Bibliotheken (\texttt{statsmodels}, \texttt{pmdarima})
fallen schlanker aus, bieten jedoch weder eine dezidierte
Fail-safe-Semantik für ungültige Eingaben noch einen fixierten
Bezeichner nach der \emph{Common Platform Enumeration} (CPE) zur
Schwachstellennachverfolgung.

\begin{figure}

\centering{

\pandocbounded{\includegraphics[keepaspectratio]{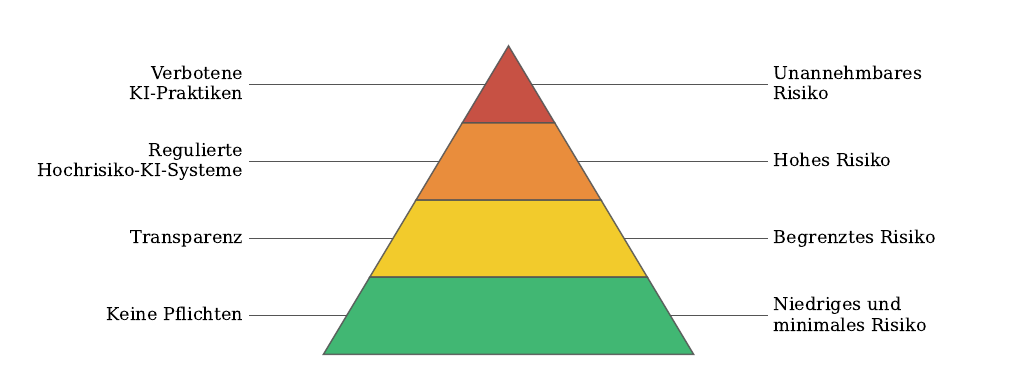}}

}

\caption{\label{fig-risk-pyramid}Die vier Risikostufen der KI-VO in der
kanonischen Pyramiden-Darstellung \citep{euro24a}. Mit der Risikohöhe
steigt die regulatorische Eingriffstiefe: Verbot an der Spitze, keine
Pflichten an der Basis. \texttt{spotforecast2-safe} ist für den Einsatz
in Systemen der zweitobersten Stufe (\emph{Hochrisiko}) konzipiert.}

\end{figure}

Wenn diese Softwarebibliotheken in sicherheitskritischen Umgebungen zum
Einsatz kommen, dann müssen die regulatorischen Anforderungen
berücksichtigt werden, die sich u.a. aus der Verordnung (EU) 2024/1689
über künstliche Intelligenz (nachfolgend als ``KI-VO'' bezeichnet,
englisch EU AI Act) ergeben. Die KI-VO verfolgt hierbei einen
\emph{risikobasierten} Ansatz: Der regulatorische Eingriff ist auf den
konkreten Risikograd zugeschnitten, nicht auf die Technologie als solche
\citep{stet24a}. Abbildung~\ref{fig-risk-pyramid} ordnet die vier
resultierenden Stufen (unannehmbares, hohes, begrenztes sowie niedriges
oder minimales Risiko) in der kanonischen Darstellung als \emph{Pyramide
der Risiken} an.

Grob vereinfacht definiert die KI-VO die vier Risikostufen wie folgt:

\begin{itemize}
\tightlist
\item
  \emph{Unannehmbares Risiko.} Die KI-VO verbietet manipulative
  KI-Systeme mit unterschwelligen Techniken, die gezielte Ausnutzung
  schutzwürdiger Gruppen, Social Scoring durch Behörden sowie die
  biometrische Echtzeit-Fernidentifizierung in öffentlich zugänglichen
  Räumen zu Strafverfolgungszwecken (mit engen Ausnahmen).
\item
  \emph{Hohes Risiko.} Hierunter fallen KI-Systeme als
  Sicherheitsbauteil oder als Produkt, das unter bestehender
  EU-Produktsicherheitsgesetzgebung geregelt ist (Spielzeug, Luftfahrt,
  Kraftfahrzeuge, Medizinprodukte, Aufzüge), sowie Systeme in den in
  Anhang III der KI-VO benannten Einsatzgebieten. Anbieter müssen eine
  Ex-ante-Konformitätsbewertung durchführen, das System in einer
  EU-Datenbank registrieren und sieben Anforderungen umsetzen: EU-1.
  menschliche Aufsicht, EU-2. technische Robustheit und Sicherheit,
  EU-3. Privatheit und Daten-Governance, EU-4. Transparenz, EU-5.
  Diversität sowie Nichtdiskriminierung und Fairness, EU-6. soziales und
  ökologisches Wohlergehen sowie EU-7. Rechenschaftspflicht. Diese
  sieben Anforderungen gehen auf die \emph{Ethics Guidelines for
  Trustworthy AI} der High-Level Expert Group on Artificial Intelligence
  (AI-HLEG) zurück \citep{hleg19a} und sind durch die \emph{Assessment
  List for Trustworthy AI} (ALTAI) \citep{hleg20a} operationalisiert
  \citep{stet24a}.
\item
  \emph{Begrenztes Risiko.} Für Chatbots, Emotionserkennungssysteme,
  biometrische Kategorisierung sowie für Systeme, die Bild-, Audio- oder
  Videoinhalte erzeugen oder manipulieren (sog. \emph{Deepfakes}),
  gelten Transparenzpflichten: Nutzer müssen erkennen können, dass sie
  mit einem KI-System interagieren.
\item
  \emph{Niedriges oder minimales Risiko.} Hier gibt es keine Pflichten.
  Die KI-VO fördert freiwillige Verhaltenskodizes, mit denen Anbieter
  von Nicht-Hochrisiko-Systemen die Anforderungen an Hochrisiko-Systeme
  übernehmen können.
\end{itemize}

\begin{longtable}[]{@{}
  >{\raggedright\arraybackslash}p{(\linewidth - 4\tabcolsep) * \real{0.3529}}
  >{\raggedright\arraybackslash}p{(\linewidth - 4\tabcolsep) * \real{0.2353}}
  >{\raggedright\arraybackslash}p{(\linewidth - 4\tabcolsep) * \real{0.4118}}@{}}
\caption{Semantische Zuordnung der KI-VO-Artikel zu den sieben
Hochrisiko-Anforderungen. EU-5 (Diversität und Nichtdiskriminierung) und
EU-6 (soziales und ökologisches Wohlergehen) sind nicht durch eine
eigene KI-VO-Zeile adressiert, weil sie keine unmittelbare
Umsetzungspflicht für das hier vorgestellte Softwarepaket
auslösen.}\label{tbl-kivo-eu-mapping}\tabularnewline
\toprule\noalign{}
\begin{minipage}[b]{\linewidth}\raggedright
\textbf{KI-VO-Artikel}
\end{minipage} & \begin{minipage}[b]{\linewidth}\raggedright
\textbf{Hochrisiko-Anforderung}
\end{minipage} & \begin{minipage}[b]{\linewidth}\raggedright
\textbf{Begründung}
\end{minipage} \\
\midrule\noalign{}
\endfirsthead
\toprule\noalign{}
\begin{minipage}[b]{\linewidth}\raggedright
\textbf{KI-VO-Artikel}
\end{minipage} & \begin{minipage}[b]{\linewidth}\raggedright
\textbf{Hochrisiko-Anforderung}
\end{minipage} & \begin{minipage}[b]{\linewidth}\raggedright
\textbf{Begründung}
\end{minipage} \\
\midrule\noalign{}
\endhead
\bottomrule\noalign{}
\endlastfoot
Art. 9 Risikomanagement & EU-7 & Rechenschaftspflicht:
Risikomanagementsystem als prozessualer Nachweisrahmen \\
Art. 10 Daten-Governance & EU-3 & Privatheit und Daten-Governance \\
Art. 11 technische Dokumentation & EU-7 & Rechenschaftspflicht durch
Dokumentation \\
Art. 12 Aufzeichnungspflichten & EU-7 & Rechenschaftspflicht durch
Protokolle \\
Art. 13 Transparenz & EU-4 & Transparenz \\
Art. 14 menschliche Aufsicht & EU-1 & Menschliche Aufsicht \\
Art. 15 Genauigkeit, Robustheit, Cybersicherheit & EU-2 & Technische
Robustheit und Sicherheit \\
Art. 17 Qualitätsmanagementsystem & EU-7 & Rechenschaftspflicht
(Dach) \\
Art. 18--21 Aufzeichnungspflichten Anbieter & EU-7 &
Rechenschaftspflicht \\
Art. 72 Beobachtung nach Inverkehrbringen & EU-7 &
Rechenschaftspflicht \\
Art. 73 Meldung & EU-7 & Rechenschaftspflicht \\
\end{longtable}

Die in diesem Bericht vorgestellte Open-Source-Softwarebibliothek
\texttt{spotforecast2-safe} ist als Komponente für Prognoseanwendungen
in sicherheitskritischen Umgebungen konzipiert, etwa für die
Lastprognose im Netzbetrieb. Ob eine solche Anwendung ein
Hochrisiko-System im Sinne der KI-VO ist, hängt allerdings vom konkreten
Einsatz ab. Tabelle~\ref{tbl-kivo-eu-mapping} zeigt die semantische
Zuordnung der KI-VO-Artikel zu den sieben Hochrisiko-Anforderungen (EU-1
bis EU-7). Die KI-VO \citep{euro24a} verpflichtet den Anbieter eines
Hochrisiko-KI-Systems zum Nachweis der Anforderungen aus Kapitel III
Abschnitt 2 (Art. 16 lit. a KI-VO). Den Betreiber treffen die
eigenständigen Pflichten des Art. 26 KI-VO. Beim Einsatz der im Bereich
der KI-Forschung etablierten Softwarepakete erbt ein
Betreiber\footnote{Je nach Kontext übernimmt ein Betreiber auch
  zusätzlich die Rolle eines Anbieters.} keinen dieser Nachweise aus der
Softwarebibliothek selbst. Diese Nachweise müssen u.U. mit erheblichem
Aufwand neu erbracht werden.

Hingegen verlagert die Software \texttt{spotforecast2-safe} diese
Nachweisführung in die Softwarebibliothek hinein, sodass der Betreiber
sie erbt. Dazu werden die in Kapitel~\ref{sec-principles} beschriebenen
Entwurfsgrundsätze definiert. Die Disziplin, d.h. die Übersetzung
regulatorischer Bestimmungen in implementierbare, prüfbare
Softwareanforderungen bei gleichzeitiger Aufrechterhaltung einer
bidirektionalen Rückverfolgbarkeit zwischen jeder Bestimmung und dem sie
erfüllenden Mechanismus, ist als \emph{Legal-Requirements-Engineering}
(LRE) bekannt \citep{otto07a, brea08a}. Einen systematischen Überblick
über das Gebiet gibt \citet{kose24a}. LRE stellt sicher, dass alle
relevanten Gesetze, Richtlinien, Motive, Hintergrundinformationen und
Standards in die Softwareentwicklung integriert werden, um nachhaltige
Rechtskonformität zu gewährleisten \citep{hens24a}. LRE bedeutet, dass
bereits im Entwurfsstadium für eine KI die rechtlichen Erfordernisse von
Beginn an mit berücksichtigt werden sollten. Dieses sogenannte
\emph{Compliance-by-Design} erspart in späteren Stadien der Entwicklung
mühsame und teure Anpassungen. Kapitel~\ref{sec-lre} zeigt, wie die
Software \texttt{spotforecast2-safe} die LRE-Disziplin anwendet.

Compliance-Monitoring-Bibliotheken wie \texttt{Deepchecks}
\citep{chor22a} und \texttt{Evidently\ AI}\footnote{\url{https://github.com/evidentlyai/evidently}}
decken die Evaluations-Seite der Robustheitsanforderung nach Art. 15
KI-VO teilweise ab. Sie setzen aber auf einer bereits fertiggestellten
Prognosesoftware auf und verändern weder Herkunftsnachweis,
Determinismus noch den Abhängigkeitsfußabdruck der Software selbst.
Kapitel~\ref{sec-deepchecks} und Kapitel~\ref{sec-evidently} (im Anhang)
zeigen auf, wie die hier vorgestellte Software
\texttt{spotforecast2-safe} mit den
Compliance-Monitoring-Funktionalitäten von \texttt{Deepchecks} und
\texttt{Evidently\ AI} kombiniert werden kann.

Eine Softwarebibliothek, die auf einem Software-Stack mit Dutzenden
nicht fixierter Abhängigkeiten beruht, ist zwar prinzipiell nicht
unzertifizierbar. Die Kosten, den verpflichtenden Prüfpfad herzustellen,
wachsen mit jeder Abhängigkeit.

\begin{figure}

\centering{

\pandocbounded{\includegraphics[keepaspectratio]{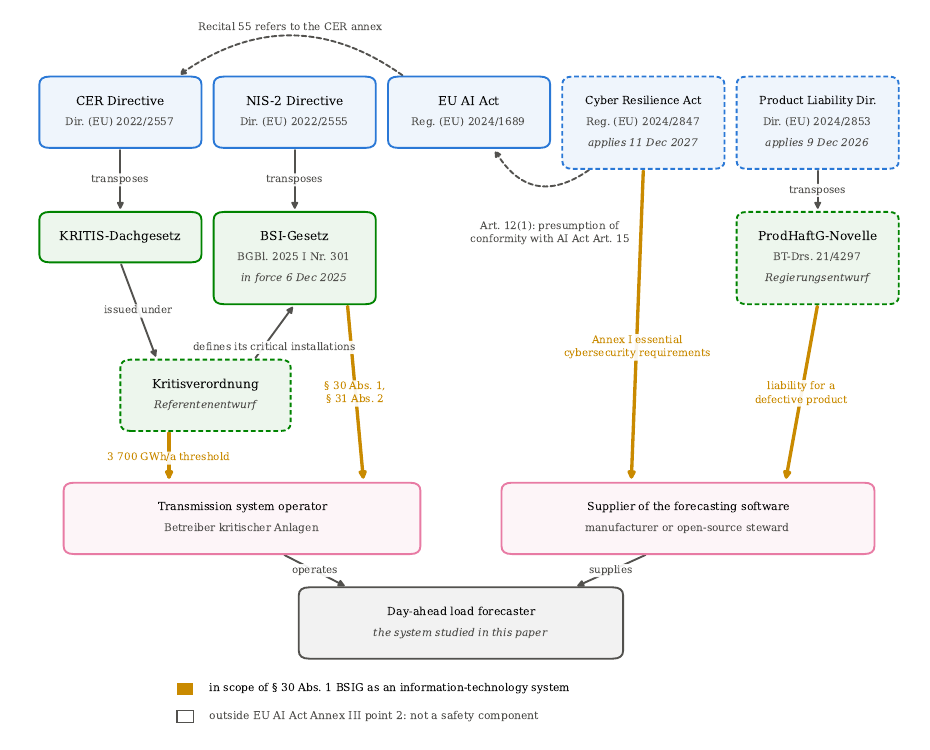}}

}

\caption{\label{fig-regulatory-map}Vereinfachte Darstellung ausgewählter
Rechtsakte, die auf ein KI-Lastprognosesystem einwirken, und die
Akteure, die sie binden. \emph{Blau} kennzeichnet Unionsrechtsakte,
\emph{Grün} ihre deutschen Entsprechungen, \emph{Rosa} die beiden
Akteure. Durchgezogene Pfeile bezeichnen eine Ableitung, d.h. eine in
deutsches Recht umgesetzte Richtlinie oder eine auf Grundlage eines
Gesetzes erlassene Rechtsverordnung. Die KI-VO und die
Cyberresilienz-Verordnung haben keine deutsche Entsprechung, weil eine
EU-Verordnung in EU-Mitgliedsstaaten unmittelbar gilt. Von ihnen führt
deshalb kein Pfeil nach unten. Gestrichelte Pfeile bezeichnen die
Bezugnahme eines Rechtsakts auf einen anderen. \emph{Ockerfarbene}
Pfeile bezeichnen den Ansatzpunkt einer Pflicht und sind jeweils mit der
Vorschrift beschriftet, die sie begründet. Eine \emph{gestrichelte}
Umrandung kennzeichnet einen Rechtsakt, der noch Entwurf oder noch nicht
anwendbar ist. Quelle: \citet{bart26o}.}

\end{figure}%

Die von der KI-VO definierten Anforderungen EU-1 bis EU-7 sind indessen
nicht die einzigen regulatorischen Kenngrößen, an denen sich eine
KI-Prognosebibliothek messen lassen muss, wenn sie in
sicherheitskritischen Umgebungen eingesetzt werden soll.
Abbildung~\ref{fig-regulatory-map}, entnommen aus \citet{bart26o}, zeigt
stark vereinfacht die maßgeblichen regulatorischen Instrumente. Im
Folgenden seien insbesondere die \emph{Cyberresilienz-Verordnung} (CRA,
engl. Cyber Resilience Act) und die \emph{Produkthaftungsrichtlinie}
(PLD, engl. Product Liability Directive) hervorgehoben, die beide in der
EU unmittelbar auf Software- und KI-Systeme wirken.

\emph{Cyberresilienz-Verordnung}. In industriellen Einsatzkontexten
läuft eine Prognosebibliothek als Software-Komponente auf Ebene 2 oder 3
des Purdue-Referenzmodells\footnote{Ein Referenzmodell für Automations-
  und Industrienetze.} (ISA-95 / IEC 62264), wo der dominierende
Cybersicherheitsrahmen die Normreihe ISA/IEC 62443 \citep{iec62443_1_1}
ist\footnote{Die Normreihe verteilt die Verantwortlichkeiten auf
  Anlagenbetreiber (62443-2-1 \citep{iec62443_2_1}), Dienstleister
  (62443-2-4 \citep{iec62443_2_4}), Systemintegratoren (62443-3-2
  \citep{iec62443_3_2}, 62443-3-3 \citep{iec62443_3_3}) sowie
  Komponentenhersteller (62443-4-1 \citep{iec62443_4_1}, 62443-4-2
  \citep{iec62443_4_2}). Die beiden letztgenannten Teile sind für einen
  Softwareanbieter unmittelbar einschlägig.}. Innerhalb der Europäischen
Union gilt die 62443-Reihe, insbesondere EN IEC 62443-4-1, als
aussichtsreiche Grundlage für die harmonisierten Normen, die nach dem
Normungsauftrag M/606 des Durchführungsbeschlusses C(2025)618 der
Kommission vom 3. Februar 2025 \citep{euM606} unter der mittlerweile in
Kraft getretenen Cyberresilienz-Verordnung \citep{euCRA24} zu erarbeiten
sind. Eine förmliche Benennung bestehender Normen enthält der Beschluss
nicht. Die 62443-Reihe prägt zudem die Risikomanagementpflichten nach
der Richtlinie (EU) 2022/2555 (``NIS-2-Richtlinie'') \citep{euNIS22}
sowie der Richtlinie zur Resilienz kritischer Einrichtungen
(CER-Richtlinie, engl. Critical Entities Resilience) \citep{euRCE22}.
\citet{fluc24a} bezeichnet 62443-2-1 sinngemäß als ``Äquivalent der
ISO/IEC 27001 für cyberphysische Systeme (CPS)''. Der für
\texttt{spotforecast2-safe} unmittelbar bindende Teil ist indessen
62443-4-1.

\emph{Zivilrechtliche Produkthaftung}. Über die öffentlich-rechtlichen
Konformitäts- und Cybersicherheitsrahmen hinaus erweitert die neue
Produkthaftungsrichtlinie (Richtlinie (EU) 2024/2853) \citep{euPLD24}
die verschuldensunabhängige Produkthaftung auf Software und KI-Systeme:
Diese sind erstmals ausdrücklich als ``Produkt'' im Sinne der Haftung
erfasst, einschließlich integrierter digitaler Dienste und
sicherheitsrelevanter Updates nach Inverkehrbringen. Die Richtlinie
ersetzt die Produkthaftungsrichtlinie 85/374/EWG von 1985 und gilt für
Produkte, die nach dem 9. Dezember 2026 in Verkehr gebracht oder in
Betrieb genommen werden. Bis zu diesem Datum müssen die Mitgliedstaaten
auch die nationale Umsetzung abgeschlossen haben. In Deutschland wird
sie durch den Entwurf eines Gesetzes zur Modernisierung des
Produkthaftungsrechts \citep{breg26a} umgesetzt, der sich zum
Berichtszeitpunkt im parlamentarischen Verfahren befindet
(Bundestags-Drucksache 21/4297, erste Lesung am 4. März 2026).
Open-Source-Software, die außerhalb einer Geschäftstätigkeit entwickelt
oder bereitgestellt wird, bleibt von der Produkthaftung ausgenommen. Für
eine unter der Affero General Public License (AGPL-3.0-or-later)
veröffentlichten Bibliothek wie \texttt{spotforecast2-safe} entsteht die
Haftung erst dort, wo sie im Zuge einer gewerblichen Tätigkeit auf dem
Unionsmarkt bereitgestellt wird. Zum aktuellen Zeitpunkt der Erstellung
dieses Berichts band keines der beiden Instrumente das Paket: Der CRA
ist erst ab dem 11. Dezember 2027 anzuwenden, und die
Produkthaftungsrichtlinie erfasst Produkte, die nach dem 9. Dezember
2026 in Verkehr gebracht werden.

\emph{Open-Source-Sonderregelungen.} Es gibt einen fast unüberschaubaren
Dschungel an Vorschriften, die ein Anbieter und Betreiber von Modellen
oder Systemen der KI eventuell zu berücksichtigen hätten. Als
prominenteste seien hier die Art. 50 und 51 KI-VO, die
Datenschutz-Grundverordnung (DSGVO) der EU, das Datengesetz (Data Act),
die Verletzung von Urheber-, Persönlichkeits- oder Beschäftigtenrechten
zu nennen. Dieser Bericht beschränkt sich bewusst für einen ersten
Aufschlag von freier und Open-Source-Software (Free and Open-Source
Software, FOSS) in KRITIS auf die eingangs erwähnten Regularien.

\emph{KI-VO.} Über die FOSS-Ausnahme der PLD hinaus enthalten die KI-VO
und der CRA jeweils eigene, voneinander unabhängige Sonderregelungen für
FOSS, die für \texttt{spotforecast2-safe} als unter AGPL-3.0-or-later
veröffentlichte Bibliothek einschlägig sind. Art. 2 Abs. 12 KI-VO
\citep{euro24a} nimmt unter freier oder Open-Source-Lizenz
veröffentlichte KI-Systeme grundsätzlich vom Geltungsbereich der
Verordnung aus, hebt diese Ausnahme aber für Hochrisiko-KI-Systeme im
Sinne von Anhang III, für verbotene Praktiken nach Art. 5 KI-VO und für
transparenzpflichtige Systeme nach Art. 50 KI-VO wieder auf. Ob diese
Ausnahme im typischen KRITIS-Einsatzfall einer Lastprognose entfällt,
hängt allein davon ab, ob das System im Sinne von Anhang III Nr. 2 KI-VO
als \emph{Sicherheitsbauteil} bei der Verwaltung und dem Betrieb der
Elektrizitätsversorgung eingesetzt wird. Erwägungsgrund 55 KI-VO fasst
diesen Begriff eng: Sicherheitsbauteile schützen unmittelbar die
physische Unversehrtheit kritischer Infrastruktur oder die Gesundheit
und Sicherheit von Personen, sind für das Funktionieren des Systems
selbst aber gerade nicht erforderlich. Als Beispiele können hier die
Überwachung des Wasserdrucks und der Betrieb von Feuermeldeanlagen in
Cloud-Computing-Zentren genannt werden. \emph{Eine
Day-ahead-Lastprognose erfüllt in der Regel keines der beiden Merkmale.
Eine solche Prognose steht damit neben der kritischen Infrastruktur,
nicht innerhalb ihres Sicherheitsperimeters.} Im Regelfall ist sie daher
kein Hochrisiko-System nach Anhang III Nr. 2, die FOSS-Ausnahme des Art.
2 Abs. 12 KI-VO bleibt anwendbar, und die in Kapitel~\ref{sec-ki-vo-9}
bis Kapitel~\ref{sec-ki-vo-15} erörterten materiellen Anforderungen
treffen \texttt{spotforecast2-safe} nicht kraft Gesetzes. Die
Beurteilung des konkreten Einsatzfalls bleibt gleichwohl Aufgabe des
Anbieters: Wird die Software-Bibliothek in ein System eingebettet, das
die Merkmale des Erwägungsgrundes 55 erfüllt, etwa in einen
automatischen Netz- oder Lastabwurfschutz\footnote{Netzschutz bezeichnet
  die Schutztechnik im Übertragungs- und Verteilnetz: Schutzrelais
  (Distanz-, Differential-, Überstromschutz), die einen Fehler wie einen
  Kurzschluss innerhalb von Millisekunden erkennen und die betroffene
  Leitung oder das Betriebsmittel über Leistungsschalter automatisch vom
  Netz trennen, bevor Menschen gefährdet werden oder Anlagen Schaden
  nehmen. Automatischer Lastabwurf ist die letzte Verteidigungslinie
  gegen einen Netzzusammenbruch: Fällt die Netzfrequenz unter kritische
  Schwellen, trennen Relais automatisch und stufenweise Verbraucherlast
  vom Netz, um das Gleichgewicht zwischen Erzeugung und Verbrauch
  wiederherzustellen und einen Blackout zu verhindern. Dies geschieht
  \emph{autonom}, ohne menschliche Entscheidung im Moment des Eingriffs.
  Beide Automatismen erfüllen die beiden Merkmale des Erwägungsgrundes
  55: Sie schützen unmittelbar die physische Unversehrtheit des Netzes
  und die Sicherheit von Personen, sind aber für das Funktionieren des
  Systems selbst nicht erforderlich, denn das Netz läuft im
  Normalbetrieb ohne jeglichen Einsatz des Schutzes. Dieser wirkt
  definitionsgemäß nur im Störfall. Sie wären daher Sicherheitsbauteile
  im Sinne von Anhang III Nr. 2 KI-VO, während die
  Day-ahead-Lastprognose spiegelbildlich liegt: Sie schützt nichts
  unmittelbar und ist für den Normalbetrieb erforderlich.}, so entfällt
die Ausnahme, und die Pflichten treffen den Anbieter dieses Systems.
Auch ohne diese Einbettung informiert eine Lastprognose die
Entscheidungen der Betreiber kritischer Anlagen, und Determinismus,
Reproduzierbarkeit und Auditierbarkeit sind genau die Eigenschaften, auf
denen ein Audit eines solchen Systems ruhen würde.
\texttt{spotforecast2-safe} erfüllt die genannten Anforderungen deshalb
bewusst \emph{im Vorgriff} und unabhängig davon, ob sie im Einzelfall
bereits geschuldet sind. Genau darin liegt der
Compliance-by-Design-Ansatz des Pakets. Eigenständig daneben tritt Art.
25 Abs. 4 Unterabs. 1 Satz 2 KI-VO, der die Pflicht zur schriftlichen
Wertschöpfungsketten-Vereinbarung zwischen Komponentenanbieter und
Anbieter eines Hochrisiko-Systems ausdrücklich nicht auf solche unter
freier oder Open-Source-Lizenz bereitgestellte Werkzeuge, Dienste,
Prozesse oder Komponenten erstreckt, sofern es sich nicht um Modelle mit
allgemeinem Verwendungszweck (General-Purpose AI, GPAI) handelt. Die
Bibliothek schuldet ihrem nachgelagerten Anbieter daher keine
vertraglich geregelte Kooperation, sondern stellt die
Mitwirkungs-Voraussetzungen über offene Provenienz, dokumentierte
Schnittstellen und reproduzierbare Releases zur Verfügung (siehe
Kapitel~\ref{sec-principles}). Die GPAI-spezifischen FOSS-Ausnahmen aus
Art. 53 Abs. 2 KI-VO und Art. 54 Abs. 6 KI-VO sind auf eine klassische
Regressionsbibliothek wie \texttt{spotforecast2-safe} nicht anwendbar,
weil \texttt{spotforecast2-safe} kein GPAI-Modell im Sinne von Art. 3
Nr. 63 KI-VO in den Verkehr bringt.

\emph{CRA.} Der CRA \citep{euCRA24} ergänzt diese Systematik um Art. 24
CRA, der für ``Open-Source Software Stewards'' (typischerweise
gemeinnützige oder gemeinwirtschaftliche Organisationen, die
FOSS-Projekte langfristig pflegen) gegenüber den Hersteller-Pflichten
der Art. 13 ff.~CRA abgesenkte Pflichten vorsieht (eine dokumentierte
Cybersicherheits-Policy, koordinierte Schwachstellen-Behandlung,
Kooperation mit der Marktüberwachung) und die Verhängung von
Verwaltungsbußgeldern mit Ausnahme der Meldepflichten ausschließt.

\emph{NIS-2.} Die NIS-2 \citep{euNIS22} sieht keine ausdrückliche
FOSS-Ausnahme vor. Ihre Lieferketten-Pflicht aus Art. 21 Abs. 2 lit. d
NIS-2 trifft den regulierten KRITIS-Betreiber und nicht den
Open-Source-Maintainer. Sie wirkt allerdings faktisch über vertragliche
Anforderungen in das FOSS-Ökosystem hinein. Die einschlägige
rechtswissenschaftliche Literatur arbeitet dieses Geflecht systematisch
auf \citep{colo25a, vsch25a, kike24a}.

\emph{Vertrauenswürdigkeit.} Eine eigenständige Forschungslinie
behandelt, \emph{wie} die Vertrauenswürdigkeit eines
Hochrisiko-KI-Systems zu beurteilen ist. \citet{stet24a} schlagen einen
Rahmen zur Vertrauenswürdigkeits-Assurance vor, der die Anforderungen
der KI-VO in einem sechsstufigen, iterativ ausgelegten Prozess in
prüfbare Eigenschaften zerlegt und die KI-System-Technik mit der
Assurance-Argumentation der funktionalen Sicherheit verbindet. Der
vorliegende Bericht ist mit diesem Rahmen vereinbar: Jede Prozessregel
in Kapitel~\ref{sec-principles} ist ein Argumentationsschritt, und die
Spalte \emph{Mechanismus} der Tabelle~\ref{tbl-compliance} ist die
Substantiation.

\emph{Künstliche Intelligenz und Grundrechte}. Der
rechtswissenschaftliche Hintergrund wird von \citet{raue25a}
aufgearbeitet, die die Wechselwirkung der KI-VO mit der Charta der
Grundrechte der Europäischen Union, dem Antidiskriminierungsrecht und
dem Datenschutzrecht untersucht. Diese Arbeit ist die maßgebliche
Bezugsquelle, wenn ein Betreiber die hier dokumentierten technischen
Nachweise in eine Grundrechte-Folgenabschätzung (Fundamental Rights
Impact Assessment) nach Art. 27 KI-VO zu übertragen hat.

Tabelle~\ref{tbl-normen-einleitung} fasst die in den vorstehenden
Absätzen angeführten Rechtsakte, Normen und Referenzmodelle nach
Regulierungsebene zusammen.

{

\begin{longtable}{p{0.15\textwidth} p{0.21\textwidth} p{0.56\textwidth}}

\caption{\label{tbl-normen-einleitung}Systematische Übersicht der angeführten Rechtsakte, Normen und Referenzmodelle. KI-VO = Verordnung (EU) 2024/1689; CRA = Verordnung (EU) 2024/2847 (Cyberresilienz-Verordnung); NIS-2-RL = Richtlinie (EU) 2022/2555; CER-RL = Richtlinie (EU) 2022/2557; CPS = cyberphysisches System; ISMS = Informationssicherheits-Managementsystem; IACS = Industrial Automation and Control Systems.}

\tabularnewline

\\
\hline
\textbf{Bereich} & \textbf{Rechtsakt / Norm} & \textbf{Gegenstand} \\
\hline
\endfirsthead
\multicolumn{3}{l}{\itshape (Fortsetzung von \tablename~\ref{tbl-normen-einleitung})}\\
\hline
\textbf{Bereich} & \textbf{Rechtsakt / Norm} & \textbf{Gegenstand} \\
\hline
\endhead
\hline
\multicolumn{3}{r}{\itshape (Fortsetzung auf der nächsten Seite)}\\
\endfoot
\hline
\endlastfoot
EU-Regulierung (KI) & KI-VO & Hochrisiko-KI-Systeme: Daten-Governance, Transparenz, Genauigkeit, Robustheit und Cybersicherheit. \\
EU-Regulierung (Produktsicherheit) & Cyberresilienz-Verordnung (CRA) & Produkte mit digitalen Elementen: grundlegende Cybersicherheitsanforderungen (Anhang~I Teil~I) und Schwachstellenbehandlung über den Unterstützungszeitraum von grundsätzlich mindestens fünf Jahren (Art.~13 Abs.~8, Anhang~I Teil~II). \\
EU-Regulierung (Netz- und Informationssicherheit) & NIS-2-Richtlinie & Risikomanagementpflichten wesentlicher und wichtiger Einrichtungen. \\
EU-Regulierung (Resilienz) & CER-Richtlinie & Resilienz kritischer Einrichtungen (Energie, Verkehr, Wasser u.\,a.). \\
EU-Regulierung (Zivilhaftung) & Produkthaftungs-Richtlinie (PLD, Richtlinie (EU) 2024/2853) & Verschuldensunabhängige Haftung für fehlerhafte Produkte; erstmals ausdrücklich Software und KI-Systeme erfasst; Umsetzungs- und Geltungsbeginn 9.\,Dezember 2026. Deutsche Umsetzung durch den Entwurf eines Gesetzes zur Modernisierung des Produkthaftungsrechts (BT-Drs.\ 21/4297). \\
KI-Ethik-Leitlinien & Ethics Guidelines for Trustworthy AI \citep{hleg19a} & Drei-Komponenten-Rahmen (rechtlich, ethisch, robust); sieben Anforderungen, die den Hochrisiko-Pflichten EU-1 bis EU-7 der KI-VO zugrunde liegen. \\
KI-Ethik-Leitlinien & Assessment List for Trustworthy AI \citep[ALTAI,][]{hleg20a} & Operationalisierung der sieben Anforderungen als Selbstbewertungs-Checkliste. \\
KI-Normen (ISO/IEC) & Sechs ISO/IEC-Kernnormen zur KI (TS 4213, 5338, 23894, 24027, 38507, 42001) & Operationalisierung der sieben Hochrisiko-Anforderungen durch technische und Management-System-Normen; detaillierte Auflistung in Tabelle~\ref{tbl-ki-vo-iso} und Tabelle~\ref{tbl-core-standards}. \\
Funktionale Sicherheit & IEC 61508-3 & Software sicherheitsbezogener elektrischer, elektronischer und programmierbarer Systeme; Grundlage der SIL-Einstufung. \\
Funktionale Sicherheit & ISO 26262-6 & Straßenfahrzeugspezifische Adaption von IEC 61508 auf Software-Ebene; Grundlage der ASIL-Einstufung. \\
Industrielle Cybersicherheit & ISA/IEC 62443-1-1 & Terminologie, Konzepte und Modelle der Normreihe. \\
Industrielle Cybersicherheit & ISA/IEC 62443-2-1 & Cybersicherheits-Managementsystem des Anlagenbetreibers (CPS-Pendant zu ISO/IEC 27001). \\
Industrielle Cybersicherheit & ISA/IEC 62443-2-4 & Sicherheitsanforderungen an IACS-Dienstleister. \\
Industrielle Cybersicherheit & ISA/IEC 62443-3-2 & Risikobewertung und Systemauslegung für IACS. \\
Industrielle Cybersicherheit & ISA/IEC 62443-3-3 & Systemweite Sicherheitsanforderungen (sieben Grundanforderungen, vier Sicherheitsstufen). \\
Industrielle Cybersicherheit & ISA/IEC 62443-4-1 & Sicherer Produktentwicklungszyklus des Komponentenherstellers (für \texttt{spotforecast2-safe} unmittelbar einschlägig). \\
Industrielle Cybersicherheit & ISA/IEC 62443-4-2 & Technische Sicherheitsanforderungen an Komponenten (Software-Application-Profil). \\
Referenzmodell & ISA-95 / IEC 62264 & Funktionale Hierarchie industrieller Automatisierung (Purdue-Modell, Ebenen 0--4). \\
Managementsystem & ISO/IEC 27001 & Informationssicherheits-Managementsystem (ISMS); im Text als Pendant zu 62443-2-1 herangezogen. \\
\hline

\end{longtable}

}

Das Paket \texttt{spotforecast2-safe} versucht, den eingangs erwähnten
inhärenten Zielkonflikt zwischen Forschung und Sicherheit zu lösen.
Ausgangspunkt für die Entwicklung von \texttt{spotforecast2-safe} ist
\texttt{skforecast} \citep{scip24a}. Ungefähr ein Drittel der
sicherheitskritischen Codezeilen sind ein direkter Port aus dem
\texttt{skforecast}-Quellcode. Ergänzender Code setzt die vier in
Kapitel~\ref{sec-code-rules} beschriebenen Code-Entwicklungsregeln um
(``No dead code'', ``Deterministische Transformationen'',
``Fail-safe-Verarbeitung'' und ``Minimaler Umfang'') sowie die vier in
Kapitel~\ref{sec-process-rules} beschriebenen Prozessregeln. Durch die
Berücksichtigung dieser Regeln ist der Funktionsumfang der Software
reduziert. Um diese gewollte Reduktion auszugleichen, wird
\texttt{spotforecast2-safe} als Teilmenge (``core'') des umfangreicheren
Open-Source-Schwesterpakets \texttt{spotforecast2} \citep{spotforecast2}
veröffentlicht. Das \texttt{spotforecast2} Paket erhält zusätzliche
Funktionen: interaktive Visualisierung auf Grundlage von \texttt{plotly}
und \texttt{matplotlib}, Hyperparametertuning mittels \texttt{optuna}
und \texttt{spotoptim}, sowie Modellerklärungswerkzeuge.

Der Rest dieses Berichts\footnote{Dieser Bericht erfüllt ähnliche
  Anforderungen wie die in ihm vorgestellte Software: Jeder Codeblock
  wird zum Zeitpunkt der Dokumentenerstellung ausgeführt und das gesamte
  Dokument inkl. Literaturverzeichnis steht unter Versionskontrolle.}
beschreibt die von \texttt{spotforecast2-safe} implementierte Teilmenge.
Der Bericht verfolgt drei Ziele. Erstens verortet er das Paket innerhalb
der Landschaft der Python-Zeitreihenwerkzeuge und erläutert, welche
Funktionen in der sicherheitskritischen Teilmenge aus welchem Grund
bewusst fehlen (Kapitel~\ref{sec-related}). Zweitens dokumentiert er die
Architektur (Kapitel~\ref{sec-architecture}), die formalen
Vorverarbeitungs- und Prognosealgorithmen
(Kapitel~\ref{sec-preprocessing} bis Kapitel~\ref{sec-testen}) sowie die
in den jüngsten Versionen ausgelieferten sicherheitsrelevanten
Entwurfsentscheidungen (Kapitel~\ref{sec-principles} bis
Kapitel~\ref{sec-safety}). Drittens bildet er die Entwurfsentscheidungen
auf einzelne Bestimmungen der KI-VO und der Normen zur funktionalen
Sicherheit ab (Kapitel~\ref{sec-ki-vo-andere-normen}) und schließt mit
einem kurzen Beispiel zur elektrischen Lastprognose
(Kapitel~\ref{sec-example}).

\section{Softwarepakete zur Zeitreihenprognose}\label{sec-related}

Das Python-Ökosystem der Zeitreihenprognose ist im vergangenen Jahrzehnt
stark gewachsen und hat auch Implementationen der statistischen
Programmiersprache R übernommen. Grundlegend ist das Paket
\texttt{statsmodels}, welches ARIMA-, SARIMAX- und
Exponentialglättungsmodelle in Anlehnung an die klassische Darstellung
von \citet{box15a} bereitstellt. \texttt{scikit-learn} \citep{pedr11a}
verzichtet bewusst auf zeitreihenspezifische Funktionen und delegiert
diese an nachgelagerte Bibliotheken. Die wichtigsten dieser Bibliotheken
werden kurz vorgestellt.

\begin{itemize}
\tightlist
\item
  \texttt{skforecast} \citep{scip24a, rodr25a} stellt eine
  \texttt{Forecaster}-Abstraktion bereit, welche jeden mit scikit-learn
  kompatiblen Regressor in einen rekursiven, direkten oder mehrreihigen
  Forecaster einhüllt. Es ist der engste konzeptionelle Verwandte des
  hier dokumentierten Pakets, weil \texttt{spotforecast2-safe} auf
  \texttt{skforecast} aufbaut. Wir übernehmen daraus die Architektur der
  rekursiven Prognose, den Rolling-Origin-Splitter und wesentliche Teile
  der Module \texttt{\_binner}, \texttt{\_differentiator} sowie die
  Residuen-Infrastruktur. Nicht übernommen werden das
  Visualisierungs-Subsystem (das auf \texttt{matplotlib} und
  \texttt{plotly} angewiesen ist) und das Hyperparameter-Tuning mittels
  \texttt{bayesian\_search\_forecaster} (das auf \texttt{optuna} beruht
  und nichtdeterministische Optimierungsspuren erzeugt). Diese Tools
  sind im Paket \texttt{spotforecast2} enthalten.
\item
  Die vereinheitlichte
  \texttt{Forecaster}/\texttt{BaseEstimator}-Programmierschnittstelle
  (Application Programming Interface, API) von \texttt{sktime}
  \citep{loni19a}, die eine Vielzahl von Modellen und
  Validierungsschemata bereitstellt, ist die umfangreichste aller hier
  betrachteten Bibliotheken. Allerdings ist die API nicht vollständig
  kompatibel mit der von \texttt{scikit-learn}, so dass die Integration
  von Regressoren aus diesem Ökosystem nicht ohne weiteres möglich ist.
  Zudem sind viele Modelle in \texttt{sktime} auf
  \texttt{statsmodels}-Implementationen angewiesen, welche nicht für die
  Anforderungen der funktionalen Sicherheit optimiert sind (z.B. keine
  fail-safe-Verarbeitung ungültiger Eingaben). Gerade die Breite des
  Funktionsumfangs vergrößert zudem die Abhängigkeits- und damit die
  Prüffläche.
\item
  Das Nixtla-Ökosystem, das auf eine Verbesserung des
  Laufzeitdurchsatzes abzielt, implementiert drei komplementäre
  Bibliotheken: \texttt{statsforecast} \citep{garz22a} stellt
  vektorisierte Implementierungen klassischer Modelle bereit und tauscht
  dabei Modellflexibilität gegen aggressive Laufzeitoptimierung mittels
  Numba \citep{lam15a}. \texttt{mlforecast} erweitert dieselbe API auf
  Gradient-Boosting-Verfahren. \texttt{neuralforecast} bietet
  Deep-Learning-Architekturen über eine einheitliche Schnittstelle. Das
  Deep-Learning-Backend verletzt die Regel des minimalen Umfangs
  unmittelbar.
\item
  \texttt{Darts} \citep{herz22a} führt eine vereinheitlichte API für
  verschiedene Zeitreihenmodelle ein. Der Installationsumfang wird von
  optionalen Abhängigkeiten dominiert, deren Verhalten die Verfügbarkeit
  von Modellklassen zum Importzeitpunkt beeinflusst. Ähnliches gilt für
  \texttt{tsai}, \texttt{Merlion} und \texttt{Kats}.
\end{itemize}

Keine der genannten Bibliotheken ist softwaretechnisch mangelhaft. Im
Gegenteil, sie stellen herausragende (Forschungs-) Software dar, sind
aber gute Beispiele für den oben erwähnten inhärenten Zielkonflikt
zwischen Forschung und Sicherheit. \texttt{spotforecast2-safe} versucht,
diesen Konflikt aufzulösen, indem Sicherheit \emph{per Design} von
Anfang an mitgedacht wird. Der Betreiber erbt die erforderlichen
Sicherheitsnachweise aus der Software und muss sie nicht selbst
nachträglich implementieren.

\section{\texorpdfstring{Das Paket
\texttt{spotforecast2-safe}}{Das Paket spotforecast2-safe}}\label{sec-package}

\subsection{Architektur}\label{sec-architecture}

Das \texttt{spotforecast2-safe} Paket ist in drei Schichten gegliedert.
Die unterste Schicht (\texttt{forecaster/}, \texttt{preprocessing/},
\texttt{splitter/}, \texttt{backtesting/}, \texttt{data/}) enthält die
Schätzer-Wrapper, Transformatoren, Splitter und den Backtesting-Treiber,
die am unmittelbarsten aus \texttt{skforecast} hervorgehen. Eine
mittlere Schicht (\texttt{manager/}) fasst einen Schätzer zusammen mit
einem \texttt{ExogBuilder}, einem Konfigurator, einem Logger und einer
Persistenz-Schicht zu einer Bereitstellungseinheit zusammen. Die oberste
Schicht (\texttt{multitask/}, \texttt{processing/} und \texttt{tasks/})
stellt End-to-End-Pipelines bereit: \texttt{multitask/} orchestriert
Mehrziel-Prognosen über eine Task-Hierarchie, deren Dispatcher
\texttt{MultiTask} zwischen Lazy-, Default-, Predict- und Aufräum-Läufen
wählt, \texttt{processing/} liefert die zugehörigen Vorhersage-,
Blending- und Bewertungsfunktionen, und \texttt{tasks/} komponiert die
unteren Schichten zu Konsolenskripten. Das Modul
\texttt{submission\_cli.py} ergänzt Prüf- und Abbruch-Helfer für
automatisierte Submissions-Workflows.

\dirtree{%
.1 src/spotforecast2\_safe/.
.2 forecaster/.
.3 recursive/, wrappers/.
.2 preprocessing/.
.3 repeating\_basis\_function.py.
.3 \_binner.py, \_differentiator.py.
.3 linearly\_interpolate\_ts.py.
.3 exog\_builder.py, imputation.py, outlier.py.
.2 splitter/.
.3 split\_ts\_cv.py, split\_one\_step.py.
.2 backtesting/.
.3 validation.py.
.2 manager/.
.3 logger.py, demo\_metrics.py, features.py.
.3 persistence.py, trainer.py, predictor.py.
.2 multitask/.
.3 base.py, multi.py, runner.py.
.3 strategies.py, guards.py.
.2 configurator/, data/, calendar/, stats/.
.2 processing/, tasks/.
.2 datasets/, downloader/, weather/.
.2 security/, utils/.
.2 exceptions.py, submission\_cli.py.
}

Die öffentliche Schnittstelle wird durch \texttt{\_\_all\_\_} auf der
obersten Paketebene festgelegt und umfasst zehn Symbole:
\texttt{Period}, \texttt{RepeatingBasisFunction}, \texttt{ExogBuilder},
\texttt{LinearlyInterpolateTS}, die vier
\texttt{ForecasterRecursive*}-Modellklassen
(\texttt{ForecasterRecursiveModel}, \texttt{ForecasterRecursiveLGBM},
\texttt{ForecasterRecursiveXGB}, \texttt{ForecasterRecursiveCatBoost}),
\texttt{ConfigEntsoe} sowie den Alias \texttt{Config}. Die Paketversion
ist zusätzlich als Modulattribut \texttt{\_\_version\_\_} abrufbar. Jede
API-brechende Änderung an diesen Symbolen ist per Vereinbarung an einen
Conventional-Commit-Betreff \texttt{feat!:} und an einen
Major-Versionssprung gebunden. Private Module\footnote{Alles mit
  führendem Unterstrich oder in einem nicht auf oberster Ebene
  re-exportierten Untermodul.} sind ausdrücklich instabil und von dieser
Zusicherung ausgenommen.

Das Hinzufügen eines neuen Schätzer-Wrappers erfolgt demnach im
Verzeichnis \texttt{forecaster/wrappers/}: Der Wrapper erbt von
\texttt{ForecasterRecursiveModel}, der selbst \texttt{fit},
\texttt{predict}, \texttt{backtest} sowie das Speichern und Laden
orchestriert. Der Wrapper ist ausschließlich dafür verantwortlich, den
Backend-Regressor und etwaige backend-spezifische Serialisierungshaken
zu deklarieren. Die exogene Merkmalsschicht, die
Rolling-Origin-Validierung und das Audit-Logging werden vererbt.

Das Schwesterpaket \texttt{spotforecast2} \citep{spotforecast2} (Version
10.9.0) deklariert \texttt{spotforecast2-safe} als Abhängigkeit und
steuert genau die Erweiterungen bei, die aus der sicheren Teilmenge
bewusst ausgeschlossen sind: die Auto-Tuning-Tasks (\texttt{OptunaTask},
\texttt{SpotOptimTask}) samt Suchräumen, Full-Varianten der vier
Forecaster-Klassen mit Optuna-Tuning und SHAP-Merkmalswichtigkeit,
Plot-Funktionen sowie zusätzliche Statistik- und
Modellselektions-Werkzeuge. Den vollständigen Prognosekern,
einschließlich des Abrufs von ENTSO-E-Daten (European Network of
Transmission System Operators for Electricity) und Wetterdaten
(\texttt{downloader/}, \texttt{weather/}), auf dem das Prüfnarrativ aus
Kapitel~\ref{sec-safety} aufbaut, enthält allein
\texttt{spotforecast2-safe}. Die Abhängigkeitsrichtung verläuft
einseitig: Der \texttt{spotforecast2-safe} Kern importiert nicht aus
\texttt{spotforecast2} und ist eigenständig installier- und testbar.
Diese Aufteilung ermöglicht es, die sichere Teilmenge in einem eigenen
Takt zu veröffentlichen und unabhängig zu zertifizieren, während
\texttt{spotforecast2} für den allgemeineren, nicht regulierten Einsatz
funktional eine Obermenge bildet.

\subsection{Vorverarbeitung}\label{sec-preprocessing}

Das Teilpaket \texttt{preprocessing} enthält vier Arten von
Transformatoren:

\begin{enumerate}
\def\labelenumi{\arabic{enumi}.}
\tightlist
\item
  Zyklische Kodierer überführen periodische Kalendermerkmale (Stunde des
  Tages, Tag der Woche, Tag des Jahres) in eine glatte Darstellung.
\item
  Interpolatoren rekonstruieren kurze Lücken unter ausdrücklicher
  Zusicherung.
\item
  Ein Quantilbinner diskretisiert ein stetiges Merkmal in gleich häufig
  belegte Bins.
\item
  Ein Differenzierer wendet einen Finite-Differenzen-Operator an.
\end{enumerate}

Alle vier Transformatoren erben von
\texttt{sklearn.base.TransformerMixin} und sind damit in einer
\texttt{Pipeline} komponierbar.

\subsection{Prognose}\label{sec-forecasting}

Die Prognoseschicht implementiert die rekursive Mehrschrittstrategie:
Ein einzelnes skalares Ein-Schritt-voraus-Modell wird trainiert und
sodann iteriert, um Vorhersagehorizonte beliebiger Länge zu erzeugen
\citep{bont13a, scip24a}. Prognoseintervalle werden per Voreinstellung
durch Bootstrap-Resampling der In-Sample-Residuen berechnet. Es handelt
sich um dasselbe Verfahren wie in \texttt{skforecast} \citep{rodr25a}.
Bei gegebenem Seed ist es deterministisch.

Jeder scikit-learn-kompatible Schätzer kann durch direkte Instanziierung
von \texttt{ForecasterRecursiveModel} eingesetzt werden. Die mit dem
Paket ausgelieferten Backend-Regressoren sind LightGBM \citep{ke17a},
XGBoost \citep{chen16a} und CatBoost \citep{prok18a}, implementiert als
\texttt{ForecasterRecursiveLGBM}, \texttt{ForecasterRecursiveXGB} und
\texttt{ForecasterRecursiveCatBoost}. Gradient-Boosting-Verfahren dieser
Art erzielen auf M5-artigen Datensätzen \citep{makr22a} gute Ergebnisse.
Die Entscheidung, die Backends des Pakets auf LightGBM, XGBoost,
CatBoost und scikit-learn-Regressoren zu beschränken (und die drei
genannten Wrapper als Standard-Modelle auszuliefern), folgt unmittelbar
aus dieser empirischen Evidenz.

\subsection{Splitter zur Erzeugung von Trainings- und
Testdaten}\label{sec-testen}

Eine naive Anwendung von \texttt{sklearn.model\_selection.KFold} auf
eine zeitlich indexierte Reihe würde unzulässigerweise Informationen aus
der Zukunft in den Trainingssatz einfügen. Zeitreihenbewusste Splitter
schaffen Abhilfe, indem sie den Trainingsabschnitt zwingen, dem
Testabschnitt zeitlich vorauszugehen. Das Paket stellt zwei solcher
Splitter sowie einen Backtesting-Treiber bereit.

\begin{figure}

\centering{

\pandocbounded{\includegraphics[keepaspectratio]{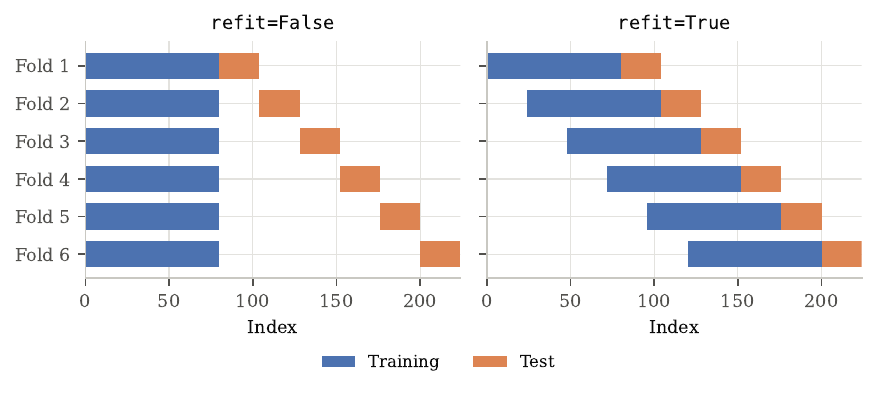}}

}

\caption{\label{fig-folds}Rolling-Origin-Backtest-Folds von
\texttt{TimeSeriesFold} (initiales Training 80, Schrittweite 24, sechs
Folds). Trainingsbereiche sind blau dargestellt, ausgehaltene
Testbereiche orange. Links (\texttt{refit=False}) wird der Forecaster
einmal auf dem Anfangsfenster trainiert, danach rückt nur das
Testfenster um \texttt{steps} Beobachtungen vor. Rechts
(\texttt{refit=True}) wird an jedem Ursprung auf einem gleitenden
Fenster konstanter Länge neu trainiert.}

\end{figure}

\texttt{TimeSeriesFold} realisiert das Rolling-Origin-Protokoll. Der
Splitter zerlegt die Reihe in \emph{Folds}, also zeitlich geordnete
Train-Test-Paare, und liefert die Testfenster. Abbildung~\ref{fig-folds}
stellt zwei mögliche Einstellungen gegenüber.

\texttt{OneStepAheadFold} ist eine schlanke Spezialisierung mit
\texttt{steps=1}, die üblicherweise zur Evaluation des bedingten
Mittelwertregressors selbst (statt der Mehrschritt-Punktprognose)
verwendet wird. \texttt{backtesting\_forecaster} komponiert einen der
beiden Splitter mit einem \texttt{ForecasterRecursive*} sowie einer
Liste von Metriken (\texttt{mean\_absolute\_error},
\texttt{mean\_squared\_error}, \texttt{mean\_absolute\_scaled\_error}
\citep{hynd06a}) und liefert einen pandas-DataFrame der Kennzahlen je
Fold zusammen mit dem verketteten Prognosevektor. Parallelisierte
Refit-Ausführung wird über \texttt{joblib} angeboten und verpflichtet
den Aufrufer, \texttt{n\_jobs} zu fixieren. Dies ist die einzige Stelle
im Paket, an der Parallelität ausgesetzt wird, und die Fixierung ist
zwingend, weil OpenMP-Scheduling eine bekannte Quelle
plattformübergreifenden Nichtdeterminismus ist.

\section{Entwurfsgrundsätze}\label{sec-principles}

Die Entwurfsgrundsätze gliedern sich in zwei parallele Kategorien.

\begin{itemize}
\tightlist
\item
  Die \emph{Code-Entwicklungsregeln} beschränken, welche Bestandteile
  der Software-Quelltext enthalten darf. Diese werden zum
  Commit-Zeitpunkt durch einen Test oder einen Linter\footnote{Ein
    Linter ist ein statisches Analysewerkzeug, das den Quelltext ohne
    Ausführung auf Stilverletzungen, verdächtige Muster und Verstöße
    gegen Programmierkonventionen prüft.} durchgesetzt.
\item
  Die \emph{Prozessregeln} beschränken, wie das Paket entwickelt,
  ausgeliefert und betrieben wird. Sie werden durch
  \texttt{MODEL\_CARD.md}, das dokumentierte Release-Verfahren und ein
  versioniertes Logging-Schema erfüllt.
\end{itemize}

Beide Kategorien sind erforderlich: Die Code-Entwicklungsregeln allein
genügen nicht, um die regulatorischen Pflichten zu Rückverfolgbarkeit,
Bedrohungsmodellierung, Daten-Governance, Lieferketten-Integrität und
betrieblicher Beobachtbarkeit zu erfüllen.

\subsection{Code-Entwicklungsregeln}\label{sec-code-rules}

Die vier nicht verhandelbaren Code-Regeln, die zugleich die
Entwicklungspraktiken für sichere Software, welche IEC 62443-4-1
\citep{iec62443_4_1} fordert, erfüllen, lauten:

\begin{tcolorbox}[enhanced jigsaw, arc=.35mm, bottomrule=.15mm, bottomtitle=1mm, breakable, colback=white, colbacktitle=quarto-callout-note-color!10!white, colframe=quarto-callout-note-color-frame, coltitle=black, left=2mm, leftrule=.75mm, opacityback=0, opacitybacktitle=0.6, rightrule=.15mm, title=\textcolor{quarto-callout-note-color}{\faInfo}\hspace{0.5em}{Code-Regeln}, titlerule=0mm, toprule=.15mm, toptitle=1mm]

\begin{itemize}
\tightlist
\item
  CR-1. \emph{Kein toter Code (``No dead code'').} ``Dead code'' ist
  Code, dessen Entfernung das Verhalten des Programms nicht beeinflusst.
  Keine Funktion, Klasse oder Verzweigung wird ausgeliefert, ohne dass
  ein Test oder ein ausführbares Docstring-Beispiel sie erreicht.
\item
  CR-2. \emph{Deterministische Transformationen.} Die gleiche Eingabe
  generiert die gleiche bitweise Ausgabe. Durch die folgenden Maßnahmen
  wird dies umgesetzt: Zufallszahlengeneratoren sind geseedet, die
  Dictionary-Iterationsreihenfolge wird nicht herangezogen und die
  \texttt{joblib}-Parallelität ist auf ein festes \texttt{n\_jobs}
  Argument fixiert.
\item
  CR-3. \emph{Fail-safe-Verarbeitung.} Ungültige Eingaben lösen eine
  explizite Ausnahme aus, statt stillschweigend imputiert oder
  konvertiert zu werden. Zum Beispiel lösen NaN-Werte oder fehlende
  Indizes explizite Exceptions vom Typ \texttt{ValueError} oder
  \texttt{TypeError} aus. Stille Imputation ist nur hinter einem
  typisierten
  \texttt{Literal{[}"raise",\ "ffill\_bfill",\ "passthrough"{]}}-Schalter
  verfügbar, der ausdrücklich gesetzt werden muss.
\item
  CR-4. \emph{Minimale CVE-Angriffsfläche (CVE: Common Vulnerabilities
  and Exposures).} Eine kurze, versionierte Sperrliste (Negativliste)
  verbotener Abhängigkeiten (\texttt{plotly}, \texttt{matplotlib},
  \texttt{spotoptim}, \texttt{optuna}, \texttt{torch},
  \texttt{tensorflow}) wird durch einen Test umgesetzt, der die
  Lock-Dateien mit \texttt{grep} prüft.
\end{itemize}

\end{tcolorbox}

\subsection{Prozessregeln}\label{sec-process-rules}

Vier weitere Regeln bestimmen, wie das Software-Paket entwickelt,
ausgeliefert und betrieben wird. Jede wird zum Prüfungs- oder
Release-Zeitpunkt kontrolliert und durch ein Artefakt erfüllt
(\texttt{MODEL\_CARD.md}, die in \texttt{RELEASING.md} dokumentierten
\texttt{Makefile}-Ziele oder ein versioniertes Logging-Schema). Diese
Aufteilung entspricht dem \emph{Trustworthiness-Assurance}-Prozess, den
\citet{stet24a} vorschlagen und in dem KI-spezifische regulatorische
Anforderungen zu prüfbaren Eigenschaften verfeinert werden, deren
Nachweise entlang des gesamten Entwicklungslebenszyklus gesammelt
werden.

Die Durchführung der Prozessregeln lässt sich durch
Continuous-Integration-Jobs (CI) automatisieren. Etablierte Werkzeuge
stehen z.B. auf GitHub oder GitLab zur Verfügung, um den größten Teil
dieser Arbeit umzusetzen. Die folgenden Regeln nennen deshalb jeweils
das geforderte Artefakt, den Mechanismus, mit dem dieses Paket es
erzeugt, und den etablierten Weg, die Regel maschinell zu prüfen.

Das Paket \texttt{spotforecast2-safe} selbst setzt diese Prüfungen
bewusst nicht auf einer CI-Plattform auf, sondern verwendet die lokale
Entwicklungsumgebung mittels Make \citep{free23a}. Die gesamte Prüfkette
hinter den Prozessregeln läuft über dokumentierte Make-Ziele
(\texttt{make\ verify}, \texttt{make\ release}, \texttt{make\ ship})
vollständig lokal und ist damit von jedem Auditor ohne Plattform-Zugang
reproduzierbar. GitLab dient ausschließlich dem Hosting von Quelltext
und Releases, nicht der Sicherheits- oder Konformitätsprüfung. Das
Schwesterpaket \texttt{spotforecast2} folgt demselben Modell. Der Preis
dieser Unabhängigkeit ist offen ausgewiesen: Ohne CI-Identität gibt es
kein schlüsselloses Signieren und keine unabhängige Ausführung der
Testsuite auf einer zweiten Plattform (siehe
Kapitel~\ref{sec-ki-vo-13}).

\begin{tcolorbox}[enhanced jigsaw, arc=.35mm, bottomrule=.15mm, bottomtitle=1mm, breakable, colback=white, colbacktitle=quarto-callout-note-color!10!white, colframe=quarto-callout-note-color-frame, coltitle=black, left=2mm, leftrule=.75mm, opacityback=0, opacitybacktitle=0.6, rightrule=.15mm, title=\textcolor{quarto-callout-note-color}{\faInfo}\hspace{0.5em}{Prozessregeln}, titlerule=0mm, toprule=.15mm, toptitle=1mm]

\begin{itemize}
\tightlist
\item
  PR-1. \emph{Rückverfolgbarkeit.} Jedes öffentliche Symbol trägt einen
  Docstring, der in der generierten API-Referenz erscheint, und die
  Ausführung der Make-Pipeline (\texttt{make\ guard}) schlägt fehl, wenn
  ein öffentliches Symbol undokumentiert oder sein generierter Stub
  veraltet ist. Die Verknüpfung eines Symbols mit einer dokumentierten
  Anforderung in \texttt{MODEL\_CARD.md} (Abschnitte 2 bis 5) oder mit
  einer bestimmten Klausel der Tabelle~\ref{tbl-compliance} entsteht zum
  Prüfungszeitpunkt. Damit werden IEC 61508-3 Abschnitt 7.2
  (Spezifikation der Software-Sicherheitsanforderungen)
  \citep{iec61508}, ISO 26262-6 Abschnitt 6.4 (Spezifikation der
  Software-Sicherheitsanforderungen, ASIL-abhängig nach Abschnitt 4.4)
  \citep{iso26262} sowie Art. 9 KI-VO (Risikomanagement über den
  gesamten Lebenszyklus) \citep{euro24a} unterstützt. Verwaiste Symbole,
  also solche ohne Anforderungsverweis und ohne
  \texttt{MODEL\_CARD}-Eintrag, werden in der Code-Prüfung beanstandet.
\item
  PR-2. \emph{Dokumentiertes Bedrohungsmodell.} Im Modul-Docstring jedes
  netzseitigen Moduls wird eine STRIDE-Tabelle\footnotemark{} gepflegt,
  derzeit in \texttt{downloader/entsoe.py},
  \texttt{downloader/resilience.py} und
  \texttt{weather/\_\_init\_\_.py}. Jede Änderung der Angriffsfläche
  erfordert einen aktualisierten Bedrohungseintrag. Die Regel steht in
  \texttt{CONTRIBUTING.md} und wird zum Review-Zeitpunkt kontrolliert.
  Der etablierte Weg, sie maschinell zu erzwingen, ist ein
  pfadgefilterter CI-Job, der einen Merge-Request zurückweist, wenn ein
  netzseitiges Modul geändert wird, ohne dass der zugehörige
  Bedrohungseintrag aktualisiert ist. Damit werden IEC 62443-4-1 SR-1
  und SR-2 (bedrohungsmodellgetriebene Sicherheitsanforderungen)
  \citep{iec62443_4_1} erfüllt.
\item
  PR-3. \emph{Lieferketten-Integrität.} Die Regel verlangt zwei
  Artefakte: eine Software-Stückliste (Software Bill of Materials, SBOM)
  und eine Attestierung, die sie an das tatsächlich ausgelieferte
  Artefakt bindet. Die SBOM erzeugt \texttt{make\ sbom} als
  CycloneDX-Dokument. Die Attestierung ist derzeit nicht erbracht, denn
  Releases tragen keine kryptografische Signatur (siehe
  Kapitel~\ref{sec-ki-vo-13}).
\item
  PR-4. \emph{Strukturiertes Audit-Protokoll.} Jede betriebliche Aktion
  erzeugt einen Protokollsatz gemäß einem fixierten JSON-Schema unter
  \texttt{\textasciitilde{}/spotforecast2\_safe\_models/logs/}. Das
  Schema ist versioniert: \texttt{audit\_log\_schema.json} führt seine
  Version in der \texttt{\$id}, \texttt{SCHEMA\_VERSION} ist eine
  fixierte Konstante, und \texttt{tests/manager/test\_logger\_schema.py}
  sichert die Übereinstimmung beider ab. Seit Release 26.0.0 sind die
  Sätze zusätzlich Hash-verkettet: Jeder Satz trägt im Pflichtfeld
  \texttt{prev\_hash} den SHA-256-Wert der zuvor geschriebenen Zeile,
  und die Funktion \texttt{verify\_audit\_log()} prüft die Kette offline
  und meldet den ersten Bruch mit Zeilennummer. Seit Release 27.0.0
  weist ihr Prüfbericht zudem aus, ob der Kettenkopf extern verankert
  wurde (siehe Kapitel~\ref{sec-ki-vo-12}).
\end{itemize}

\end{tcolorbox}

\footnotetext{STRIDE ist ein von Microsoft eingeführtes
Klassifikationsschema für Bedrohungen. Das Akronym steht für
\emph{Spoofing} (Identitätsvortäuschung), \emph{Tampering}
(Manipulation), \emph{Repudiation} (Abstreitbarkeit), \emph{Information
Disclosure} (Informationsoffenlegung), \emph{Denial of Service}
(Dienstverweigerung) und \emph{Elevation of Privilege}
(Rechteausweitung). Eine STRIDE-Tabelle zählt je Modul und Datenfluss
auf, welche dieser sechs Kategorien zutreffen, welche Gegenmaßnahme
greift und in welcher Quelldatei diese realisiert ist.}

Eine weitere, gesondert geführte Regel betrifft die Daten-Governance:
Das Paket \texttt{spotforecast2-safe} liefert selbst keine
Trainingsdaten aus. Jedes Forecaster-Objekt zeichnet seine Provenienz
(Erstellungs- und Trainingszeitpunkt, Bibliotheks- und Python-Version)
in seinem persistierten Zustand auf. Einen Inhalts-Hash der
Trainingsdaten berechnet das Paket nicht. Dies unterstützt die
Provenienz-Anforderung aus Art. 10 KI-VO (Daten und Daten-Governance)
\citep{euro24a} so weit, wie eine Bibliothek ohne Einsicht in die
tatsächlichen Trainingsdaten des Betreibers reichen kann. Die
Dokumentation der Datenquellen und der Abrufzeitpunkte wie auch die
Beurteilung von Repräsentativität, Verzerrung und Eignung für den
bestimmungsgemäßen Zweck obliegen weiterhin dem Betreiber.

Ein weiterer, nicht strikt zu den vier Regeln zählender, aber aus ihnen
folgender Grundsatz ist, dass jede Quelldatei einen SPDX-Kopf (Software
Package Data Exchange) im von REUSE v3.0 \citep{reuse30} empfohlenen
Format trägt. Portierte \texttt{skforecast}-Dateien tragen sowohl den
ursprünglichen BSD-3-Clause-Vermerk als auch die hinzugefügte
AGPL-3.0-or-later-Angabe. Die lizenzrechtliche Auflösung als
Doppellizenz-Ableitung wird ermöglicht, und es entsteht nicht der
Eindruck der Nachlizenzierung fremder Software.

\section{Sicherheitskritische Umsetzung}\label{sec-safety}

\subsection{Standards der
Open-Source-Community}\label{standards-der-open-source-community}

Neben den in Kapitel~\ref{sec-principles} aufgeführten vier Code- und
vier Prozessregeln verwendet \texttt{spotforecast2-safe} mehrere
community-geprägte Qualitätsmerkmale. Diese werden über
maschinenauswertbare Statusabzeichen (Badges) in der Datei
\texttt{README.md} des Pakets publiziert. Sie verweisen teils auf
unabhängige externe Prüfstellen (PyPI, pepy.tech, bestpractices.dev),
teils auf die maßgeblichen Artefakte im Repositorium.
Tabelle~\ref{tbl-badges} gibt einen vollständigen Überblick.

\begin{longtable}[]{@{}
  >{\raggedright\arraybackslash}p{(\linewidth - 6\tabcolsep) * \real{0.1304}}
  >{\raggedright\arraybackslash}p{(\linewidth - 6\tabcolsep) * \real{0.2609}}
  >{\raggedright\arraybackslash}p{(\linewidth - 6\tabcolsep) * \real{0.3913}}
  >{\raggedright\arraybackslash}p{(\linewidth - 6\tabcolsep) * \real{0.2174}}@{}}
\caption{Vollständige Liste der Status-Abzeichen (Badges) in der
\protect\texttt{README.md} von \protect\texttt{spotforecast2-safe},
gruppiert nach den im Repository verwendeten Kategorien. Die Gruppen
\emph{Version \& License} und \emph{Downloads} dienen ausschließlich der
Paketauffindung und -nachverfolgung. Die Gruppen \emph{Quality},
\emph{Scores} und \emph{Status} sind Konformitäts- bzw.
Qualitätssignale. OpenSSF = Open Source Security Foundation; SPDX =
Software Package Data Exchange. CR-X verweist auf die Code-Regel X in
Kapitel~\ref{sec-code-rules}.}\label{tbl-badges}\tabularnewline
\toprule\noalign{}
\begin{minipage}[b]{\linewidth}\raggedright
\textbf{Gruppe}
\end{minipage} & \begin{minipage}[b]{\linewidth}\raggedright
\textbf{Abzeichen}
\end{minipage} & \begin{minipage}[b]{\linewidth}\raggedright
\textbf{Bedeutung}
\end{minipage} & \begin{minipage}[b]{\linewidth}\raggedright
\textbf{Quelle / Referenz}
\end{minipage} \\
\midrule\noalign{}
\endfirsthead
\toprule\noalign{}
\begin{minipage}[b]{\linewidth}\raggedright
\textbf{Gruppe}
\end{minipage} & \begin{minipage}[b]{\linewidth}\raggedright
\textbf{Abzeichen}
\end{minipage} & \begin{minipage}[b]{\linewidth}\raggedright
\textbf{Bedeutung}
\end{minipage} & \begin{minipage}[b]{\linewidth}\raggedright
\textbf{Quelle / Referenz}
\end{minipage} \\
\midrule\noalign{}
\endhead
\bottomrule\noalign{}
\endlastfoot
Version \& License & Python Version & Python-Mindestversion für die
Paketinstallation & \texttt{pyproject.toml}; python.org \\
Version \& License & PyPI Version & Aktuelle Paket-Version im Python
Package Index & PyPI \\
Version \& License & License & Lizenzhinweis (AGPL-3.0-or-later) &
\texttt{LICENSE}; SPDX \\
Downloads & PyPI Downloads & Monatliche Download-Zahlen & PyPI \\
Downloads & Total Downloads & Kumulative Download-Statistik &
pepy.tech \\
Quality & EU AI Act & Regulatorische Einsetzbarkeit unter der KI-VO &
\texttt{MODEL\_CARD.md} \\
Quality & Dependencies & Minimierte Abhängigkeitsfläche (CR-4) &
\texttt{pyproject.toml} \\
Quality & Audit & Whitebox-Audit-Status & \texttt{MODEL\_CARD.md} \\
Quality & Reliability & Fail-safe-Verhalten (CR-3) &
\texttt{MODEL\_CARD.md} \\
Quality & Security & Dokumentierte Sicherheitsrichtlinie &
Sicherheitsrichtlinie der Paket-Dokumentation \\
Quality & Documentation & Veröffentlichte API-Referenz und Dokumentation
& Dokumentations-Site des Pakets \\
Scores & OpenSSF Best Practices & Best-Practices-Reifegrad
(\emph{passing} / \emph{silver} / \emph{gold}); Projekt-Kennung 11932 &
bestpractices.dev \\
Status & Maintenance & Aktive Pflege des Projekts &
Projekt-Repositorium \\
Status & Code style: black & Deterministischer Quelltext-Stilformatierer
(\texttt{black}) & github.com/psf/black \\
\end{longtable}

\subsection{Qualität}\label{qualituxe4t}

\subsubsection{Fail-safe-NaN-Behandlung}\label{sec-fail-safe-nan}

Ein tabellarischer Messwert kann fehlen, z.B. weil eine Wetterstation
eine Stunde lang keine Daten liefert, weil ein Stromzähler neu gestartet
wurde oder weil ein Importfehler eine Zelle leer gelassen hat. In Python
wird eine solche Lücke durch den Sonderwert \texttt{NaN} (``Not a
Number'') markiert. Für die Weiterverarbeitung ist entscheidend, wie die
nachgelagerte Software auf ein solches Feld reagiert: Soll sie den
leeren Wert schätzen und weiterrechnen oder soll sie die Bearbeitung
stoppen und eine Entscheidung des Betreibers anfordern?

Die in der Python-Welt etablierten Zeitreihen-Bibliotheken folgen dabei
überwiegend dem Muster ``stillschweigend weitermachen''. Für
exploratives Arbeiten ist diese Haltung zweckmäßig, weil sie zügiges
Prototyping mit unsauberen Daten ermöglicht. Für den Einsatz in
sicherheitskritischen Umgebungen, etwa der elektrischen Lastprognose für
einen Netzbetreiber, ist sie jedoch genau der Fall, den Art. 15 KI-VO,
IEC 61508 und ISO 26262 vermeiden wollen: Eine Bedienperson kann auf der
Grundlage einer scheinbar vollständigen Prognose disponieren, ohne zu
bemerken, dass ein Teil der Eingangsdaten gefehlt hat und ohne Nachfrage
vom Paket ersetzt wurde.

Ein konkretes Beispiel macht den Unterschied deutlich:

\begin{example}[Fail-safe
vs.~stillschweigend]\protect\hypertarget{exm-fail-safe-nan}{}\label{exm-fail-safe-nan}

Angenommen, ein Netzdisponent soll für den kommenden Tag die stündliche
Last prognostizieren und die Wetter-API liefert für eine einzelne
Stunde, z.B. für 03:00 Uhr, keinen Temperaturwert. Nach der
stillschweigenden Vorgehensweise würde das Prognose-Paket den Wert
(linear) zwischen 02:00 und 04:00 interpolieren, die Lastprognose mit
scheinbar normalem Konfidenzintervall zurückliefern und der Disponent
würde die 03:00-Stunde regulär einplanen. Die Lücke in den Messwerten
bliebe nur im Vorverarbeitungs-Log erkennbar. Es muss sich zudem jemand
die Mühe machen, diese Log-Dateien zu lesen. Nach dem Fail-safe-Ansatz
hingegen wird die Lücke zum sichtbaren Ereignis: Der Wetter-Adapter
lehnt die unvollständige Wetterzeile mit einer expliziten Fehlermeldung
ab, die CI-Pipeline schlägt Alarm und der Betreiber muss eine zu
dokumentierende Entscheidung treffen: Er kann auf ein Fallback-Modell
umschalten, die 03:00-Stunde als ungültig markieren, oder die Lücke
bewusst schließen und diese Entscheidung im Audit-Protokoll festhalten.

\end{example}

\texttt{spotforecast2-safe} kehrt den Default der oben genannten
Bibliotheken um. Der Abbruch ist das gewünschte Ergebnis und wird nicht
als Störung gesehen: Der Fehler wird sichtbar, erreicht die CI-Pipeline
und zwingt den Betreiber zu einer ausdrücklichen Entscheidung. Der
zentrale Punkt ist, dass der Betreiber in einem Audit nicht behaupten
kann, er habe die Datenlücke übersehen. Damit adressiert der Mechanismus
unmittelbar Art. 15 KI-VO (Genauigkeit und Robustheit) \citep{euro24a},
das Fail-safe-Prinzip der IEC 61508-3 \citep{iec61508} sowie die
Entwurfsprinzipien für die Software-Einheitsimplementierung nach ISO
26262-6 Abschnitt 8.4 \citep{iso26262}. Wie eine Lücke nach der
ausdrücklichen Entscheidung behandelt wird, zeigt Schritt 3 in
Kapitel~\ref{sec-example}.

\subsection{Prüfkette und
Konformität}\label{pruxfcfkette-und-konformituxe4t}

\subsubsection{Verifikations-Prüfkette}\label{verifikations-pruxfcfkette}

Die Prüfkette des Pakets ist in \texttt{RELEASING.md} dokumentiert und
im \texttt{Makefile}-Ziel \texttt{make\ verify} zusammengefasst, das
vier Tore nacheinander durchläuft: \texttt{guard}
(API-Dokumentations-Abdeckung, Aktualität der generierten Stubs,
REUSE-Lint und Sperrlisten-Prüfung), \texttt{lint} (deterministische
Formatierung und statische Analyse), \texttt{test} (vollständige
Testsuite mit erzwungener Abdeckungs-Untergrenze) und \texttt{doc}
(Quarto-Rendern der Dokumentation einschließlich der ausführbaren
Beispiele). Jedes Release setzt einen vollständigen
\texttt{verify}-Durchlauf voraus, und \texttt{make\ release} verweigert
die Ausführung auf einem unsauberen Arbeitsbaum. CI ist der etablierte
Weg, dieselbe Prüfkette zusätzlich an jeden git Push zu binden.

\subsubsection{Zeilenabdeckung}\label{zeilenabdeckung}

Die Zeilenabdeckung wird bei jedem Durchlauf von \texttt{make\ test}
gemessen und gegen die in \texttt{pyproject.toml} festgelegte
Untergrenze \texttt{fail\_under\ =\ 68} erzwungen, unterhalb derer der
Testlauf fehlschlägt. Die Testsuite umfasst mehr als 3000 Tests und
erreicht derzeit 75,99 \% Zeilenabdeckung. \texttt{CONTRIBUTING.md}
nennt 80 \% Zeilenabdeckung als Zielwert für neuen Code.

\subsubsection{SPDX- und REUSE-Konformität}\label{sec-spdx-reuse}

\emph{SPDX} ist ein standardisiertes, maschinenlesbares Format für
Lizenz- und Urheberrechts-Metadaten in Quelldateien. Die Spezifikation
wurde von der Linux Foundation entwickelt und als Norm ISO/IEC 5962:2021
\citep{iso5962} ratifiziert. SPDX-Kopfzeilen wie
\texttt{SPDX-FileCopyrightText} und \texttt{SPDX-License-Identifier}
erlauben es einer SBOM-Werkzeugkette, die Lizenzlage eines Pakets ohne
manuelle Sichtung aus den Quellen zu rekonstruieren. Die \emph{REUSE
Specification} der Free Software Foundation Europe, hier in der Version
3.0 verwendet \citep{reuse30}, operationalisiert SPDX zu drei prüfbaren
Anforderungen:

\begin{itemize}
\tightlist
\item
  \begin{enumerate}
  \def\labelenumi{(\roman{enumi})}
  \tightlist
  \item
    jede Datei trägt im Kopf einen \texttt{SPDX-FileCopyrightText}- und
    einen \texttt{SPDX-License-Identifier}-Eintrag,
  \end{enumerate}
\item
  \begin{enumerate}
  \def\labelenumi{(\roman{enumi})}
  \setcounter{enumi}{1}
  \tightlist
  \item
    die Volltexte aller im Projekt verwendeten Lizenzen werden im
    Verzeichnis \texttt{LICENSES/} abgelegt und
  \end{enumerate}
\item
  \begin{enumerate}
  \def\labelenumi{(\roman{enumi})}
  \setcounter{enumi}{2}
  \tightlist
  \item
    die Einhaltung wird durch das Werkzeug \texttt{reuse\ lint}
    automatisiert verifiziert.
  \end{enumerate}
\end{itemize}

Die so garantierte Lizenz-Rückverfolgbarkeit adressiert die
Dokumentationspflichten zu Softwarekomponenten und Werkzeugen Dritter
aus Art. 11 i.V.m. Anhang IV Nr. 1 KI-VO \citep{euro24a} und stützt die
Dokumentations-Praxis SM (Security Management, Inventar von
Drittanbieter-Komponenten) des IEC 62443-4-1 \citep{iec62443_4_1}. Der
REUSE-Lint ist zudem Teil des \texttt{guard}-Tors von
\texttt{make\ verify}, sodass die Lizenzkonformität bei jedem
Prüfdurchlauf und vor jedem Release erneut verifiziert wird.

\subsection{Bewertungen}\label{bewertungen}

\subsubsection{OpenSSF Best Practices Badge und
Scorecard}\label{openssf-best-practices-badge-und-scorecard}

Die Open Source Security Foundation (OpenSSF)\footnote{\url{https://openssf.org}}
unterhält zwei unabhängige Programme zur Bewertung der Sicherheitsreife
quelloffener Projekte.

\begin{itemize}
\tightlist
\item
  Das \emph{OpenSSF Best Practices Badge}-Programm\footnote{\url{https://www.bestpractices.dev}}
  prüft projektseitig eingereichte Nachweise zu Dokumentation,
  Zugänglichkeit des Quelltextes, Sicherheitsmeldewegen,
  Abhängigkeitsverwaltung sowie Test- und Buildautomatisierung auf den
  Reifegraden \emph{passing}, \emph{silver} und \emph{gold}.
  \texttt{spotforecast2-safe} ist unter der Projekt-Kennung 11932
  registriert und weist den Reifegrad \emph{passing} nach.
\item
  Die \emph{OpenSSF Scorecard}\footnote{\url{https://scorecard.dev}}
  ergänzt dieses selbstauskunftsbasierte Programm um eine automatisierte
  Bewertung der Projekt-Lieferkette anhand einer festen Liste von
  Prüfpunkten, darunter Branch-Schutz, Token-Scoping, automatisierte
  Abhängigkeits-Aktualisierung, statische Sicherheitsanalyse und
  signierte Releases. Für \texttt{spotforecast2-safe} liegt derzeit
  keine Scorecard-Bewertung vor. Der Nachweis der Lieferketten-Praxis
  stützt sich stattdessen auf das Best-Practices-Badge und die Zuordnung
  zum Security Development Lifecycle (SDL).
\end{itemize}

\subsection{Projektstatus}\label{projektstatus}

Die Badges der Gruppe \emph{Status} (Wartungsaktivität und festgelegter
Quelltextformatierer \texttt{black}\footnote{\url{https://github.com/psf/black}})
sind kein Konformitätsnachweis im Sinne der KI-VO oder der IEC 62443.
Sie machen allerdings sichtbar, dass das Paket aktiv gepflegt wird und
sein Quelltext einem deterministischen Stilformatierer unterliegt, was
bereits in Kapitel~\ref{sec-principles} als Bestandteil der
Code-Entwicklungsregeln verankert ist.

\subsection{Quarantäne beschädigter
Caches}\label{quarantuxe4ne-beschuxe4digter-caches}

Der Wetterdaten-Client speichert API-Antworten als Parquet-Dateien in
einem durch den Nutzer konfigurierbaren Verzeichnis. Der Cache-Leser
fängt \texttt{(OSError,\ ValueError)} Fehler ab, schreibt eine Meldung
auf dem Level \texttt{WARNING} und benennt die beschädigte Datei in
\texttt{\textless{}cache\textgreater{}.corrupt-\textless{}epoch\textgreater{}}
um, so dass ein Betreiber sie forensisch wiederherstellen kann. Eine
fehlende Cache-Datei bleibt stumm und liefert \texttt{None}.

\subsection{CPE-Kennung zur SBOM-Integration}\label{sec-cpe-sbom}

Eine SBOM ist eine maschinenlesbare Liste aller Komponenten eines
Software-Pakets\footnote{Vergleichbar mit der Zutatenliste auf einer
  Lebensmittelverpackung.}. Sie führt auf, welche Dritt-Bibliotheken in
welchen Versionen und unter welchen Lizenzen in der ausgelieferten
Software enthalten sind. Auf dieser Grundlage kann ein Betreiber ohne
Quelltext-Analyse feststellen, welche Komponenten in seiner Installation
stecken. Dies ist Voraussetzung für spätere Schwachstellenbewertungen.

Die CPE ist die standardisierte Kennung, unter der ein einzelnes
Software-Produkt in dieser Stückliste erscheint\footnote{Vergleichbar
  mit einem Barcode.}. Die Spezifikation CPE 2.3 des US-amerikanischen
National Institute of Standards and Technology \citep{nistir7695}
schreibt ein festes Format vor:
\texttt{cpe:2.3:a:\textless{}hersteller\textgreater{}:\textless{}produkt\textgreater{}:\textless{}version\textgreater{}:...}.
Eine typische Kennung für das vorliegende Paket lautet damit:\\
\texttt{cpe:2.3:a:sequential\_parameter\_optimization:spotforecast2\_safe:*:*:*:*:*:*:*:*}.\\
Mit einer stabilen CPE-Kennung kann ein Schwachstellen-Scanner
automatisch feststellen, dass eine bestimmte Version eines Produkts
betroffen ist, sobald eine CVE unter derselben CPE gemeldet wird,
insbesondere durch den von der US-amerikanischen Cybersecurity and
Infrastructure Security Agency (CISA) gepflegten Katalog bekannter
ausgenutzter Schwachstellen (Known Exploited Vulnerabilities, KEV)
\citep{cisaKEV}.

Die Besonderheit von \texttt{spotforecast2-safe} ist, dass die
CPE-Kennung nicht von nachgelagerten Werkzeugen erraten wird, sondern im
Paket selbst festgelegt ist. Die meisten Python-Pakete verlassen sich
auf die Heuristiken von Scannern wie \texttt{pip-audit} oder
\texttt{safety} und deklarieren keine CPE. Im Gegensatz dazu enthält die
Datei \texttt{src/spotforecast2\_safe/utils/cpe.py} die kanonische
CPE-Zeichenkette, und \texttt{tests/test\_cpe.py} prüft sie im
Round-Trip (Serialisierung und anschließende Deserialisierung liefern
dasselbe Objekt). Damit wird die Kennung zu einem erstklassigen
API-Vertrag: Eine wesentliche Software-Änderung benötigt einen
Conventional-Commit-Betreff \texttt{feat!:}, der einen
Major-Versions-Sprung erzwingt und als Eintrag in der
\texttt{CHANGELOG.md} erscheint. Ein Betreiber kann sich also darauf
verlassen, dass die in seiner SBOM stehende CPE über die gesamte
Minor-Versions-Linie hinweg stabil bleibt.

Aus regulatorischer Sicht entspricht dieses Vorgehen den drei Pflichten,
die die fixierte CPE unmittelbar bedient:

\begin{itemize}
\tightlist
\item
  Anhang I Teil II Nummer 1 CRA \citep{euCRA24} verlangt vom Hersteller
  eines Produkts mit digitalen Elementen die Erstellung einer
  Software-Stückliste.
\item
  Art. 72 KI-VO \citep{euro24a} verpflichtet Anbieter von
  Hochrisiko-KI-Systemen zur Beobachtung nach dem Inverkehrbringen, die
  eine verlässliche Rückführung von CVE-Meldungen auf das eigene Produkt
  voraussetzt.
\item
  Das Prozessgebiet \emph{SM} aus IEC 62443-4-1 \citep{iec62443_4_1}
  verlangt ein Inventar der Dritt-Komponenten, das in diesem Mechanismus
  die SBOM ist.
\end{itemize}

\section{KI-VO-Anforderungen und einschlägige
Normen}\label{sec-ki-vo-andere-normen}

Die folgende Analyse beschreibt die Anforderungen, die die KI-VO an
Hochrisiko-Systeme und deren Anbieter stellt. Wie in
Kapitel~\ref{sec-introduction} dargelegt, treffen diese Pflichten
\texttt{spotforecast2-safe} im einfachen Prognosefall nicht kraft
Gesetzes. Das Paket erfüllt die Anforderungen im Vorgriff und aus
Entwurfsentscheidung. Adressat der Pflichten ist stets der Anbieter des
Systems, in das die Bibliothek eingebettet wird, falls dieses System die
Merkmale eines Hochrisiko-Systems erfüllt. Die nachfolgenden Abschnitte
sind in diesem Sinne zu lesen.

\begin{figure}

\centering{

\includegraphics[width=0.9\linewidth,height=\textheight,keepaspectratio,alt={Venn-Diagramm zweier Mengen operationalisierungswesentlicher und eignungswesentlicher KI-Standards. In der Schnittmenge stehen sechs Kernnormen: ISO/IEC 4213, ISO/IEC 5338, ISO/IEC 23894, ISO/IEC 24027, ISO/IEC 38507 und ISO/IEC 42001.}]{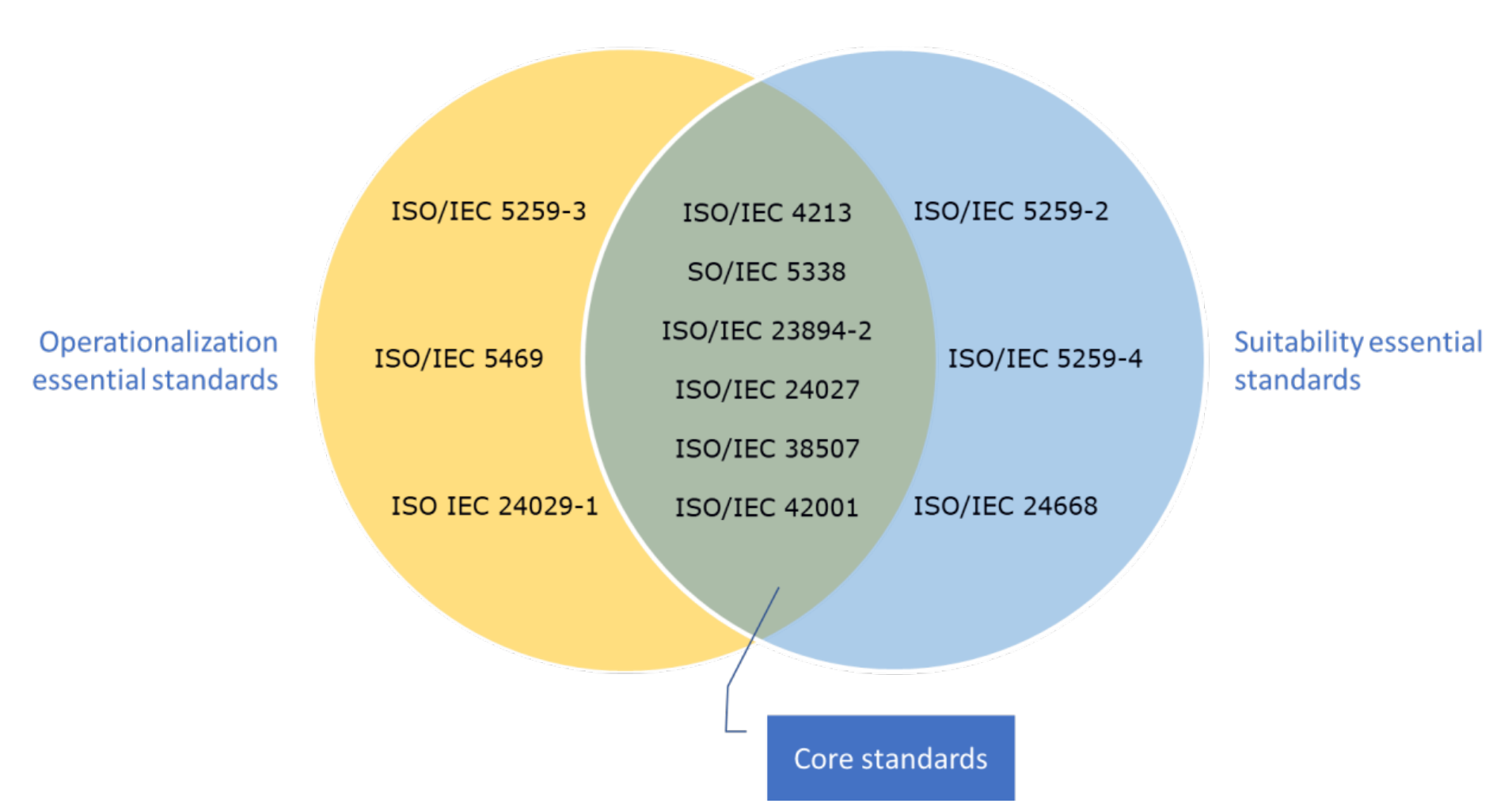}

}

\caption{\label{fig-core-standards}Zusammenhang der Gruppen
operationalisierungswesentlicher und eignungswesentlicher KI-Standards
sowie der zentralen Kernmenge. Die sechs Core Standards bilden die
Schnittmenge der beiden essentiellen Gruppen (je neun Standards).
Quelle: \citet{publ21a}, Abb. 18, S. 52; © European Union 2021,
lizenziert unter Creative Commons Attribution 4.0 International (CC BY
4.0); Originalabbildung übernommen mit Quellenangabe gemäß
Lizenzbedingungen.}

\end{figure}

Basierend auf \citet{publ21a} stellen \citet{stet24a} zusammen, welche
Normen der ISO/IEC und des Europäischen Instituts für
Telekommunikationsnormen (ETSI) die einzelnen Anforderungen der KI-VO an
Hochrisiko-Systeme operationalisieren. Tabelle~\ref{tbl-ki-vo-iso}
übernimmt diese Bestandsaufnahme. Aus dieser Bestandsaufnahme lässt sich
eine engere Kernmenge destillieren: \citet{publ21a} verdichten die zwölf
operationalisierungs- und eignungswesentlichen Standards auf sechs
sogenannte \emph{Core Standards} (Abbildung~\ref{fig-core-standards}),
die in beiden Essentiellen-Gruppen enthalten sind und damit den
kleinsten gemeinsamen Nenner der durchgeführten Reifegrad- und
Operationalisierungsanalyse bilden. Es handelt sich um die folgenden
sechs Normen: ISO/IEC 4213, ISO/IEC 5338, ISO/IEC 23894, ISO/IEC 24027,
ISO/IEC 38507 und ISO/IEC 42001. Tabelle~\ref{tbl-core-standards}
erläutert die sechs Core Standards und verweist jeweils auf die
ISO-/IEC-Primärquelle.

\begin{longtable}{p{0.33\columnwidth} p{0.6\columnwidth}}

\caption{\label{tbl-ki-vo-iso}Zuordnung der Anforderungen der KI-VO an Hochrisiko-KI-Systeme zu einschlägigen ISO/IEC- und ETSI-Operationalisierungsnormen. Die Auflistung folgt \citet{stet24a}, Tabelle~1 (dort nach dem JRC-Bericht 125952 \citep{publ21a}). Viele der genannten Normen befinden sich noch in der Ausarbeitung. TS = Technical Specification; SAI = Securing Artificial Intelligence (ETSI Industry Specification Group).}

\tabularnewline

\\
\hline
\textbf{KI-VO-Anforderung} & \textbf{Einschlägige Operationalisierungs-Normen} \\
\hline
\endfirsthead
\multicolumn{2}{l}{\itshape (Fortsetzung von \tablename~\ref{tbl-ki-vo-iso})}\\
\hline
\textbf{KI-VO-Anforderung} & \textbf{Einschlägige Operationalisierungs-Normen} \\
\hline
\endhead
\hline
\multicolumn{2}{r}{\itshape (Fortsetzung auf der nächsten Seite)}\\
\endfoot
\hline
\endlastfoot
Art.~9: Risikomanagementsystem & ISO/IEC 5338; ISO/IEC 5469; ISO/IEC 23894.2; ISO/IEC 38507; ISO/IEC 42001 \\
Art.~10: Daten und Daten-Governance & ISO/IEC TS 4213; ISO/IEC 5259-2, -3, -4; ISO/IEC 5338; ISO/IEC 5469; ISO/IEC 23894.2; ISO/IEC 24027; ISO/IEC 24029-1; ISO/IEC 24668; ISO/IEC 38507; ISO/IEC 42001; ETSI SAI 002, 005 \\
Art.~11: Technische Dokumentation & ISO/IEC 23894.2; ISO/IEC 24027; ISO/IEC 42001 \\
Art.~12: Aufzeichnungspflichten & ISO/IEC 23894.2 \\
Art.~13: Transparenz gegenüber Nutzern & ISO/IEC 23894.2; ISO/IEC 24028; ISO/IEC 38507; ISO/IEC 42001 \\
Art.~14: Menschliche Aufsicht & ISO/IEC 23894.2; ISO/IEC 38507; ISO/IEC 42001 \\
Art.~15: Genauigkeit, Robustheit, Cybersicherheit & ISO/IEC TS 4213; ISO/IEC 5338; ISO/IEC 5469; ISO/IEC 23894.2; ISO/IEC 24029-1; ISO/IEC 24668; ISO/IEC 42001; ETSI SAI 002, 003, 005, 006 \\
Art.~17: Qualitätsmanagementsystem & ISO/IEC 5259-3, -4; ISO/IEC 5338; ISO/IEC 23894.2; ISO/IEC 24029-1; ISO/IEC 38507; ISO/IEC 42001 \\

\end{longtable}

\begin{longtable}[]{@{}
  >{\raggedright\arraybackslash}p{(\linewidth - 4\tabcolsep) * \real{0.2644}}
  >{\raggedright\arraybackslash}p{(\linewidth - 4\tabcolsep) * \real{0.5517}}
  >{\raggedright\arraybackslash}p{(\linewidth - 4\tabcolsep) * \real{0.1839}}@{}}
\caption{Erläuterung der sechs Core Standards aus \citet{publ21a}. Die
in der Schnittmenge der operationalisierungs- und eignungswesentlichen
Gruppen auftretenden Normen stellen den kleinsten gemeinsamen Nenner der
für die KI-VO als besonders einschlägig identifizierten ISO/IEC-Normen
dar. SC 42 = ISO/IEC JTC 1/SC 42 \emph{Artificial intelligence}; SC 40 =
JTC 1/SC 40 \emph{IT Service Management and IT Governance}; PDCA =
Plan--Do--Check--Act.}\label{tbl-core-standards}\tabularnewline
\toprule\noalign{}
\begin{minipage}[b]{\linewidth}\raggedright
\textbf{Norm}
\end{minipage} & \begin{minipage}[b]{\linewidth}\raggedright
\textbf{Erläuterung}
\end{minipage} & \begin{minipage}[b]{\linewidth}\raggedright
\textbf{Referenz}
\end{minipage} \\
\midrule\noalign{}
\endfirsthead
\toprule\noalign{}
\begin{minipage}[b]{\linewidth}\raggedright
\textbf{Norm}
\end{minipage} & \begin{minipage}[b]{\linewidth}\raggedright
\textbf{Erläuterung}
\end{minipage} & \begin{minipage}[b]{\linewidth}\raggedright
\textbf{Referenz}
\end{minipage} \\
\midrule\noalign{}
\endhead
\bottomrule\noalign{}
\endlastfoot
ISO/IEC TS 4213:2022 & Standardisierte Methoden zur Messung der
Klassifikationsleistung von Machine-Learning-Modellen, -Systemen und
-Algorithmen (Technical Specification, 33 Seiten). & \citet{isoi22a} \\
ISO/IEC 5338:2023 & Lebenszyklus-Prozesse für KI-Systeme; übernimmt
ISO/IEC/IEEE 15288 und 12207 und ergänzt KI-spezifische Prozesse aus
ISO/IEC 22989 und 23053. & \citet{isoi23a} \\
ISO/IEC 23894:2023 & Leitfaden zum KI-Risikomanagement, methodisch an
ISO 31000 angelehnt; mit konkreten Umsetzungsbeispielen über den
gesamten KI-Lebenszyklus. & \citet{isoi23b} \\
ISO/IEC TR 24027:2021 & Technischer Bericht zu Bias in KI-Systemen und
KI-gestützter Entscheidungsfindung; Messverfahren und Typologie der
Bias-Quellen (kognitive, Daten-, Engineering-Bias). & \citet{isoi21a} \\
ISO/IEC 38507:2022 & IT-Governance-Implikationen der KI-Nutzung durch
Organisationen; adressiert das Leitungsgremium und wird abweichend von
den übrigen Kern-Normen vom SC 40 (nicht SC 42) gepflegt. &
\citet{isoi22b} \\
ISO/IEC 42001:2023 & Zertifizierbare Management-System-Norm für KI
(PDCA-basiert); Anforderungen an Policies, Risikobewertung,
Lifecycle-Steuerung und Lieferantenaufsicht, das ``ISO 27001 für KI''. &
\citet{isoi23c} \\
\end{longtable}

Tabelle~\ref{tbl-core-standards-mapping} bildet jeden der sechs Core
Standards auf die Entwurfsgrundsätze aus Kapitel~\ref{sec-principles}
(CR-1--CR-4 Code-Regeln, PR-1--PR-4 Prozessregeln) und auf die tragenden
\texttt{spotforecast2-safe}-Artefakte ab. Die Zuordnung verweist
ausschließlich auf Mechanismen, die bereits in diesem Bericht
dokumentiert sind. Sie erfindet keine neuen Kontrollen, sondern verlinkt
die bestehende Nachweisführung auf das ISO/IEC-Vokabular.

\begin{longtable}[]{@{}
  >{\raggedright\arraybackslash}p{(\linewidth - 4\tabcolsep) * \real{0.2727}}
  >{\raggedright\arraybackslash}p{(\linewidth - 4\tabcolsep) * \real{0.2727}}
  >{\raggedright\arraybackslash}p{(\linewidth - 4\tabcolsep) * \real{0.4545}}@{}}
\caption{Abbildung der sechs Core Standards aus
Tabelle~\ref{tbl-core-standards} auf die Entwurfsgrundsätze in
Kapitel~\ref{sec-principles} und die tragenden
\protect\texttt{spotforecast2-safe}-Artefakte. ISO/IEC TS 4213 und
ISO/IEC TR 24027 sind klassifikationsbezogen. Für die
Regressions-Prognose sind nur ihre Methodik bzw. ihre
Bias-Quellen-Taxonomie übertragbar. ISO/IEC 38507 adressiert das
Leitungsgremium (Governing Body) des Betreibers. Die gelisteten
Paket-Artefakte liefern das Substrat, auf dem Governance-Entscheidungen
gefällt werden können, nicht die Governance-Struktur selbst. CR =
Code-Regel; PR = Prozessregel; SBOM = Software Bill of Materials; CPE =
Common Platform Enumeration; STRIDE = Spoofing, Tampering, Repudiation,
Information disclosure, Denial of service, Elevation of
privilege.}\label{tbl-core-standards-mapping}\tabularnewline
\toprule\noalign{}
\begin{minipage}[b]{\linewidth}\raggedright
\textbf{Norm}
\end{minipage} & \begin{minipage}[b]{\linewidth}\raggedright
\textbf{Tragende Regeln}
\end{minipage} & \begin{minipage}[b]{\linewidth}\raggedright
\textbf{Paket-Mechanismen und Evidenz}
\end{minipage} \\
\midrule\noalign{}
\endfirsthead
\toprule\noalign{}
\begin{minipage}[b]{\linewidth}\raggedright
\textbf{Norm}
\end{minipage} & \begin{minipage}[b]{\linewidth}\raggedright
\textbf{Tragende Regeln}
\end{minipage} & \begin{minipage}[b]{\linewidth}\raggedright
\textbf{Paket-Mechanismen und Evidenz}
\end{minipage} \\
\midrule\noalign{}
\endhead
\bottomrule\noalign{}
\endlastfoot
ISO/IEC 42001:2023 \citep{isoi23c} & PR-1--PR-4, CR-1, CR-4 & Policy-
und Zielrahmen in \texttt{MODEL\_CARD.md} (Abschnitte 1 und 4) und
\texttt{CONTRIBUTING.md}; Risikobewertung via STRIDE (PR-2); technische
Kontrollen via CR-1--CR-4; PDCA-Zyklus über die in \texttt{RELEASING.md}
dokumentierten \texttt{Makefile}-Ziele \texttt{verify},
\texttt{preflight}, \texttt{release} und \texttt{ship};
Lieferantenaufsicht über \texttt{uv.lock} und
\texttt{tests/test\_prohibited\_dependencies.py}; Lebenszyklus-Nachweis
über Release-Tags und Audit-Protokoll (\texttt{manager/logger.py},
\texttt{audit\_log\_schema.json}). \\
ISO/IEC 23894:2023 \citep{isoi23b} & PR-2, CR-3, PR-1, PR-4 &
Risikoidentifikation durch modulweises STRIDE in
\texttt{CONTRIBUTING.md} und der Sicherheitsrichtlinie der
Paket-Dokumentation; Risikoanalyse durch CR-3 fail-safe-Verträge
(typisierter \texttt{on\_missing}-Schalter) und CR-2 Determinismus;
Risikobehandlung durch CR-4 Abhängigkeits-Sperrliste; Monitoring durch
das strukturierte Audit-Protokoll (PR-4); Kommunikation über
\texttt{MODEL\_CARD.md} Abschnitt 6 und \texttt{CHANGELOG.md}. \\
ISO/IEC 5338:2023 \citep{isoi23a} & CR-1, CR-2, PR-1, PR-3, PR-4 &
Technical-Management-Prozesse über das dokumentierte Release-Verfahren
und Conventional-Commits; Entwicklung über die pre-commit-Kette
(\texttt{black}, \texttt{isort}, \texttt{ruff}) und
\texttt{test\_docstring\_examples\_*.py} (CR-1); Verifikation über die
erzwungene Abdeckungs-Untergrenze (\texttt{fail\_under\ =\ 68}) und
\texttt{test\_linearly\_interpolate\_ts.py}; Deployment über
versioniertes PyPI-Wheel plus CycloneDX-SBOM (PR-3, unsigniert);
Operation über Audit-Protokoll (PR-4); Deprecation über
\texttt{feat!:}-Major-Sprünge im \texttt{CHANGELOG.md}. \\
ISO/IEC TS 4213:2022 \citep{isoi22a} & CR-2, PR-1 & Methodische
Disziplin bei der Leistungsmessung: Rolling-Origin-Evaluation
(Abbildung~\ref{fig-folds}), Fehlermaß-Reporting
(Abbildung~\ref{fig-residuals}) und dokumentierte Evaluierungsmetrik in
\texttt{MODEL\_CARD.md} Abschnitt 7; Reproduzierbarkeit durch
CR-2-Determinismus; PR-1 verknüpft jede Metrik mit einer
\texttt{MODEL\_CARD}-Anforderung. \\
ISO/IEC TR 24027:2021 \citep{isoi21a} & CR-2, CR-3, PR-1, PR-4 &
Engineering-Bias eliminiert durch CR-2 Determinismus; Daten-Bias
adressiert durch CR-3 Verbot stiller Imputation
(\texttt{on\_missing="raise"}-Default); kognitive Bias-Quellen in
\texttt{MODEL\_CARD.md} Abschnitt 6 (bekannte Grenzen, kontraindizierte
Anwendungen); Bias-Messung über Residualanalyse
(Abbildung~\ref{fig-residuals}) als Regressions-Pendant zur
24027-Messvorgabe für Klassifikatoren. \\
ISO/IEC 38507:2022 \citep{isoi22b} & PR-1, PR-3, PR-4 (indirekt) & Das
Paket liefert das Substrat der Governance-Entscheidung, nicht die
Governance selbst: \texttt{MODEL\_CARD.md} Abschnitte 6 bis 9 als
Entscheidungsgrundlage; Audit-Protokoll (PR-4) als operative
Nachvollziehbarkeit für das Leitungsgremium; CPE-Kennung in
\texttt{utils/cpe.py} plus CycloneDX-SBOM aus \texttt{make\ sbom} für
das Lieferketten-Risiko; die Verantwortlichkeits-Tabelle in
\texttt{MODEL\_CARD.md} Abschnitt 1 macht Rollen sichtbar. \\
\end{longtable}

\subsection{Die Pflichten der KI-VO im
Einzelnen}\label{die-pflichten-der-ki-vo-im-einzelnen}

Die Code- und Prozessregeln aus Kapitel~\ref{sec-principles} bilden auf
einzelne Bestimmungen der KI-VO, der IEC 61508-3, der ISO 26262-6, der
Normreihe ISA/IEC 62443 und des CRA ab. Anhand der Artikel aus der KI-VO
lassen sich die Pflichten des Anbieters eines Hochrisiko-KI-Systems auf
die sechs Core Standards aus Tabelle~\ref{tbl-core-standards} abbilden.
Diese Zuordnungen werden in den folgenden Abschnitten erläutert. Es
handelt sich dabei im Einzelnen um

\begin{itemize}
\tightlist
\item
  Artikel 4 KI-VO (KI-Kompetenz), der in Kapitel~\ref{sec-ki-vo-4}
  erläutert wird,
\item
  Artikel 9 KI-VO (Risikomanagement), der in Kapitel~\ref{sec-ki-vo-9}
  erläutert wird,
\item
  Artikel 10 KI-VO (Daten und Daten-Governance), der in
  Kapitel~\ref{sec-ki-vo-10} erläutert wird,
\item
  Artikel 11 KI-VO (Technische Dokumentation), der in
  Kapitel~\ref{sec-ki-vo-11} erläutert wird,
\item
  Artikel 12 KI-VO (Aufzeichnungspflichten), der in
  Kapitel~\ref{sec-ki-vo-12} erläutert wird,
\item
  Artikel 13 KI-VO (Transparenz gegenüber Nutzern), der in
  Kapitel~\ref{sec-ki-vo-13} erläutert wird,
\item
  Artikel 14 KI-VO (Menschliche Aufsicht), der in
  Kapitel~\ref{sec-ki-vo-14} erläutert wird,
\item
  Artikel 15 KI-VO (Genauigkeit, Robustheit, Cybersicherheit), der in
  Kapitel~\ref{sec-ki-vo-15} erläutert wird,
\item
  Artikel 16 KI-VO (Pflichten der Anbieter), der in
  Kapitel~\ref{sec-ki-vo-16} erläutert wird,
\item
  Artikel 17 KI-VO (Qualitätsmanagementsystem), der in
  Kapitel~\ref{sec-ki-vo-17} erläutert wird,
\item
  Artikel 18 KI-VO (Aufbewahrung der Dokumentation), der in
  Kapitel~\ref{sec-ki-vo-18} erläutert wird,
\item
  Artikel 72 KI-VO (Beobachtung nach dem Inverkehrbringen), der in
  Kapitel~\ref{sec-ki-vo-72} erläutert wird und
\item
  Artikel 73 KI-VO (Meldung von Vorkommnissen), der in
  Kapitel~\ref{sec-ki-vo-73} erläutert wird.
\end{itemize}

\subsubsection{Artikel 4 KI-VO (KI-Kompetenz)}\label{sec-ki-vo-4}

Artikel 4 KI-VO \citep{euro24a} verpflichtet Anbieter und Betreiber,
sicherzustellen, dass ihr Personal und andere Personen, die in ihrem
Auftrag ein KI-System betreiben oder verwenden, über ein hinreichendes
Maß an KI-Kompetenz verfügen, das auf den Nutzungskontext und die
Personengruppen, auf die das System einwirkt, abgestimmt ist.
\texttt{spotforecast2-safe} ist so entworfen, dass eine
Prozessingenieurin ohne formale Ausbildung im maschinellen Lernen diese
Anforderungen durch Lektüre der Paketdokumentation selbst erreichen
kann. Drei Mechanismen tragen dazu bei:

\begin{itemize}
\tightlist
\item
  Jeder öffentliche Transformator und jeder Forecaster führt ausführbare
  Docstring-Beispiele mit, so dass das beabsichtigte Verhalten durch das
  Ausführen kleiner, in sich geschlossener Ausschnitte erlernt werden
  kann und nicht auf das Studium externer Anleitungen angewiesen ist.
\item
  Typisierte Verträge machen die Vor- und Nachbedingungen jedes Aufrufs
  an der Aufrufstelle sichtbar. Damit sinkt das Fachwissen, das zum
  korrekten Lesen eines Trainingsskripts erforderlich ist.
\item
  Das strukturierte Audit-Protokoll (PR-4, Kapitel~\ref{sec-principles})
  legt zeilenweise die Abfolge der Entscheidungen offen, die ein
  laufender Forecaster getroffen hat, so dass das Personal des
  Betreibers das betriebliche Verhalten nachträglich überprüfen kann.
\end{itemize}

Die modellbezogene \texttt{MODEL\_CARD.md} ergänzt diese Mechanismen,
indem sie den bestimmungsgemäßen Einsatzrahmen und die bekannten Grenzen
in einer Sprache festhält, mit der auch nicht spezialisierte
Bedienpersonen arbeiten können. Zusammengenommen senken diese
Entwurfsentscheidungen die KI-Kompetenzschwelle, welche der Betreiber
zur verantwortungsvollen Beaufsichtigung des Systems erreichen muss, und
stützen damit unmittelbar die Pflicht aus Art. 4 KI-VO.

\subsubsection{Artikel 9 KI-VO (Risikomanagement)}\label{sec-ki-vo-9}

Artikel 9 KI-VO \citep{euro24a} verpflichtet den Anbieter eines
Hochrisiko-KI-Systems, ein Risikomanagementsystem (Risk Management
System, RMS) einzurichten, umzusetzen, zu dokumentieren und zu pflegen
(Abs. 1). Dieses ist als kontinuierlicher, iterativer Prozess über den
gesamten Lebenszyklus hinweg zu verstehen und umfasst vier Schritte
(Abs. 2): Identifikation und Analyse bekannter und vernünftigerweise
absehbarer Risiken (Art. 9 Abs. 2 lit. a KI-VO), Abschätzung und
Bewertung von Risiken in der bestimmungsgemäßen Verwendung und bei
vernünftigerweise absehbarem Missbrauch (Art. 9 Abs. 2 lit. b KI-VO),
Bewertung weiterer Risiken aus dem Beobachtungssystem nach Art. 72 KI-VO
(Art. 9 Abs. 2 lit. c KI-VO) sowie die Ergreifung geeigneter,
zielgerichteter Risikomanagementmaßnahmen (Art. 9 Abs. 2 lit. d KI-VO).
Die Maßnahmen müssen das kombinierte Zusammenspiel der Anforderungen aus
Kapitel III Abschnitt 2 berücksichtigen (Abs. 4) und so beschaffen sein,
dass die verbleibenden Restrisiken vertretbar sind (Abs. 5). Das System
ist durch Tests gegen vorab definierte Metriken und probabilistische
Schwellen zu erproben (Abs. 6 und 8), insbesondere im Hinblick auf
minderjährige und andere vulnerable Personengruppen (Abs. 9). Bei
paralleler Regulierung darf es mit bestehenden unionsrechtlichen
Risikomanagement-Prozessen kombiniert werden (Abs. 10).

Für eine als Komponente eingebundene Prognosebibliothek gilt hier wie in
allen folgenden Artikeln eine \emph{geteilte Verantwortung}: Der
Anbieter oder Betreiber des eingebundenen Hochrisiko-Systems bleibt
primärer Adressat der jeweiligen Pflicht, während die Bibliothek die
technische Infrastruktur beisteuert, mit der sich die Pflicht erfüllen
lässt. Für das RMS heißt das: \texttt{spotforecast2-safe} deckt die
beiden anbieterseitigen Pflichten ab, die eine Bibliothek allein
adressieren kann, nämlich die risikomindernde Gestaltung ihres eigenen
Codes (Art. 9 Abs. 3 und Art. 9 Abs. 5 lit. a KI-VO) und die
Bereitstellung der technischen Informationen, auf denen der Anbieter
sein eigenes RMS aufbaut (Art. 9 Abs. 5 lit. c KI-VO). Drei Mechanismen
tragen diese Arbeit.

\emph{Bedrohungsmodell-basierte Risikoidentifikation} (Art. 9 Abs. 2
lit. a und b KI-VO). Für jedes netzseitige Modul
(\texttt{downloader/entsoe.py}, \texttt{downloader/resilience.py},
\texttt{weather/\_\_init\_\_.py}) wird im Modul-Docstring eine
STRIDE-Tabelle geführt, deren Regel in \texttt{CONTRIBUTING.md}
festgehalten ist. Jede Änderung der Angriffsfläche erfordert im selben
Merge-Request einen aktualisierten Bedrohungseintrag. Die maschinelle
Erzwingung dieser Regel über einen pfadgefilterten CI-Job ist in
Kapitel~\ref{sec-process-rules} beschrieben. Das bildet die
Identifikations- und Bewertungsschritte für die bestimmungsgemäße
Verwendung und den vernünftigerweise absehbaren Missbrauch des
Netzwerk-vermittelten Teils der Bibliothek ab. Für den datenseitigen
Teil wirkt der typisierte \texttt{on\_missing}-Vertrag (siehe
Kapitel~\ref{sec-code-rules}, CR-3) als Missbrauchs-Früherkennung: Eine
fehlende oder nicht-numerische Eingabe wird nicht stillschweigend
imputiert, sondern zur dokumentationspflichtigen Ausnahme erhoben.

\emph{Designseitige Risikominderung} (Art. 9 Abs. 2 lit. d und Art. 9
Abs. 5 lit. a und b KI-VO). Die vier Code-Regeln aus
Kapitel~\ref{sec-principles} (CR-1 \emph{Kein toter Code}, CR-2
\emph{Determinismus}, CR-3 \emph{Fail-safe}, CR-4 \emph{Minimale
CVE-Angriffsfläche}) und die vier Prozessregeln (PR-1
\emph{Rückverfolgbarkeit}, PR-2 \emph{Bedrohungsmodell}, PR-3
\emph{Lieferketten-Integrität}, PR-4 \emph{Strukturiertes
Audit-Protokoll}) sind die konkreten, im Quellcode und in der
Entwicklungspipeline verankerten Risikomanagementmaßnahmen. Sie
eliminieren bestimmte Risiko-Klassen über die Gestaltung der Bibliothek
(Art. 9 Abs. 5 lit. a KI-VO) statt über nachgelagerte Bedienungshinweise
und sind im Sinne von Art. 9 Abs. 4 KI-VO explizit auf das Zusammenspiel
der übrigen Anforderungen aus Kapitel III Abschnitt 2 KI-VO abgestimmt:
Die Zuordnung ist in Tabelle~\ref{tbl-compliance} zeilenweise
aufgetragen und in Abbildung~\ref{fig-compliance-map} visuell
zusammengefasst.

\emph{Laufende Prüfung und Rückführung aus der Beobachtung} (Art. 9 Abs.
2 lit. c und Art. 9 Abs. 6 und 8 KI-VO). Die in
Kapitel~\ref{sec-ki-vo-72} beschriebene Log-Instrumentierung speist
beobachtete Ereignisse nach dem Inverkehrbringen in die Risikoanalyse
zurück. Die über \texttt{tests/test\_docstring\_examples\_*.py}
ausgeführten ausführbaren Docstrings, die in \texttt{pyproject.toml}
festgeschriebene und durch \texttt{make\ test} erzwungene
Abdeckungs-Untergrenze (\texttt{fail\_under\ =\ 68}) und die
Rolling-Origin-Kreuzvalidierung \texttt{split\_ts\_cv} bilden zusammen
die von Abs. 6 und 8 KI-VO geforderte Testebene gegen vorab definierte
Metriken. Statische Sicherheitsanalyse und automatisierte
Abhängigkeits-Überwachung sind die etablierten Werkzeuge, um diese
Testebene um sicherheitsorientierte Prüfungen über den gesamten
Lebenszyklus zu erweitern.

\subsubsection{Artikel 10 KI-VO (Daten-Governance)}\label{sec-ki-vo-10}

Artikel 10 KI-VO \citep{euro24a} verpflichtet den Anbieter eines
Hochrisiko-KI-Systems zu Daten-Governance-Maßnahmen über Trainings-,
Validierungs- und Testdatensätze (Abs. 1--4) und enthält eine
Ausnahmeregel für besondere Kategorien personenbezogener Daten (Abs. 5)
sowie eine Einschränkung des Anwendungsbereichs für Systeme, die ohne
Modelltraining entwickelt werden (Abs. 6). Für eine Prognosebibliothek,
die selbst keine Trainingsdaten ausliefert, greift die geteilte
Verantwortung aus Kapitel~\ref{sec-ki-vo-9}: Das Paket verhindert vor
allem, dass die materiellen Anforderungen stillschweigend umgangen
werden. Drei Mechanismen tragen diese Unterstützung.

\emph{Dokumentierte Entwurfsentscheidungen und Annahmen (Art. 10 Abs. 2
lit. a und d KI-VO).} Der strukturierte \texttt{MODEL\_CARD.md}-Rahmen
(siehe Kapitel~\ref{sec-ki-vo-11}) mit den Abschnitten \emph{Intended
Use and Scope}, \emph{Data and Operational Design Domain} und
\emph{Evaluation} zwingt den Betreiber, die relevanten
Entwurfsentscheidungen und die Annahmen über die Eingangsdaten
schriftlich festzuhalten.

\emph{Datenherkunft und Vorverarbeitung (Art. 10 Abs. 2 lit. b und c
KI-VO).} Jedes \texttt{Forecaster}-Objekt persistiert in seinem Zustand
die Metadaten des Anpassungsvorgangs: Erstellungs- und
Trainingszeitpunkt, Trainingszeitraum sowie Bibliotheks- und
Python-Version (siehe Kapitel~\ref{sec-example}, Schritt 9). Für
Live-Datenpfade stehen die Adapter \texttt{downloader/entsoe.py} für
ENTSO-E \citep{entsoe24} und \texttt{weather/client.py} für Open-Meteo
\citep{openmeteo24} zur Verfügung. Vorverarbeitung mit
Datenänderungs-Semantik (lineare Interpolation, Füll-Imputation,
Ausreißer-Markierung) wird nie automatisch in die Pipeline eingefügt,
sondern erst durch einen ausdrücklichen Aufruf des Betreibers wirksam
(siehe Kapitel~\ref{sec-code-rules} CR-3). Die Füll-Imputation
\texttt{get\_missing\_weights} liefert dabei eine Gewichtsserie mit, die
gefüllte Zeilen und ihren Lag-Nachlauf mit dem Stichprobengewicht null
vom Modelltraining ausschließt (siehe Kapitel~\ref{sec-example}, Schritt
3). Annotation und Labelling fallen im regressionsbasierten
Prognose-Einsatzfall nicht an.

\emph{Erkennung von Datenlücken (Art. 10 Abs. 2 lit. h KI-VO).} Die
Fail-safe-Prüfung \texttt{check\_y} weist jede Zielreihe mit fehlenden
Werten vor dem Anpassen zurück und macht so aus einer stillschweigend
umgangenen Datenlücke ein auffindbares, dokumentationspflichtiges
Ereignis. Die dadurch erzwungene Betreiber-Entscheidung
(Kapitel~\ref{sec-fail-safe-nan}) operationalisiert die in Art. 10 Abs.
2 lit. h KI-VO geforderte ``Identifizierung relevanter Datenlücken oder
Unzulänglichkeiten''.

Für die Verzerrungsanforderungen/Bias (Art. 10 Abs. 2 lit. f und g
KI-VO) und die Repräsentativitätspflicht (Art. 10 Abs. 3 KI-VO) liegt
die Beurteilung beim Betreiber, weil Verzerrungen erst aus dem
Zusammenspiel von Trainingsdaten und Einsatzszenario entstehen. Die
Bibliothek stützt die Prüfbarkeit durch deterministische
Transformationen (CR-2, siehe Kapitel~\ref{sec-ki-vo-15}), so dass
wiederholte Experimente auf demselben Datensatz bitweise gleiche
Resultate liefern und Abweichungen ausschließlich auf Datenunterschiede
zurückführbar sind. Die Repräsentativität kann der Betreiber mit der
portierten Rolling-Origin-Kreuzvalidierung \texttt{split\_ts\_cv}
empirisch prüfen. Besondere Kategorien personenbezogener Daten (Art. 10
Abs. 5 KI-VO) fallen im typischen Anwendungsfall der Lastprognose nicht
an. Wird das Paket ausschließlich zur Inferenz mit einem bereits
angelernten Modell eingesetzt, verengt sich nach Art. 10 Abs. 6 KI-VO
der Anwendungsbereich von Art. 10 Abs. 2 bis 5 KI-VO auf den
Testdatensatz. Dies ist eine Konfiguration, die der
\texttt{Forecaster}-Persistenzmechanismus mit serialisiertem
Trainingszeitraum und festgehaltenen Versionsständen ausdrücklich
unterstützt.

\subsubsection{Artikel 11 KI-VO (Technische
Dokumentation)}\label{sec-ki-vo-11}

Das Open-Source-Werkzeug \texttt{quartodoc} \citep{chow25a} erzeugt aus
den Docstrings einer Python-Bibliothek eine durchsuchbare
HTML-API-Referenz innerhalb des Quarto-Publikationssystems
\citep{alla26a} und stellt bei \texttt{spotforecast2-safe} das
Bindeglied zwischen den ausführbaren Docstrings im Quellcode und der für
Anhang IV KI-VO geforderten technischen Dokumentation her.

Artikel 11 KI-VO \citep{euro24a} in Verbindung mit Anhang IV KI-VO
verlangt, dass für jedes Hochrisiko-KI-System vor dem Inverkehrbringen
eine technische Dokumentation erstellt und fortlaufend gepflegt wird,
die einer zuständigen Behörde oder einer benannten Stelle die
Konformitätsbewertung ermöglicht. Bei \texttt{spotforecast2-safe} tragen
drei Artefakte diese Pflicht: die im Repository versionierte
\texttt{MODEL\_CARD.md} nach der Taxonomie des
Hugging-Face-Model-Card-Guidebook \citep{ozon22a} mit den Abschnitten
\emph{Uses}, \emph{Bias, Risks, and Limitations}, \emph{How to Get
Started}, \emph{Evaluation}, \emph{Environmental Impact},
\emph{Glossary}, \emph{Citation} sowie der Abbildung auf KI-VO und IEC
62443; die automatisch per \texttt{quartodoc} aus Docstrings generierte
API-Referenz unter \texttt{docs/}; sowie der vorliegende Bericht selbst,
der die architektonischen Entwurfsentscheidungen und ihre Abbildung auf
die regulatorischen Bestimmungen festhält. Die narrative Dokumentation
ergänzt die maschinenprüfbaren Nachweise aus Kapitel~\ref{sec-ki-vo-13}
und Kapitel~\ref{sec-ki-vo-15}, ohne sie zu ersetzen.

\subsubsection{Artikel 12 KI-VO
(Aufzeichnungspflichten)}\label{sec-ki-vo-12}

Artikel 12 KI-VO \citep{euro24a} verpflichtet den Anbieter, ein
Hochrisiko-KI-System so zu gestalten, dass es Ereignisse (``Logs'')
technisch automatisch und über den gesamten Lebenszyklus aufzeichnet
(Abs. 1). Die Protokollierung muss diejenigen Ereignisse erfassen, die
(Abs. 2) zur Identifikation von Risiken im Sinne von Art. 79 Abs. 1
KI-VO und von substanziellen Modifikationen dienen (Art. 12 Abs. 2 lit.
a KI-VO), die Beobachtung nach dem Inverkehrbringen gemäß Art. 72 KI-VO
(Art. 12 Abs. 2 lit. b KI-VO) ermöglichen und die Beaufsichtigung der
von Art. 26 Abs. 5 KI-VO erfassten Systeme (Art. 12 Abs. 2 lit. c KI-VO)
unterstützen. Für die besondere Gruppe der biometrischen
Fernidentifizierungssysteme (Anhang III Nr. 1 lit. a KI-VO) verlangt
Abs. 3 darüber hinaus Mindestinhalte: Nutzungsdauer, Referenzdatenbank,
Eingabedaten eines Treffers und beteiligte Personen.

Die Umsetzung im Paket ist in Prozessregel PR-4 (siehe
Kapitel~\ref{sec-process-rules}) als strukturiertes Audit-Protokoll
verankert und im Modul
\texttt{src/spotforecast2\_safe/manager/logger.py} konkret ausgeführt.
Einstiegspunkt ist die Funktion \texttt{setup\_logging()}, die stets
einen \emph{Konsolen-Handler} mit Plaintext-Formatter für die
menschliche Lesbarkeit aufsetzt (konfigurierbarer Level, Standard
\texttt{INFO}) und, sobald der Aufrufer ein Protokollverzeichnis
\texttt{log\_dir} übergibt, zusätzlich einen \emph{Datei-Handler} mit
dem zur Klasse gehörigen \texttt{JsonAuditFormatter}, der unabhängig von
der Konsolen-Ausführlichkeit dauerhaft auf \texttt{INFO}-Level
persistiert. Die Log-Datei wird mit einem Zeitstempel im Namen
(\texttt{sf2-safe-logger\_YYYYMMDD\_HHMMSS.log}) in diesem Verzeichnis
abgelegt, so dass jeder Lauf ein eigenes, nach Beginn sortierbares
Protokoll erzeugt. Die mitgelieferten Aufgaben übergeben standardmäßig
\texttt{\textasciitilde{}/spotforecast2\_safe\_models/logs/}. Kann die
Log-Datei nicht angelegt werden, bricht \texttt{setup\_logging()} mit
einem \texttt{OSError} ab, statt stillschweigend auf reine
Konsolenausgabe zurückzufallen. In einem sicherheitskritischen Ablauf
ist ein fehlender Audit-Trail kein hinnehmbarer Zustand. Dies erfüllt
Art. 12 Abs. 1 KI-VO (automatische Protokollierung über den
Lebenszyklus) in der vom Betreiber reproduzierbaren Form einer
JSON-Zeilen-Datei.

Das Schema der einzelnen Protokollsätze ist in
\texttt{src/spotforecast2\_safe/manager/audit\_log\_schema.json}
(Version 2.0.0) fixiert und umfasst sieben Pflichtfelder:
\texttt{schema\_version} (Konstante ``2.0.0''), \texttt{prev\_hash}
(SHA-256-Wert der vorangehenden Protokollzeile, im ersten Satz einer
Datei ein dokumentierter Genesis-Wert aus 64 Nullen),
\texttt{timestamp\_utc} (ISO 8601 UTC mit Mikrosekunden-Genauigkeit und
abschließendem ``Z''), \texttt{logger} (Python-Logger-Name),
\texttt{level}
(\texttt{DEBUG}/\texttt{INFO}/\texttt{WARNING}/\texttt{ERROR}/\texttt{CRITICAL}),
\texttt{event} (kurzer Slug wie \texttt{task\_start}, \texttt{fit},
\texttt{predict}, \texttt{fetch}) und \texttt{message} (formatierter
Klartext); hinzu kommen die optionalen Felder \texttt{task} (Name der
auslösenden Top-Level-Aufgabe), \texttt{context} (strukturierter Zusatz
aus \texttt{extra=\{...\}}) und \texttt{exception} (Traceback, sofern
\texttt{exc\_info} gesetzt war). Risiko-Ereignisse nach Art. 12 Abs. 2
lit. a KI-VO werden in diesem Schema durch die Kombination aus
\texttt{level\ \textgreater{}=\ ERROR}, einem beschreibenden
\texttt{event}-Slug und dem \texttt{exception}-Feld ausgewiesen. Jede
Änderung des Schemas selbst ist eine \emph{substanzielle Modifikation}
im Sinne der Bestimmung und ist per Vereinbarung an einen
Conventional-Commit-Betreff \texttt{feat!:} und einen
Major-Versionssprung gebunden (siehe Kapitel~\ref{sec-ki-vo-13}). Diese
Bindung wird zum Review-Zeitpunkt von Hand eingehalten. Ein
Vergleichs-Job, der die Schema-Version gegen die Paket-Version prüft,
kann sie maschinell erzwingen. Damit ist die in Art. 12 Abs. 2 lit. a
KI-VO geforderte Nachvollziehbarkeit substanzieller Modifikationen aus
dem Release-Pfad auf das Log-Schema abgebildet. Die UTC-Zeitstempel mit
Mikrosekunden-Genauigkeit und die konstante \texttt{schema\_version}
erlauben die in Art. 12 Abs. 2 lit. b KI-VO verlangte
deployment-übergreifende Korrelation der Ereignisse für die Beobachtung
nach dem Inverkehrbringen (Art. 72 KI-VO). Das \texttt{task}-Feld
ermöglicht die in Art. 12 Abs. 2 lit. c KI-VO verlangte Rekonstruktion
des operativen Betriebs.

Das Schema sichert die Form der Protokollsätze, nicht ihren Inhalt. Ein
Betreiber, der die Dateien hält, könnte einzelne Sätze nachträglich
ändern, löschen oder umsortieren, und gerade er hat nach Art. 26 Abs. 5
KI-VO die Aufsicht über den Betrieb, aus dem eine Behörde das
Systemverhalten rekonstruieren soll. Seit Release 26.0.0 begegnet das
Paket dieser Lücke mit einer Hash-Kette: Das Pflichtfeld
\texttt{prev\_hash} verkettet jeden Satz mit den exakten Bytes seines
Vorgängers, und die öffentliche Funktion \texttt{verify\_audit\_log()}
läuft eine Datei offline ab und meldet den ersten Bruch mit
Zeilennummer. Damit sind Änderung, Löschung und Umsortierung innerhalb
einer aufbewahrten Datei erkennbar. Die Grenze des Verfahrens ist ebenso
dokumentiert: Eine ungeschlüsselte Kette kann von einem Angreifer, der
die gesamte Datei neu schreiben kann, neu berechnet werden. Beweiskraft
gegenüber Dritten entsteht erst, wenn der Betreiber Dateiname, Satzzahl
und Kettenkopf in einem unabhängig kontrollierten Speicher ablegt, etwa
über einen RFC-3161-Zeitstempel oder ein Transparenzlog. Diese
Verankerung bleibt, wie die Aufbewahrung nach Art. 18 KI-VO
(Kapitel~\ref{sec-ki-vo-18}), bewusst eine Betreiberpflicht, denn die
Bibliothek ist offline-deterministisch und führt keine Netzwerkaufrufe
aus. Seit Release 27.0.0 behandelt der Prüfbericht diese Grenze als
eigenen Befund: \texttt{verify\_audit\_log()} nimmt einen zuvor
hinterlegten Kettenkopf als Anker entgegen und meldet dessen Status, und
ohne vorgelegten Anker benennt der Bericht die fehlende Verankerung
ausdrücklich, statt allein die intakte Kette zu bescheinigen. Die
Make-Ziele \texttt{anchor} und \texttt{anchor-verify} des Repositoriums
automatisieren die Hinterlegung über einen RFC-3161-Zeitstempel, wobei
der Netzwerkaufruf im Betreiber-Werkzeug liegt und nicht in der
Bibliothek. Beispiel~\ref{exm-fahrtenbuch} veranschaulicht Mechanismus
und Grenze.

\begin{example}[Fahrtenbuch mit
Fingerabdrücken]\protect\hypertarget{exm-fahrtenbuch}{}\label{exm-fahrtenbuch}

Das Hash-verkettete Audit-Protokoll gleicht einem Fahrtenbuch, in dem
jede neue Zeile einen Fingerabdruck der vorherigen Zeile enthält. Wird
eine Zeile nachträglich geändert, gelöscht oder umsortiert, passen die
Fingerabdrücke nicht mehr zusammen, und die Prüfung schlägt an.
Geschützt ist damit nur das Innere des Buchs. Wer das ganze Buch
besitzt, kann es vollständig neu schreiben, mit lauter frischen,
zueinander passenden Fingerabdrücken. Dagegen hilft, den jeweils letzten
Fingerabdruck außerhalb des Fahrtenbuchs zu hinterlegen, etwa durch die
Veröffentlichung als Zeitungsanzeige oder durch die Hinterlegung bei
einem Notar. Diese Hinterlegung ist die externe Verankerung des
Kettenkopfs.

\end{example}

Artikel 12 Abs. 3 KI-VO ist inhaltlich nicht einschlägig, weil
\texttt{spotforecast2-safe} eine regressionsbasierte Prognosebibliothek
ist und kein biometrisches Fernidentifizierungssystem nach Anhang III
Nr. 1 lit. a KI-VO. Sollte ein Betreiber für sein eigenes Auditprofil
eine intervall-orientierte Erfassung wünschen, so stellt das Schema eine
technische Grundlage dafür bereit.

\subsubsection{Artikel 13 KI-VO (Transparenz)}\label{sec-ki-vo-13}

Artikel 13 KI-VO \citep{euro24a} verpflichtet den Anbieter eines
Hochrisiko-KI-Systems, das System so zu gestalten, dass der Betreiber
seine Ausgabe sachgerecht interpretieren und verwenden kann, und eine
Gebrauchsanweisung mitzuliefern, die u.a. Zweckbestimmung,
Leistungsmerkmale, vorbestimmte Änderungen und notwendige Wartungs- und
Aktualisierungsmaßnahmen beschreibt. Das Paket behandelt diese Pflicht
als ausführbare und nicht als rein narrative Spezifikation. Vier
Mechanismen bilden zusammen den maschinenprüfbaren Teil des Prüfpfads
(audit trail).

\emph{Ausführbare Docstrings.} Jedes öffentliche Symbol wird mit einem
Docstring ausgeliefert, dessen Beispiele im Rahmen der Testsuite über
\texttt{test\_docstring\_examples\_*.py} bei jedem Testlauf ausgeführt
werden. Damit wird ein verbreiteter Fehlermodus ausgeschlossen, bei dem
Dokumentation und Verhalten unbemerkt auseinanderdriften. Ausführbare
Docstrings dokumentieren die Zweckbestimmung und das erwartete Verhalten
jedes öffentlichen Symbols und adressieren damit unmittelbar Art. 13
Abs. 3 lit. b Ziff. i KI-VO (Zweckbestimmung) sowie Ziff. vii
(Information zur sachgerechten Interpretation der Ausgabe).

\emph{Produkt-Identität über die CPE-Kennung.} Die in
Kapitel~\ref{sec-cpe-sbom} detailliert beschriebene, im Paket
festgelegte CPE-2.3-Kennung \citep{nistir7695} identifiziert Hersteller
und Produkt in jeder abgeleiteten SBOM und erfüllt so den Identitätsteil
von Art. 13 Abs. 3 lit. a KI-VO. Die darin liegende
Cybersicherheitsleistung (automatisierte Auflösung gegen CVE-Meldungen
und den CISA-KEV-Katalog) wird unter Kapitel~\ref{sec-ki-vo-15}
gemeinsam mit den übrigen Cybersicherheitsmechanismen erörtert.

\emph{Conventional Commits und Release-Verfahren.} Das durch pre-commit
erzwungene Commit-Nachrichtenformat entspricht Conventional Commits. Ein
Betreff \texttt{fix:} zeigt per Konvention einen Patch-Versionssprung
an, \texttt{feat:} einen Minor-Versionssprung, \texttt{feat!:} oder
\texttt{fix!:} einen Major-Versionssprung. Das Release selbst ist eine
dokumentierte Folge von drei \texttt{Makefile}-Zielen,
\texttt{make\ preflight}, \texttt{make\ release\ VERSION=X.Y.Z} und
\texttt{make\ ship\ VERSION=X.Y.Z}, die \texttt{RELEASING.md}
vollständig beschreibt. Sie prüft den Arbeitsbaum, verweigert den
Release, sobald ein Commit seit dem letzten Tag nicht als Conventional
Commit lesbar ist, regeneriert die \texttt{CHANGELOG.md}, setzt den Tag
und veröffentlicht ein Wheel auf PyPI. Diese Prüfkette adressiert Art.
13 Abs. 3 lit. c KI-VO (vorbestimmte Änderungen durch den Anbieter im
Zeitpunkt der Konformitätsbewertung) und Art. 13 Abs. 3 lit. e KI-VO
(Software-Updates als Bestandteil der Wartungs- und Pflegemaßnahmen).
Gleichzeitig operationalisiert sie IEC 62443-4-1 SUM-4 (sichere
Auslieferung von Updates) \citep{iec62443_4_1}, allerdings nur
teilweise, denn das Wheel ist unsigniert.

\emph{Narrative Dokumentation als Ergänzung.} Die maschinenprüfbaren
Mechanismen (Docstrings, CPE, Conventional Commits) werden durch
narrative Dokumentation ergänzt, nicht ersetzt: den vorliegenden
Bericht, die \texttt{MODEL\_CARD.md} und die \texttt{CHANGELOG.md}
(siehe Kapitel~\ref{sec-ki-vo-11}). Die ausgelieferte
\texttt{MODEL\_CARD.md} enthält unter anderem die Abschnitte
\emph{Intended Use and Scope}, \emph{Data and Operational Design
Domain}, \emph{Evaluation}, \emph{Model Transparency} und \emph{How to
Audit} und ist damit selbst ein Transparenz-Artefakt im Sinne des Art.
13 KI-VO.

\subsubsection{Artikel 14 KI-VO (Menschliche
Aufsicht)}\label{sec-ki-vo-14}

Dieselben Mechanismen wie in Kapitel~\ref{sec-ki-vo-4} adressieren zudem
das anbieterseitige Gegenstück in Art. 14 Abs. 4 KI-VO \citep{euro24a},
wonach ein Hochrisiko-KI-System so bereitgestellt werden muss, dass die
mit der menschlichen Aufsicht betrauten Personen in die Lage versetzt
werden, die Fähigkeiten und Grenzen des Systems zu verstehen (Art. 14
Abs. 4 lit. a KI-VO), seinen Betrieb zu überwachen und Anomalien,
Fehlfunktionen sowie unerwartete Leistungen zu erkennen (ebenfalls Art.
14 Abs. 4 lit. a KI-VO) sowie seine Ausgabe zutreffend zu interpretieren
(Art. 14 Abs. 4 lit. c KI-VO).

Ausführbare Docstrings, typisierte Verträge und \texttt{MODEL\_CARD.md}
erfüllen die Verstehens- und Interpretationspflichten. Das strukturierte
Audit-Protokoll erfüllt die Überwachungs- und Anomalieerkennungspflicht.
Art. 4 KI-VO begründet die betreiberseitige Kompetenzpflicht, Art. 14
Abs. 4 KI-VO die anbieterseitige Gestaltungspflicht, die diese Kompetenz
betrieblich erst erreichbar macht. Die Artefakte von
\texttt{spotforecast2-safe} erfüllen beide Pflichten aus derselben
Quellenbasis.

\subsubsection{Artikel 15 KI-VO (Genauigkeit, Robustheit und
Cybersicherheit)}\label{sec-ki-vo-15}

Artikel 15 KI-VO \citep{euro24a} beschreibt die Genauigkeit, Robustheit
und Cybersicherheit für Hochrisiko-KI-Systeme. Abs. 1 verlangt, dass
diese Eigenschaften ``throughout their lifecycle'' konsistent erbracht
werden, Abs. 4 fordert Widerstandsfähigkeit gegen Fehler, Ausfälle und
Inkonsistenzen. Auf der Implementierungsseite von
\texttt{spotforecast2-safe} wird diese Konsistenz-Anforderung durch
einen konsequent deterministischen Ausführungspfad erfüllt, der auf drei
Ebenen abgesichert ist. Auf Schätzerebene werden LightGBM und XGBoost
stets mit einem vom Wrapper durchgereichten \texttt{random\_state}
instanziiert. Zusätzlich werden die Optionen \texttt{deterministic=True}
und \texttt{force\_col\_wise=True} für LightGBM gesetzt, um den
Unterschied zwischen zeilen- und spaltenweisem Histogramm-Aufbau auf
Mehrkernsystemen zu beseitigen. Auf Merkmalsebene ist die Reihenfolge
der Verkettung exogener Merkmale Python-Versions-stabil, weil der
\texttt{ExogBuilder} seine Bestandteile in einer Liste und nicht in
einem Dictionary ablegt. Auf E/A-Ebene bewahrt ein Parquet-Round-Trip
die dtypes, nicht jedoch die Frequenz des DatetimeIndex. Der Cache-Leser
vergleicht daher wertebasiert und nicht frequenzbasiert, und die eigens
dafür vorgesehene Testsuite
\texttt{tests/test\_weather\_client.py::TestWeatherServiceCache} fixiert
dieses Verhalten. Damit wird die von Art. 15 Abs. 1 KI-VO geforderte
konsistente Leistung über den Lebenszyklus in einer maschinenprüfbaren,
vom Betreiber reproduzierbaren Form belegt.

Ergänzend zur Robustheit adressiert die Cybersicherheit nach Art. 15
Abs. 5 KI-VO die Abwehr und Nachverfolgung von Schwachstellen. Die in
Kapitel~\ref{sec-cpe-sbom} detailliert beschriebene, im Paket
festgelegte CPE-2.3-Kennung \citep{nistir7695} ist das technische
Bindeglied, das nachgelagerte Schwachstellen-Scanner nutzen, um eine
unter derselben CPE gemeldete CVE deterministisch der installierten
Paket-Version zuzuordnen. Ohne diese Fixierung würden Scanner auf
heuristische Paketnamen-Matches angewiesen bleiben, mit den bekannten
Fehltreffer- und Übersehensrisiken.

Jenseits der bindenden Pflichten der Verordnung liefert die
KI-spezifische Ergänzung der Normreihe \emph{Systems and software
Quality Requirements and Evaluation} (SQuaRE), ISO/IEC 25059:2023
\citep{isoiec25059}, das Qualitäts\-vokabular, das die nach Art. 40
KI-VO erwarteten harmonisierten Normen des Europäischen Komitees für
Normung (CEN) und des Europäischen Komitees für elektrotechnische
Normung (CENELEC) voraussichtlich übernehmen werden. Ihr Qualitätsmodell
verfeinert die Unterteilmerkmale funktionale Korrektheit, Robustheit und
Steuerbarkeit durch den Benutzer der ISO/IEC 25010 für KI-Systeme und
bildet damit das natürliche technische Pendant zu Art. 15 KI-VO. Die
Fassung 2023 wird derzeit überarbeitet: Das Umfrageverfahren zum Draft
International Standard ISO/IEC 25059:2025 ist abgeschlossen, und eine an
die KI-VO angepasste europäische Fassung (Preliminary European Norm,
prEN ISO/IEC 25059) wurde über das gemeinsame technische Komitee
CEN-CENELEC/JTC 21 im Februar 2026 in die öffentliche Umfrage gegeben.

\subsubsection{Artikel 16 KI-VO (Pflichten der
Anbieter)}\label{sec-ki-vo-16}

Artikel 16 KI-VO fasst die anbieterseitigen Pflichten für
Hochrisiko-KI-Systeme in einem Katalog zusammen. Der Katalog kombiniert
materielle Anforderungen an das Produkt selbst, organisatorische
Anforderungen an den Anbieter sowie prozedurale Anforderungen an den
Inverkehrbringungs- und Nachmarktpfad. Hinzu tritt Art. 16 lit. b KI-VO,
der eine identifizierende Anbieter-Kennzeichnung am Produkt selbst, an
seiner Verpackung oder an der begleitenden Dokumentation verlangt.

Weil der Katalog überwiegend auf andere Artikel der Verordnung verweist,
ist Art. 16 lit. a KI-VO, soweit seine Pflichten überhaupt greifen, für
\texttt{spotforecast2-safe} durch die bereits behandelten Abschnitte zu
Art. 9 KI-VO (Kapitel~\ref{sec-ki-vo-9}), Art. 10 KI-VO
(Kapitel~\ref{sec-ki-vo-10}), Art. 11 KI-VO
(Kapitel~\ref{sec-ki-vo-11}), Art. 12 KI-VO
(Kapitel~\ref{sec-ki-vo-12}), Art. 13 KI-VO (Kapitel~\ref{sec-ki-vo-13})
und Art. 15 KI-VO (Kapitel~\ref{sec-ki-vo-15}) erfüllt. Die Pflicht aus
Art. 16 lit. c KI-VO durch Kapitel~\ref{sec-ki-vo-17}, die aus Art. 16
lit. d KI-VO durch Kapitel~\ref{sec-ki-vo-18} und die aus Art. 16 lit. j
KI-VO durch Kapitel~\ref{sec-ki-vo-73}. Im Rahmen der geteilten
Verantwortung (Kapitel~\ref{sec-ki-vo-9}) deckt das Paket drei
Pflicht-Cluster unmittelbar auf Komponentenebene ab.

\emph{Identität und Kontakt} (Art. 16 lit. b KI-VO). Der Paket-Name, die
AGPL-3.0-or-later-Lizenz, die Maintainer-Adresse und der
Schwachstellen-Meldekanal sind in \texttt{pyproject.toml}, in der
PyPI-Listung sowie in der Sicherheitsrichtlinie der Paket-Dokumentation
festgeschrieben und werden mit jedem Release neu publiziert. Die in
Kapitel~\ref{sec-cpe-sbom} beschriebene CPE-2.3-Kennung ist die
maschinenlesbare Form derselben Angabe und erscheint in jeder
abgeleiteten Software-Stückliste. Ein Anbieter, der die Bibliothek in
ein Hochrisiko-KI-System einbaut, kann die nach Art. 16 lit. b KI-VO
erforderliche Herkunftsangabe deterministisch aus dem Paket übernehmen,
statt sie händisch zu pflegen.

\emph{Nachweisführung auf Anforderung} (Art. 16 lit. k im Zusammenspiel
mit Art. 16 lit. a KI-VO). Die in Kapitel~\ref{sec-ki-vo-11}
beschriebene technische Dokumentation (\texttt{MODEL\_CARD.md}, die
\texttt{quartodoc}-API-Referenz und der vorliegende Bericht), die
zeilenweise Abbildung in Tabelle~\ref{tbl-compliance} und die grafische
Zusammenfassung in Abbildung~\ref{fig-compliance-map} liefern dem
Anbieter die auf behördliche Aufforderung nach Art. 16 lit. k KI-VO
vorzuweisenden Nachweise. Weil sämtliche Nachweise unter
Versionskontrolle stehen und über das annotierte Release-Tag
reproduzierbar sind, kann der Anbieter zu jeder ausgelieferten
Paket-Version die dazugehörige Konformitäts-Evidenz rekonstruieren.

\emph{Korrekturen und Barrierefreiheit der technischen Kommunikation}
(Art. 16 lit. j und Art. 16 lit. l KI-VO). Das in
Kapitel~\ref{sec-ki-vo-13} beschriebene Release-Verfahren liefert die in
Art. 16 lit. j KI-VO geforderten Korrekturmaßnahmen in Form zeitlich
rückverfolgbarer, getaggter Releases mit zugehörigem
\texttt{CHANGELOG.md}-Eintrag. Für die Barrierefreiheit der technischen
Kommunikation nach Art. 16 lit. l KI-VO wird die mit \texttt{quartodoc}
erzeugte API-Dokumentation als reine HTML-Seite mit semantischer
Auszeichnung (Überschriftsebenen, Alternativtexte an Grafiken,
durchsuchbarer Klartext) ausgeliefert. Ausführbare Docstrings stellen
die Funktionsbeschreibung zusätzlich in durchsuchbarer Klartext-Form
bereit. Damit unterstützt das Paket die beiden
Barrierefreiheits-Richtlinien (EU) 2016/2102 (öffentliche Stellen) und
(EU) 2019/882 (European Accessibility Act) auf der Doku-Ebene, die eine
Bibliothek allein erreichen kann.

Einige Pflichten des Art. 16 KI-VO liegen jenseits der Reichweite einer
Bibliothek und bleiben dem Anbieter vorbehalten: Art. 16 lit. e KI-VO
betrifft die betriebliche Aufbewahrung der vom eingesetzten System
automatisch erzeugten Protokolle (das Paket erzeugt diese über das
Audit-Log-Schema, siehe Kapitel~\ref{sec-ki-vo-12}; ihre Aufbewahrung
liegt beim Anbieter); Art. 16 lit. f KI-VO verlangt das Durchlaufen
eines Konformitätsbewertungsverfahrens nach Art. 43 KI-VO für das
eingebettete Gesamt-System; Art. 16 lit. g KI-VO die
EU-Konformitätserklärung nach Art. 47 KI-VO; Art. 16 lit. h KI-VO die
CE-Kennzeichnung nach Art. 48 KI-VO; Art. 16 lit. i KI-VO die
Registrierung in der EU-Datenbank nach Art. 49 Abs. 1 KI-VO. Diese
Pflichten setzen ein vollständiges Produkt voraus und können von einer
Komponente nicht erfüllt werden. Die oben beschriebene maschinenlesbare
Konformitäts-Evidenz beschleunigt die Durchführung dieser Verfahren
durch den Anbieter.

\subsubsection{Artikel 17 KI-VO
(Qualitätsmanagementsystem)}\label{sec-ki-vo-17}

Die KI-VO organisiert die anbieterseitigen Pflichten um Art. 17 KI-VO
herum. Dieser verlangt ein dokumentiertes Qualitätsmanagementsystem
(QMS), welches die einzelnen materiellen Pflichten (Risikomanagement in
Art. 9 KI-VO, Daten-Governance in Art. 10 KI-VO, technische
Dokumentation in Art. 11 KI-VO, Aufzeichnungspflichten in Art. 12 KI-VO,
Transparenz in Art. 13 KI-VO, Genauigkeit und Robustheit in Art. 15
KI-VO) mit den anbieterseitigen Prozesspflichten (Aufzeichnungspflicht
des Anbieters nach Art. 18 KI-VO, Beobachtung nach dem Inverkehrbringen
nach Art. 72 KI-VO und Meldung schwerwiegender Vorfälle nach Art. 73
KI-VO) integriert. Tabelle~\ref{tbl-compliance} enthält daher eine
``KI-VO Art.-17-Zeile'', deren Mechanismus-Spalte auf die zur Erfüllung
der jeweiligen Teilpflicht dienenden Mechanismen verweist und die
benachbarten Zeilen querverweist. Ein Betreiber, der eine
Konformitätsbewertung vorbereitet, kann die ``KI-VO Art.-17-Zeile'' als
Index in die übrige Tabelle lesen.

Ein Mechanismus, der von Art. 17 KI-VO zwar nicht eigens verlangt, aber
für das QMS eines Software-Anbieters unverzichtbar ist, ist die
Lizenz-Rückverfolgbarkeit. \texttt{spotforecast2-safe} erfüllt sie durch
REUSE-v3.0-Konformität (detailliert in Kapitel~\ref{sec-spdx-reuse}):
Jede Quelldatei, jedes Daten-Fixture und jeder Test trägt einen
SPDX-Kopf; \texttt{uv\ run\ reuse\ lint} ist in pre-commit eingebunden
und wird bei jedem Commit ausgeführt; die \texttt{REUSE.toml} löst die
Lizenz der wenigen extern lizenzierten Dateien auf. Aus dieser
Konformität kann der Betreiber über \texttt{reuse\ spdx} jederzeit ein
\texttt{SPDX-3.0-JSON}-Dokument erzeugen und in sein eigenes QMS
einspeisen. Art. 13 KI-VO erhebt die Lizenzfrage nicht selbst. Ein
nachgelagerter Betreiber muss sie vor der Integration der Bibliothek
klären, und das QMS des Anbieters liefert ihm die Antwort schon zum
Release-Zeitpunkt.

\subsubsection{Artikel 18 KI-VO (Aufbewahrung der
Dokumentation)}\label{sec-ki-vo-18}

Die Aufbewahrungspflicht aus Art. 18 KI-VO bezieht sich in erster Linie
auf die technische Dokumentation gem. Art. 11 KI-VO und das QMS aus Art.
17 KI-VO. Die für die Marktüberwachung zuständigen Behörden
(Bundesnetzagentur) sollen jederzeit das Hochrisiko-KI-System prüfen
können. Die Dauer der Aufbewahrung beträgt 10 Jahre nach
Inverkehrbringen, bzw. Inbetriebnahme (Art. 3 Nr. 9 bzw. Nr. 11 KI-VO).
In Anbetracht des kommenden Produkthaftungsgesetzes empfehlen sich
längere Aufbewahrungszeiten.

\subsubsection{Artikel 72 KI-VO (Beobachtung nach dem
Inverkehrbringen)}\label{sec-ki-vo-72}

Artikel 72 KI-VO \citep{euro24a} verpflichtet den Anbieter eines
Hochrisiko-KI-Systems, ein Beobachtungssystem einzurichten und zu
dokumentieren, das dem Risiko und der Technologie angemessen ist (Abs.
1), aktiv und systematisch Leistungsdaten über die gesamte Lebensdauer
des Systems sammelt, dokumentiert und analysiert (Abs. 2), auf einem
schriftlichen Beobachtungsplan beruht, der Bestandteil der technischen
Dokumentation nach Anhang IV ist (Abs. 3), und sich in bestehende
Überwachungssysteme aus anderen EU-Harmonisierungsrechtsakten einbinden
lässt (Abs. 4). Das Beobachtungssystem muss dem Anbieter eine laufende
Beurteilung der Konformität mit den Anforderungen aus Kapitel III
Abschnitt 2 KI-VO ermöglichen. Operativ sensible Daten einer
Strafverfolgungsbehörde als Betreiber sind ausgenommen.

Im Rahmen der geteilten Verantwortung (Kapitel~\ref{sec-ki-vo-9})
liefert \texttt{spotforecast2-safe} die technische Infrastruktur, aus
der das geforderte Beobachtungssystem gespeist werden kann, gegliedert
in drei aufeinander abgestimmte Datenquellen.

\begin{enumerate}
\def\labelenumi{\arabic{enumi}.}
\item
  \emph{Betriebsseitige Log-Instrumentierung.} Das in
  Kapitel~\ref{sec-ki-vo-12} beschriebene Audit-Log-Schema v2.0.0
  erzeugt je Prognoselauf eine zeitstempel-benannte JSON-Datei unter
  \texttt{\textasciitilde{}/spotforecast2\_safe\_models/logs/} mit
  \texttt{timestamp\_utc} in Mikrosekunden-Auflösung, Ereignis-Slug
  (\texttt{event}), Log-Level, dem Namen der auslösenden Aufgabe
  (\texttt{task}) sowie einem freigeformten Kontext-Dictionary. Ein
  Beobachtungssystem kann diese Dateien über beliebige
  Log-Aggregations-Werkzeuge (ELK-Stack, Grafana Loki, Splunk)
  einsammeln und damit die in Art. 72 Abs. 2 KI-VO geforderte aktive und
  systematische Leistungserfassung über die Lebensdauer umsetzen, ohne
  dass das Paket eine eigene Telemetrie-Infrastruktur mitbringt.
\item
  \emph{Lebenszyklus-Metadaten der Ausprägung.} Jedes
  \texttt{Forecaster}-Objekt persistiert seine Anpassungs-Metadaten und
  die Modell-Hyperparameter (siehe Kapitel~\ref{sec-ki-vo-10}); die
  CPE-Kennung (siehe Kapitel~\ref{sec-cpe-sbom}) identifiziert die
  verwendete Paket-Version in jeder abgeleiteten SBOM; das
  Release-Verfahren (siehe Kapitel~\ref{sec-ki-vo-13}) dokumentiert in
  der \texttt{CHANGELOG.md} maschinenlesbar, wie sich
  aufeinanderfolgende Versionen voneinander unterscheiden. Zusammen
  liefern diese Artefakte dem Beobachtungssystem die Angaben, die für
  die in Art. 72 Abs. 2 KI-VO verlangte laufende Konformitätsbewertung
  notwendig sind, weil der Anbieter zu jedem gespeicherten Log-Eintrag
  die genaue Kombination aus Paket-Version, Trainingsdaten und
  Konfiguration rekonstruieren kann.
\item
  \emph{Beobachtung der Lieferkette.} Die CPE-Kennung (siehe
  Kapitel~\ref{sec-cpe-sbom}) und die CycloneDX-SBOM aus
  \texttt{make\ sbom} erlauben es, das Paket und seinen aufgelösten
  Abhängigkeitsgraphen fortlaufend gegen die National Vulnerability
  Database und den CISA-KEV-Katalog aufzulösen. Statische
  Sicherheitsanalyse und automatisierte Abhängigkeits-Überwachung binden
  diese Auflösung üblicherweise an jeden Push. Diese Beobachtung bildet
  die in Art. 72 Abs. 2 KI-VO als Analyse von ``Interaktion mit anderen
  KI-Systemen'' mitgedachte Dimension ab, soweit diese über
  Drittkomponenten vermittelt ist, und speist dieselbe Ereigniskette,
  die auch Art. 73 KI-VO bedient. Für den Anbieter bedeutet das, dass
  die Beobachtungspflicht aus Art. 72 KI-VO und die Meldepflicht aus
  Art. 73 KI-VO auf derselben Ereigniskette aufsetzen und nicht in zwei
  voneinander getrennten Systemen geführt werden müssen.
\end{enumerate}

Der nach Art. 72 Abs. 3 KI-VO verlangte Beobachtungsplan ist Bestandteil
der technischen Dokumentation und findet in der \texttt{MODEL\_CARD.md}
(siehe Kapitel~\ref{sec-ki-vo-11}) seinen schriftlichen Niederschlag.
Das Gesetz zur Durchführung der Verordnung über künstliche Intelligenz
(KI-Marktüberwachungs- und Innovationsförderungsgesetz) befindet sich
(Stand 1.4.2026) nach der Unterrichtung durch die Bundesregierung in der
Stellungnahme des Bundesrates. Mit einer Verabschiedung wird bis Ende
2026 gerechnet. Im Rahmen der geplanten zur Verfügung gestellten
Informationen könnte es ein Template für Art. 72 Abs. 3 KI-VO geben.

Die Kommission hat von der Ermächtigung aus Art. 72 Abs. 3 KI-VO, ein
Muster zu erstellen, keinen Gebrauch gemacht. Es gibt Überlegungen, im
Zuge einer Novelle der KI-VO diese Ermächtigungsgrundlage zu streichen.
Hintergrund ist die Idee, größtmögliche Flexibilität für die Entwicklung
und den Einsatz von Hochrisiko-KI zu schaffen.

Für einen Anbieter, der zugleich Pflichten aus anderen
EU-Harmonisierungsrechtsakten trägt (etwa der NIS-2-Richtlinie oder bei
medizinischen Hochrisiko-Produkten der Medizinprodukte-Verordnung (MDR)
bzw. der Verordnung über In-vitro-Diagnostika (IVDR)), erlaubt Abs. 4
die Integration der KI-VO-Vorgaben in sein bestehendes
Beobachtungssystem bzw. die Integration der QMS aus anderen sektoralen
Rechtsvorschriften in das QMS der KI-VO, um eine Doppeldokumentation zu
vermeiden (Erwägungsgrund 81 KI-VO). Die einheitliche Log- und
Release-Spur des Pakets unterstützt diese Integration, weil sie
maschinenlesbar ist und sich in jedes bestehende System für
Security-Information-and-Event-Management (SIEM) oder
Device-Surveillance einpflegen lässt.

\subsubsection{Artikel 73 KI-VO (Meldung schwerwiegender
Vorfälle)}\label{sec-ki-vo-73}

Artikel 73 KI-VO \citep{euro24a} verpflichtet den Anbieter eines
Hochrisiko-KI-Systems, jeden schwerwiegenden Vorfall der
Marktüberwachungsbehörde des betroffenen Mitgliedstaats zu melden (Abs.
1). Die Fristen sind abgestuft: spätestens 15 Tage nach Kenntnisnahme im
Regelfall (Abs. 2), spätestens 2 Tage bei weitverbreiteten Verstößen
oder Vorfällen im Sinne von Art. 3 Nr. 49 lit. b KI-VO (Abs. 3),
spätestens 10 Tage im Todesfall (Abs. 4). Ein zunächst unvollständiger
Erstbericht darf durch einen vollständigen Folgebericht ergänzt werden
(Abs. 5). Nach der Meldung führt der Anbieter unverzüglich eigene
Untersuchungen (Risikobewertung und Korrekturmaßnahmen) durch und
kooperiert mit den zuständigen Behörden. Unilaterale Veränderungen am
System, die eine nachträgliche Ursachenermittlung beeinträchtigen
könnten, sind untersagt (Art. 73 Abs. 6 KI-VO).

Primärer Adressat der Meldepflicht ist im Rahmen der geteilten
Verantwortung (Kapitel~\ref{sec-ki-vo-9}) der Anbieter des eingebundenen
Systems, regelmäßig der Betreiber-Integrator, der die Bibliothek
einsetzt. Das Paket stellt drei Mechanismen bereit, die es ihm
ermöglichen, die Fristen des Art. 73 KI-VO technisch einzuhalten und die
in Art. 73 Abs. 6 KI-VO geforderte Untersuchungs- und Korrekturkette
nachweisbar abzuwickeln.

\begin{enumerate}
\def\labelenumi{\arabic{enumi}.}
\item
  \emph{Vertraulicher Meldekanal und Versionspolitik.} Die
  Sicherheitsrichtlinie der Paket-Dokumentation etabliert die
  E-Mail-Adresse des Maintainers als primären Kanal für die vertrauliche
  Meldung. Als zweiter Weg stehen vertrauliche Issues im
  Projekt-Repositorium zur Verfügung. Öffentliche Issues werden
  ausdrücklich ausgeschlossen. Die veröffentlichten Reaktionszusagen
  (Bestätigung binnen 24 Stunden, Ersteinschätzung binnen 3 Werktagen,
  koordinierte öffentliche Offenlegung erst nach Bereitstellung einer
  Korrektur) sind mit den 2-Tage- und 10-Tage-Fristen des Art. 73 Abs. 3
  und 4 KI-VO zeitlich kompatibel. Die Sicherheitsrichtlinie
  dokumentiert zudem die Versionspolitik (Sicherheits-Updates erhält
  ausschließlich das jeweils aktuelle Release, einen
  Long-Term-Support-Zweig gibt es nicht), so dass der Anbieter vor der
  Meldung klären kann, auf welcher Version die Untersuchung ansetzt. Die
  Verantwortlichkeits-Tabelle in \texttt{MODEL\_CARD.md} Abschnitt 1
  benennt den freigabezuständigen Maintainer und macht die von Art. 73
  Abs. 6 KI-VO verlangte Kooperationsachse nachvollziehbar.
\item
  \emph{Früherkennungssignale.} Die CPE-verankerte SBOM (siehe
  Kapitel~\ref{sec-cpe-sbom}) macht das Paket und seine Abhängigkeiten
  für Schwachstellen-Scanner deterministisch auflösbar und ist damit die
  Datengrundlage, aus der Anbieter und Maintainer Vorwarnsignale
  beziehen, bevor ein Vorfall eine Marktüberwachung erreicht.
  Fortlaufend ausgeführt wird diese Auflösung üblicherweise durch
  statische Sicherheitsanalyse und automatisierte Abhängigkeits-Alarme.
  In IEC 62443-4-1-Terminologie füllt diese Früherkennung das
  Prozessgebiet \emph{DM} (Defect Management). In der KI-VO-Terminologie
  schafft sie den Wissensstand, auf dessen Basis der Anbieter den Beginn
  der Meldefristen überhaupt datieren kann.
\item
  \emph{Nachweisbare Korrekturmaßnahme.} Nach der Triage einer Meldung
  erzeugt das Release-Verfahren (siehe Kapitel~\ref{sec-ki-vo-13}) einen
  neuen Release mit zugehöriger Korrektur. Die begleitende
  CycloneDX-SBOM (siehe Kapitel~\ref{sec-cpe-sbom}) erlaubt dem
  Empfänger den Abgleich mit der gemeldeten Schwachstelle. Releases
  tragen derzeit keine kryptografische Signatur. Der
  \texttt{CHANGELOG.md}-Eintrag verweist auf die Kennung der Meldung, so
  dass der von Art. 73 Abs. 6 KI-VO verlangte Nachweis einer
  untersuchungsbasierten Korrekturkette (Commit-Trail über Release zu
  Veröffentlichungsprotokoll) ohne zusätzliche Handarbeit entsteht.
\end{enumerate}

Für die Fälle, in denen der Anbieter oder Betreiber gleichzeitig
Pflichten aus anderen EU-Rechtsakten trägt (NIS-2-Richtlinie, MDR/IVDR
bei medizinischen Hochrisiko-Produkten, CRA-Anhang II bei Produkten mit
digitalen Elementen) und der KI-VO entsprechenden Meldepflichten,
reduziert Art. 73 Abs. 9 und 10 KI-VO den Meldegegenstand auf Art. 3 Nr.
49 lit. c KI-VO. Die oben beschriebenen Mechanismen decken diesen
verringerten Rahmen gleichermaßen ab, weil Advisory-, Release- und
SBOM-Spur identisch bleiben. Was die Bibliothek nicht leisten kann, ist
die inhaltliche Entscheidung darüber, ob ein konkreter Vorfall einen
schwerwiegenden Vorfall im Sinne von Art. 3 Nr. 49 KI-VO darstellt.

\subsection{Verhältnis zur Cyberresilienz-Verordnung und zur
NIS-2-Richtlinie}\label{sec-cra-nis2}

Der CRA \citep{euCRA24}, in Kraft getreten am 10. Dezember 2024 und mit
ihren wesentlichen Pflichten ab dem 11. Dezember 2027 anzuwenden, belegt
Hersteller von ``Produkten mit digitalen Elementen'' mit grundlegenden
Cybersicherheitsanforderungen (Anhang I Teil I) und einer über den
Unterstützungszeitraum von grundsätzlich mindestens fünf Jahren
laufenden Pflicht zur Behandlung von Schwachstellen (Art. 13 Abs. 8
i.V.m. Anhang I Teil II). Eine als Wheel auf PyPI ausgelieferte
Prognosebibliothek ist ein CRA-Produkt mit digitalen Elementen, sobald
sie im Zuge einer gewerblichen Tätigkeit auf dem Unionsmarkt
bereitgestellt wird. Der Durchführungsbeschluss C(2025)618 der
Kommission vom 3. Februar 2025 (Normungsauftrag M/606) beauftragt CEN,
CENELEC und ETSI mit der Ausarbeitung von 41 harmonisierten Normen;
Typ-C-Produktkategorie-Normen sind bis zum 30. Oktober 2026,
Typ-B-technische-Maßnahmen-Normen bis zum 30. Oktober 2027 vorgesehen;
zwei horizontale Normen (Anhang-I-Einträge 1 und 15) sind bereits zum
30. August 2026 fällig. EN IEC 62443-4-1 ist unter den erwarteten
Typ-B-Bezugsnormen. Eine heute bestehende Anpassung an 62443-4-1 ist
demnach zugleich der kostengünstigste Weg zur Konformitätsbewertung nach
CRA im Jahr 2027. Die NIS-2 \citep{euNIS22} gilt bereits. Art. 21 Abs. 2
lit. e NIS-2 verpflichtet wesentliche und wichtige Einrichtungen,
Maßnahmen zur ``Sicherheit bei Erwerb, Entwicklung und Wartung von Netz-
und Informationssystemen'' zu ergreifen.

Für den deutschen KRITIS-Kontext lässt sich diese Pflichtenlage
konkretisieren. Die NIS-2-Richtlinie reguliert Einrichtungen, nicht
Systeme, und stellt an ein Prognosemodell als solches keine Anforderung.
Ein Übertragungsnetzbetreiber oberhalb des Schwellenwerts des Entwurfs
der Kritisverordnung \citep{bund26a} ist jedoch Betreiber kritischer
Anlagen im Sinne des BSI-Gesetzes \citep{deut25a}. § 30 Abs. 1 BSIG, die
deutsche Umsetzung des Risikomanagement-Katalogs aus Art. 21 NIS-2,
verpflichtet ihn, Störungen der Verfügbarkeit, Integrität und
Vertraulichkeit sämtlicher informationstechnischer Systeme, Komponenten
und Prozesse zu vermeiden, die er für die Erbringung seiner Dienste
nutzt. Ein produktiv eingesetzter Day-ahead-Prognosedienst ist ein
solches System. Ist er für die Funktionsfähigkeit einer kritischen
Anlage maßgeblich, tritt die Pflicht zum Einsatz von Systemen zur
Angriffserkennung aus § 31 Abs. 2 BSIG hinzu. Die Pflichten treffen den
Betreiber und erreichen den Lieferanten der Prognosebibliothek über die
Sicherheit der Lieferkette, die derselbe Maßnahmenkatalog in das
Risikomanagement des Betreibers einbezieht (Art. 21 Abs. 2 lit. d
NIS-2). In dieser Lesart sind die von den Prozessregeln aus
Kapitel~\ref{sec-process-rules} geforderten Artefakte genau die
lieferantenseitige Evidenz, die ein solcher Betreiber anfordern würde:
die Software-Stückliste (PR-3), das dokumentierte Bedrohungsmodell
(PR-2) und das strukturierte Audit-Protokoll (PR-4). Im hier
dokumentierten Zeitraum band keine dieser Pflichten das Paket selbst,
denn \texttt{spotforecast2-safe} wird außerhalb einer gewerblichen
Tätigkeit quelloffen bereitgestellt, und sein Maintainer ist keine
Einrichtung im Sinne der Richtlinie. Diese Lesart ist damit im selben
Sinne antizipatorisch wie die Einordnung unter CRA und
Produkthaftungsrichtlinie in Kapitel~\ref{sec-introduction}.

\subsection{Legal-Requirements-Engineering}\label{sec-lre}

\begin{figure}

\centering{

\pandocbounded{\includegraphics[keepaspectratio]{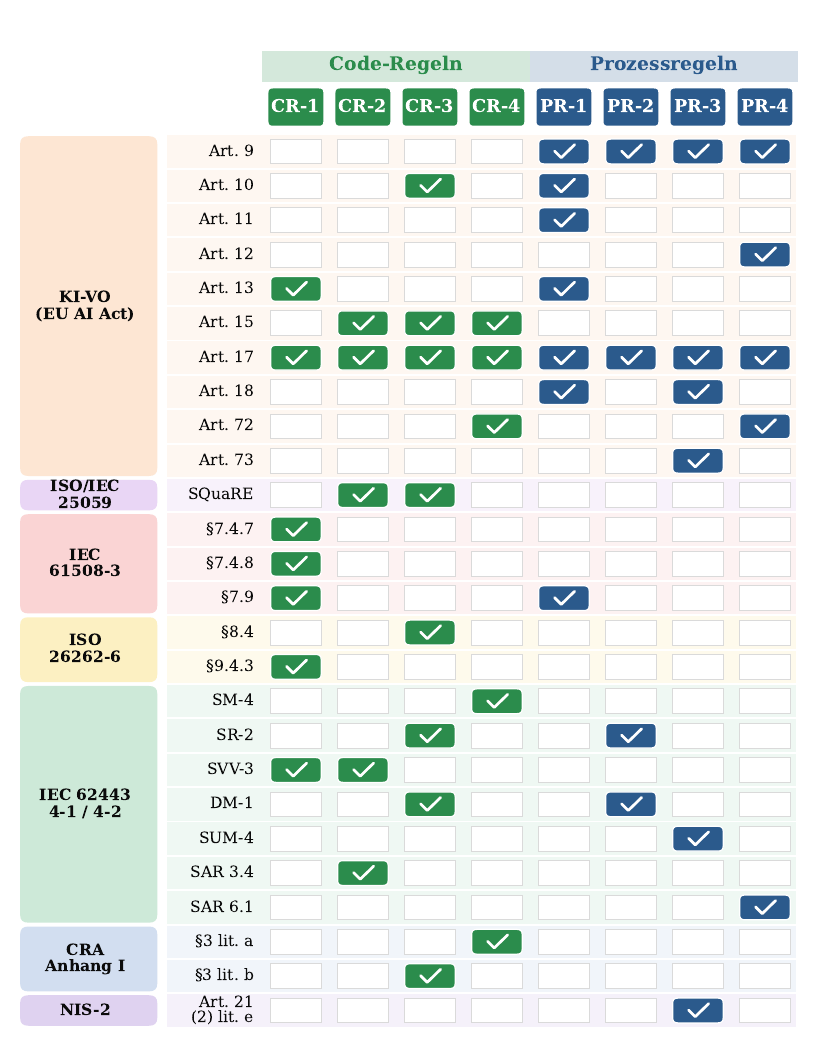}}

}

\caption{\label{fig-compliance-map}Konformitätsmatrix: Abbildung der
Code- (CR) und Prozessregeln (PR) aus Kapitel~\ref{sec-principles} auf
die regulatorischen Bestimmungen der vorhergehenden Tabelle. Ein Häkchen
in einer Zelle bedeutet, dass die in der Mechanismus-Spalte von
Tabelle~\ref{tbl-compliance} genannte Umsetzung diese Regel
instanziiert. Grün markierte Zeilen kennzeichnen Code-Regeln, blau die
Prozessregeln; die farbigen Bänder am oberen Rand fassen die
Bestimmungen nach Regelfamilie zusammen.}

\end{figure}

Tabelle~\ref{tbl-compliance} stellt die LRE-Rückverfolgungsmatrix für
das vorliegende Paket dar. Jede Tabellenzeile nennt eine Klausel oder
Anforderung sowie den Mechanismus im Paket, der sie adressiert.
Abbildung~\ref{fig-compliance-map} fasst die Zuordnung visuell zusammen.

{\small

\begin{longtable}{p{0.28\textwidth} p{0.64\textwidth}}

\caption{\label{tbl-compliance}Abbildung regulatorischer Bestimmungen auf Paket-Mechanismen. KI-VO = Verordnung (EU) 2024/1689; IEC61508 = IEC 61508-3:2010; ISO26262 = ISO 26262-6:2018; IEC62443 = Normreihe ISA/IEC 62443; CRA = Verordnung (EU) 2024/2847 (Cyberresilienz-Verordnung); NIS-2-RL = Richtlinie (EU) 2022/2555.}

\tabularnewline

\\
\hline
\textbf{Bestimmung} & \textbf{Mechanismus} \\
\hline
\endfirsthead
\multicolumn{2}{l}{\itshape (Fortsetzung von \tablename~\ref{tbl-compliance})}\\
\hline
\textbf{Bestimmung} & \textbf{Mechanismus} \\
\hline
\endhead
\hline
\multicolumn{2}{r}{\itshape (Fortsetzung auf der nächsten Seite)}\\
\endfoot
\hline
\endlastfoot
KI-VO Art.~9 (Risikomanagement über den gesamten Lebenszyklus) & Die vier Prozessregeln aus Kapitel~\ref{sec-principles} bilden den Risikomanagementrahmen; STRIDE-Bedrohungsmodell je netzseitigem Modul; Bedrohungseinträge werden im selben Merge-Request wie die zugehörige Codeänderung aktualisiert. \\
KI-VO Art.~10 (Daten-Governance) & Ausdrücklicher \texttt{on\_missing}-Vertrag; keine stille Imputation; Provenienz der Trainingsdaten in \texttt{MODEL\_CARD.md} festgehalten. \\
KI-VO Art.~11 und Anhang~IV (technische Dokumentation) & \texttt{MODEL\_CARD.md} wird im Repository zu Zweckbestimmung, Datenprovenienz, Metriken und Grenzen gepflegt; automatisch generierte API-Dokumentation mittels quartodoc. \\
KI-VO Art.~12 (Aufzeichnungspflichten / automatische Protokolle) & Doppel-Handler-Logger: Konsole plaintext, Datei-Senke JSON nach fixiertem Schema (\texttt{audit\_log\_schema.json}); \texttt{SCHEMA\_VERSION} ist aus dem Schema abgeleitet und per Test fixiert. Schemaänderungen sind per Vereinbarung an einen Conventional-Commits-Betreff \texttt{feat!:} und einen Major-Sprung gebunden; die Bindung wird zum Review-Zeitpunkt eingehalten. \\
KI-VO Art.~13 (Transparenz) & Ausführbare Docstrings für jedes öffentliche Symbol; festgelegte öffentliche API in \texttt{\_\_all\_\_}; CHANGELOG durch \texttt{git-cliff} im Zuge von \texttt{make release} erzeugt. \\
KI-VO Art.~15 (Genauigkeit, Robustheit, Cybersicherheit) & Deterministische Transformationen; Sperrliste verbotener Abhängigkeiten; CPE-Kennung für SBOM. \\
ISO/IEC 25059:2023 (SQuaRE-Qualitätsmodell für KI-Systeme, technisches Pendant zu Art.~15 KI-VO) & Funktionale Korrektheit durch deterministische Transformationen; Robustheit durch typisierte Konfiguration und fail-safe \texttt{on\_missing}-Vertrag; Steuerbarkeit durch das in Kapitel~\ref{sec-principles} dokumentierte Verbot stiller Auto-Anpassung. \\
KI-VO Art.~17 (Qualitätsmanagementsystem als Dachbestimmung) & QMS dokumentiert als: (i) Compliance-Strategie über die Prozessregeln aus Kapitel~\ref{sec-principles}; (ii) Entwurfskontrolle in \texttt{CONTRIBUTING.md}; (iii) Verifikation und Validierung in der \texttt{make verify}-Prüfkette; (iv) Datenmanagement in der Art.~10-Zeile; (v) Risikomanagement in der Art.~9-Zeile; (vi) Beobachtung nach dem Inverkehrbringen und Vorfallsmeldung in den Zeilen zu Art.~72 und Art.~73; (vii) Aufzeichnungspflichten in den Zeilen zu Art.~12 und Art.~18; (viii) Ressourcen- und Verantwortungsrahmen in \texttt{MODEL\_CARD.md} Abschnitt 1. \\
KI-VO Art.~18 (Aufzeichnungspflichten des Anbieters: Dokumentation, Protokolle, Konformitätsnachweise) & Das Repository selbst ist das Aufzeichnungsarchiv: Conventional-Commits-Historie, Release-Tags, \texttt{CHANGELOG.md}, release-spezifische \texttt{MODEL\_CARD.md}, \texttt{uv.lock}-Momentaufnahme je Tag. Aufzeichnungen werden über die zehnjährige Frist des Art.~18 KI-VO mittels Git-Historie und der veröffentlichten PyPI-Releases vorgehalten. \\
KI-VO Art.~72 (Beobachtung nach dem Inverkehrbringen) & Strukturiertes Audit-Protokoll unter \texttt{\~{}/spotforecast2\_safe\_models/logs/}; CPE-Kennung ermöglicht die Schwachstellenverfolgung nach Release. \\
KI-VO Art.~73 (Meldung schwerwiegender Vorfälle an die Behörden) & Vertraulicher Meldekanal nach der Sicherheitsrichtlinie der Paket-Dokumentation (E-Mail, vertrauliche Issues); CPE-verankerte SBOM als Früherkennungs-Datengrundlage; CHANGELOG-Einträge verweisen auf die Kennung der Meldung; koordinierte Offenlegung mit Auslieferung der Korrektur über PyPI. \\
IEC 61508-3 Abschnitt 7.4.7 (Anforderungen an das Testen von Softwaremodulen) & Je Modul eine Testdatei; Abdeckungs-Untergrenze \texttt{fail\_under = 68}, durch \texttt{make test} erzwungen; 80\,\% Zielwert für neuen Code laut \texttt{CONTRIBUTING.md}. \\
IEC 61508-3 Abschnitt 7.4.8 (Anforderungen an Software-Integrationstests) & Konsolen-Skripte der obersten Ebene (\texttt{tasks/}) werden in der Testsuite End-to-End ausgeübt. \\
IEC 61508-3 Abschnitt 7.9 (Software-Verifikation) & Ausführbare Docstrings; \texttt{make verify} schlägt fehl, wenn Docstring-Beispiele fehlschlagen. \\
ISO 26262-6 Abschnitt 8.4 (Entwurfsprinzipien, u.a. Initialisierung von Variablen) & Fail-safe: nicht initialisierte oder NaN-Eingaben lösen eine Ausnahme aus, statt standardmäßig auf Null gesetzt zu werden. \\
ISO 26262-6 Abschnitt 9.4.3 (Komponenten-Tests) & Je Modul eine Testdatei; pytest-Konfiguration in \texttt{pyproject.toml} im strengen Modus. \\
IEC 62443-4-1 SM-4 (Komponenten Dritter) & Fixierte \texttt{uv.lock}; Sperrliste verbotener Abhängigkeiten; CPE-Kennung; doppelte SPDX-Köpfe auf portiertem \texttt{skforecast}-Code. \\
IEC 62443-4-1 SR-2 (bedrohungsmodellgetriebene Anforderungen) & Fail-safe \texttt{on\_missing}-Vertrag; enge Ausnahmebehandlung an der Netzwerkgrenze. \\
IEC 62443-4-1 SVV-3 (Sicherheitstests) & Ausführbare Docstrings; modulweise Unit-Tests; Abdeckungs-Untergrenze \texttt{fail\_under = 68}, durch \texttt{make test} erzwungen. \\
IEC 62443-4-1 DM-1 (Management sicherheitsrelevanter Meldungen) & Vertraulicher Meldeweg nach der Sicherheitsrichtlinie der Paket-Dokumentation; enge, CVE-relevante Ausnahmebehandlung; Nachverfolgung über \texttt{CHANGELOG.md}. \\
IEC 62443-4-1 SUM-4 (Auslieferung sicherheitsrelevanter Updates) & \emph{Teilweise:} versioniertes PyPI-Wheel über \texttt{make release}/\texttt{ship}; Conventional-Commits-Tor; CHANGELOG. Releases tragen derzeit keine kryptografische Signatur. \\
IEC 62443-4-2 SAR 3.4 (Integrität der Software) & Deterministische Transformationen; fixierte CPE; Parquet-Round-Trip-Vertrag. \\
IEC 62443-4-2 SAR 6.1 (Zugänglichkeit des Audit-Protokolls) & Doppel-Handler-Logger mit dateibasierter Audit-Senke unter \texttt{\~{}/spotforecast2\_safe\_models/logs/}. \\
CRA Anhang I Nr.~1 Abs.~3 lit.~a (keine bekannten ausnutzbaren Schwachstellen) & CPE-Kennung; CycloneDX-SBOM für den Abgleich mit der National Vulnerability Database; \texttt{grep}-Prüfung der Lock-Dateien gegen verbotene Abhängigkeiten. \\
CRA Anhang I Nr.~1 Abs.~3 lit.~b (sicher als Standardeinstellung) & Fail-safe-Standardwert \texttt{on\_missing="raise"}; keine impliziten Netzwerkaufrufe beim Import. \\
NIS-2-RL Art.~21 Abs.~2 lit.~e (Sicherheit bei Erwerb, Entwicklung, Wartung) & SDL-Nachweis, zusammengefasst in \texttt{MODEL\_CARD.md} Abschnitt 10. \\

\end{longtable}

}

\section{Beispiel: Prognose der elektrischen Last}\label{sec-example}

Dieser Abschnitt führt die vollständige
\texttt{spotforecast2-safe}-Pipeline an einer synthetischen Lastreihe
vor, die die täglichen und wöchentlichen Muster einer ENTSO-E-Marktzone
nachbildet. Synthetische statt Live-Daten werden verwendet, damit die
Dokumentenerstellung deterministisch und offlinefähig bleibt. Das Modul
\texttt{downloader/entsoe.py} und ein Open-Meteo-Adapter
\citep{openmeteo24} stellen die korrespondierenden Live-Datenpfade für
den produktiven Einsatz bereit.

Jeder der folgenden neun Schritte besteht aus einer kurzen Erläuterung,
was passiert und warum, sowie einem Code-Block, der beim Render des
Berichts live ausgeführt wird. Wo der produktive Pfad einen Netzzugriff
erfordert, zeigt ein zusätzliches, nicht ausgeführtes Listing den realen
Aufruf. Für alle Berechnungen verwenden wir ausschließlich die
Schnittstelle der Bibliothek \texttt{spotforecast2-safe}. Das Training
läuft über denselben Konfigurations- und Pipeline-Einstieg
(\texttt{ConfigMulti}, \texttt{MultiTask}), den auch der produktive
Betrieb verwendet, so dass der eigentliche Schätzer (LightGBM) nie
direkt aufgerufen werden muss. Jeder Schritt schließt mit einem Satz,
der ihn einer Code- oder Prozessregel und ihrer Rechtsgrundlage
zuordnet. Tabelle~\ref{tbl-step-compliance} fasst diese Zuordnung am
Ende des Abschnitts zusammen.

Die Abbildungen dieses Berichts entstehen dagegen mit den
Plot-Funktionen des Schwesterpakets \texttt{spotforecast2}, die auf
\texttt{matplotlib} und \texttt{plotly} aufsetzen. Beide Bibliotheken
stehen auf der Abhängigkeits-Sperrliste der Code-Regel CR-4 (siehe
Kapitel~\ref{sec-code-rules}), Plot-Funktionalität ist aus
\texttt{spotforecast2-safe} deshalb ausgeschlossen. Die Dokumentations-
und Auswertungsschicht, zu der dieser Bericht gehört, ist nicht Teil des
sicheren Kerns und darf das Schwesterpaket nutzen.
\texttt{spotforecast2-safe} selbst bleibt plotting-frei, und die
Sperrlisten-Prüfung aus CR-4 stellt sicher, dass das so bleibt.

\subsection{Schritt 1: Audit-Protokoll
einschalten}\label{schritt-1-audit-protokoll-einschalten}

Die Bibliothek stellt für Aufzeichnungen einen strukturierten Audit-Sink
bereit. Eine spätere Aufsichtsbehörde kann anhand dieser Aufzeichnungen
rekonstruieren, welche Daten wann verarbeitet wurden. Übergibt der
Aufrufer ein Protokollverzeichnis \texttt{log\_dir}, schreibt die
Bibliothek zusätzlich zur Konsolenausgabe eine Datei, in der jede Zeile
ein JSON-Objekt nach einem versionierten Schema ist (Pflichtfelder unter
anderem \texttt{schema\_version}, \texttt{prev\_hash},
\texttt{timestamp\_utc} und \texttt{event}). Die Sätze sind über das
Feld \texttt{prev\_hash} zu einer Hash-Kette verbunden, deren Prüfung
Schritt 9 zeigt. Ohne diese Angabe protokolliert sie ausschließlich auf
die Konsole. Der Datei-Sink ist fest auf die Stufe INFO gebunden, damit
die Audit-Spur auch dann vollständig bleibt, wenn der Betreiber die
Konsole leiser stellt. Im produktiven Einsatz liegt das Verzeichnis
standardmäßig unter
\texttt{\textasciitilde{}/spotforecast2\_safe\_models/logs/}. Hier
verwenden wir ein lokales Arbeitsverzeichnis, das zu Beginn geleert
wird, damit jeder Render bei Null beginnt. Eine eingebaute Rotation oder
Aufbewahrungsfrist hat die Bibliothek bewusst nicht: Die Aufbewahrung
der Protokolle bleibt eine Betreiberpflicht (Art. 18 KI-VO,
Kapitel~\ref{sec-ki-vo-18}). Fachliche Ereignisse trägt der Aufrufer
über das \texttt{extra}-Argument ein. Wir markieren den Beginn des
Tutorial-Laufs mit einem eigenen Ereignis und lesen zur Kontrolle den
ersten Eintrag der Datei zurück. Diese Praxis instantiiert die
Aufzeichnungspflicht aus Art. 12 KI-VO (Kapitel~\ref{sec-ki-vo-12}) und
Prozessregel PR-4. Der Modul-Docstring des Loggers benennt zusätzlich
IEC 62443-4-2 SAR 6.1 (Zugänglichkeit des Audit-Protokolls) und IEC
61508-3 §7.4.7 als einschlägige Normanforderungen.

\begin{Shaded}
\begin{Highlighting}[]
\ImportTok{import}\NormalTok{ json}
\ImportTok{import}\NormalTok{ logging}
\ImportTok{import}\NormalTok{ shutil}
\ImportTok{from}\NormalTok{ pathlib }\ImportTok{import}\NormalTok{ Path}
\ImportTok{from}\NormalTok{ spotforecast2\_safe.manager.logger }\ImportTok{import}\NormalTok{ setup\_logging}
\NormalTok{artifacts }\OperatorTok{=}\NormalTok{ Path(}\StringTok{"\_paper\_artifacts"}\NormalTok{)}
\NormalTok{shutil.rmtree(artifacts, ignore\_errors}\OperatorTok{=}\VariableTok{True}\NormalTok{)}
\NormalTok{logger, log\_path }\OperatorTok{=}\NormalTok{ setup\_logging(level}\OperatorTok{=}\NormalTok{logging.INFO, log\_dir}\OperatorTok{=}\NormalTok{artifacts }\OperatorTok{/} \StringTok{"logs"}\NormalTok{)}
\NormalTok{logger.info(}
    \StringTok{"Tutorial{-}Lauf beginnt."}\NormalTok{,}
\NormalTok{    extra}\OperatorTok{=}\NormalTok{\{}\StringTok{"event"}\NormalTok{: }\StringTok{"tutorial\_start"}\NormalTok{, }\StringTok{"task"}\NormalTok{: }\StringTok{"lazy"}\NormalTok{,}
           \StringTok{"context"}\NormalTok{: \{}\StringTok{"seed"}\NormalTok{: }\DecValTok{2026}\NormalTok{, }\StringTok{"horizon\_h"}\NormalTok{: }\DecValTok{24}\NormalTok{\}\},}
\NormalTok{)}
\NormalTok{record }\OperatorTok{=}\NormalTok{ json.loads(log\_path.read\_text(encoding}\OperatorTok{=}\StringTok{"utf{-}8"}\NormalTok{).splitlines()[}\DecValTok{0}\NormalTok{])}
\BuiltInTok{print}\NormalTok{(}\SpecialStringTok{f"Audit{-}Datei:    }\SpecialCharTok{\{}\NormalTok{log\_path}\SpecialCharTok{\}}\SpecialStringTok{"}\NormalTok{)}
\BuiltInTok{print}\NormalTok{(}\SpecialStringTok{f"Schema{-}Version: }\SpecialCharTok{\{}\NormalTok{record[}\StringTok{\textquotesingle{}schema\_version\textquotesingle{}}\NormalTok{]}\SpecialCharTok{\}}\SpecialStringTok{, Event: }\SpecialCharTok{\{}\NormalTok{record[}\StringTok{\textquotesingle{}event\textquotesingle{}}\NormalTok{]}\SpecialCharTok{\}}\SpecialStringTok{"}\NormalTok{)}
\end{Highlighting}
\end{Shaded}

\begin{verbatim}
2026-08-13 18:58:13,818 - sf2-safe-logger - INFO - Persistent logging initialized at: _paper_artifacts/logs/sf2-safe-logger_20260813_185813.log
2026-08-13 18:58:13,818 - sf2-safe-logger - INFO - Tutorial-Lauf beginnt.
Audit-Datei:    _paper_artifacts/logs/sf2-safe-logger_20260813_185813.log
Schema-Version: 2.0.0, Event: audit_log_init
\end{verbatim}

\subsection{Schritt 2: Datengrundlage}\label{schritt-2-datengrundlage}

Der \texttt{spotforecast2-safe} Vorhersageprozess erzeugt keine Daten,
er bezieht sie. Die Ziellast stammt als aggregierte Ist-Last von der
ENTSO-E-Transparenzplattform. Die Wetter-Kovariaten stammen aus dem
Open-Meteo-Archiv und aus der Open-Meteo-Wettervorhersage: Über dem
Horizont sind Kovariaten selbst Prognosen, keine Messungen. Das folgende
Listing zeigt den produktiven Bezugspfad\footnote{Es wird beim Render
  bewusst nicht ausgeführt, weil beide Aufrufe Netzzugriff und für
  ENTSO-E einen API-Schlüssel erfordern.}:

\begin{Shaded}
\begin{Highlighting}[]
\ImportTok{import}\NormalTok{ os}
\ImportTok{from}\NormalTok{ spotforecast2\_safe.downloader.entsoe }\ImportTok{import}\NormalTok{ download\_new\_data}
\ImportTok{from}\NormalTok{ spotforecast2\_safe.data.fetch\_data }\ImportTok{import}\NormalTok{ fetch\_weather\_data}
\NormalTok{download\_new\_data(api\_key}\OperatorTok{=}\NormalTok{os.environ[}\StringTok{"ENTSOE\_API\_KEY"}\NormalTok{], country\_code}\OperatorTok{=}\StringTok{"FR"}\NormalTok{)}
\NormalTok{weather }\OperatorTok{=}\NormalTok{ fetch\_weather\_data(cov\_start}\OperatorTok{=}\StringTok{"2025{-}01{-}01"}\NormalTok{, cov\_end}\OperatorTok{=}\StringTok{"2025{-}03{-}31"}\NormalTok{)}
\end{Highlighting}
\end{Shaded}

Für den Bericht ersetzen wir diesen Bezugspfad durch ein synthetisches
Abbild: eine Lastreihe als Summe von vier einfachen Komponenten, nämlich
einem leichten Aufwärtstrend (etwa: zunehmender Stromverbrauch über das
Quartal), einem Tagesgang mit Spitze am Mittag, einem Wochentagszuschlag
(an Werktagen ist die Last höher als am Wochenende) und einem
normalverteilten Messrauschen. Die Reihe stammt aus der Paketfunktion
\texttt{make\_synthetic\_load} und wird mit einem festen Seed
(Startwert) erzeugt, so dass sie bei jedem Render byteweise identisch
ist und der Bericht offline entsteht. Der Index ist eine durchgehende
UTC-Stundenfolge und damit genau das Datenformat, das
\texttt{spotforecast2-safe} erwartet. Die dokumentierte Wahl der
Datengrundlage operationalisiert die Daten-Governance aus Art. 10 KI-VO
(Kapitel~\ref{sec-ki-vo-10}). Der feste Seed setzt Code-Regel CR-2
\emph{Determinismus} um.

\begin{Shaded}
\begin{Highlighting}[]
\ImportTok{import}\NormalTok{ numpy }\ImportTok{as}\NormalTok{ np}
\ImportTok{import}\NormalTok{ pandas }\ImportTok{as}\NormalTok{ pd}
\ImportTok{from}\NormalTok{ spotforecast2\_safe.data }\ImportTok{import}\NormalTok{ make\_synthetic\_load}
\NormalTok{y }\OperatorTok{=}\NormalTok{ make\_synthetic\_load(seed}\OperatorTok{=}\DecValTok{2026}\NormalTok{)}
\NormalTok{y.head(}\DecValTok{3}\NormalTok{)}
\end{Highlighting}
\end{Shaded}

\begin{verbatim}
2025-01-01 00:00:00+00:00    47.103439
2025-01-01 01:00:00+00:00    47.757509
2025-01-01 02:00:00+00:00    47.089588
Freq: h, Name: load, dtype: float64
\end{verbatim}

\subsection{Schritt 3: Datenlücken explizit
behandeln}\label{schritt-3-datenluxfccken-explizit-behandeln}

Reale Sensordaten enthalten gelegentlich Lücken, etwa wenn ein Zähler
neu startet oder eine Übertragung ausfällt. Eine sicherheitskritische
Bibliothek darf solche Lücken nicht stillschweigend überspielen.
\texttt{spotforecast2-safe} verweigert deshalb das Anpassen auf einer
unvollständigen Zielreihe: Die Eingangsprüfung \texttt{check\_y}, die
jedem Trainingslauf vorgeschaltet ist, weist fehlende Werte mit einer
expliziten Ausnahme zurück (Kapitel~\ref{sec-fail-safe-nan}, CR-3
\emph{Fail-safe}). Wir simulieren einen Messausfall am Vormittag des 14.
Februar 2025 und arbeiten mit dieser Reihe weiter:

\begin{Shaded}
\begin{Highlighting}[]
\ImportTok{from}\NormalTok{ spotforecast2\_safe.preprocessing.checking }\ImportTok{import}\NormalTok{ check\_y}
\NormalTok{y\_gapped }\OperatorTok{=}\NormalTok{ y.copy()}
\NormalTok{y\_gapped.loc[}\StringTok{"2025{-}02{-}14 09:00"}\NormalTok{:}\StringTok{"2025{-}02{-}14 14:00"}\NormalTok{] }\OperatorTok{=}\NormalTok{ np.nan}
\ControlFlowTok{try}\NormalTok{:}
\NormalTok{    check\_y(y\_gapped)}
\ControlFlowTok{except} \PreprocessorTok{ValueError} \ImportTok{as}\NormalTok{ err:}
    \BuiltInTok{print}\NormalTok{(}\SpecialStringTok{f"ValueError: }\SpecialCharTok{\{}\NormalTok{err}\SpecialCharTok{\}}\SpecialStringTok{"}\NormalTok{)}
\end{Highlighting}
\end{Shaded}

\begin{verbatim}
ValueError: `y` has missing values.
\end{verbatim}

Der Abbruch zwingt den Betreiber zu einer ausdrücklichen Entscheidung,
wie die Lücke zu behandeln ist. An dieser Stelle unterscheidet sich
\texttt{spotforecast2-safe} grundsätzlich von der verbreiteten
Standard-Imputation, wie sie etwa imputeTS \citep{mori17a}, mice
\citep{buur11a} oder der \texttt{SimpleImputer} von scikit-learn
\citep{pedr11a} anbieten. \citet{mori15a} vergleichen solche Verfahren
für univariate Zeitreihen. Diese Werkzeuge geben allein die imputierte
Reihe zurück: Die eingesetzten Werte sind erfundene, nie gemessene
Größen, die von echten Messwerten anschließend nicht mehr zu
unterscheiden sind und als scheinbare Beobachtungen in das
Modelltraining eingehen. Die statistische Literatur behandelt fehlende
Werte aus diesem Grund als eigenständiges Inferenzproblem, dessen
Behandlung vom Fehlmechanismus abhängt und dokumentiert werden muss
\citep{rubi76a, littl19a}.

\texttt{spotforecast2-safe} trennt dagegen das Imputieren vom Verwenden.
Die Funktion \texttt{get\_missing\_weights} füllt die Lücke ebenfalls,
denn die Merkmalskonstruktion benötigt einen lückenlosen Zeitindex.
Zusätzlich liefert sie aber eine Gewichtsserie zurück, die innerhalb der
Lücke und für \texttt{window\_size} nachfolgende Zeilen null ist, weil
die Lag-Merkmale dieser Zeilen sonst gefüllte Werte aufnehmen würden.
Diese Serie geht als Stichprobengewicht in das Training ein: Eine
imputierte Stunde trägt kein Trainingssignal, bleibt aber im
Audit-Protokoll zählbar. Genau dieser Mechanismus läuft in der Pipeline
aus Schritt 5 automatisch mit: Die voreingestellte Imputationsmethode
\texttt{"weighted"} ruft \texttt{get\_missing\_weights} intern auf und
übergibt die Gewichtsserie als \texttt{weight\_func} an den Forecaster.
Wir zeigen den Mechanismus vorab an der Reihe mit dem Messausfall. Bei
den 168 Lags des Modells aus Schritt 5 entwertet die Lücke zusätzlich
die 168 nachfolgenden Stunden:

\begin{Shaded}
\begin{Highlighting}[]
\ImportTok{from}\NormalTok{ spotforecast2\_safe.preprocessing }\ImportTok{import}\NormalTok{ get\_missing\_weights}
\NormalTok{y\_gapped\_filled, gap\_weights }\OperatorTok{=}\NormalTok{ get\_missing\_weights(}
\NormalTok{    y\_gapped.to\_frame(), window\_size}\OperatorTok{=}\DecValTok{168}
\NormalTok{)}
\BuiltInTok{print}\NormalTok{(}\SpecialStringTok{f"Fehlende Stunden:      }\SpecialCharTok{\{}\BuiltInTok{int}\NormalTok{(y\_gapped.isna().}\BuiltInTok{sum}\NormalTok{())}\SpecialCharTok{\}}\SpecialStringTok{"}\NormalTok{)}
\BuiltInTok{print}\NormalTok{(}\SpecialStringTok{f"Stunden mit Gewicht 0: }\SpecialCharTok{\{}\BuiltInTok{int}\NormalTok{((gap\_weights }\OperatorTok{==} \DecValTok{0}\NormalTok{).}\BuiltInTok{sum}\NormalTok{())}\SpecialCharTok{\}}\SpecialStringTok{ "}
      \SpecialStringTok{f"(Lücke plus 168 Folgestunden)"}\NormalTok{)}
\end{Highlighting}
\end{Shaded}

\begin{verbatim}
Fehlende Stunden:      6
Stunden mit Gewicht 0: 174 (Lücke plus 168 Folgestunden)
\end{verbatim}

\begin{figure}

\centering{

\pandocbounded{\includegraphics[keepaspectratio]{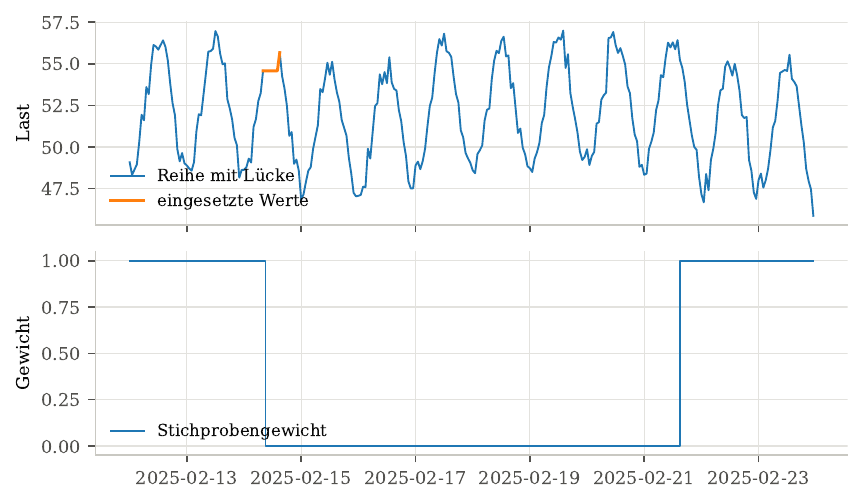}}

}

\caption{\label{fig-imputation-weights}Explizite Lückenbehandlung mit
\texttt{get\_missing\_weights} am simulierten sechsstündigen Messausfall
(14. Februar 2025, 09:00 bis 14:00 UTC). Oberes Panel: Reihe mit Lücke
(\emph{blau}) und die von der Füllung eingesetzten Werte
(\emph{orange}). Unteres Panel: zugehörige Gewichtsserie; das
Stichprobengewicht ist innerhalb der Lücke und für die 168 nachfolgenden
Stunden (das Lag-Fenster des Modells aus Schritt 5) null, so dass
eingesetzte Werte kein Trainingssignal tragen.}

\end{figure}

Abbildung~\ref{fig-imputation-weights} stellt beide Rückgabewerte dar:
oben die Reihe mit der Lücke und den eingesetzten Werten, unten die
zugehörige Gewichtsserie, deren Nullzone die Lücke und das nachfolgende
Lag-Fenster abdeckt. Der sechsstündige Messausfall entwertet damit gut
eine Woche Trainingsdaten. Genau diese Konsequenz verschweigt eine
Standard-Imputation.

Eine gesonderte Vorab-Imputation ist danach nicht nötig, und dieser
Schritt endet deshalb bewusst ohne einen weiteren Verarbeitungsaufruf:
Die Lücke bleibt in der Reihe stehen, und die Pipeline aus Schritt 5
behandelt sie über ihre voreingestellte Imputationsmethode automatisch
nach genau dem gezeigten Mechanismus. Die Entscheidung des Betreibers
besteht in der dokumentierten Wahl dieser Methode, nicht im Erfinden von
Werten. Diese Disziplin operationalisiert Art. 10 KI-VO
(Daten-Governance, Kapitel~\ref{sec-ki-vo-10}) und CR-3
\emph{Fail-safe}.

\subsection{Schritt 4: Prognoseursprung
festlegen}\label{schritt-4-prognoseursprung-festlegen}

Bei Zeitreihen darf die Aufteilung in Trainings- und Evaluationsdaten
nicht zufällig erfolgen, weil sonst zukünftige Information ins Training
einsickern würde. Im produktiven Betrieb gibt es darüber hinaus gar
keinen statischen Train-Test-Schnitt: Das Modell wird täglich neu auf
einem rollierenden Fenster der jüngsten Historie trainiert (in
\texttt{ConfigMulti} steuert \texttt{train\_size} diese Spanne,
voreingestellt drei Jahre), und der ``Testsatz'' ist stets der morgige
Zieltag. Das Tutorial friert deshalb einen einzelnen Prognoseursprung
reproduzierbar ein: Trainiert wird chronologisch auf den ersten zwei
Monaten, die Zieltage des März bleiben für die Bewertung in den
Schritten 7 und 8 zurückbehalten. Der Messausfall aus Schritt 3 liegt
damit im Trainingsbereich. Diese chronologische Aufteilung ist die
einfachste Variante der in Art. 10 Abs. 2 lit. f und g KI-VO
(Kapitel~\ref{sec-ki-vo-10}) geforderten Repräsentativitätsprüfung.

\begin{Shaded}
\begin{Highlighting}[]
\NormalTok{y\_train }\OperatorTok{=}\NormalTok{ y\_gapped.loc[}\StringTok{"2025{-}01{-}01"}\NormalTok{:}\StringTok{"2025{-}03{-}01"}\NormalTok{]}
\NormalTok{y\_eval  }\OperatorTok{=}\NormalTok{ y\_gapped.loc[}\StringTok{"2025{-}03{-}02"}\NormalTok{:]}
\BuiltInTok{print}\NormalTok{(}\SpecialStringTok{f"Trainingsbereich:   }\SpecialCharTok{\{}\NormalTok{y\_train}\SpecialCharTok{.}\NormalTok{index[}\DecValTok{0}\NormalTok{]}\SpecialCharTok{.}\NormalTok{date()}\SpecialCharTok{\}}\SpecialStringTok{ bis }\SpecialCharTok{\{}\NormalTok{y\_train}\SpecialCharTok{.}\NormalTok{index[}\OperatorTok{{-}}\DecValTok{1}\NormalTok{]}\SpecialCharTok{.}\NormalTok{date()}\SpecialCharTok{\}}\SpecialStringTok{ "}
      \SpecialStringTok{f"(}\SpecialCharTok{\{}\BuiltInTok{len}\NormalTok{(y\_train)}\SpecialCharTok{\}}\SpecialStringTok{ Stunden)"}\NormalTok{)}
\BuiltInTok{print}\NormalTok{(}\SpecialStringTok{f"Evaluationsbereich: }\SpecialCharTok{\{}\NormalTok{y\_eval}\SpecialCharTok{.}\NormalTok{index[}\DecValTok{0}\NormalTok{]}\SpecialCharTok{.}\NormalTok{date()}\SpecialCharTok{\}}\SpecialStringTok{ bis }\SpecialCharTok{\{}\NormalTok{y\_eval}\SpecialCharTok{.}\NormalTok{index[}\OperatorTok{{-}}\DecValTok{1}\NormalTok{]}\SpecialCharTok{.}\NormalTok{date()}\SpecialCharTok{\}}\SpecialStringTok{ "}
      \SpecialStringTok{f"(}\SpecialCharTok{\{}\BuiltInTok{len}\NormalTok{(y\_eval)}\SpecialCharTok{\}}\SpecialStringTok{ Stunden)"}\NormalTok{)}
\end{Highlighting}
\end{Shaded}

\begin{verbatim}
Trainingsbereich:   2025-01-01 bis 2025-03-01 (1440 Stunden)
Evaluationsbereich: 2025-03-02 bis 2025-03-31 (720 Stunden)
\end{verbatim}

\subsection{Schritt 5: Pipeline konfigurieren und
trainieren}\label{schritt-5-pipeline-konfigurieren-und-trainieren}

In der Produktion wird das Training nicht von Hand verdrahtet, sondern
über denselben Konfigurations-Einstieg gestartet, den auch der tägliche
Betrieb nutzt: Ein \texttt{ConfigMulti}-Objekt bündelt alle
Entscheidungen (Horizont, Merkmalsfenster, Seed, Cache-Verzeichnis), und
die Klasse \texttt{MultiTask} führt darauf die Vorbereitungs-, Prüf- und
Trainingsstufen aus. Für den Ein-Aufruf-Einsatz kapselt
\texttt{multitask.runner.run(...)} dieselbe Kette zusätzlich in einer
einzigen Funktion. Wir konfigurieren nur zwei fachliche Größen: den
Prognosehorizont von 24 Stunden und das Merkmalsfenster von 168 Stunden.
\emph{Lags} sind die zurückliegenden Werte, die das Modell als Eingabe
verwendet: mit \texttt{lags=168} sieht das Modell für jede Stunde die
letzte volle Woche als Kontext, was für die wöchentliche Saisonalität
(Werktag versus Wochenende) entscheidend ist. ``Recursive'' heißt, dass
die Vorhersage einer Stunde wiederum als Eingabe für die nächste Stunde
dient. Alles Weitere sind Bibliotheks-Defaults: der LightGBM-Schätzer
mit festem Seed und fixierter Parallelität (\texttt{n\_jobs=1}), die
Ausreißer-Erkennung, deren Befunde rein protokollierend in das
Pipeline-Protokoll eingehen und keine Werte verändern und die gewichtete
Imputation aus Schritt 3, die den dort belassenen Messausfall jetzt
automatisch füllt und über die Gewichtsserie aus dem Training
heraushält. Eine Hyperparameter-Suche findet bewusst nicht statt. Die
tägliche Optimierung des produktiven Betriebs bleibt außerhalb dieses
Tutorials. Dieser Schritt setzt CR-2 um (geseedeter Schätzer,
\texttt{n\_jobs=1}) und adressiert die Genauigkeits- und
Robustheitsanforderung aus Art. 15 KI-VO (Kapitel~\ref{sec-ki-vo-15}).

\begin{Shaded}
\begin{Highlighting}[]
\ImportTok{from}\NormalTok{ spotforecast2\_safe.configurator }\ImportTok{import}\NormalTok{ ConfigMulti}
\ImportTok{from}\NormalTok{ spotforecast2\_safe.multitask }\ImportTok{import}\NormalTok{ MultiTask}
\NormalTok{cfg }\OperatorTok{=}\NormalTok{ ConfigMulti(}
\NormalTok{    predict\_size}\OperatorTok{=}\DecValTok{24}\NormalTok{,}
\NormalTok{    lags\_consider}\OperatorTok{=}\NormalTok{[}\DecValTok{168}\NormalTok{],}
\NormalTok{    window\_size}\OperatorTok{=}\DecValTok{168}\NormalTok{,}
\NormalTok{    use\_exogenous\_features}\OperatorTok{=}\VariableTok{False}\NormalTok{,}
\NormalTok{    use\_outlier\_detection}\OperatorTok{=}\VariableTok{True}\NormalTok{,}
\NormalTok{    auto\_save\_models}\OperatorTok{=}\VariableTok{False}\NormalTok{,}
\NormalTok{    data\_frame\_name}\OperatorTok{=}\StringTok{"synthetic\_load"}\NormalTok{,}
\NormalTok{    cache\_home}\OperatorTok{=}\NormalTok{artifacts }\OperatorTok{/} \StringTok{"cache"}\NormalTok{,}
\NormalTok{)}
\NormalTok{mt }\OperatorTok{=}\NormalTok{ MultiTask(cfg, task}\OperatorTok{=}\StringTok{"lazy"}\NormalTok{,}
\NormalTok{               dataframe}\OperatorTok{=}\NormalTok{y\_train.to\_frame(), data\_test}\OperatorTok{=}\NormalTok{y\_eval.to\_frame())}
\NormalTok{mt.prepare\_data().detect\_outliers().impute().build\_exogenous\_features()}
\NormalTok{result }\OperatorTok{=}\NormalTok{ mt.run(task}\OperatorTok{=}\StringTok{"lazy"}\NormalTok{)}
\NormalTok{fc }\OperatorTok{=}\NormalTok{ mt.results[}\StringTok{"lazy"}\NormalTok{][}\StringTok{"load"}\NormalTok{][}\StringTok{"forecaster"}\NormalTok{]}
\BuiltInTok{print}\NormalTok{(}\SpecialStringTok{f"Trainingsfenster: }\SpecialCharTok{\{}\NormalTok{fc}\SpecialCharTok{.}\NormalTok{training\_range\_[}\DecValTok{0}\NormalTok{]}\SpecialCharTok{.}\NormalTok{date()}\SpecialCharTok{\}}\SpecialStringTok{ bis "}
      \SpecialStringTok{f"}\SpecialCharTok{\{}\NormalTok{fc}\SpecialCharTok{.}\NormalTok{training\_range\_[}\DecValTok{1}\NormalTok{]}\SpecialCharTok{.}\NormalTok{date()}\SpecialCharTok{\}}\SpecialStringTok{"}\NormalTok{)}
\BuiltInTok{print}\NormalTok{(}\SpecialStringTok{f"Schätzer:         }\SpecialCharTok{\{}\BuiltInTok{type}\NormalTok{(fc.estimator)}\SpecialCharTok{.}\VariableTok{\_\_name\_\_}\SpecialCharTok{\}}\SpecialStringTok{, "}
      \SpecialStringTok{f"Lags: }\SpecialCharTok{\{}\NormalTok{fc}\SpecialCharTok{.}\NormalTok{lags}\SpecialCharTok{.}\NormalTok{size}\SpecialCharTok{\}}\SpecialStringTok{, Seed: }\SpecialCharTok{\{}\NormalTok{cfg}\SpecialCharTok{.}\NormalTok{random\_state}\SpecialCharTok{\}}\SpecialStringTok{"}\NormalTok{)}
\BuiltInTok{print}\NormalTok{(}\SpecialStringTok{f"Lückengewichte:   }\SpecialCharTok{\{}\StringTok{\textquotesingle{}übergeben\textquotesingle{}} \ControlFlowTok{if}\NormalTok{ mt}\SpecialCharTok{.}\NormalTok{weight\_func }\KeywordTok{is} \KeywordTok{not} \VariableTok{None} \ControlFlowTok{else} \StringTok{\textquotesingle{}keine\textquotesingle{}}\SpecialCharTok{\}}\SpecialStringTok{"}\NormalTok{)}
\end{Highlighting}
\end{Shaded}

\begin{verbatim}
Trainingsfenster: 2025-01-01 bis 2025-03-01
Schätzer:         LGBMRegressor, Lags: 168, Seed: 314159
Lückengewichte:   übergeben
\end{verbatim}

Mit \texttt{use\_exogenous\_features=False} bleibt der Lauf bewusst
zielgrößen-only und damit offlinefähig. In der Produktion baut dieselbe
Pipeline-Stufe \texttt{build\_exogenous\_features} die Kovariaten
bibliotheksintern auf, nämlich Wetter von Open-Meteo sowie Kalender-,
Feiertags- und Sonnenstands-Merkmale samt zyklischer Kodierung. Das
folgende, nicht ausgeführte Listing zeigt die entsprechenden
Konfigurationsschalter:

\begin{Shaded}
\begin{Highlighting}[]
\NormalTok{cfg\_prod }\OperatorTok{=}\NormalTok{ ConfigMulti(}
\NormalTok{    use\_exogenous\_features}\OperatorTok{=}\VariableTok{True}\NormalTok{,      }\CommentTok{\# Wetter (Open{-}Meteo) und Kalender}
\NormalTok{    include\_holiday\_features}\OperatorTok{=}\VariableTok{True}\NormalTok{,    }\CommentTok{\# Feiertagskalender}
\NormalTok{    include\_ephemeris\_features}\OperatorTok{=}\VariableTok{True}\NormalTok{,  }\CommentTok{\# Sonnenstand, Tageslichtdauer}
\NormalTok{    on\_weather\_failure}\OperatorTok{=}\StringTok{"raise"}\NormalTok{,       }\CommentTok{\# CR{-}3: kein stiller Weiterlauf}
\NormalTok{)}
\end{Highlighting}
\end{Shaded}

\subsection{Schritt 6: Vorhersage
erzeugen}\label{schritt-6-vorhersage-erzeugen}

Der trainierte Forecaster berechnet eine Tages-Vorhersage. Das Modell
startet mit den letzten Trainings-Stunden als Anfangs-Lag-Fenster und
schiebt diese Stunde für Stunde rekursiv weiter. Die Pipeline hat
dieselbe Vorhersage bereits als Teil ihres Ergebnis-Pakets erzeugt
(\texttt{future\_pred}, zusammen mit In-Sample-Anpassung, Metriken und
Modellidentität): Ein Betreiber erhält nie eine einfache Zahlenreihe,
sondern ein interpretierbares Paket, also die Informationsbasis, die
Art. 13 KI-VO (Kapitel~\ref{sec-ki-vo-13}) für die Transparenz gegenüber
dem Betreiber verlangt. Der Vergleich beider Aufrufe zeigt zugleich den
Determinismus aus CR-2 im Vollzug: gleicher Zustand, gleiche Ausgabe.

\begin{Shaded}
\begin{Highlighting}[]
\NormalTok{y\_eval\_24h }\OperatorTok{=}\NormalTok{ y\_eval.iloc[:}\DecValTok{24}\NormalTok{]}
\NormalTok{y\_pred }\OperatorTok{=}\NormalTok{ fc.predict(steps}\OperatorTok{=}\DecValTok{24}\NormalTok{)}
\NormalTok{y\_pred.name }\OperatorTok{=} \StringTok{"forecast"}
\BuiltInTok{print}\NormalTok{(y\_pred.head(}\DecValTok{3}\NormalTok{))}
\BuiltInTok{print}\NormalTok{(}\SpecialStringTok{f"Identisch mit future\_pred aus Schritt 5: "}
      \SpecialStringTok{f"}\SpecialCharTok{\{}\BuiltInTok{bool}\NormalTok{((y\_pred }\OperatorTok{==}\NormalTok{ result[}\StringTok{\textquotesingle{}future\_pred\textquotesingle{}}\NormalTok{]).}\BuiltInTok{all}\NormalTok{())}\SpecialCharTok{\}}\SpecialStringTok{"}\NormalTok{)}
\end{Highlighting}
\end{Shaded}

\begin{verbatim}
2025-03-02 00:00:00+00:00    48.093535
2025-03-02 01:00:00+00:00    48.347808
2025-03-02 02:00:00+00:00    47.748086
Freq: h, Name: forecast, dtype: float64
Identisch mit future_pred aus Schritt 5: True
\end{verbatim}

\subsection{Schritt 7: Genauigkeit
messen}\label{schritt-7-genauigkeit-messen}

\begin{figure}

\centering{

\pandocbounded{\includegraphics[keepaspectratio]{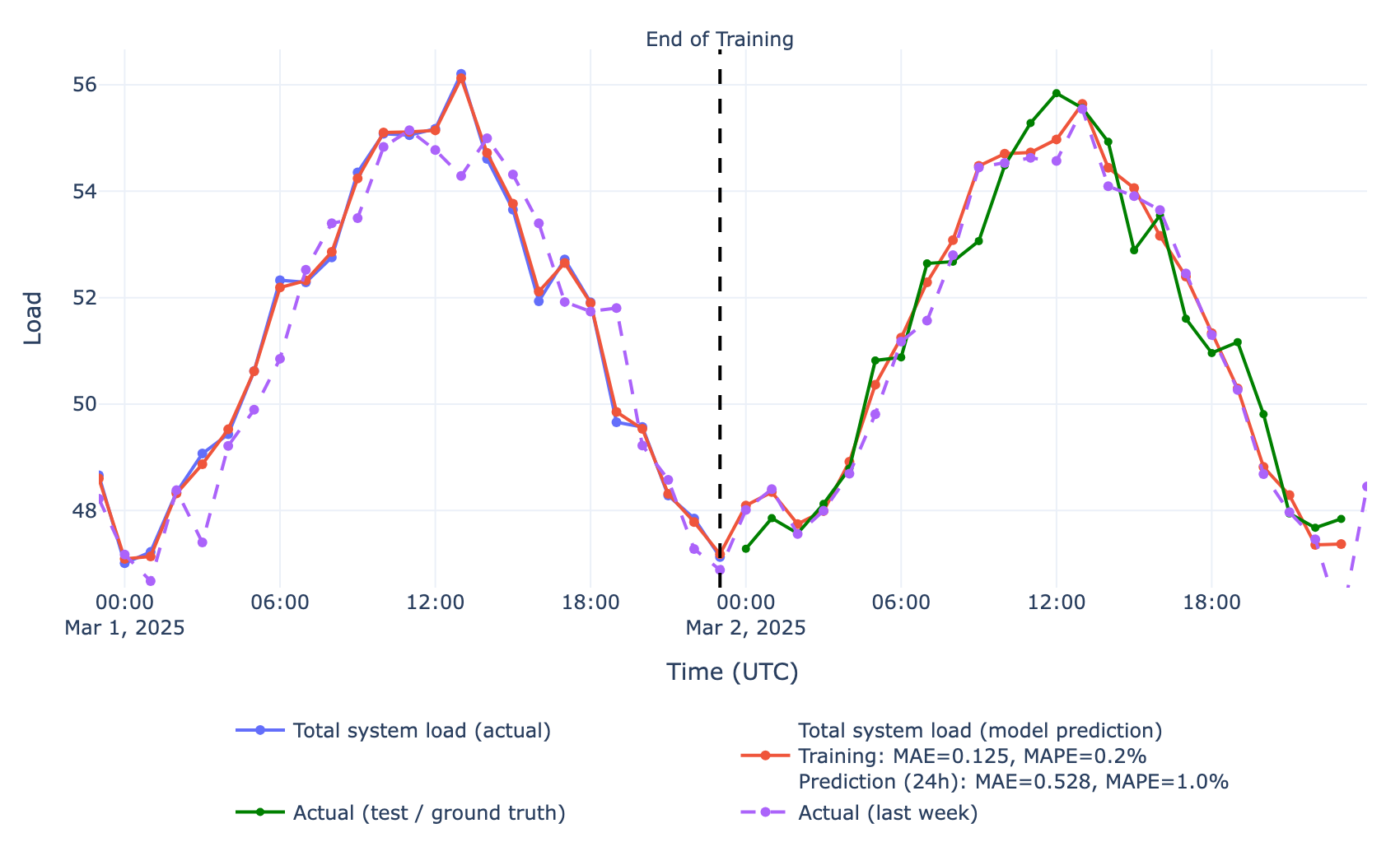}}

}

\caption{\label{fig-forecast24}24-Stunden-Lastprognose als unveränderte
Report-Abbildung des Produktionssystems, erzeugt mit \texttt{make\_plot}
aus dem Ergebnis-Paket der Pipeline aus Schritt 5. Das Fenster zeigt die
letzten 24 Trainingsstunden und den 24-Stunden-Horizont, die
\emph{gestrichelte} vertikale Linie (End of Training) markiert das
Trainingsende. Total system load (actual): beobachtete Last bis zum
Trainingsende. Total system load (model prediction): In-Sample-Anpassung
und anschließende 24-h-Vorhersage. Actual (test / ground truth): die
erst nach dem Zieltag verfügbare Ist-Last, gegen die das
Day-after-Scoring rechnet. Actual (last week): Last der Vorwoche als
Saisonalitätskontext. Die Kennzahlen der Legende stammen unverändert aus
dem Ergebnis-Paket (\texttt{metrics\_train} und
\texttt{metrics\_future}), MAE in Lasteinheiten, MAPE in Prozent.}

\end{figure}

Drei Standard-Maße quantifizieren die Abweichung zwischen Prognose und
tatsächlich beobachteter Last. Der mittlere absolute Fehler (Mean
Absolute Error, MAE) gibt die durchschnittliche Abweichung in den
Einheiten der Lastreihe an. Die Wurzel des mittleren quadratischen
Fehlers (Root Mean Squared Error, RMSE) bestraft große Abweichungen
überproportional und ist ebenfalls in Lasteinheiten skaliert. Der
mittlere absolute prozentuale Fehler (Mean Absolute Percentage Error,
MAPE) gibt die mittlere prozentuale Abweichung an. Die Paketfunktion
\texttt{score\_forecasts} liefert diese Maße mit einer Zeile je
Verfahren und der Zahl der ausgewerteten Stunden (\texttt{n}). Die
Legende von Abbildung~\ref{fig-forecast24} rechnet dagegen nichts nach,
sondern übernimmt ihre Kennzahlen unverändert aus dem Ergebnis-Paket der
Pipeline. Im laufenden Betrieb ist dieser Schritt zeitversetzt: Bewertet
wird erst, wenn die Ist-Last publiziert ist, also am Folgetag, und
nachträgliche Revisionen der Ist-Werte führen über ein mehrwöchiges
Fenster zu erneutem Scoring. Für diese tagesweise Bewertung stellt die
Bibliothek \texttt{score\_forecasts\_by\_period} und
\texttt{aggregate\_period\_scores} bereit. Hier genügt
\texttt{score\_forecasts} über den eingefrorenen Zieltag, ergänzt um das
Scoring, das die Pipeline selbst bereits mitliefert (bei einem
24-Stunden-Horizont stimmt das tagesbezogene
\texttt{metrics\_future\_one\_day} mit \texttt{metrics\_future}
überein). Die Maße dienen der Genauigkeits-Anforderung aus Art. 15 KI-VO
(Kapitel~\ref{sec-ki-vo-15}). Die tägliche Nachbewertung gegen
publizierte Ist-Werte setzt zugleich die in Art. 72 KI-VO geforderte
Beobachtung nach dem Inverkehrbringen um (Kapitel~\ref{sec-ki-vo-72}).

\begin{Shaded}
\begin{Highlighting}[]
\ImportTok{from}\NormalTok{ spotforecast2\_safe.processing }\ImportTok{import}\NormalTok{ score\_forecasts}
\NormalTok{scores }\OperatorTok{=}\NormalTok{ score\_forecasts(}
\NormalTok{    \{}\StringTok{"Modellprognose"}\NormalTok{: y\_pred\}, y\_eval\_24h, metrics}\OperatorTok{=}\NormalTok{(}\StringTok{"mae"}\NormalTok{, }\StringTok{"rmse"}\NormalTok{, }\StringTok{"mape"}\NormalTok{)}
\NormalTok{)}
\BuiltInTok{print}\NormalTok{(scores.}\BuiltInTok{round}\NormalTok{(}\DecValTok{3}\NormalTok{))}
\BuiltInTok{print}\NormalTok{(}\SpecialStringTok{f"Pipeline{-}eigenes Scoring (metrics\_future): "}
      \SpecialStringTok{f"MAE }\SpecialCharTok{\{}\NormalTok{result[}\StringTok{\textquotesingle{}metrics\_future\textquotesingle{}}\NormalTok{][}\StringTok{\textquotesingle{}mae\textquotesingle{}}\NormalTok{]}\SpecialCharTok{:.3f\}}\SpecialStringTok{"}\NormalTok{)}
\end{Highlighting}
\end{Shaded}

\begin{verbatim}
                  mae   rmse   mape   n
Modellprognose  0.528  0.626  1.026  24
Pipeline-eigenes Scoring (metrics_future): MAE 0.528
\end{verbatim}

Abbildung~\ref{fig-forecast24} ist keine eigens für diesen Bericht
gebaute Auswertung, sondern die Report-Abbildung des Produktionssystems:
\texttt{make\_plot} aus \texttt{spotforecast2.plots} erzeugt sie
unmittelbar aus dem Ergebnis-Paket aus Schritt 5, und das Dokument
rechnet dafür keine einzige Kennzahl nach. Die in Schritt 6 beschriebene
Paket-Idee wird damit visuell eingelöst, denn Trainingsanpassung,
24-h-Vorhersage, Metriken und Vergleichskontext erscheinen gemeinsam in
einer Darstellung (Art. 13 KI-VO, Kapitel~\ref{sec-ki-vo-13}).
Strukturell folgt die Abbildung \citet{chag24a}. Die gestrichelte Spur
Actual (last week) zeigt die Last der Vorwoche als unbewerteten
Saisonalitätskontext, die grüne Spur Actual (test / ground truth) die
erst nach dem Zieltag publizierte Ist-Last. Die englischen
Beschriftungen bleiben unverändert, weil dieselbe Abbildung im Betrieb
als Report ausgeliefert wird.

\subsection{Schritt 8:
Rolling-Origin-Backtest}\label{schritt-8-rolling-origin-backtest}

Ein einziger Train-Test-Schnitt prüft die Modellgüte nur an einer
einzigen Stelle der Zeitachse. Ein \emph{Rolling-Origin-Backtest}
schiebt den Trennpunkt schrittweise durch die Zeitreihe und führt für
jeden Schritt eine 24-Stunden-Vorhersage aus. Daraus entsteht eine ganze
Verteilung von Genauigkeitsmessungen, die robuster gegen einzelne
Ausreißer ist.

\begin{figure}

\centering{

\pandocbounded{\includegraphics[keepaspectratio]{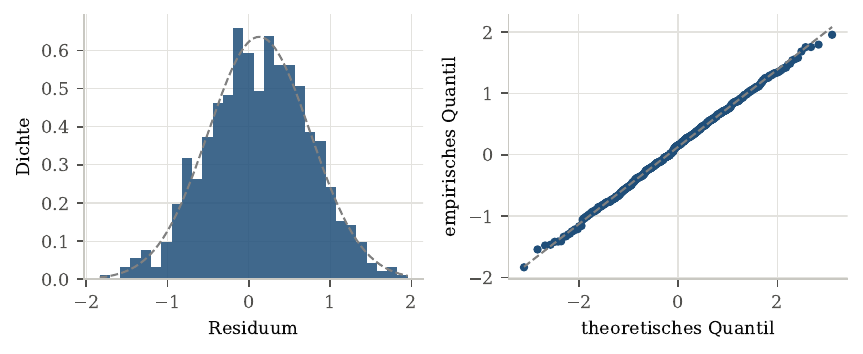}}

}

\caption{\label{fig-residuals}Residuen der 24-Stunden-Prognosen aus dem
Rolling-Origin-Backtest in Schritt 8. Links: Histogramm mit überlagerter
Normaldichte, angepasst an die empirischen Momente. Rechts: sortierte
Residuen gegen Standardnormal-Quantile (QQ-Diagramm). Die Konzentration
der Punkte um die Diagonale spricht für eine annähernd normalverteilte
Fehlerstruktur.}

\end{figure}

Im realen System läuft diese Prüfung nicht im operativen Scoring,
sondern innerhalb der täglichen Modellauswahl: Dort bildet eine
Rolling-Origin-Kreuzvalidierung auf dem Trainingsfenster die innere
Zielfunktion der Hyperparameter-Suche. Jeder Ursprung verwendet
ausschließlich Daten, die ihm zeitlich vorausgehen. Das Tutorial zeigt
sie als Validierungsinstrument über den gesamten Evaluationsbereich. Als
Eingabe dient die von der Pipeline gefüllte Trainingsreihe, verlängert
um die unveränderten Evaluationsdaten, denn der Backtest benötigt eine
lückenlose Reihe. Der Forecaster trägt seine Gewichtsfunktion aus
Schritt 5 mit sich, so dass die gefüllten Stunden auch beim erneuten
Anpassen im ersten Fold kein Trainingssignal erhalten. Die Prüfung
unterstützt damit das Risikomanagement nach Art. 9 KI-VO
(Kapitel~\ref{sec-ki-vo-9}) und die in Art. 10 Abs. 3 KI-VO
(Kapitel~\ref{sec-ki-vo-10}) geforderte Repräsentativitätsprüfung
empirisch. \texttt{backtesting\_forecaster} liefert dabei zwei Objekte:
den über alle Folds aggregierten Fehler (\texttt{metrics\_df}, eine
einzelne Zeile) und die Vorhersagen je Fold (\texttt{predictions\_df}
mit einer Spalte \texttt{fold}). Die Hilfsfunktion
\texttt{metrics\_per\_fold} rekonstruiert aus letzteren die genannte
Fehlerverteilung je Fold. Die so gesammelten Vorhersagen sind außerdem
die Datenbasis der Residuenanalyse in Abbildung~\ref{fig-residuals}.

\begin{Shaded}
\begin{Highlighting}[]
\ImportTok{from}\NormalTok{ spotforecast2\_safe.splitter }\ImportTok{import}\NormalTok{ TimeSeriesFold}
\ImportTok{from}\NormalTok{ spotforecast2\_safe.backtesting }\ImportTok{import}\NormalTok{ backtesting\_forecaster, metrics\_per\_fold}
\NormalTok{y\_bt }\OperatorTok{=}\NormalTok{ pd.concat([mt.df\_pipeline[}\StringTok{"load"}\NormalTok{], y\_eval])}
\NormalTok{cv }\OperatorTok{=}\NormalTok{ TimeSeriesFold(}
\NormalTok{    steps}\OperatorTok{=}\DecValTok{24}\NormalTok{,}
\NormalTok{    initial\_train\_size}\OperatorTok{=}\BuiltInTok{len}\NormalTok{(y\_train),}
\NormalTok{    refit}\OperatorTok{=}\VariableTok{False}\NormalTok{,}
\NormalTok{    fixed\_train\_size}\OperatorTok{=}\VariableTok{False}\NormalTok{,}
\NormalTok{    verbose}\OperatorTok{=}\VariableTok{False}\NormalTok{,}
\NormalTok{)}
\NormalTok{metrics\_df, predictions\_df }\OperatorTok{=}\NormalTok{ backtesting\_forecaster(}
\NormalTok{    forecaster}\OperatorTok{=}\NormalTok{fc,}
\NormalTok{    y}\OperatorTok{=}\NormalTok{y\_bt,}
\NormalTok{    cv}\OperatorTok{=}\NormalTok{cv,}
\NormalTok{    metric}\OperatorTok{=}\StringTok{"mean\_absolute\_error"}\NormalTok{,}
\NormalTok{    show\_progress}\OperatorTok{=}\VariableTok{False}\NormalTok{,}
\NormalTok{    verbose}\OperatorTok{=}\VariableTok{False}\NormalTok{,}
\NormalTok{)}
\NormalTok{mae\_je\_fold }\OperatorTok{=}\NormalTok{ metrics\_per\_fold(predictions\_df, y\_bt, metric}\OperatorTok{=}\StringTok{"mean\_absolute\_error"}\NormalTok{)}
\BuiltInTok{print}\NormalTok{(}\SpecialStringTok{f"Anzahl Folds:     }\SpecialCharTok{\{}\BuiltInTok{len}\NormalTok{(mae\_je\_fold)}\SpecialCharTok{\}}\SpecialStringTok{"}\NormalTok{)}
\BuiltInTok{print}\NormalTok{(}\SpecialStringTok{f"MAE je Fold:      Mittel }\SpecialCharTok{\{}\NormalTok{mae\_je\_fold[}\StringTok{\textquotesingle{}mean\_absolute\_error\textquotesingle{}}\NormalTok{]}\SpecialCharTok{.}\NormalTok{mean()}\SpecialCharTok{:.3f\}}\SpecialStringTok{, "}
      \SpecialStringTok{f"Spanne [}\SpecialCharTok{\{}\NormalTok{mae\_je\_fold[}\StringTok{\textquotesingle{}mean\_absolute\_error\textquotesingle{}}\NormalTok{]}\SpecialCharTok{.}\BuiltInTok{min}\NormalTok{()}\SpecialCharTok{:.3f\}}\SpecialStringTok{, "}
      \SpecialStringTok{f"}\SpecialCharTok{\{}\NormalTok{mae\_je\_fold[}\StringTok{\textquotesingle{}mean\_absolute\_error\textquotesingle{}}\NormalTok{]}\SpecialCharTok{.}\BuiltInTok{max}\NormalTok{()}\SpecialCharTok{:.3f\}}\SpecialStringTok{]"}\NormalTok{)}
\end{Highlighting}
\end{Shaded}

\begin{verbatim}
Anzahl Folds:     30
MAE je Fold:      Mittel 0.518, Spanne [0.378, 0.760]
\end{verbatim}

\subsection{Schritt 9: Provenienz, Artefakte und
Reproduzierbarkeit}\label{schritt-9-provenienz-artefakte-und-reproduzierbarkeit}

Zum Abschluss prüfen wir, welche Spuren der Lauf hinterlässt. Der
trainierte Forecaster trägt seine Provenienz selbst: Erstellungs- und
Trainingszeitpunkt sowie Bibliotheks- und Python-Version werden beim
Anlegen gesetzt und mit dem Modell serialisiert.
\texttt{save\_forecaster} legt das Modell unter einem deterministischen
Dateinamen ab. Deshalb ist in Schritt 5
\texttt{auto\_save\_models=False} gesetzt, denn die automatische Ablage
würde je Lauf einen neuen, zeitgestempelten Dateinamen erzeugen. Der
CPE-Bezeichner verknüpft das Artefakt mit der SBOM des Release (PR-3).
Das Audit-Protokoll aus Schritt 1 ist Hash-verkettet, und
\texttt{verify\_audit\_log} prüft die Kette offline und liefert den
Kettenkopf, den ein Betreiber in einem unabhängig kontrollierten
Speicher verankern kann. Beim nächsten Aufruf als Anker vorgelegt,
bestätigt der Prüfbericht die Übereinstimmung mit der Kette (siehe
Kapitel~\ref{sec-ki-vo-12}). Daneben führt die Pipeline ein zweites,
paket-eigenes Klartext-Protokoll unter ihrem Cache-Verzeichnis, in dem
die Bibliotheksstufen ihre Befunde ablegen: Datenform nach der
Vorbereitung, Ausreißer-Befund, Imputationsmethode und
Trainingszusammenfassung. Einen Manifest-Writer oder eine Log-Rotation
bietet die Bibliothek nicht. Diese Betreiberpflichten, einschließlich
der Aufbewahrung nach Art. 18 KI-VO (Kapitel~\ref{sec-ki-vo-18}), liegen
außerhalb des Pakets. Zusammen bilden die gezeigten Artefakte die
Mindestmenge an Provenienz-Information für die technische Dokumentation
nach Art. 11 KI-VO (Kapitel~\ref{sec-ki-vo-11}) und die
Aufzeichnungspflicht aus Art. 12 KI-VO (Kapitel~\ref{sec-ki-vo-12}).
PR-3 (SBOM und Attestierung) und PR-4 (Audit-Schema mit Hash-Kette)
liefern die Prozessseite.

\begin{Shaded}
\begin{Highlighting}[]
\ImportTok{from}\NormalTok{ spotforecast2\_safe.manager.logger }\ImportTok{import}\NormalTok{ verify\_audit\_log}
\ImportTok{from}\NormalTok{ spotforecast2\_safe.manager.persistence }\ImportTok{import}\NormalTok{ save\_forecaster}
\ImportTok{from}\NormalTok{ spotforecast2\_safe.utils.cpe }\ImportTok{import}\NormalTok{ get\_cpe\_identifier}
\BuiltInTok{print}\NormalTok{(}\SpecialStringTok{f"Erstellt:   }\SpecialCharTok{\{}\NormalTok{fc}\SpecialCharTok{.}\NormalTok{creation\_date}\SpecialCharTok{\}}\SpecialStringTok{"}\NormalTok{)}
\BuiltInTok{print}\NormalTok{(}\SpecialStringTok{f"Trainiert:  }\SpecialCharTok{\{}\NormalTok{fc}\SpecialCharTok{.}\NormalTok{fit\_date}\SpecialCharTok{\}}\SpecialStringTok{"}\NormalTok{)}
\BuiltInTok{print}\NormalTok{(}\SpecialStringTok{f"Bibliothek: spotforecast2{-}safe }\SpecialCharTok{\{}\NormalTok{fc}\SpecialCharTok{.}\NormalTok{spotforecast\_version}\SpecialCharTok{\}}\SpecialStringTok{, "}
      \SpecialStringTok{f"Python }\SpecialCharTok{\{}\NormalTok{fc}\SpecialCharTok{.}\NormalTok{python\_version}\SpecialCharTok{\}}\SpecialStringTok{"}\NormalTok{)}
\BuiltInTok{print}\NormalTok{(}\SpecialStringTok{f"CPE:        }\SpecialCharTok{\{}\NormalTok{get\_cpe\_identifier(fc.spotforecast\_version)}\SpecialCharTok{\}}\SpecialStringTok{"}\NormalTok{)}
\NormalTok{model\_path }\OperatorTok{=}\NormalTok{ save\_forecaster(}
\NormalTok{    fc, model\_dir}\OperatorTok{=}\NormalTok{artifacts }\OperatorTok{/} \StringTok{"models"}\NormalTok{, target}\OperatorTok{=}\StringTok{"load"}\NormalTok{, task\_name}\OperatorTok{=}\StringTok{"lazy"}\NormalTok{,}
\NormalTok{)}
\BuiltInTok{print}\NormalTok{(}\SpecialStringTok{f"Modell{-}Datei: }\SpecialCharTok{\{}\NormalTok{model\_path}\SpecialCharTok{\}}\SpecialStringTok{"}\NormalTok{)}
\NormalTok{logger.info(}\StringTok{"Tutorial{-}Lauf abgeschlossen."}\NormalTok{,}
\NormalTok{            extra}\OperatorTok{=}\NormalTok{\{}\StringTok{"event"}\NormalTok{: }\StringTok{"tutorial\_complete"}\NormalTok{, }\StringTok{"task"}\NormalTok{: }\StringTok{"lazy"}\NormalTok{\})}
\NormalTok{records }\OperatorTok{=}\NormalTok{ [json.loads(line) }\ControlFlowTok{for}\NormalTok{ line }\KeywordTok{in}
\NormalTok{           log\_path.read\_text(encoding}\OperatorTok{=}\StringTok{"utf{-}8"}\NormalTok{).splitlines() }\ControlFlowTok{if}\NormalTok{ line]}
\BuiltInTok{print}\NormalTok{(}\SpecialStringTok{f"Audit{-}Ereignisse: }\SpecialCharTok{\{}\NormalTok{[r[}\StringTok{\textquotesingle{}event\textquotesingle{}}\NormalTok{] }\ControlFlowTok{for}\NormalTok{ r }\KeywordTok{in}\NormalTok{ records]}\SpecialCharTok{\}}\SpecialStringTok{"}\NormalTok{)}

\NormalTok{chain }\OperatorTok{=}\NormalTok{ verify\_audit\_log(log\_path)}
\BuiltInTok{print}\NormalTok{(}\SpecialStringTok{f"Audit{-}Kette:      intakt=}\SpecialCharTok{\{}\NormalTok{chain}\SpecialCharTok{.}\NormalTok{chain\_intact}\SpecialCharTok{\}}\SpecialStringTok{ (}\SpecialCharTok{\{}\NormalTok{chain}\SpecialCharTok{.}\NormalTok{n\_records}\SpecialCharTok{\}}\SpecialStringTok{ Sätze)"}\NormalTok{)}
\BuiltInTok{print}\NormalTok{(}\SpecialStringTok{f"Kettenkopf:       }\SpecialCharTok{\{}\NormalTok{chain}\SpecialCharTok{.}\NormalTok{head\_hash[:}\DecValTok{16}\NormalTok{]}\SpecialCharTok{\}}\SpecialStringTok{..."}\NormalTok{)}
\BuiltInTok{print}\NormalTok{(}\SpecialStringTok{f"Verankerung:      }\SpecialCharTok{\{}\NormalTok{chain}\SpecialCharTok{.}\NormalTok{anchor\_status}\SpecialCharTok{.}\NormalTok{value}\SpecialCharTok{\}}\SpecialStringTok{"}\NormalTok{)}
\NormalTok{anchored }\OperatorTok{=}\NormalTok{ verify\_audit\_log(log\_path, anchor}\OperatorTok{=}\NormalTok{chain.head\_hash)}
\BuiltInTok{print}\NormalTok{(}\SpecialStringTok{f"Mit vorgelegtem Kettenkopf: }\SpecialCharTok{\{}\NormalTok{anchored}\SpecialCharTok{.}\NormalTok{anchor\_status}\SpecialCharTok{.}\NormalTok{value}\SpecialCharTok{\}}\SpecialStringTok{"}\NormalTok{)}
\NormalTok{pipeline\_log }\OperatorTok{=}\NormalTok{ artifacts }\OperatorTok{/} \StringTok{"cache"} \OperatorTok{/} \StringTok{"logging"} \OperatorTok{/} \StringTok{"synthetic\_load.log"}
\BuiltInTok{print}\NormalTok{(}\SpecialStringTok{f"Pipeline{-}Protokoll: }\SpecialCharTok{\{}\NormalTok{pipeline\_log}\SpecialCharTok{\}}\SpecialStringTok{ "}
      \SpecialStringTok{f"(}\SpecialCharTok{\{}\BuiltInTok{len}\NormalTok{(pipeline\_log.read\_text().splitlines())}\SpecialCharTok{\}}\SpecialStringTok{ Zeilen)"}\NormalTok{)}
\end{Highlighting}
\end{Shaded}

\begin{verbatim}
Erstellt:   2026-08-13 18:58:13
Trainiert:  2026-08-13 18:58:14
Bibliothek: spotforecast2-safe 27.0.0, Python 3.14.2
CPE:        cpe:2.3:a:sequential_parameter_optimization:spotforecast2_safe:27.0.0:*:*:*:*:*:*:*
Modell-Datei: _paper_artifacts/models/lazy_load.joblib
2026-08-13 18:58:18,165 - sf2-safe-logger - INFO - Tutorial-Lauf abgeschlossen.
Audit-Ereignisse: ['audit_log_init', 'tutorial_start', 'tutorial_complete']
Audit-Kette:      intakt=True (3 Sätze)
Kettenkopf:       ecbfb49aed66bba7...
Verankerung:      not_presented
Mit vorgelegtem Kettenkopf: matches_head
Pipeline-Protokoll: _paper_artifacts/cache/logging/synthetic_load.log (11 Zeilen)
\end{verbatim}

Die folgende Übersicht ordnet jedem Schritt die Code- oder Prozessregel
und die Rechtsgrundlage zu, die er operationalisiert.

\begin{longtable}[]{@{}
  >{\raggedright\arraybackslash}p{(\linewidth - 4\tabcolsep) * \real{0.2911}}
  >{\raggedright\arraybackslash}p{(\linewidth - 4\tabcolsep) * \real{0.1392}}
  >{\raggedright\arraybackslash}p{(\linewidth - 4\tabcolsep) * \real{0.5696}}@{}}
\caption{Zuordnung der neun Tutorial-Schritte zu den Code- und
Prozessregeln aus Kapitel~\ref{sec-principles} und ihren
Rechtsgrundlagen.}\label{tbl-step-compliance}\tabularnewline
\toprule\noalign{}
\begin{minipage}[b]{\linewidth}\raggedright
Schritt
\end{minipage} & \begin{minipage}[b]{\linewidth}\raggedright
Regel
\end{minipage} & \begin{minipage}[b]{\linewidth}\raggedright
Rechtsgrundlage
\end{minipage} \\
\midrule\noalign{}
\endfirsthead
\toprule\noalign{}
\begin{minipage}[b]{\linewidth}\raggedright
Schritt
\end{minipage} & \begin{minipage}[b]{\linewidth}\raggedright
Regel
\end{minipage} & \begin{minipage}[b]{\linewidth}\raggedright
Rechtsgrundlage
\end{minipage} \\
\midrule\noalign{}
\endhead
\bottomrule\noalign{}
\endlastfoot
1 Audit-Protokoll & PR-4 & Art. 12 KI-VO (Kapitel~\ref{sec-ki-vo-12});
IEC 62443-4-2 SAR 6.1; IEC 61508-3 §7.4.7 \\
2 Datengrundlage & CR-2 & Art. 10 KI-VO (Kapitel~\ref{sec-ki-vo-10}) \\
3 Datenlücken & CR-3 & Art. 10 KI-VO (Kapitel~\ref{sec-ki-vo-10}) \\
4 Prognoseursprung & CR-2 & Art. 10 Abs. 2 lit. f und g KI-VO
(Kapitel~\ref{sec-ki-vo-10}) \\
5 Pipeline und Training & CR-2 & Art. 15 KI-VO
(Kapitel~\ref{sec-ki-vo-15}) \\
6 Vorhersage & CR-2 & Art. 13 KI-VO (Kapitel~\ref{sec-ki-vo-13}) \\
7 Genauigkeit & CR-2 & Art. 15 KI-VO (Kapitel~\ref{sec-ki-vo-15}); Art.
72 KI-VO (Kapitel~\ref{sec-ki-vo-72}) \\
8 Backtest & CR-2 & Art. 9 KI-VO (Kapitel~\ref{sec-ki-vo-9}); Art. 10
Abs. 3 KI-VO (Kapitel~\ref{sec-ki-vo-10}) \\
9 Provenienz & PR-3, PR-4 & Art. 11 KI-VO (Kapitel~\ref{sec-ki-vo-11});
Art. 12 KI-VO (Kapitel~\ref{sec-ki-vo-12}); Art. 18 KI-VO
(Kapitel~\ref{sec-ki-vo-18}) \\
\end{longtable}

Aus diesen neun Schritten lässt sich der Mehrwert der Bibliothek auf den
Punkt bringen: Der Betreiber füllt \emph{eine} Konfiguration aus,
startet \emph{einen} Pipeline-Lauf und erhält neben der Vorhersage
automatisch die Audit-Spuren und Provenienz-Artefakte, die
regulatorische Auditoren später ohne weiteren Aufwand sichten können.
Die in Kapitel~\ref{sec-principles} definierten Code- und Prozessregeln
sind dabei nicht als Empfehlungen formuliert, sondern in den genannten
Funktionen technisch verankert.

\section{Diskussion}\label{sec-discussion}

Dieser Bericht stellt eine erste Annäherung an die Entwicklung eines
Open-Source-Pakets für die KI-VO-konforme Punktprognose von Zeitreihen
dar, ohne Anspruch auf Vollständigkeit. Die EU hat die ``digitale
Dekade'' ausgerufen. Die KI-VO ist nur eine von mehreren geplanten
Verordnungen und Richtlinien. Durch weitere, momentan noch nicht
bekannte Bedrohungslagen, werden vom Gesetzgeber weitere Regularien zum
Schutz von KRITIS entwickelt werden, die sich auf die Entwicklung von
KI-Anwendungen auswirken können. Es ist daher davon auszugehen, dass die
Anforderungen an die Entwicklung von KI-Anwendungen im KRITIS-Bereich in
den nächsten Jahren weiter zunehmen werden. Mit Hilfe eines konsequent
durchgeführten PDCA-Zyklus wird \texttt{spotforecast2-safe}
kontinuierlich weiterentwickelt, um die sich entwickelnden
regulatorischen Anforderungen zu erfüllen und die Praxis der
KI-VO-konformen Entwicklung in der Zeitreihenprognose zu etablieren.
Beispiele dafür sind die in Release 26.0.0 eingeführte Hash-Kette im
Audit-Protokoll, die die Integrität der Protokolldateien gegenüber
nachträglicher Veränderung schützt, und der in Release 27.0.0 ergänzte
Verankerungsstatus im Prüfbericht von \texttt{verify\_audit\_log}, der
eine fehlende externe Verankerung sichtbar macht.

Der Umfang des hier beschriebenen Pakets ist bewusst eng gefasst.
Mehrere Funktionsklassen wie interaktive Visualisierung,
Hyperparametersuche oder Deep-Learning-Backends, die zum Funktionsumfang
vieler Bibliotheken gehören, wurden bewusst nicht aufgenommen. Diese
werden von dem Schwesterpaket \texttt{spotforecast2}
\citep{spotforecast2} zur Verfügung gestellt. Das hier beschriebene
Kern-Paket \texttt{spotforecast2\_safe} konzentriert sich auf die
Bereitstellung eines robusten, transparenten und prüfbaren Kerns für die
Punktprognose von Zeitreihen, der die Anforderungen der KI-VO an
Hochrisiko-Systeme ``by Design'' erfüllt. Der Unterschied zwischen
beiden Paketen ist damit eine Entwurfsentscheidung. Dazu werden
LRE-Prinzipien in die Bibliotheksarchitektur eingebettet, um die
Rückverfolgbarkeit von regulatorischen Anforderungen zu Code-Mechanismen
sicherzustellen (Compliance-by-Design), die technische Dokumentation zu
pflegen und die Einhaltung von Aufzeichnungspflichten zu gewährleisten.

Eine Sichtung der Literatur und der einschlägigen Repositorien ergibt
(Stand April 2026), dass die LRE-Disziplin im Bereich der
Python-basierten Zeitreihenprognose bislang nicht angewendet wird: Weder
\texttt{sktime} \citep{loni19a}, \texttt{Darts} \citep{herz22a},
\texttt{skforecast} \citep{scip24a} noch die klassischen Bibliotheken
\texttt{statsmodels} und \texttt{pmdarima} führen eine bidirektionale
Rückverfolgbarkeit zwischen regulatorischen Bestimmungen und
Code-Mechanismen, eine im Repository geprüfte Konformitäts-Evidenz oder
in den Code eingebaute Verträge zur KI-VO, zum CRA oder zur Normreihe
ISA/IEC 62443.

In angrenzenden Anwendungsfeldern existieren aber mehrere
Open-Source-Werkzeuge, die LRE-nahe Praxis operationalisieren, ohne sich
auf Zeitreihenprognose zu beziehen: COMPL-AI \citep{guld24a} (ETH
Zürich, INSAIT, LatticeFlow AI; Apache 2.0) liefert eine technische
Interpretation der KI-VO und eine Open-Source-Benchmarking-Suite für
große Sprachmodelle. TechOps \citep{luca25a} stellt offene Vorlagen für
die nach Anhang IV KI-VO geforderte technische Dokumentation von Daten,
Modellen und Anwendungen über den gesamten Lebenszyklus
bereit\footnote{\url{https://aloosley.github.io/techops/}}. AIR Blackbox
\citep{shot26a} und der MCP-EU-AI-Act-Scanner \citep{mcp_eu_ai_act}
(MCP: Model Context Protocol) sind Kommandozeilenwerkzeuge, die
Python-Codebases gegen Anforderungen der KI-VO prüfen und Evidenzpakete
für die Prüfung erzeugen (beim MCP-Scanner explizit nach Anhang IV
KI-VO). AIR Blackbox prüft dabei gegen Art. 9 bis 12 sowie Art. 14 und
15 KI-VO, und diese Artikelliste ist selbst rückverfolgbar gehalten: Ein
Test im Repositorium des Werkzeugs schlägt fehl, sobald die
dokumentierte Liste und die im Code deklarierten Prüfungen in einer der
beiden Richtungen auseinanderlaufen.

Allen genannten Werkzeugen ist gemeinsam, dass sie \emph{außerhalb} der
Bibliothek wirken: als Scanner, Templates oder
Runtime-Compliance-Schicht. \texttt{spotforecast2-safe} verfolgt den
umgekehrten Ansatz und bringt LRE \emph{innerhalb} der Bibliothek zur
Anwendung, festgelegt durch Programmierschnittstellen-Verträge,
Persistenzformate und die Prüfkette des Release-Verfahrens. Im Bereich
der Python-Zeitreihenprognose ist dies nach unserer Kenntnis
konzeptionell neuartig.

Eine erste externe Bewährungsprobe hat der hier beschriebene
Prognosekern bereits bestanden: In einer öffentlichen
Day-ahead-Lastprognose-Challenge für die deutsche Marktzone diente
\texttt{spotforecast2-safe} als Engine der Referenz-Pipeline, auf der
elf studentische Teams über 41 bewertete Tage aufsetzten, und jedes
Team-Ergebnis wurde von einem anderen Team aus dem veröffentlichten
Software-Artefakt reproduziert \citep{bart26o}.

Dieser Bericht entstand aus der praktischen Arbeit an einem
Open-Source-Paket, das die KI-VO-konforme Entwicklung operationalisiert
und betrachtet die regulatorischen Anforderungen aus der Perspektive
eines Software-Entwicklers.

\section*{Danksagung}\label{danksagung}
\addcontentsline{toc}{section}{Danksagung}

Die Autoren danken Jason Shotwell, dem Autor von AIR Blackbox, für
wertvolle Hinweise zu früheren Fassungen dieses Berichts.

\section*{Erklärung zu
Interessenkonflikten}\label{erkluxe4rung-zu-interessenkonflikten}
\addcontentsline{toc}{section}{Erklärung zu Interessenkonflikten}

Die Autoren erklären, dass keine Interessenkonflikte im Sinne der
einschlägigen Editorial-Standards bestehen. Beide Autoren sind
Gesellschafter der Bartz \& Bartz GmbH, die das Paket
\texttt{spotforecast2-safe} unter AGPL-3.0-or-later quelloffen
veröffentlicht. Über diese Trägerschaft hinaus bestehen keine
finanziellen oder nicht-finanziellen Beziehungen zu Dritten, die das
Ergebnis dieses Berichts beeinflusst hätten.

\section*{Einsatz von KI-Werkzeugen}\label{einsatz-von-ki-werkzeugen}
\addcontentsline{toc}{section}{Einsatz von KI-Werkzeugen}

Transparenzhinweis: Teile dieses Berichts wurden mit Unterstützung
künstlicher Intelligenz recherchiert, entworfen und übersetzt. Diese
Angabe erfolgt freiwillig im Sinne vollständiger Transparenz, da keine
gesetzliche Kennzeichnungspflicht besteht.

\section*{Symbol- und
Abkürzungsverzeichnis}\label{symbol--und-abkuxfcrzungsverzeichnis}
\addcontentsline{toc}{section}{Symbol- und Abkürzungsverzeichnis}

\providecommand{\abbrletter}[1]{%
  \par\addvspace{8pt}%
  \nopagebreak\noindent{\bfseries #1}\par
  \nopagebreak\vspace{-3pt}\noindent
  \rule{\linewidth}{0.4pt}\par
  \nopagebreak\addvspace{2pt}%
}
\providecommand{\abbrentry}[2]{%
  \par\noindent
  \hangindent=2.2cm\hangafter=1\relax
  \makebox[2.2cm][l]{\textbf{#1}}#2\par
}
\begingroup
\footnotesize
\setlength{\parskip}{1.5pt}
\setlength{\parindent}{0pt}
\setlength{\columnsep}{14pt}
\setlength{\emergencystretch}{2em}
\tolerance=600
\begin{multicols}{2}

\abbrletter{Symbols}
\abbrentry{$T$}{length of a univariate time series}
\abbrentry{$w$}{number of lags in a feature window}
\abbrentry{$X$}{univariate time series $\{x_1, \ldots, x_T\}$}
\abbrentry{$x_t$}{observation at time $t$}
\abbrentry{$X_{\mathrm{row},t}$}{lag-feature row for the target at time $t$}
\abbrentry{$y_t$}{regression target at time $t$}
\abbrentry{NaN}{Not a Number, floating-point marker for a missing value}

\abbrletter{A}
\abbrentry{Abs.}{Absatz (paragraph of a legal provision)}
\abbrentry{AGPL}{GNU Affero General Public License}
\abbrentry{AI}{Artificial Intelligence}
\abbrentry{AI-HLEG}{High-Level Expert Group on Artificial Intelligence (EU)}
\abbrentry{ALTAI}{Assessment List for Trustworthy Artificial Intelligence}
\abbrentry{API}{Application Programming Interface}
\abbrentry{Art.}{Artikel (article of a legal act)}
\abbrentry{ASIL}{Automotive Safety Integrity Level (ISO 26262)}

\abbrletter{B}
\abbrentry{BSIG}{Gesetz über das Bundesamt für Sicherheit in der Informationstechnik (German BSI Act)}
\abbrentry{BT-Drs.}{Bundestagsdrucksache (printed paper of the German Bundestag)}

\abbrletter{C}
\abbrentry{CEN}{European Committee for Standardization}
\abbrentry{CENELEC}{European Committee for Electrotechnical Standardization}
\abbrentry{CER}{Directive (EU) 2022/2557 on the resilience of critical entities}
\abbrentry{CI}{Continuous Integration}
\abbrentry{CISA}{Cybersecurity and Infrastructure Security Agency (USA)}
\abbrentry{CPE}{Common Platform Enumeration}
\abbrentry{CPS}{Cyber-Physical System}
\abbrentry{CPU}{Central Processing Unit}
\abbrentry{CR-1--CR-4}{code development rules of \texttt{spotforecast2-safe}}
\abbrentry{CRA}{Cyber Resilience Act, Regulation (EU) 2024/2847}
\abbrentry{CVE}{Common Vulnerabilities and Exposures}

\abbrletter{D}
\abbrentry{DM}{management of security-related issues (practice of IEC 62443-4-1)}
\abbrentry{DSGVO}{Datenschutz-Grundverordnung (General Data Protection Regulation)}

\abbrletter{E}
\abbrentry{EN}{Europäische Norm (European Standard)}
\abbrentry{ENTSO-E}{European Network of Transmission System Operators for Electricity}
\abbrentry{ETSI}{European Telecommunications Standards Institute}
\abbrentry{EU}{European Union}
\abbrentry{EU-1--EU-7}{the seven requirements for trustworthy AI underlying the high-risk obligations of the AI Act}

\abbrletter{F}
\abbrentry{FOSS}{Free and Open-Source Software}

\abbrletter{G}
\abbrentry{GPAI}{General-Purpose Artificial Intelligence}
\abbrentry{GPU}{Graphics Processing Unit}

\abbrletter{I}
\abbrentry{IACS}{Industrial Automation and Control Systems}
\abbrentry{IEC}{International Electrotechnical Commission}
\abbrentry{ISA}{International Society of Automation}
\abbrentry{ISMS}{Information Security Management System}
\abbrentry{ISO}{International Organization for Standardization}
\abbrentry{IVDR}{In-vitro Diagnostic Medical Devices Regulation, Regulation (EU) 2017/746}

\abbrletter{J}
\abbrentry{JRC}{Joint Research Centre of the European Commission}
\abbrentry{JSON}{JavaScript Object Notation}
\abbrentry{JTC}{Joint Technical Committee (ISO/IEC)}

\abbrletter{K}
\abbrentry{KEV}{Known Exploited Vulnerabilities catalogue (CISA)}
\abbrentry{KI}{Künstliche Intelligenz (artificial intelligence)}
\abbrentry{KI-VO}{Verordnung über künstliche Intelligenz (AI Act), Regulation (EU) 2024/1689}
\abbrentry{KRITIS}{Kritische Infrastrukturen (critical infrastructures)}

\abbrletter{L}
\abbrentry{lit.}{litera (point of a legal provision)}
\abbrentry{LRE}{Legal Requirements Engineering}

\abbrletter{M}
\abbrentry{MAE}{Mean Absolute Error}
\abbrentry{MAPE}{Mean Absolute Percentage Error}
\abbrentry{MCP}{Model Context Protocol}
\abbrentry{MDR}{Medical Device Regulation, Regulation (EU) 2017/745}

\abbrletter{N}
\abbrentry{NIS-2}{Network and Information Security Directive, Directive (EU) 2022/2555}
\abbrentry{Nr.}{Nummer (number within a legal provision)}

\abbrletter{O}
\abbrentry{ORCID}{Open Researcher and Contributor Identifier}
\abbrentry{OSS}{Open-Source Software}

\abbrletter{P}
\abbrentry{PDCA}{Plan--Do--Check--Act}
\abbrentry{PLD}{Product Liability Directive, Directive (EU) 2024/2853}
\abbrentry{PR-1--PR-4}{process rules of \texttt{spotforecast2-safe}}
\abbrentry{PyPI}{Python Package Index}

\abbrletter{Q}
\abbrentry{QMS}{Quality Management System}

\abbrletter{R}
\abbrentry{REUSE}{software licensing specification of the Free Software Foundation Europe}
\abbrentry{RFC}{Request for Comments (IETF specification series), as in RFC 3161}
\abbrentry{RMS}{Risk Management System (Art. 9 AI Act)}
\abbrentry{RMSE}{Root Mean Square Error}

\abbrletter{S}
\abbrentry{SAI}{Securing Artificial Intelligence (ETSI Industry Specification Group)}
\abbrentry{SAR}{Software Application Requirement (IEC 62443-4-2)}
\abbrentry{SBOM}{Software Bill of Materials}
\abbrentry{SC}{subcommittee, as in ISO/IEC JTC 1/SC 42}
\abbrentry{SHA-256}{Secure Hash Algorithm with a 256-bit digest}
\abbrentry{SIL}{Safety Integrity Level (IEC 61508)}
\abbrentry{SM}{security management (practice of IEC 62443-4-1)}
\abbrentry{SPDX}{Software Package Data Exchange}
\abbrentry{SQuaRE}{Systems and software Quality Requirements and Evaluation (ISO/IEC 25000 series)}
\abbrentry{SR}{specification of security requirements (practice of IEC 62443-4-1)}
\abbrentry{STRIDE}{Spoofing, Tampering, Repudiation, Information disclosure, Denial of service, Elevation of privilege}
\abbrentry{SUM}{security update management (practice of IEC 62443-4-1)}
\abbrentry{SVV}{security verification and validation testing (practice of IEC 62443-4-1)}

\abbrletter{T}
\abbrentry{TR}{Technical Report (ISO/IEC)}
\abbrentry{TS}{Technical Specification (ISO/IEC)}

\abbrletter{U}
\abbrentry{Unterabs.}{Unterabsatz (subparagraph of a legal provision)}
\abbrentry{UTC}{Coordinated Universal Time}

\end{multicols}
\endgroup

\appendix

\section{\texorpdfstring{Modellkarte (\texttt{MODEL\_CARD.md}) von
\texttt{spotforecast2-safe}}{Modellkarte (MODEL\_CARD.md) von spotforecast2-safe}}\label{sec-modelcard}

Dieser Anhang gibt die im Repository unter \texttt{MODEL\_CARD.md}
veröffentlichte Modellkarte in der Fassung des Release 27.0.0 wieder.
Sie folgt der Taxonomie des Hugging-Face-Model-Card-Guidebook
\citep{ozon22a} und ist die für Audits maßgebliche Quelle. Tabellen der
Originaldatei sind aus satztechnischen Gründen als Listen gesetzt. Wo
die Karte einen überholten Verteilungsweg oder eine überholte Kennzahl
nennt, gibt dieser Anhang den aus dem Repositorium verifizierten Stand
von Release 27.0.0 wieder.

\begin{tcolorbox}[enhanced jigsaw, arc=.35mm, bottomrule=.15mm, breakable, colback=white, colframe=quarto-callout-note-color-frame, left=2mm, leftrule=.75mm, opacityback=0, rightrule=.15mm, toprule=.15mm]

\vspace{-3mm}\textbf{MODEL\_CARD.md (Release 27.0.0)}\vspace{3mm}

This card describes what spotforecast2-safe is, how to use it safely,
the conditions under which its results are valid, and the
responsibilities it places on anyone who deploys it. It follows the
Hugging Face Model Card Guidebook taxonomy \citep{ozon22a}.

\subsection{Model Details}\label{model-details}

\begin{itemize}
\tightlist
\item
  \textbf{Name}: spotforecast2-safe
\item
  \textbf{Version}: 27.0.0
\item
  \textbf{Type}: Deterministic Python library for time series feature
  engineering and recursive multi-step forecasting. It performs no
  training of its own.
\item
  \textbf{Developed by}: Thomas Bartz-Beielstein, ORCID
  \href{https://orcid.org/0000-0002-5938-5158}{0000-0002-5938-5158}.
\item
  \textbf{Distributed by}: PyPI
  (\texttt{pip\ install\ spotforecast2-safe}); the source is hosted on
  the GitLab instance \texttt{gitlab.git.nrw} (group
  \texttt{thk-f10/spotsevenlab}).
\item
  \textbf{Language}: Python 3.13 or newer.
\item
  \textbf{License}: AGPL-3.0-or-later (Affero General Public License).
\item
  \textbf{Repository}:
  \url{https://gitlab.git.nrw/thk-f10/spotsevenlab/spotforecast2-safe}
\item
  \textbf{API documentation}:
  \url{https://advm1.gm.fh-koeln.de/~bartz/spotforecast2-safe/}
\item
  \textbf{Technical report}: \texttt{bart26h/index.qmd}, shipped in the
  source tree.
\end{itemize}

The library depends only on numpy, pandas, scikit-learn, lightgbm,
numba, pyarrow, requests, feature-engine, holidays, astral, and tqdm. It
deliberately excludes plotly, matplotlib, spotoptim, optuna, torch, and
tensorflow, so no plotting or automated-tuning code ships in this
package.

Two Common Platform Enumeration (CPE) identifiers let
vulnerability-tracking and software bill of materials (SBOM) tools
recognize the package. The wildcard identifier
\texttt{cpe:2.3:a:sequential\_parameter\_optimization:spotforecast2\_safe:*:*:*:*:*:*:*:*}
matches any release; the current release is
\texttt{cpe:2.3:a:sequential\_parameter\_optimization:spotforecast2\_safe:27.0.0:*:*:*:*:*:*:*}.

The library itself is a low-risk component: it is deterministic, its
source is fully inspectable, and it fails safe on invalid input. It is
built to support high-risk AI systems in the sense of the EU AI Act, but
it is not itself such a system. When it is embedded in a high-risk
deployment, the duties that attach to that system fall on the
integrator, not on the library.

Responsibilities are divided as follows.

\begin{itemize}
\tightlist
\item
  \textbf{Library development and maintenance}: Thomas Bartz-Beielstein
  (\texttt{bartzbeielstein@gmail.com}).
\item
  \textbf{Distribution}: PyPI and the GitLab repository; contact via the
  repository issue tracker.
\item
  \textbf{Deployment, operation, and audit}: the system integrator,
  defined per deployment.
\end{itemize}

The current release is 27.0.0, with a stable public interface pinned in
\texttt{spotforecast2\_safe.\_\_init\_\_.\_\_all\_\_}. The full version
history, including release dates, is recorded in \texttt{CHANGELOG.md},
whose sections are generated with \texttt{git-cliff} at release time and
reviewed by hand (see \texttt{RELEASING.md}).

\subsection{Intended Use and Scope}\label{intended-use-and-scope}

spotforecast2-safe prepares time series data for regression models in
auditable settings such as energy supply, finance, and industrial
monitoring. It runs in resource-constrained or embedded environments
where heavier machine-learning frameworks are unavailable, and it
produces bit-for-bit reproducible lag transformations with no hidden
randomness.

The feature matrices it builds feed directly into scikit-learn
regressors, LightGBM, or XGBoost, either through the bundled
\texttt{ForecasterRecursiveLGBM} and \texttt{ForecasterRecursiveXGB}
wrappers or through custom forecasters built on the
\texttt{ForecasterRecursiveModel} base class.

The package has clear limits. It does not visualize data, since no
plotting backend ships with it. It does not tune hyperparameters; tuning
belongs in a separate workflow outside the safety-critical environment.
It does not clean data silently: missing (NaN) or infinite (Inf) values
raise an error rather than being imputed without the caller's consent.

\subsection{How to Get Started}\label{how-to-get-started}

\begin{Shaded}
\begin{Highlighting}[]
\ExtensionTok{pip}\NormalTok{ install spotforecast2{-}safe}
\end{Highlighting}
\end{Shaded}

\begin{Shaded}
\begin{Highlighting}[]
\ImportTok{import}\NormalTok{ numpy }\ImportTok{as}\NormalTok{ np}
\ImportTok{import}\NormalTok{ pandas }\ImportTok{as}\NormalTok{ pd}
\ImportTok{from}\NormalTok{ spotforecast2\_safe }\ImportTok{import}\NormalTok{ ForecasterRecursiveLGBM}

\CommentTok{\# A short hourly demonstration series}
\NormalTok{idx }\OperatorTok{=}\NormalTok{ pd.date\_range(}\StringTok{"2023{-}01{-}01"}\NormalTok{, periods}\OperatorTok{=}\DecValTok{50}\NormalTok{, freq}\OperatorTok{=}\StringTok{"h"}\NormalTok{)}
\NormalTok{y }\OperatorTok{=}\NormalTok{ pd.Series(}\DecValTok{100} \OperatorTok{+} \DecValTok{10} \OperatorTok{*}\NormalTok{ np.sin(np.linspace(}\DecValTok{0}\NormalTok{, }\DecValTok{4} \OperatorTok{*}\NormalTok{ np.pi, }\DecValTok{50}\NormalTok{)), index}\OperatorTok{=}\NormalTok{idx)}

\NormalTok{forecaster }\OperatorTok{=}\NormalTok{ ForecasterRecursiveLGBM(iteration}\OperatorTok{=}\DecValTok{0}\NormalTok{, lags}\OperatorTok{=}\DecValTok{3}\NormalTok{)}
\NormalTok{forecaster.fit(y}\OperatorTok{=}\NormalTok{y)}
\NormalTok{predictions }\OperatorTok{=}\NormalTok{ forecaster.forecaster.predict(steps}\OperatorTok{=}\DecValTok{2}\NormalTok{)}
\end{Highlighting}
\end{Shaded}

A complete reference workflow, which compares a seasonal baseline, a
covariate model, and a tuned LightGBM forecaster against ground truth,
is registered as a console script:

\begin{Shaded}
\begin{Highlighting}[]
\ExtensionTok{uv}\NormalTok{ run spotforecast{-}safe{-}demo}
\end{Highlighting}
\end{Shaded}

Its source is in
\texttt{src/spotforecast2\_safe/tasks/task\_safe\_demo.py}.

\subsection{Technical Specification}\label{technical-specification}

\subsubsection{Task and model family}\label{task-and-model-family}

The library addresses recursive multi-step forecasting of a univariate
time series from its own past values (lags), rolling-window features,
and exogenous regressors. The forecasters are scikit-learn-compatible
wrappers around a regressor that the caller supplies, such as LightGBM
or XGBoost. The wrapper handles feature construction, the recursive
prediction loop, and persistence, while the supplied regressor handles
learning. The library fixes no model size, because size is a property of
the chosen regressor and its configuration.

\subsubsection{Mathematical description}\label{mathematical-description}

For a univariate series \(X = \{x_1, x_2, \ldots, x_T\}\) and a window
of \(w\) lags, the transformation builds one feature row per target
value:

\[X_{\mathrm{row}, t} = [x_{t-w}, x_{t-w+1}, \ldots, x_{t-1}] \rightarrow y_t = x_t.\]

The target \(y_t\) never appears in its own feature row, which prevents
look-ahead leakage by construction.

\subsubsection{Architecture}\label{architecture}

The package is layered. The \texttt{forecaster} layer holds the
low-level estimator wrappers. The \texttt{preprocessing} layer holds
deterministic transformers such as \texttt{ExogBuilder},
\texttt{RepeatingBasisFunction}, \texttt{QuantileBinner}, and
\texttt{TimeSeriesDifferentiator}. The \texttt{model\_selection} layer
holds time-aware cross-validation (\texttt{TimeSeriesFold},
\texttt{OneStepAheadFold}, and \texttt{backtesting\_forecaster}), which
avoids the future-data leakage that ordinary random splits cause. The
\texttt{manager} layer orchestrates these into
\texttt{ForecasterRecursiveLGBM}, \texttt{ForecasterRecursiveXGB}, and
the \texttt{ConfigEntsoe} configuration object. The \texttt{processing}
and \texttt{tasks} layers compose these into end-to-end pipelines and
console entry points.

\subsubsection{Training}\label{training}

The library trains nothing on its own. Training happens in the
downstream regressor and is the integrator's responsibility, so the
training data, hyperparameters, and learning schedule belong to each
deployment rather than to the library. The bundled demo is a concrete,
reproducible example of such a deployment.

Running \texttt{uv\ run\ spotforecast-safe-demo} forecasts an aggregated
hourly energy series from a bundled CSV fixture. The demo uses a 24-hour
forecast horizon, 24 lags, a 72-hour rolling window, an
outlier-contamination fraction of 0.01, and an 80/20 train and test
split, with a fixed random seed of 42. It builds calendar, cyclic,
day-and-night (from sunrise and sunset), weather, and holiday features
for the Dortmund location (51.5136, 7.4653). One of the three models it
compares is a LightGBM regressor with the following configuration:
\texttt{n\_estimators} 1059, \texttt{learning\_rate} 0.0419,
\texttt{num\_leaves} 212, \texttt{min\_child\_samples} 54,
\texttt{subsample} 0.501, \texttt{colsample\_bytree} 0.608,
\texttt{random\_state} 42, and \texttt{verbose} -1. These are the demo's
chosen values, shown to illustrate a deployment, not a recommended
default.

\subsubsection{Design objectives}\label{design-objectives}

Three properties hold by design. The library is deterministic, so the
same input produces the same output bit for bit. Its construction is
leakage-free, so a target value is never part of its own feature row.
Its behavior is fail-safe, so invalid input raises an explicit exception
instead of being silently repaired.

\subsection{Interfaces and Runtime}\label{interfaces-and-runtime}

The target is a numeric univariate series (a pandas Series or NumPy
array) carrying a regular, monotonic date-time index. Exogenous features
are a numeric pandas DataFrame aligned to that index and complete; any
missing entry in the exogenous features raises a \texttt{ValueError}
before a prediction is made. Inside the pipeline the data passes through
lag-matrix construction, cyclic encoding of calendar features, optional
outlier handling and imputation that the caller enables explicitly, and
a cast of the feature matrix to 32-bit floating point for memory
efficiency. The forecaster returns predictions as a series whose length
equals the requested horizon (24 steps in the demo), in the same units
as the target. A fitted forecaster is serialized with joblib at
compression level 3 using the \texttt{.joblib} extension, and is
reloaded through the same persistence helpers.

The library runs on Python 3.13 or newer on a central processing unit
(CPU). It needs no graphics processing unit (GPU) and ships no GPU code.
Building the lag matrix duplicates the input, so peak memory grows in
proportion to the series length times the window size. Bit-for-bit
reproducibility assumes deterministic regressor settings: the bundled
LightGBM wrapper enables LightGBM's deterministic and column-wise flags,
and single-threaded execution removes any remaining floating-point
reordering.

All runtime dependencies carry permissive licenses, which keeps the
combined distribution simple for integrators. The library itself is
distributed under the copyleft AGPL-3.0-or-later license. The dependency
licenses are: numpy, pandas, scikit-learn, and feature-engine under
BSD-3-Clause; numba under BSD; lightgbm and holidays under MIT; pyarrow,
requests, and astral under Apache-2.0; tqdm under MPL-2.0 and MIT.

Because the library performs no training and uses no GPU, its own energy
cost is small. Runtime cost is dominated by vector operations during
feature engineering and by whatever regressor the caller trains. A
typical LightGBM fit on an hourly series of about 100,000 rows completes
in seconds on one commodity CPU core. No pretrained weights ship with
the package, so there are no embedded training emissions to report.

\subsection{Data and Operational Design
Domain}\label{data-and-operational-design-domain}

The fixtures under \texttt{src/spotforecast2\_safe/datasets/csv/} and
the demo dataset support reproducible documentation and tests;
production data is supplied by the integrator. Docstring examples in the
source are executed as tests, and time-aware cross-validation is used
during validation so that no future observation can influence a past
prediction.

The Operational Design Domain (ODD) is the set of conditions under which
the library's results are valid. Outside these conditions the library is
designed to raise an error rather than return an unreliable result.

\begin{itemize}
\tightlist
\item
  \textbf{Target series}: numeric, univariate, with a regular and
  monotonic date-time index; anything else raises an error.
\item
  \textbf{Exogenous features}: numeric, complete, aligned to the target
  index; a \texttt{ValueError} is raised on any missing entry.
\item
  \textbf{Sampling interval}: uniform (hourly in the demo); non-uniform
  sampling yields unreliable results.
\item
  \textbf{Minimum history}: longer than the window size plus the number
  of lags, about 96 hourly points for the demo defaults; below that the
  model cannot be called.
\item
  \textbf{Missing target values}: rejected unless the caller explicitly
  enables imputation.
\item
  \textbf{Series length}: validated to about one million rows; beyond
  about ten million the caller must process the series in chunks to
  avoid memory exhaustion.
\item
  \textbf{Numeric precision}: feature matrices are computed in 32-bit
  floating point; values needing higher precision fall outside the
  domain.
\item
  \textbf{Any invalid input}: an explicit \texttt{ValueError} or
  \texttt{TypeError}, never silent repair.
\end{itemize}

Forecast accuracy is bounded by the downstream regressor and its
training data, so concept drift, seasonal shifts, or regime changes
degrade forecasts even when the feature engineering stays correct. Users
who build lag or calendar features outside the provided builders risk
leaking a target value into its own feature row; the bundled
\texttt{ExogBuilder} and task paths are leakage-free, hand-rolled
pipelines are not. Operating states that are scarce in the training data
are forecast less reliably than common ones.

To stay inside the valid domain, validate every new deployment against
historical ground truth before it carries live traffic, build features
only through \texttt{ExogBuilder} or the bundled tasks, keep the
regressor deterministic when reproducibility is required, and process
very long series in windowed chunks.

\subsection{Evaluation}\label{evaluation}

Because no training runs inside the library, classical accuracy metrics
do not describe the library itself. The library is evaluated on
software-quality properties, while forecast accuracy is a property of
each deployment.

DataFrames that contain missing or infinite values raise a
\texttt{ValueError}. The public loaders \texttt{load\_timeseries},
\texttt{load\_timeseries\_forecast}, and
\texttt{WeatherService.get\_dataframe} refuse to return silently imputed
values unless the caller opts in. Input types are checked at runtime,
and identical input yields identical output bytes. The test suite
collects 3039 tests; \texttt{CONTRIBUTING.md} names 80 percent line
coverage as the aim for new code, the enforced floor is
\texttt{fail\_under\ =\ 68} in \texttt{pyproject.toml}, and the suite
currently measures 75.99 percent line coverage. CPE identifier
generation is tested directly. Excluding heavy dependencies keeps the
Common Vulnerabilities and Exposures (CVE) attack surface small: there
is no web server, no deep-learning runtime, and no plotting backend.

For deployments, the demo computes mean absolute error (MAE) and mean
squared error (MSE) when it compares each model against ground truth.
These metrics are deterministic and can be reproduced by running the
demo. Their numerical values depend on the data vintage and are
therefore not fixed in this card.

\subsection{Model Transparency}\label{model-transparency}

The library produces point forecasts. It does not natively quantify or
calibrate predictive uncertainty, so a deployment that needs prediction
intervals or calibrated probabilities must add them in the downstream
regressor or a wrapper around it.

The code is white-box: there are no compiled inference kernels and no
opaque weights, so every transformation can be read and audited in
source. Feature attributions are available through the downstream
regressor's own importance measures, for example LightGBM's split and
gain importances. The package ships no separate explainability backend
such as SHAP or LIME, consistent with its minimal-dependency policy.

\subsection{Operation: Monitoring and
Response}\label{operation-monitoring-and-response}

A deployment should watch the quality of incoming data (missing or
out-of-range values and gaps in the timestamps), the drift of the input
and target distributions away from the training period, and the forecast
error measured against a simple baseline. The production configuration
carries a refit cadence, with a default of seven days, and a
maximum-model-age policy that signals when retraining is due.

When monitoring crosses a deployment-defined threshold, the usual
responses are to refit or retrain the model, to fall back to the
seasonal baseline forecaster or the last known-good model, and to alert
the responsible team. A dual-handler logger writes timestamped records
to the console and, when a log directory is supplied, to JSON files in
that directory (the bundled tasks default to
\texttt{\textasciitilde{}/spotforecast2\_safe\_models/logs/}), which
supports audit retention; each file is a self-contained SHA-256 hash
chain, so an edited, deleted, or reordered record is detectable offline
with \texttt{verify\_audit\_log}. A report without a presented anchor
explicitly states that external anchoring was not established, rather
than printing an unqualified pass; \texttt{make\ anchor} /
\texttt{make\ anchor-verify} automate binding the resulting head hash to
an independently controlled record (an RFC 3161 timestamp by default),
but running them, and how often, remains an operator duty. The
thresholds and escalation steps are owned by the integrator.

\subsection{Compliance Support}\label{compliance-support}

The package is built to support the development of high-risk AI systems
under the EU AI Act. The package itself is not certified; full-system
certification is the integrator's responsibility.

It rejects missing or infinite data by default, which supports the
data-governance duty of Article 10. This card together with the
technical report forms a technical-documentation baseline for Article
11, and the CPE identifiers in §A.1 feed SBOM and vulnerability-tracking
pipelines. The logging facility supports the record-keeping duty of
Article 12, and its integrity is evidenced by a per-file SHA-256 hash
chain. The white-box code supports the transparency duty of Article 13.
The deterministic, reproducible transformations support the
accuracy-and-robustness duty of Article 15, while formal system-level
verification remains the integrator's responsibility.

The authoritative mapping to IEC 61508, ISO 26262, ISA/IEC 62443, and
the individual EU AI Act articles is maintained in the technical report
(\texttt{bart26h/index.qmd}, section \emph{Compliance Mapping}). These
references reflect the standards as of 2026-04-19; users must track
later amendments themselves.

\subsection{Glossary}\label{glossary}

\begin{itemize}
\tightlist
\item
  \textbf{EU AI Act} --- Regulation (EU) 2024/1689 on artificial
  intelligence, in force since 2024-08-01.
\item
  \textbf{IEC 61508} --- International standard for the functional
  safety of electrical, electronic, and programmable electronic
  safety-related systems.
\item
  \textbf{ISA/IEC 62443} --- Standard series for the security of
  industrial automation and control systems.
\item
  \textbf{ISO 26262} --- International standard for the functional
  safety of road vehicles.
\end{itemize}

\subsection{How to Audit}\label{how-to-audit}

An auditor can validate this package as follows.

\begin{enumerate}
\def\labelenumi{\arabic{enumi}.}
\tightlist
\item
  Inspect \texttt{pyproject.toml} to confirm that none of the prohibited
  libraries (\texttt{plotly}, \texttt{matplotlib}, \texttt{spotoptim},
  \texttt{optuna}, \texttt{torch}, \texttt{tensorflow}) are present.
\item
  Run \texttt{uv\ run\ pytest\ tests/} to confirm functional correctness
  and the full test suite.
\item
  Run \texttt{uv\ run\ pytest\ tests/test\_cpe.py} to confirm CPE
  identifier generation.
\item
  Record the CPE identifiers from §A.1 in vulnerability-tracking systems
  and supply-chain disclosures.
\item
  Read \texttt{get\_cpe\_identifier} in
  \texttt{src/spotforecast2\_safe/utils/cpe.py} for use in automated
  workflows.
\item
  Run \texttt{uv\ run\ reuse\ lint} to confirm license and
  copyright-header compliance.
\item
  Run \texttt{make\ verify} to execute all of the above as one gate, and
  \texttt{make\ sbom} to regenerate the CycloneDX 1.5 software bill of
  materials with the project CPE injected on the root component.
\item
  Run \texttt{verify\_audit\_log} from
  \texttt{spotforecast2\_safe.manager} against a retained audit log file
  to confirm its SHA-256 hash chain is intact; if no
  \texttt{AuditAnchor} is presented, the returned report explicitly
  states that external anchoring was not established rather than
  printing an unqualified pass.
\item
  Run
  \texttt{make\ anchor-verify\ LOG=\textless{}file-or-directory\textgreater{}}
  to re-verify a previously anchored log fully offline: it checks the
  stored RFC 3161 token against the stored TSA certificate chain, then
  confirms the file's current chain still extends the anchored head
  hash. See the Anchoring Runbook (\texttt{docs/anchoring.qmd}) for the
  full workflow, including \texttt{make\ anchor} (which requires an
  explicit \texttt{TSA\_URL}; there is no default).
\end{enumerate}

\subsubsection{Verification environment}\label{verification-environment}

Steps 1 to 8 run on a maintainer workstation. This package is no longer
built or tested by a hosted continuous-integration service, so there is
no independent clean-room execution of the test suite on a second
platform, and released artefacts carry no Sigstore signature or
transparency-log entry. The lockfile remains cross-platform:
\texttt{{[}tool.uv{]}\ required-environments} forces dependency
resolution for macOS (arm64, x86-64) and Linux x86-64, so the resolved
versions are guaranteed to have wheels on each target even though the
tests execute on only one of them. The audit-log hash chain itself is
unkeyed and unanchored by this package: it detects edits, deletions, and
reordering within a retained file, but an adversary able to rewrite the
whole file can recompute it, so external anchoring of the head hash
remains the operator's responsibility. Auditors who require independent
verification should run the steps above in their own environment.

\subsection{Citation, Authors, and
Contact}\label{citation-authors-and-contact}

Maintainer: Thomas Bartz-Beielstein, ORCID
\href{https://orcid.org/0000-0002-5938-5158}{0000-0002-5938-5158},
\texttt{bartzbeielstein@gmail.com}.

\begin{Shaded}
\begin{Highlighting}[]
\VariableTok{@misc}\NormalTok{\{}\OtherTok{spotforecast2safe}\NormalTok{,}
  \DataTypeTok{author}\NormalTok{       = \{Bartz{-}Beielstein, Thomas\},}
  \DataTypeTok{title}\NormalTok{        = \{\{spotforecast2{-}safe\}: Safety{-}critical Subset of \{spotforecast2\}\},}
  \DataTypeTok{year}\NormalTok{         = \{2026\},}
  \DataTypeTok{version}\NormalTok{      = \{27.0.0\},}
  \DataTypeTok{howpublished}\NormalTok{ = \{}\CharTok{\textbackslash{}url}\NormalTok{\{https://gitlab.git.nrw/thk{-}f10/spotsevenlab/spotforecast2{-}safe\}\},}
  \DataTypeTok{note}\NormalTok{         = \{Version 27.0.0, AGPL{-}3.0{-}or{-}later\}}
\NormalTok{\}}
\end{Highlighting}
\end{Shaded}

\textbf{APA}: Bartz-Beielstein, T. (2026). \emph{spotforecast2-safe:
Safety-critical subset of spotforecast2} (Version 27.0.0) {[}Computer
software{]}.
\url{https://gitlab.git.nrw/thk-f10/spotsevenlab/spotforecast2-safe}

The technical report (\texttt{bart26h/index.qmd}) is the long-form
reference for design rationale, compliance mapping, and evaluation
protocol.

\subsection{Disclaimer and Liability}\label{disclaimer-and-liability}

\textbf{Limitation of liability.} This library is designed with safety
principles and deterministic logic, but it is provided as is, without
warranty of any kind. The authors and contributors accept no liability
for any direct or indirect damage, system failure, or financial loss
arising from its use.

It is the sole responsibility of the system integrator to perform full
system-level safety validation, for example under ISO 26262, IEC 61508,
or the EU AI Act, before deploying this software in a production or
safety-critical environment.

\end{tcolorbox}

\section{Externe Validierung}\label{sec-external-validation}

\subsection{Externe Validierung mit Deepchecks}\label{sec-deepchecks}

Die Merkmalsmatrix aus Lags und dem Kodierer mit Basisfunktionen (RBF,
\texttt{RepeatingBasisFunction}) ist zeilenweise unabhängig. Damit lässt
sich \texttt{deepchecks.tabular} als externe Validierungsstufe auf
\texttt{(X\_train,\ y\_train\_vec)} und einen analog aufgebauten
Evaluationsblock anwenden. Das folgende Listing skizziert vier
repräsentative Checks. Es ist bewusst \emph{nicht ausführend} in den
Dokumentenbuild eingebunden, weil Deepchecks einen eigenen transitiven
Abhängigkeitsraum mitbringt (unter anderem \texttt{plotly},
\texttt{pkg\_resources}, spezifische \texttt{scikit-learn}-Scorernamen),
der mit der minimalen Abhängigkeitshülle von \texttt{spotforecast2-safe}
kollidiert. Die aus der Einordnung in Kapitel~\ref{sec-related} folgende
Konsequenz ist eine organisatorische: Deepchecks wird von einem
\emph{nachgelagerten} Job betrieben, der eine eigene Umgebung mitbringt
und dessen Ausgaben (HTML-Berichte und reduzierte
JSON-Zusammenfassungen) als Audit-Artefakte in die Nachweiskette
eingespeist werden.

\begin{Shaded}
\begin{Highlighting}[]
\CommentTok{\# Nicht{-}ausgeführtes Listing — Deepchecks als externe Validierungsstufe.}
\CommentTok{\# Ausführung in separater Umgebung mit eigenen Abhängigkeits{-}Pins.}

\ImportTok{from}\NormalTok{ deepchecks.tabular }\ImportTok{import}\NormalTok{ Dataset}
\ImportTok{from}\NormalTok{ deepchecks.tabular.checks }\ImportTok{import}\NormalTok{ (}
\NormalTok{    FeatureDrift,}
\NormalTok{    TrainTestPerformance,}
\NormalTok{    RegressionErrorDistribution,}
\NormalTok{    WeakSegmentsPerformance,}
\NormalTok{)}

\CommentTok{\# Feature{-}Matrix für den Evaluationsbereich analog zu X\_train aufbauen}
\NormalTok{X\_eval\_rows, y\_eval\_rows }\OperatorTok{=}\NormalTok{ [], []}
\ControlFlowTok{for}\NormalTok{ t }\KeywordTok{in} \BuiltInTok{range}\NormalTok{(}\BuiltInTok{len}\NormalTok{(y\_train), }\BuiltInTok{len}\NormalTok{(y\_all)):}
\NormalTok{    X\_eval\_rows.append(}
\NormalTok{        np.concatenate([y\_all.values[t }\OperatorTok{{-}}\NormalTok{ lags : t], exog\_rbf.iloc[t].values])}
\NormalTok{    )}
\NormalTok{    y\_eval\_rows.append(y\_all.values[t])}
\NormalTok{X\_eval }\OperatorTok{=}\NormalTok{ pd.DataFrame(np.vstack(X\_eval\_rows), columns}\OperatorTok{=}\NormalTok{feature\_names)}
\NormalTok{y\_eval\_vec }\OperatorTok{=}\NormalTok{ pd.Series(y\_eval\_rows, name}\OperatorTok{=}\StringTok{"load"}\NormalTok{)}

\NormalTok{train\_tbl }\OperatorTok{=}\NormalTok{ pd.concat(}
\NormalTok{    [X\_train.reset\_index(drop}\OperatorTok{=}\VariableTok{True}\NormalTok{), y\_train\_vec.reset\_index(drop}\OperatorTok{=}\VariableTok{True}\NormalTok{)], axis}\OperatorTok{=}\DecValTok{1}
\NormalTok{)}
\NormalTok{eval\_tbl }\OperatorTok{=}\NormalTok{ pd.concat(}
\NormalTok{    [X\_eval.reset\_index(drop}\OperatorTok{=}\VariableTok{True}\NormalTok{), y\_eval\_vec.reset\_index(drop}\OperatorTok{=}\VariableTok{True}\NormalTok{)], axis}\OperatorTok{=}\DecValTok{1}
\NormalTok{)}
\NormalTok{train\_ds }\OperatorTok{=}\NormalTok{ Dataset(train\_tbl, label}\OperatorTok{=}\StringTok{"load"}\NormalTok{, cat\_features}\OperatorTok{=}\NormalTok{[])}
\NormalTok{eval\_ds }\OperatorTok{=}\NormalTok{ Dataset(eval\_tbl, label}\OperatorTok{=}\StringTok{"load"}\NormalTok{, cat\_features}\OperatorTok{=}\NormalTok{[])}

\CommentTok{\# 1. Feature{-}Drift zwischen Trainings{-} und Evaluationsfenster}
\NormalTok{fd }\OperatorTok{=}\NormalTok{ FeatureDrift().run(train\_ds, eval\_ds)}
\CommentTok{\# fd.value: \{feature\_name: \{"Drift score": float, "Method": str\}\}}

\CommentTok{\# 2. Train{-}vs{-}Test{-}Performance — Overfitting{-}Indikator}
\NormalTok{ttp }\OperatorTok{=}\NormalTok{ TrainTestPerformance().run(train\_ds, eval\_ds, model)}
\CommentTok{\# ttp.value: DataFrame mit Spalten Dataset/Metric/Value}

\CommentTok{\# 3. Residuenverteilung mit Kurtosis (Ergänzung zu fig{-}residuals)}
\NormalTok{red }\OperatorTok{=}\NormalTok{ RegressionErrorDistribution().run(eval\_ds, model)}
\CommentTok{\# red.value: float (Kurtosis der Residuen)}

\CommentTok{\# 4. Schwache Segmente des Modells}
\NormalTok{wsp }\OperatorTok{=}\NormalTok{ WeakSegmentsPerformance().run(eval\_ds, model)}
\CommentTok{\# wsp.value["weak\_segments\_list"]: DataFrame der ungünstigsten Feature{-}Bereiche}
\end{Highlighting}
\end{Shaded}

Zeitreihenphänomene (Autokorrelation benachbarter Lags und die erwartete
Trend- bzw. Saisondrift zwischen Trainings- und Evaluationsfenster)
werden von Deepchecks als Warnungen ausgegeben und müssen durch
Auditoren domänenspezifisch eingeordnet werden. Die
Rolling-Origin-Korrektheit (Abbildung~\ref{fig-folds}), der
Determinismus (CR-2, Kapitel~\ref{sec-principles}) und das
Fail-safe-Verhalten (CR-3, Kapitel~\ref{sec-principles}) liegen
außerhalb des Deepchecks-Umfangs und werden durch die ausführbaren
Docstring-Beispiele (Kapitel~\ref{sec-testen}) und den CI-Workflow des
Pakets abgedeckt.

\subsection{Externe Validierung mit Evidently AI}\label{sec-evidently}

\texttt{Evidently\ AI} deckt dieselbe tabulare Validierungsrolle wie
Deepchecks ab, verlagert den Fokus allerdings von der
Pre-Production-Validierung auf das Produktionsmonitoring: Reports werden
typischerweise periodisch gegen ein \emph{Referenzfenster} berechnet und
speisen Dashboards (Prometheus, Grafana, MLflow). Für die
Regressions-Stufe des Beispiels sind drei Presets einschlägig:
\texttt{DataDriftPreset} und \texttt{TargetDriftPreset} für
Verteilungsverschiebungen zwischen Trainings- und Evaluationsblock,
\texttt{RegressionPreset} für RMSE, MAE, Residuenverteilung,
Predicted-vs-Actual-Darstellung und eine pro Feature aufgeschlüsselte
Fehler-Bias-Tabelle. Ein Vorteil gegenüber Deepchecks: Evidently erkennt
über ein \texttt{ColumnMapping(datetime=...)} die Zeitordnung der
Evaluationsreihe und zeichnet Residuen zeitannotiert, was saisonale
Fehlermuster sichtbar macht, die Abbildung~\ref{fig-forecast24} und
Abbildung~\ref{fig-residuals} zusammengefasst nicht aufdecken. Das
folgende Listing ist aus denselben Gründen wie das Deepchecks-Pendant
\emph{nicht ausführend} in den Build eingebunden: Evidently bringt einen
eigenen transitiven Abhängigkeitsraum (unter anderem \texttt{plotly},
\texttt{scipy} und diverse Visualisierungs- und Textverarbeitungspakete)
mit, der mit der minimalen Abhängigkeitshülle von
\texttt{spotforecast2-safe} kollidiert.

\begin{Shaded}
\begin{Highlighting}[]
\CommentTok{\# Nicht{-}ausgeführtes Listing — Evidently AI als externe Monitoring{-}Stufe.}
\CommentTok{\# Ausführung in separater Umgebung; X\_train, X\_eval, y\_train\_vec, y\_eval\_vec}
\CommentTok{\# und model sind im Build{-}Kontext dieses Abschnitts aufgebaut.}

\ImportTok{from}\NormalTok{ evidently.report }\ImportTok{import}\NormalTok{ Report}
\ImportTok{from}\NormalTok{ evidently.metric\_preset }\ImportTok{import}\NormalTok{ (}
\NormalTok{    DataDriftPreset,}
\NormalTok{    TargetDriftPreset,}
\NormalTok{    RegressionPreset,}
\NormalTok{)}
\ImportTok{from}\NormalTok{ evidently.pipeline.column\_mapping }\ImportTok{import}\NormalTok{ ColumnMapping}

\CommentTok{\# Referenz{-} und Vergleichs{-}Fenster mit Target und Modellprediction}
\NormalTok{reference }\OperatorTok{=}\NormalTok{ X\_train.reset\_index(drop}\OperatorTok{=}\VariableTok{True}\NormalTok{).copy()}
\NormalTok{reference[}\StringTok{"target"}\NormalTok{] }\OperatorTok{=}\NormalTok{ y\_train\_vec.reset\_index(drop}\OperatorTok{=}\VariableTok{True}\NormalTok{).values}
\NormalTok{reference[}\StringTok{"prediction"}\NormalTok{] }\OperatorTok{=}\NormalTok{ model.predict(X\_train)}
\NormalTok{reference[}\StringTok{"timestamp"}\NormalTok{] }\OperatorTok{=}\NormalTok{ y\_train.index[lags:]  }\CommentTok{\# konsistent mit X\_train{-}Zeilen}

\NormalTok{current }\OperatorTok{=}\NormalTok{ X\_eval.reset\_index(drop}\OperatorTok{=}\VariableTok{True}\NormalTok{).copy()}
\NormalTok{current[}\StringTok{"target"}\NormalTok{] }\OperatorTok{=}\NormalTok{ y\_eval\_vec.reset\_index(drop}\OperatorTok{=}\VariableTok{True}\NormalTok{).values}
\NormalTok{current[}\StringTok{"prediction"}\NormalTok{] }\OperatorTok{=}\NormalTok{ model.predict(X\_eval)}
\NormalTok{current[}\StringTok{"timestamp"}\NormalTok{] }\OperatorTok{=}\NormalTok{ y\_eval.index}

\NormalTok{column\_mapping }\OperatorTok{=}\NormalTok{ ColumnMapping(}
\NormalTok{    target}\OperatorTok{=}\StringTok{"target"}\NormalTok{,}
\NormalTok{    prediction}\OperatorTok{=}\StringTok{"prediction"}\NormalTok{,}
\NormalTok{    datetime}\OperatorTok{=}\StringTok{"timestamp"}\NormalTok{,}
\NormalTok{    numerical\_features}\OperatorTok{=}\NormalTok{feature\_names,}
\NormalTok{)}

\CommentTok{\# Report 1: Daten{-} und Zieldrift (Train{-} vs. Evaluationsfenster)}
\NormalTok{drift\_report }\OperatorTok{=}\NormalTok{ Report(metrics}\OperatorTok{=}\NormalTok{[DataDriftPreset(), TargetDriftPreset()])}
\NormalTok{drift\_report.run(}
\NormalTok{    reference\_data}\OperatorTok{=}\NormalTok{reference,}
\NormalTok{    current\_data}\OperatorTok{=}\NormalTok{current,}
\NormalTok{    column\_mapping}\OperatorTok{=}\NormalTok{column\_mapping,}
\NormalTok{)}
\CommentTok{\# drift\_report.as\_dict()["metrics"]           → JSON{-}Zusammenfassung}
\CommentTok{\# drift\_report.save\_html("evidently\_drift.html")}

\CommentTok{\# Report 2: Regressions{-}Qualität mit zeitlich aufgelöstem Residualplot}
\NormalTok{regression\_report }\OperatorTok{=}\NormalTok{ Report(metrics}\OperatorTok{=}\NormalTok{[RegressionPreset()])}
\NormalTok{regression\_report.run(}
\NormalTok{    reference\_data}\OperatorTok{=}\NormalTok{reference,}
\NormalTok{    current\_data}\OperatorTok{=}\NormalTok{current,}
\NormalTok{    column\_mapping}\OperatorTok{=}\NormalTok{column\_mapping,}
\NormalTok{)}
\CommentTok{\# regression\_report.save\_html("evidently\_regression.html")}
\end{Highlighting}
\end{Shaded}

Wie bei Deepchecks bleiben Rolling-Origin-Korrektheit, Determinismus und
Fail-safe-Verhalten außerhalb des Evidently-Umfangs und sind durch die
paket-internen Mechanismen (Kapitel~\ref{sec-testen},
Kapitel~\ref{sec-principles}) abgedeckt. Die JSON-Zusammenfassungen aus
\texttt{as\_dict()} eignen sich als maschinenlesbare Audit-Artefakte für
Art. 15 KI-VO, die HTML-Reports als menschenlesbare Ergänzung der
Modellkarte.

Der in Schritt 7 ausgewiesene RMSE ist in den natürlichen Einheiten der
synthetischen Reihe angegeben. Das vorgestellte Beispiel ist hier
bewusst ohne Wettermerkmale gehalten, damit die Dokumentenerstellung
deterministisch und offlinefähig bleibt. Produktionsreife Ergebnisse
verlangen den \texttt{weather\_client}-Adapter sowie einen Live-Zugriff
auf Open-Meteo \citep{openmeteo24}.

\section*{References / Literaturverzeichnis}\label{bibliography}
\addcontentsline{toc}{section}{References / Literaturverzeichnis}
\protect\phantomsection\label{refs}
\printbibliography[heading=none]

\end{document}